\documentclass[12pt,letterpaper]{report}

\usepackage{listings}
\usepackage{booktabs}
\usepackage{subcaption}
\usepackage{hyperref}
\hypersetup{
  colorlinks, linkcolor=blue,
  pdfstartview={XYZ null null 1.00}
}
\usepackage{mathrsfs}

\usepackage{mleftright}
\usepackage{xparse}

\NewDocumentCommand{\evalat}{sO{\big}mm}{%
  \IfBooleanTF{#1}
   {\mleft. #3 \mright|_{#4}}
   {#3#2|_{#4}}%
}

\newcommand{\norm}[1]{\left\lVert#1\right\rVert}

\expandafter\def\expandafter\UrlBreaks\expandafter{\UrlBreaks
  \do\a\do\b\do\c\do\d\do\e\do\f\do\g\do\h\do\i\do\j%
  \do\k\do\l\do\m\do\n\do\o\do\p\do\q\do\r\do\s\do\t%
  \do\u\do\v\do\w\do\x\do\y\do\z\do\A\do\B\do\C\do\D%
  \do\E\do\F\do\G\do\H\do\I\do\J\do\K\do\L\do\M\do\N%
  \do\O\do\P\do\Q\do\R\do\S\do\T\do\U\do\V\do\W\do\X%
  \do\Y\do\Z}

\newcommand{\etal}{\textit{et al}. }
\newcommand{\ie}{\textit{i}.\textit{e}. }
\newcommand{\eg}{\textit{e}.\textit{g}. }

\newcommand{\thesistitle}{Evaluation of Transfer Learning for Classification of: (1) Diabetic Retinopathy by Digital Fundus Photography and (2) Diabetic Macular Edema, Choroidal Neovascularization and Drusen by Optical Coherence Tomography}
\newcommand{\thesisauthor}{Rony Gelman, M.D., M.S.}
\newcommand{\thesisadvisor}{Professor Carlos Fernandez-Granda}
\newcommand{\graddate}{January 2019}
\newcommand{\thesisdedication}{To my wife and kids, for their enduring patience while I completed the Master's degree and thesis.}

\usepackage{setspace}

\usepackage[final]{graphicx}

\usepackage{indentfirst}

\usepackage{amsthm} 
\usepackage[tbtags]{amsmath} 

\usepackage{amsfonts}
\usepackage{amssymb}

\newcommand{\R}{\ensuremath{\mathbb{R}}}
\newcommand{\Q}{\ensuremath{\mathbb{Q}}}
\newcommand{\Z}{\ensuremath{\mathbb{Z}}}
\newcommand{\N}{\ensuremath{\mathbb{N}}}
\newcommand{\T}{\ensuremath{\mathbb{T}}}

\begin{document}

\pagenumbering{roman}
%
\thispagestyle{empty}
\begin{center}
  {\large\textbf{\thesistitle}}
  \vspace{.7in}

  by
  \vspace{.7in}

  \thesisauthor
  \vfill

\begin{doublespace}
  A thesis submitted in partial fulfillment\\
  of the requirements for the degree of\\
  Master of Science\\
  Department of Mathematics\\
  New York University\\
  \graddate
\end{doublespace}
\end{center}
\vfill

\noindent\makebox[\textwidth]{\hfill\makebox[2.5in]{\hrulefill}}\\
\makebox[\textwidth]{\hfill\makebox[2.5in]{\hfill\thesisadvisor\hfill}}
\newpage
\thispagestyle{empty}
\vspace*{0in}
\newpage

\vspace*{\fill}
\begin{center}
  \thesisdedication\addcontentsline{toc}{section}{Dedication}
\end{center}
\vfill
\newpage
\section*{Acknowledgements}\addcontentsline{toc}{section}{Acknowledgements}
%
I would like to thank Professor Rob Fergus, who served as a thesis reader. His Computer Vision course at New York University built a core foundation for this work. I also would like to thank Professor Carlos Fernandez-Granda for serving as my thesis advisor and his guidance and support.\\
\indent I am also grateful for all the faculty that I have had over the years at the Courant Institute of Mathematical Sciences, as an undergraduate student in Computer Science and as a graduate student in Computer Science and Mathematics. The Courant Insitute has been an inspiring and wonderful place to challenge myself and learn.

\newpage
\section*{Abstract}\addcontentsline{toc}{section}{Abstract}
%
Deep learning has been successfully applied to a variety of image classification tasks. In the last few years, starting with the ground-breaking results of AlexNet at the ImageNet  Large Scale Visual Recognition Challenge (ILSVRC), there has been tremendous and rapid growth in deep learning and its potential applications. There has been keen interest to apply deep learning in the medical domain, particularly specialties that heavily utilize imaging, such as dermatology, pathology, radiology, and ophthalmology.\\
\indent One issue that may hinder application of deep learning to the medical domain is the vast amount of data necessary to train deep neural networks (DNNs). ImageNet comprises over 14 million labeled images, but because of regulatory and privacy issues associated with medicine, and the generally proprietary nature of data in medical domains, obtaining large datasets to train DNNs is a challenge, particularly in the ophthalmology domain.\\
\indent Transfer learning is a technique developed to address the issue of applying DNNs for domains with limited data. Prior reports on transfer learning have examined custom networks to fully train or used a particular DNN for transfer learning. However, to the best of my knowledge, no work has systematically examined a suite of DNNs for transfer learning for classification of diabetic retinopathy, diabetic macular edema, and two key features of age-related macular degneration.\\
\indent This work attempts to investigate transfer learning for classification of these ophthalmic conditions. Part \ref{part:one} gives a condensed overview of neural networks and the DNNs under evaluation. Part \ref{part:two} gives the reader the necessary background concerning diabetic retinopathy and prior work on classification using retinal fundus photographs. The methodology and results of transfer learning for diabetic retinopathy classification are presented, showing that transfer learning towards this domain is feasible, with promising accuracy.\\
\indent Part \ref{part:three} gives an overview of diabetic macular edema, choroidal neovascularization and drusen (features associated with age-related macular degeneration), and presents results for transfer learning evaluation using optical coherence tomography to classify these entities.

\newpage
\tableofcontents

\listoffigures\addcontentsline{toc}{section}{List of Figures}
\newpage

\listoftables\addcontentsline{toc}{section}{List of Tables}
\newpage

\pagenumbering{arabic} 
\section*{Introduction}\addcontentsline{toc}{section}{Introduction}
%
Recent advances in hardware, such as graphics processing units (GPUs), have made possible practical application of deep learning. Since the breakthrough work of AlexNet\cite{Krizhevsky} at the ImageNet  Large Scale Visual Recognition Challenge (ILSVRC) in 2012, there has been rapid and tremendous growth in deep learning, and attempts to apply it towards domains such as medical imaging. One practical issue is obtaining the vast amounts of data necessary for training. Unlike the ImageNet dataset, which numbers on order of tens of millions of images, there are limited open access datasets in ophthalmology. At present time, the largest public domain diabetic retinopathy photography dataset numbers on order of tens of thousands. The issue of limited open access datasets likely will not improve because of strict regulatory and privacy constraints placed on medical data.\\
\indent Transfer learning is a technique developed to address the issue of applying deep learning for domains with limited data\cite{transfer_learning}. The basic idea is to leverage the fundamental learning blocks built with a particular deep neural network (DNN), such as ResNet, and to ``re-train'' the DNN for a particular domain of interest. Thus, one can utilize the strong ``fixed feature extractor'' capabilities of a DNN built on the millions of training examples from ImageNet to detect features common to all domains, such as object edges, and then re-train just the ``top layer'' for classification with the limited training data from a particular target domain.\\
\indent In this paper, an overview of neural networks and the DNNs evaluated for transfer learning is presented in part \ref{part:one}. A condensed review is provided of the underlying algorithm and mathematics involved with back-propagation and training the neural network. In part \ref{part:two}, the reader is provided with a background of diabetic retinopathy: the prevalence numbers and magnitude of the projected numbers of an increasing population of diabetic patients, and the limited supply of trained ophthalmologists and other screening readers able to keep up with the rising demand. An overview of diabetic retinopathy classification is provided as well as the dataset used in this study (from the Kaggle Diabetic Retinopathy Detection competition). Lastly, the methodology used to evaluate transfer learning with a suite of DNNs towards diabetic retinopathy classification is presented. The results indicate that certain modes of transfer learning yield higher classification performance, and while all of the DNNs do well, certain DNNs may be better candidates for utilization in automated screening of diabetic retinopathy with deep learning.\\
\indent In part \ref{part:three}, this report details the evaluation of transfer learning for classification of diabetic macular edema, choroidal neovascularization, and drusen (features associated with age-related macular degeneration) using a different imaging modality called optical coherence tomography (OCT). OCT has emerged as a pivotal imaging modality in ophthalmology and critical for the diagnosis and managment of many retinal diseases. Methodology similar to part \ref{part:two} is used to evaluate transfer learning for classification of these disease entities. The results concur with findings in part \ref{part:two}, supporting that deep learning has strong promise for accurately classifying these eye conditions.

\part{Deep Learning\label{part:one}}%
\chapter{Background: Deep Learning\label{chap:background_DL}}

\renewcommand{\vec}[1]{\mathbf{#1}}

This chapter is a brief overview of deep learning concepts, adapted from a set of course lecture notes\cite{computer_vision_notes}. For further details, the reader is refered to the online reference by Goodfellow \etal\cite{Goodfellow}.

\section{Neural Networks}
\begin{figure}[htb!]
\centering\includegraphics[width=0.6\linewidth]{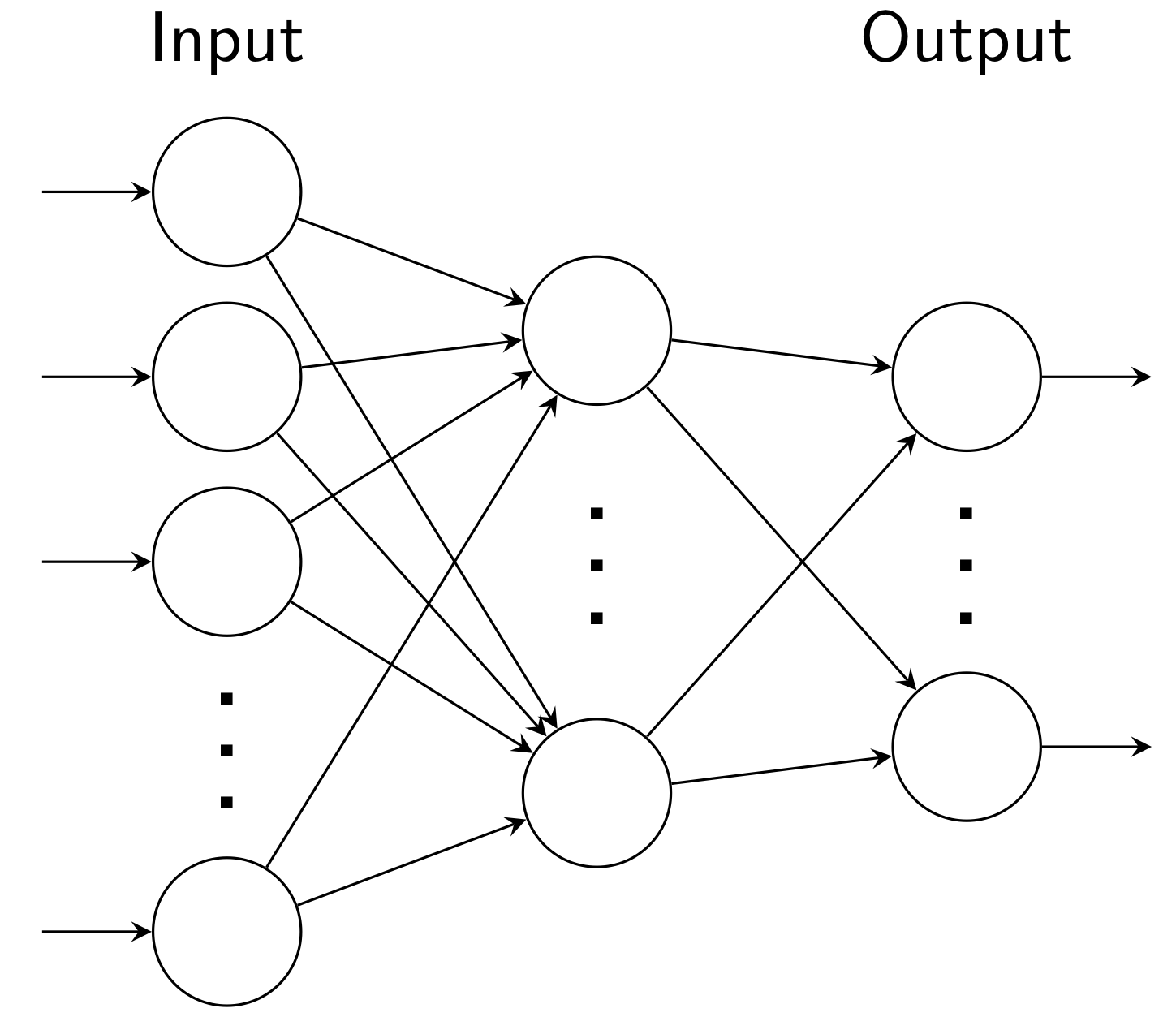}
\caption[Schematic representation of a feed-foward neural network model]{Schematic representation of a feed-foward neural network model. Reproduced from \cite{computer_vision_notes}.}
\label{fig:dl_fig1}
\end{figure}

\indent A neural network is composed of layers of artificial neurons, where each layer computes some function of the layer beneath it (figure \ref{fig:dl_fig1}). The input is mapped in a feed-forward fashion to the output. In this report, only feed-foward models are considered (no cycles).

\begin{figure}[htb!]
\centering\includegraphics[width=0.6\linewidth]{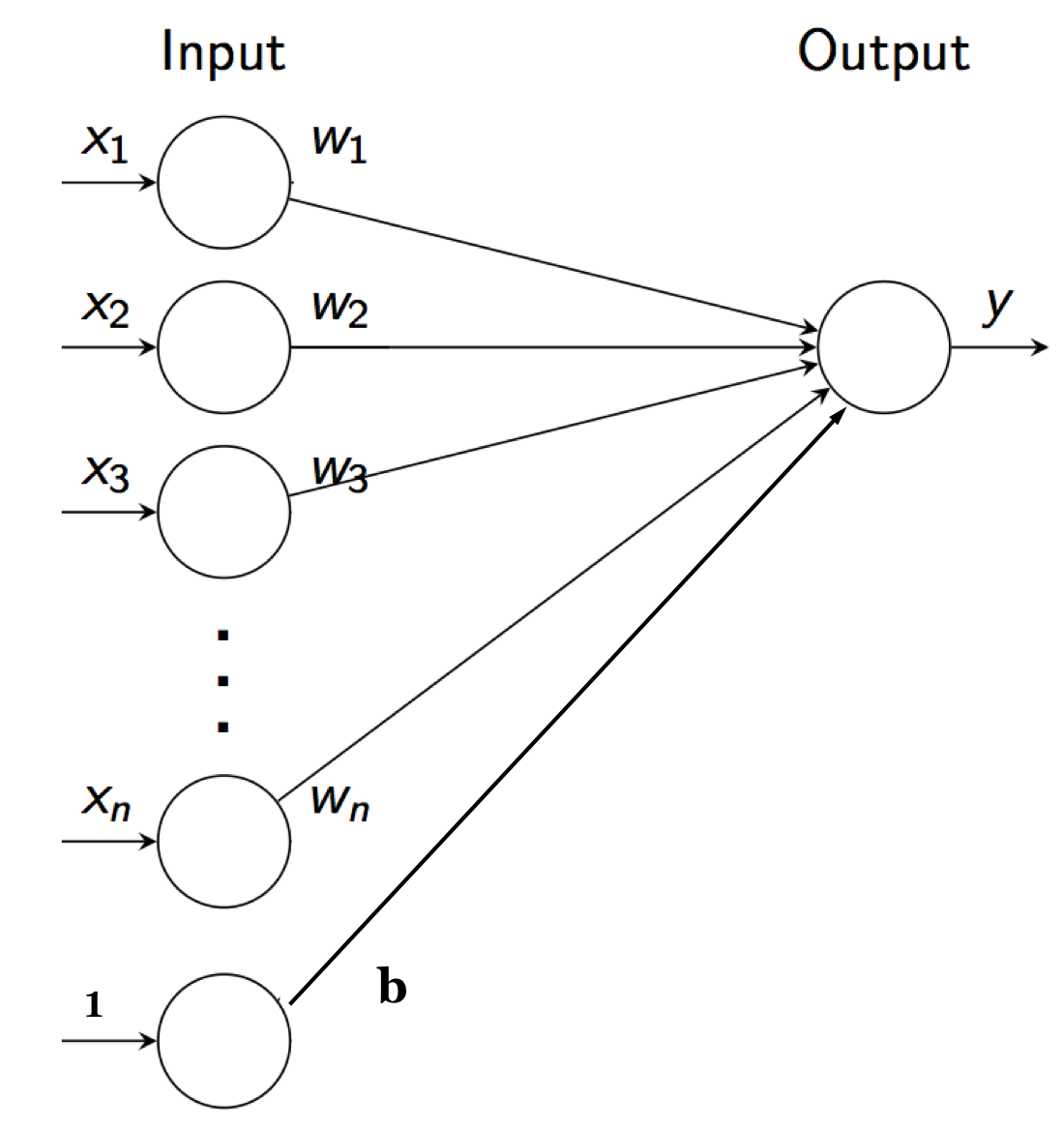}
\caption[Schematic of a feed-foward neural network with parameters and single output]{Schematic representation of a feed-foward neural network model with parameters (weights and bias) and single output. Reproduced from \cite{computer_vision_notes}.}
\label{fig:dl_fig2}
\end{figure}

\indent Neural networks have their origins in the 1940's and 1950's and in particular the Perceptron which were single layer networks with a simple learning rule (see \cite{Goodfellow}, Introduction chapter). Practical ways to train these networks were developed in the mid-1980's with the back-propogation algorithm (see \cite{Goodfellow}, chapter 6).\\
\indent In the simpliest design, the input to a neural network, denoted by $\vec{x}$, is a \texttt{[nx1]} vector (figure \ref{fig:dl_fig2}). The parameters are the weights, denoted by $\vec{w}$ (also a \texttt{[nx1]} vector), and a bias scalar, denoted by \texttt{b}. An activation, denoted by \texttt{a}, is a scalar defined by:
\begin{equation}
a = \sum_{i=1}^{n} x_i w_i + b
\end{equation}

A point-wise non-linear function, denoted by $\sigma(\cdot)$, is then applied to generate the output, $\vec{y} = f(a) = \sigma(\sum_{i=1}^{n} x_i w_i + b$).

\begin{figure}[htb!]
\centering\includegraphics[width=0.6\linewidth]{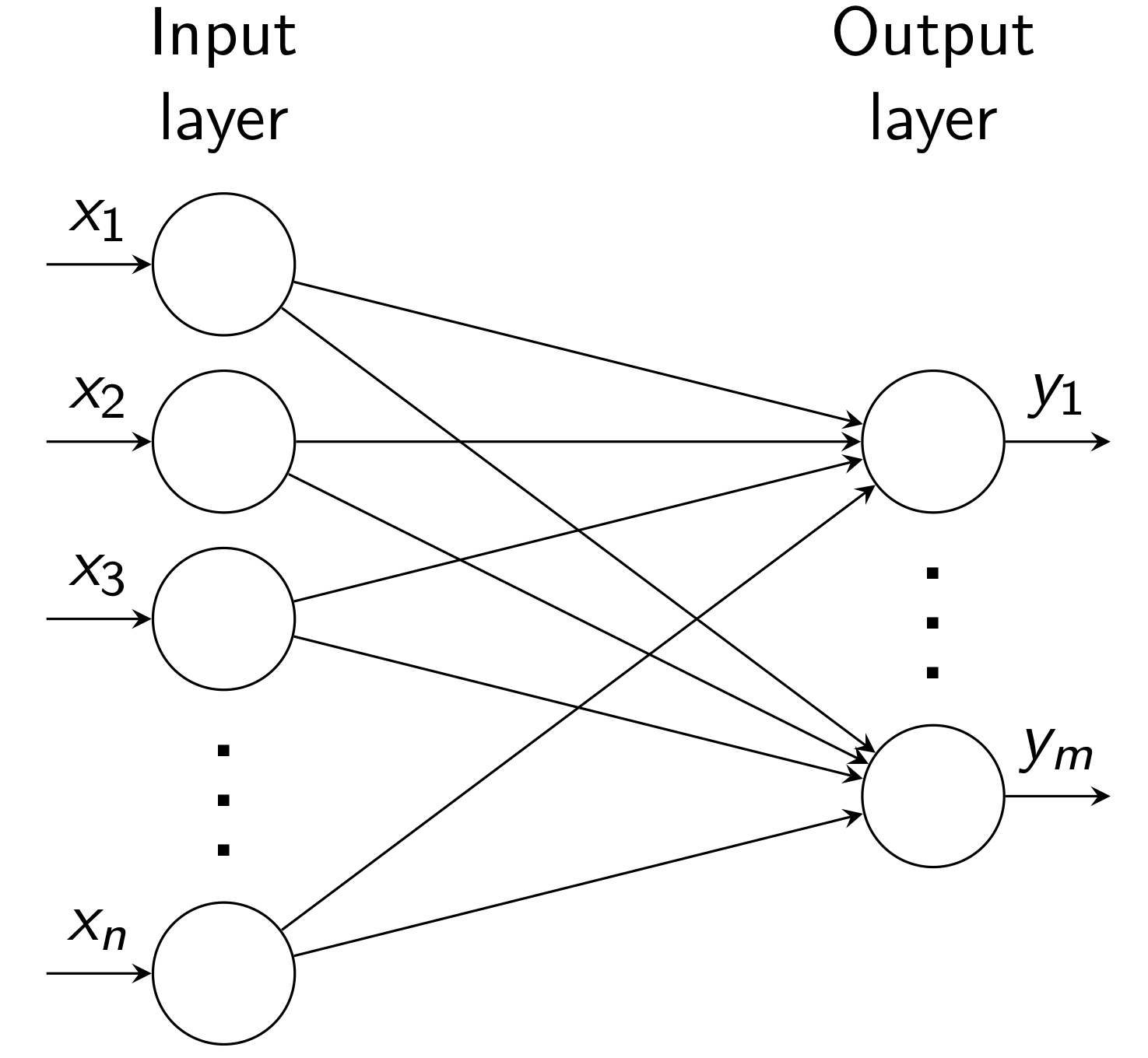}
\caption[Schematic of a feed-foward neural network with parameters and multiple outputs]{Schematic representation of a feed-foward neural network model with parameters (weights and bias) and multiple outputs. Reproduced from \cite{computer_vision_notes}.}
\label{fig:dl_fig3}
\end{figure}

The structure is modified slightly in the case of multiple outputs (figure \ref{fig:dl_fig3}). The input remains $\vec{x}$, a \texttt{[nx1]} vector. There are $m$ neurons, and thus a \texttt{[nxm]} weight matrix $\vec{W}$ and \texttt{[mx1]} bias vector $\vec{b}$. The \texttt{[mx1]} output vector $\vec{y}$ is defined by $\vec{y} = \sigma(\vec{W}\vec{x} + \vec{b})$.

Some choices for the non-linearity function $\sigma$ include: (1) the sigmoid function $\sigma(z) = \frac{1}{1+e^{-z}}$; (2) the Tanh function, $\sigma(z) = tanh(z)$; (3) the Rectified Linear (ReLU) function, $\sigma(z) = max(z,0)$; and (4) the Leaky ReLU (PReLU) $\sigma(z) = 1[z>0]max(0,z)+1[z<0]max(0,\alpha z)$. The pros and cons of each function are discussed in \cite{computer_vision_notes} and \cite{Goodfellow}.

\begin{figure}[htb!]
\centering\includegraphics[width=0.6\linewidth]{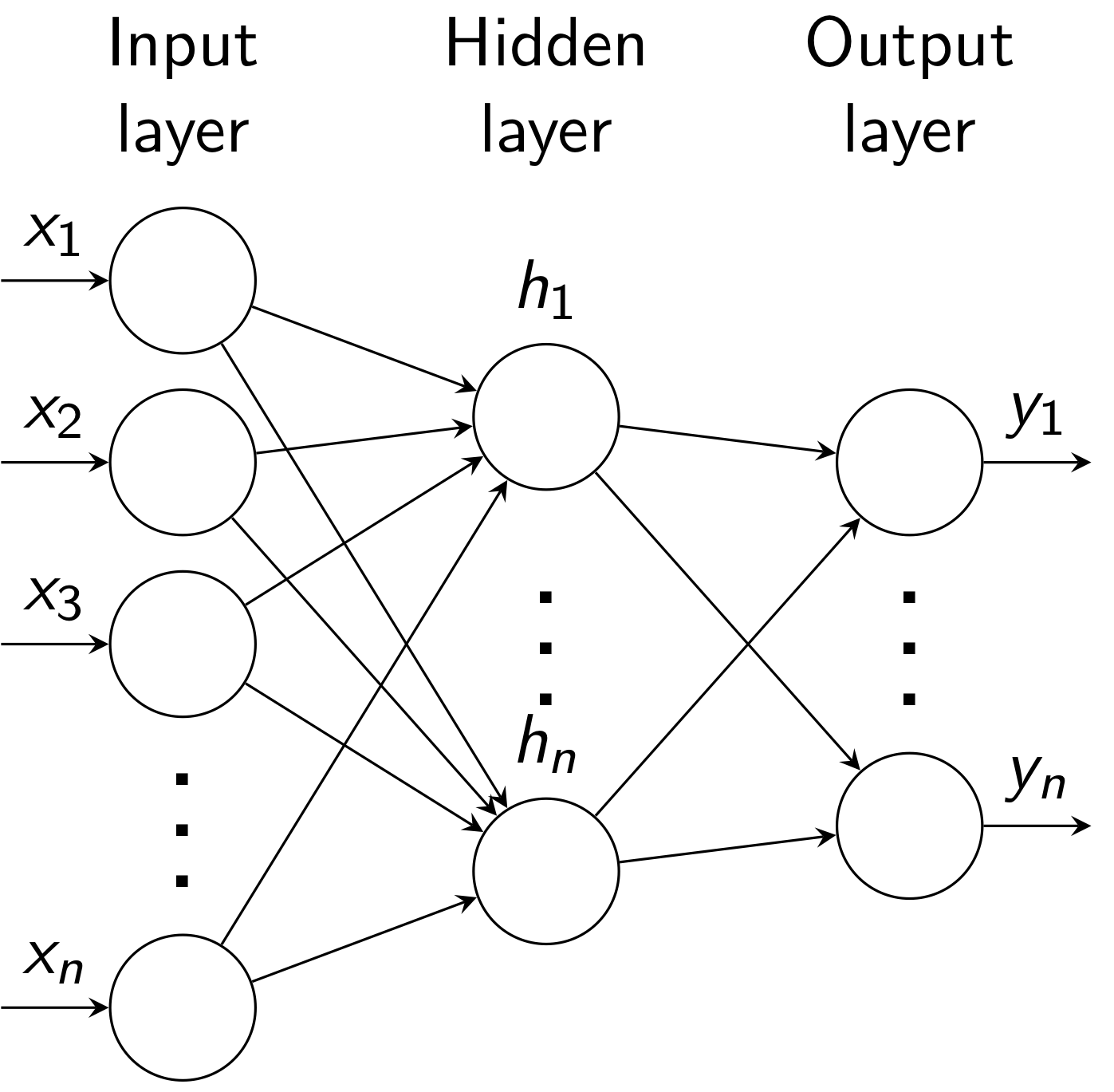}
\caption[Schematic of a feed-foward neural network with a hidden layer and multiple outputs]{Schematic representation of a feed-foward neural network model with a hidden layer and multiple outputs. Reproduced from \cite{computer_vision_notes}.}
\label{fig:dl_fig4}
\end{figure}

\begin{figure}[htb!]
\centering\includegraphics[width=0.6\linewidth]{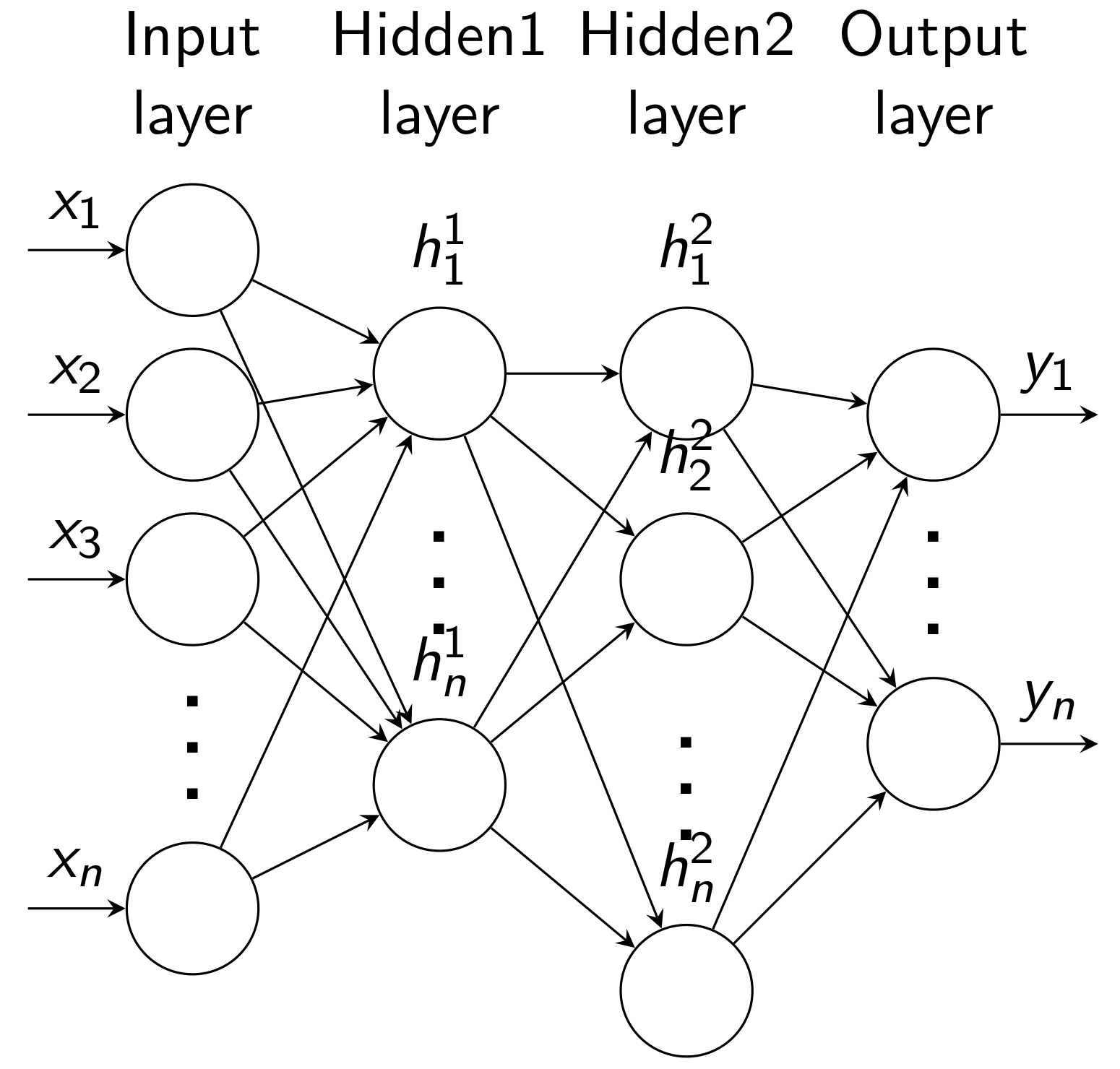}
\caption[Schematic of a feed-foward neural network with multiple hidden layers and multiple outputs]{Schematic representation of a feed-foward neural network model with multiple hidden layers and multiple outputs. Reproduced from \cite{computer_vision_notes}.}
\label{fig:dl_fig5}
\end{figure}

A neural network is typically composed of multiple layers of neurons (figure \ref{fig:dl_fig4}). This is an acyclic structure with an assumption of full connections between layers. The layer(s) between the input and output layers are called hidden. Other terminology for this type of structure includes Artificial Neural Networks (ANN), Multi-layer Perceptron (MLP), and a fully-connected network. By convention, the number of layers is equal to the number of hidden layers plus the output layer (\ie, excludes the input layer). An sample schematic of a multi-layered network is shown in figure \ref{fig:dl_fig5}.\\
\indent The issue of architecture selection, specifically how to pick the number of layers and units per layer, is difficult to determine. For fully connected models, it appears that 2-3 layers seems to be the most that can be effectively trained\cite{computer_vision_notes}. Moreover, the number of parameters grow with the square of the number of units per layer and with a large number of units per layer, overfitting may be an issue. Convolutional Neural Networks (CNNs), discussed below, address this limit in fully connected networks.


Once a model architecture has been selected, the next step is to train the model. The details of the procedure are presented in \cite{computer_vision_notes} and \cite{Goodfellow}. Briefly, the steps include:
\begin{itemize}
\item Given the dataset of input $\vec{x}$ and output $\vec{y}$, pick an appropriate cost function, $C$.
\item Forward pass the input examples throught the model to arrive at the predictions.
\item Calculate the error using the cost function $C$ to compare the predictions.
\item Apply back-propogation to pass the error back through the model, adjusting the parameters to minimize the energy $E$.
\item Once the gradients are established, use Stochastic Gradient Descent (SGD) to update the network weights.
\end{itemize}

\begin{figure}[htb!]
\centering\includegraphics[width=0.6\linewidth]{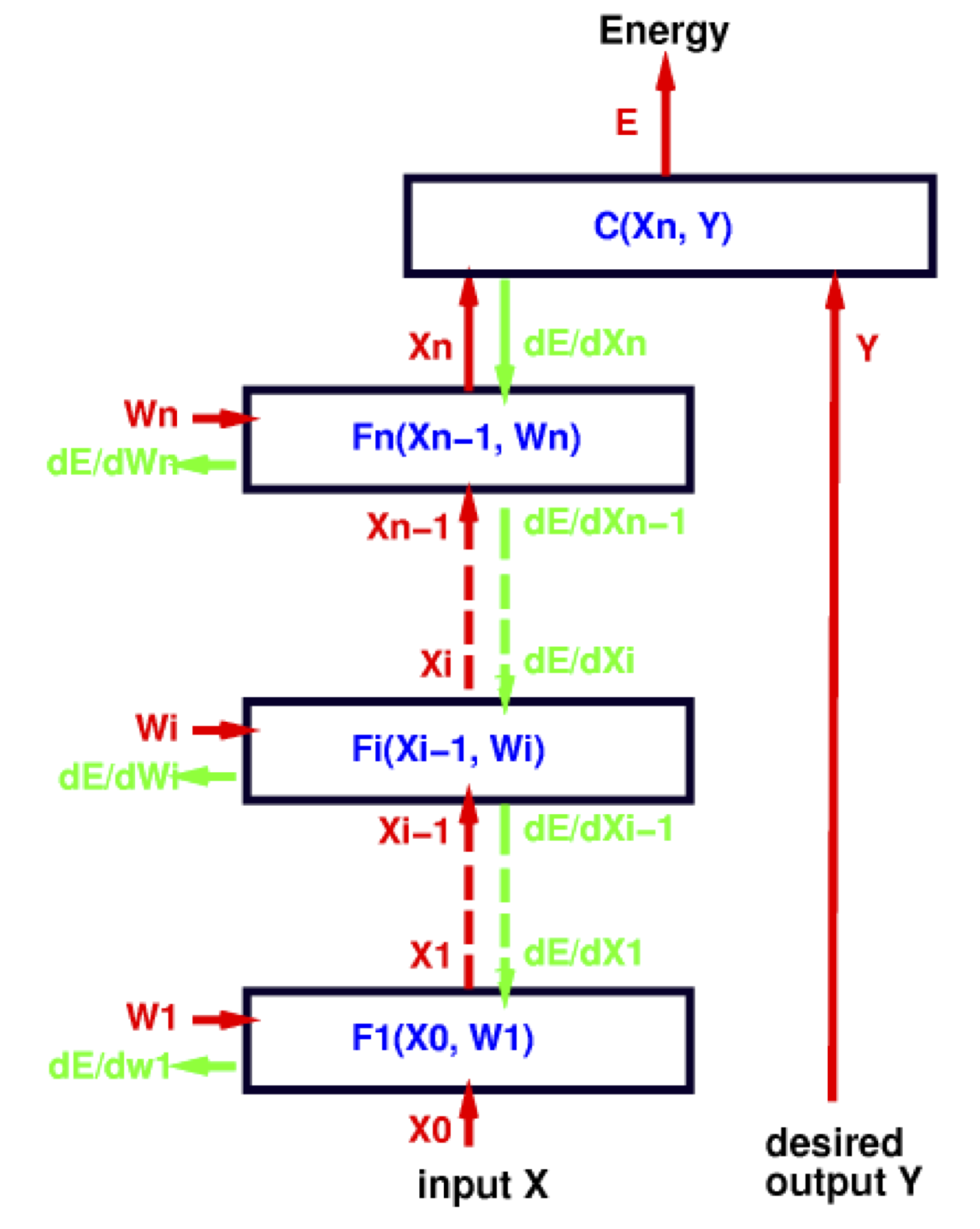}
\caption[Schematic of computing gradients in a multi-stage neural network.]{Schematic of computing gradients in a multi-stage neural network. Reproduced from \cite{computer_vision_notes}.}
\label{fig:dl_fig6}
\end{figure}

In SGD, we start with some initial set of parameters, denoted by $\theta^{0}$, and the updates are defined by $\theta^{k+1} \leftarrow\theta^{k} + \eta \Delta\theta$, where $k$ is an iteration index, $\eta$ is the learning rate, and the gradients $\Delta\theta = \frac{\partial E}{\partial \theta}$.\\
\indent Figure \ref{fig:dl_fig6} shows a schematic for computing gradients in a multi-stage architecture, where the model has $N$ layers. Each layer, $i$, has a vector of weights $\vec{W_i}$. Forward propagation takes the input $\vec{x}$ and passes it through each layer $\vec{F_i}$, such that $\vec{x_i} = \vec{F_i}(\vec{x_{i-1}},\vec{W_i})$. The prediction, $x_n$, is the output of the top layer and the cost function, $C$, compares $x_n$ to $y$. The overall energy, $E$, is defined by $E = \sum_{m=1}^{M} C(x_n^m,y^m)$, where $m$ is the number of inputs.\\
\indent Back-propagation is performed via the chain-rule. In the most general case, consider input vector $\vec{x}$, a \texttt{[nx1]} column vector:
\begin{align}
    \vec{x} &= \begin{bmatrix}
           x_{1} \\
           x_{2} \\
           \vdots \\
           x_{n}
         \end{bmatrix}
  \end{align}

\noindent and consider a function $y=\vec{F}(\vec{x})$ such that $\vec{y}$ is a \texttt{[mx1]} vector, then the \texttt{[mxn]} Jacobian matrix is given by:
\begin{equation}
\frac{\partial \vec{y}}{\partial \vec{x}} =
\begin{bmatrix}
  \frac{\partial y_1}{\partial x_1} & 
    \frac{\partial y_1}{\partial x_2} & 
    \hdots &
    \frac{\partial y_1}{\partial x_n} \\[1ex] 
   \vdots & \vdots & \vdots & \vdots \\
  \frac{\partial y_m}{\partial x_1} & 
    \frac{\partial y_m}{\partial x_2} & 
    \hdots  &
    \frac{\partial y_m}{\partial x_n} \\[1ex]
\end{bmatrix}
\end{equation}

\noindent Recall the chain rule for matrices: consider function $h(\vec{x}) = f_n(f_{n-1}(\ldots (f_1(\vec{x}))))$. Let $u_1 = f_1(\vec{x}), u_i = f_i(u_{i-1}), z = u_n = f_n(u_{n-1})$. Then, the derivative is given by a product of matrices:
\begin{equation}
\evalat[\bigg]{\frac{dz}{dx}}{x=a} = \evalat[\bigg]{\frac{dz}{du_{n-1}}}{u_{n-1}=f_{n-1}(u_{n-2})} \cdot \evalat[\bigg]{\frac{du_{n-1}}{du_{n-2}}}{u_{n-2}=f_{n-2}(u_{n-3})} \cdots \evalat[\bigg]{\frac{du_2}{du_1}}{u_1=f_1(a)} \cdot \evalat[\bigg]{\frac{du_1}{dx}}{x=a}
\end{equation}

\noindent The energy, $E$, is computed as the sum of the costs associated to each training example $x^m,y^m$:
\begin{equation}
E(\theta) = \sum_{m=1}^M C(x_n^m,y^m;\theta)
\end{equation}

\noindent and its gradient is:
\begin{equation}
\frac{\partial E}{\partial \theta_{i}} = \sum_{m=1}^M \frac{C(x_n^m,y^m;\theta)}{\partial \theta_i}
\end{equation}

\noindent Express the cost function as $C(x_n,y;\theta) = C(F_n(x_{n-1},w_n),y)$ with $\theta = [w_1, w_2, \ldots, w_n]$. Then,
\begin{equation}
\frac{\partial C}{\partial w_n} = \frac{\partial C}{\partial x_n} \cdot \frac{\partial x_n}{\partial w_n} = \frac{\partial C}{\partial x_n} \cdot \frac{\partial F_n(x_{n-1},w_n)}{\partial w_n}
\end{equation}

\noindent There are various choices for the cost function $C$ (see \cite{Goodfellow}). One common selection is the Euclidean loss: 
\begin{equation}
C(x_n,y) = \frac{1}{2}\norm{x_n - y}^2
\end{equation}

\noindent and thus the gradient is $\frac{\partial C}{\partial x_n} = x_n - y$. Apply the chain rule to compute the gradient with respect to $w_i$:
\begin{equation}
\frac{\partial C}{\partial w_i} =\frac{\partial C}{\partial x_n}\cdot \frac{\partial x_n}{\partial x_{n-1}}\cdot \frac{\partial x_{n-1}}{\partial x_{n-2}} \cdots \frac{\partial x_{i+1}}{\partial x_i} \cdot \frac{\partial x_i}{\partial w_i}
\end{equation}

\noindent Through back-propagation, if we have the value of $\frac{\partial C}{\partial x_i}$ we can compute the gradient of the layer below as $\frac{\partial C}{\partial x_{i-1}} = \frac{\partial C}{\partial x_i}\cdot \frac{\partial x_i}{\partial x_{i-1}}$. Thus, layer $i$ has two inputs, $x_{i-1}$ and $\frac{\partial C}{\partial x_i}$. At layer $i$, compute the derivatives $\frac{\partial F_i(x_{i-1},w_i)}{\partial x_{i-1}}$ and $\frac{\partial F_i(x_{i-1},w_i)}{\partial w_i}$ and obtain outputs $x_i = F_i(x_{i-1},w_i)$ and $\frac{\partial C}{\partial x_{i-1}} = \frac{\partial C}{\partial x_i} \cdot \frac{\partial F_i(x_{i-1},w_i)}{\partial x_{i-1}}$. Then, the weight update equations are applied: $\frac{\partial C}{\partial w_i} = \frac{\partial C}{\partial x_i}\cdot \frac{\partial F_i(x_{i-1},w_i)}{\partial w_i}$ and $w_i^{k+1} = w_i^k + \eta_t \frac{\partial E}{\partial w_i}$.\\
\\
\indent In summary, the backpropagation algorithm involves: (1) a forward pass where for each training example we compute the output for all the layers, $x_i = F_i(x_{i-1},w_i)$; (2) a backwards pass where we compute cost derivatives iteratively from top to bottom $\frac{\partial C}{\partial x_{i-1}} = \frac{\partial C}{\partial x_i}\cdot \frac{\partial F_i(x_{i-1},w_i)}{\partial x_{i-1}}$; and (3) compute gradients and update the weights.\\

\indent The softmax function takes an un-normalized vector and normalizes it into a probability distribution, such that after the softmax operation each element $x_i$ is in $[0,1]$ and $\sum_i x_i = 1$\cite{Goodfellow}. This function is useful for neural networks, so that the un-normalized output can be mapped to a probability distribution over predicted output classes. One softmax function can be implemented by:
\begin{equation}
\sigma(\vec{z})_j = \frac{e^{-\beta z_j}}{\sum_k^K e^{-\beta z_k}} 
\end{equation}
\noindent for $j = 1,\ldots,K$ and  $\beta \in \R$. It is often combined with the cross-entropy cost function, $E = -\sum_{c=1}^M y_{o,c}\log(p_{o,c})$, where $M$ is the number of classes, $y$ is binary indicator (0 or 1) if class label $c$ is the correct classification for observation $o$, and $p$ is the predicted probability observation $o$ is of class $c$ (\eg from the softmax).\\
\indent A constant learning rate $\eta$ is typically not optimal. Techniques to optimize the learning rate, such as annealing of learning rate, AdaGrad, RMSProp, and ADAM, are discussed further in \cite{computer_vision_notes} and \cite{Goodfellow}. A momemtum term can be added to the weight update to encourage updates to follow the previous direction. This usually helps speed up convergence. The update then become $\theta^{k+1} \leftarrow \theta^k + \alpha(\Delta\theta)^{k-1} - \eta\Delta\theta$, where $\alpha$ is typically around 0.9.\\

\section{Convolutional Neural Networks}

\begin{figure}[htb!]
\centering\includegraphics[width=0.8\linewidth]{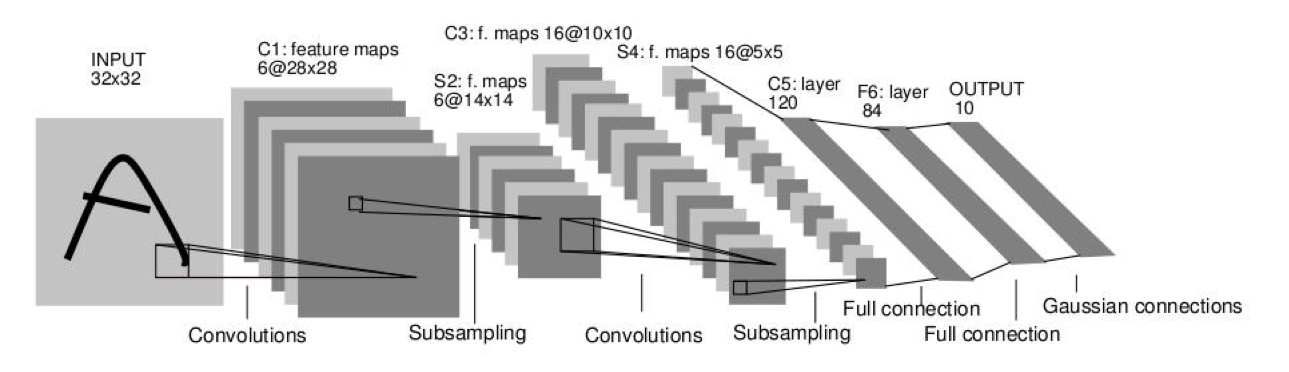}
\caption[Schematic of a CNN.]{Schematic of a CNN (LeNet-5), reproduced from \cite{LeCun}.}
\label{fig:convnet}
\end{figure}

A convolutional neural network (CNN), a form of supervised learning, is a neural network with a specialized connectivity structure where higher stages compute global, more invariant features (figure \ref{fig:convnet})\cite{computer_vision_notes}-\cite{LeCun}. The CNN model is feed-forward, where input images are fed to convolution layer(s), non-linearities, pooling layers, and finally to feature maps. They have been shown to be very successfully applied towards handwritting recognition\cite{Ciresan} and other recognition tasks (see \cite{computer_vision_notes}).
\indent The convolutional layers apply a convolution operation\footnote{See \url{https://towardsdatascience.com/intuitively-understanding-convolutions-for-deep-learning-1f6f42faee1} for an overview of the convolution operation.}  with a fixed sized filter, \eg \texttt{3x3} or \texttt{5x5}, across the image. The filter is learned during training, and options such as the stride, padding, and dilation\footnote{See the PyTorch documentation at \url{https://pytorch.org/docs/stable/nn.html} for an example implementation, \texttt{torch.nn.Conv2d}, discussing these options.} can be set. The pooling layer\footnote{See \url{http://cs231n.github.io/convolutional-networks/\#pool}} helps to reduce the spatial size of the representation, thus reducing the number of parameters and computation, and to prevent overfitting.\\
\indent ImageNet\footnote{\url{http://www.image-net.org/}} is a large scale image database with over 14 million labeled images and over 20K classes. The ImageNet  Large Scale Visual Recognition Challenge (ILSVRC) is an annual competition, which in 2012 saw a major breakthrough via the AlexNet CNN of Krizhevsky \etal\cite{Krizhevsky}, achieving a top-1 and top-5 error rates of 39.7\% and 18.9\% which was considerably better than the previous state-of-the-art results.

This section of the overview of deep learning will briefly review 5 DNNs that were evaluated in this report. These networks are AlexNet, ResNet, DenseNet, Inception\_v3, and VGG. Their variants and error rates on ImageNet are listed in table \ref{tab:networks}. The suseqeuent subsections briefly review each network.

\begin{table}[htb]
\caption[Network models evaluated.]{Networks that were evaluated in this paper. Listed are the Top-1 and Top-5 ImageNet 1-crop error rates (224x224)}\label{tab:networks}
  \begin{center}
\scalebox{0.7}{
 \begin{tabular}{|l|r|r|}

\hline
\textbf{Network} & \textbf{Top-1 error} & \textbf{Top-5 error}\\
\hline
AlexNet & 43.45 & 20.91\\
VGG-11 & 30.98 & 11.37\\
VGG-13 & 30.07 & 10.75\\
VGG-16 & 28.41 & 9.62\\
VGG-19 & 27.62 & 9.12\\
VGG-11 with batch normalization & 29.62 & 10.19\\
ResNet-18 & 30.24 & 10.92\\
ResNet-34 & 26.70 & 8.58\\
ResNet-50 & 23.85 & 7.13\\
ResNet-101 & 22.63 & 6.44\\
ResNet-152 & 21.69 & 5.94\\
Densenet-121 & 25.35 & 7.83\\
Densenet-169 & 24.00 & 7.00\\
Densenet-201 & 22.80 & 6.43\\
Densenet-161 & 22.35 & 6.20\\
Inception v3 & 22.55 & 6.44\\
\hline
\end{tabular}}
  \end{center}
\end{table}

\subsection{AlexNet}

\begin{figure}[htb!]
\centering\includegraphics[width=0.8\linewidth]{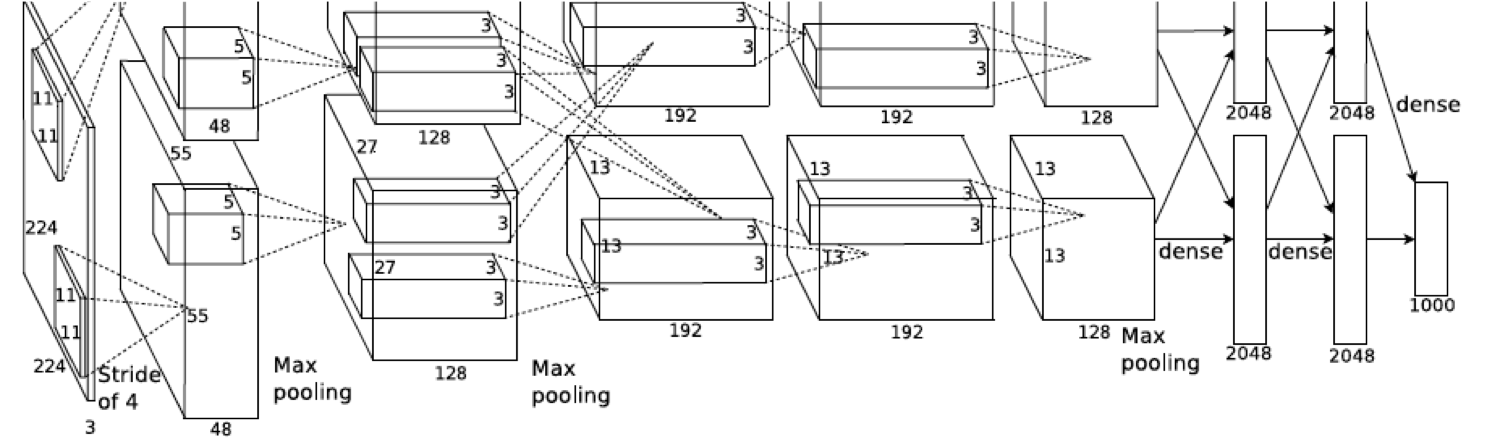}
\caption[Schematic of AlexNet.]{Schematic of AlexNet, reproduced from \cite{Krizhevsky}.}
\label{fig:alexnet}
\end{figure}

The architecture of AlexNet is shown in figure \ref{fig:alexnet}\cite{Krizhevsky}. This network has 60 million parameters and 500,000 neurons, consists of 5 convolutional layers, some of which are followed by max-pooling layers, and two globally connected layers with a final 1000-way softmax. To speed up training, they used a GPU implementation and reduced overfitting in the globally connected layers by employing a new regularization method.

\subsection{ResNet}
\begin{figure}[htb!]
\centering\includegraphics[width=0.8\linewidth]{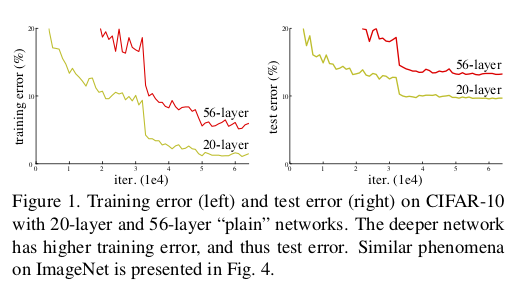}
\caption[Increasing error on CIFAR-10 with more layers.]{Increasing training and test error on CIFAR-10 with increasing network layers, reproduced from \cite{Resnet}.}
\label{fig:deeper_nets_error}
\end{figure}

\begin{figure}[htb!]
\centering\includegraphics[width=0.8\linewidth]{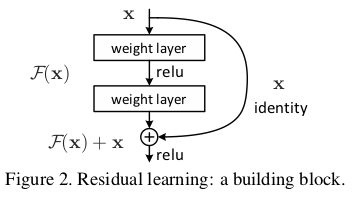}
\caption[Residual learning: a building block.]{Redisual learning: a building block, reproduced from \cite{Resnet}.}
\label{fig:residual_net_block}
\end{figure}

\begin{figure}[htb!]
\centering\includegraphics[width=0.8\linewidth]{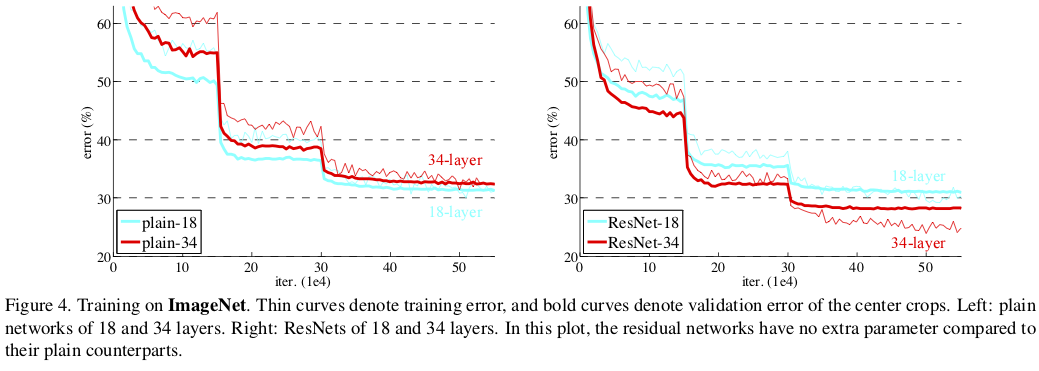}
\caption[ResNet training on ImageNet.]{ResNet training on ImageNet, reproduced from \cite{Resnet}.}
\label{fig:Resnet_results}
\end{figure}

\begin{figure}[htb!]
\centering\includegraphics[width=0.5\linewidth]{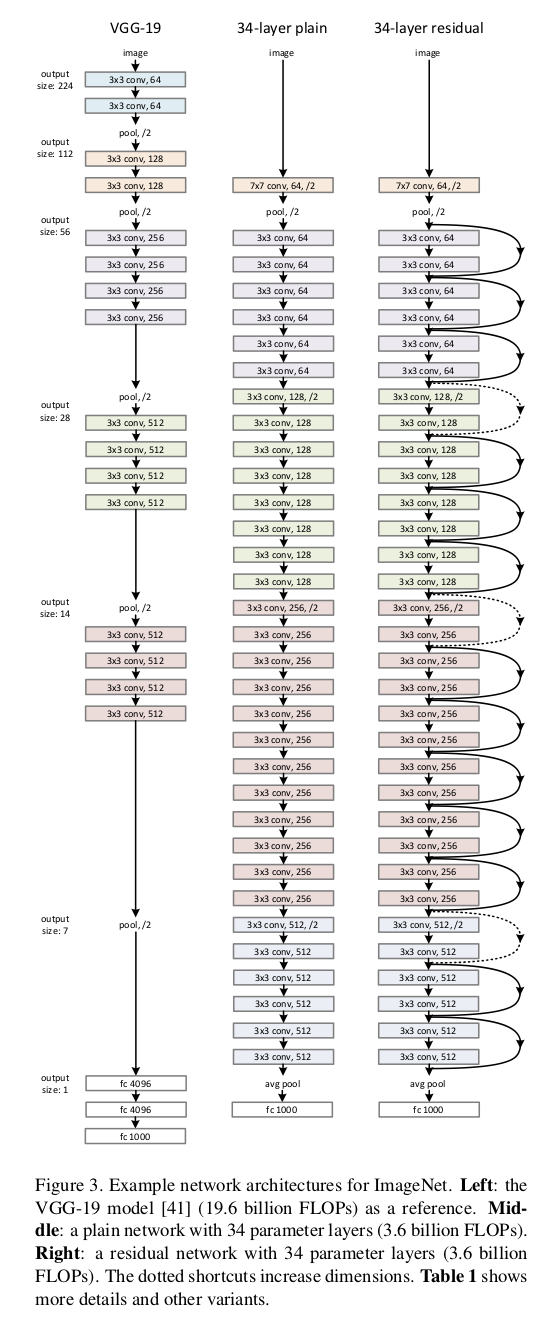}
\caption[ResNet-34 and VGG-19 architectures.]{ResNet-34 and VGG-19 architecture, reproduced from \cite{Resnet}.}
\label{fig:imagenet_arch}
\end{figure}

Deep Residual Networks (ResNets) were introduced with the observation that deeper neural networks are more difficult to train\cite{Resnet}. The ResNet authors point out that with increasing network depth, accuracy gets saturated and adding more layers to a suitably deep model leads to higher training error (see their example in figure \ref{fig:deeper_nets_error}).\\
\indent Their solution to the degradation problem is to explicitly let the stacked layers fit a residual mapping. They recast the original mapping into $\mathcal{F}(\vec{x}) + \vec{x}$ (figure \ref{fig:residual_net_block}), with the hypothesis that it is easier to optimize the residual mapping compared to the original, unreferenced mapping. They describe that their formulation of $\mathcal{F}(\vec{x}) + \vec{x}$ can be realized by ``shortcut connections,'' which skip one or more layers (in their case the shortcut connectios simply perform the identity mapping, and their outputs are added to the outputs of the stacked layers).\\
\indent The authors report that: 1) their extremely deep residual nets are easy to optimize, but the counterpart ``plain'' nets (that simply stack layers) exhibit higher training error when the depth increases (see figure \ref{fig:Resnet_results}); and 2) their deep residual nets can easily enjoy accuracy gains from greatly increased depth, producing results substantially better than previous networks. The authors state that ``this strong evidence [of excellent generalization performance in image classification, detection and localization tasks] shows that the residual learning principle is generic, and we expect that it is applicable in other vision and non-vision problems.'' This is a motivating rationale to evaluate ResNet for transfer learning in parts \ref{part:two} and \ref{part:three}. The architecture of a ResNet variant (ResNet-34) is shown in figure \ref{fig:imagenet_arch}.

\subsection{DenseNet}
\begin{figure}[htb!]
\centering\includegraphics[width=0.8\linewidth]{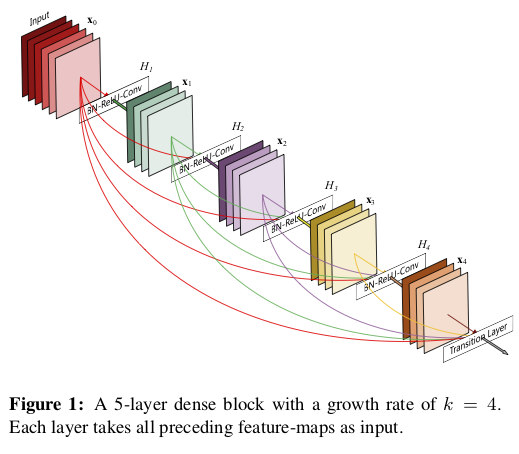}
\caption[Schematic of DenseNet.]{Schematic of DenseNet, reproduced from \cite{Densenet}.}
\label{fig:densenet}
\end{figure}

Densely Connected Convolutional Networks (DenseNets)\cite{Densenet} were recently introduced as a technique to address the ``vanishing gradient'' issue. Specifically, the DenseNet authors point out that with increasingly deep CNNs as information about the input or gradient passes through many layers, it can vanish and ``wash out'' by the time it reaches the end (or beginning) of the network. Various techniques, such as those introduced with ResNets, attempt to create short paths from early layers to later layers to deal with this issue. The designers of DenseNet created an architecture to ensure maximum information flow between layers in the network, by connecting all layers (with matching feature-map sizes) directly with each other. A schematic of their model is shown in figure \ref{fig:densenet}. In contrast to ResNet, DenseNet never combines features through summation before they are passed into a layer, rather it combines features by concatenating them. The ``denseness'' occurs because this network introduces $\frac{L(L+1)}{2}$ connections in an $L$-layer network, instead of just $L$ in traditional architectures.\\
\indent The authors point out that a possibly counter-intuitive effect of this dense connectivity pattern is that it requires fewer parameters than traditional convolutional networks, as there is no need to relearn redundant feature-maps. Moreover, they argue that one big advantage of DenseNets is their improved flow of information and gradients throughout the network, which makes them easy to train. Each layer has direct access to the gradients from the loss function and the original input signal, leading to an implicit deep supervision. This helps training of deeper network architectures. Further, they observe that dense connections have a regularizing effect, which reduces overfitting on tasks with smaller training set sizes. This potential advantage is an interesting area to investigate particularly for the relatively small traning dataset used in part \ref{part:two} of this report. In table \ref{tab:networks}, we observe that the DenseNet variants performed comparable to the ResNet variants on ImageNet.

\subsection{Inception}
\begin{figure}[htb!]
\centering\includegraphics[width=0.8\linewidth]{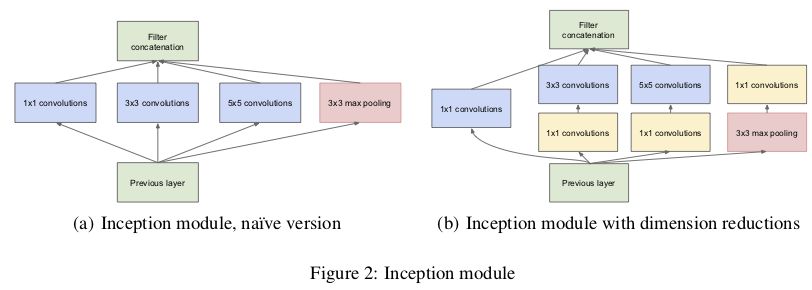}
\caption[Schematic of an Inception module.]{Schematic of an Inception module, reproduced from \cite{inception}.}
\label{fig:inception_module}
\end{figure}

The motivation for the Inception network was to perform well even under strict constraints on memory and computational budget\cite{inception}. Szegedy \etal stress that although VGG has the compelling feature of architectural simplicity, this comes at a high cost: evaluating the network requires a lot of computation. They describe that the ``main idea of the Inception architecture is based on finding out how an optimal local sparse structure in a convolutional vision network can be approximated and covered by readily available dense components.''\\
\indent One particular Inception implementation, GoogLeNet, devised a module called inception module that approximates a sparse CNN with a normal dense construction (figure \ref{fig:inception_module}). The rationale for this being that the most of the activations in a deep network are either unnecessary or redundant because of correlations between them. The Inception network keeps the width/number of the convolutional filters of a particular kernel size small, and it uses convolutions of different sizes to capture details at varied scales.

\subsection{VGG}
Simonyan and Zisserman from the Visual Geometry Group (VGG), Department of Engineering Science, University of Oxford, published the results of their CNN in 2015\cite{vgg}. They investigated the effect of the convolutional network depth on its accuracy in the large-scale image recognition setting. They noted that their ``main contribution is a thorough evaluation of networks of increasing depth using an architecture with very small (3x3) convolution filters, which shows that a significant improvement on the prior-art configurations can be achieved by pushing the depth to 16–19 weight layers.'' Their team secured the first and the second places in the localization and classification tracks respectively [ImageNet Challenge 2014]. The authors also note that their representations generalize well to other datasets, where they achieve state-of-the-art results, thus the consideration of VGG for transfer learning in this report. The architecture of the VGG-19 variant is shown in figure \ref{fig:imagenet_arch}.

\part{Transfer Learning for Classification of Diabetic Retinopathy by Digital Fundus Photography\label{part:two}}%
\chapter{Background: Diabetic Retinopathy\label{chap:background_DR}}

\section{Diabetic Retinopathy}

\subsection{Prevalence}
An estimated 25.6 million Americans aged 20 years or older have either been diagnosed or remain undiagnosed with diabetes mellitus (11\% of people in this age group) \cite{DR-fact-sheet}, and about one-third are not aware that they have the disease \cite{DM-prevalence}.\footnote{Portions of this section courtesy of Diabetic Retinopathy Preferred Practice Pattern (2017)\cite{DR-PPP}.} According to estimates based from the United States Census Bureau data, approximately one-third of Americans are at risk of developing diabetes mellitus during their lifetime \cite{DR-lifetime-risk}.\\
\indent Diabetic retinopathy (DR) is a leading cause of new cases of legal blindness among working-age Americans and represents a leading cause of blindness in this age group worldwide\cite{DR-epi}. The prevalence rate for retinopathy for all adults with diabetes aged 40 and older in the United States is 28.5\% (4.2 million people); worldwide, the prevalence rate has been estimated at 34.6\% (93 million people). An estimate of the prevalence rate for vision-threatening DR (VTDR) in the United States is 4.4\% (0.7 million people). Worldwide, this prevalence rate has been estimated at 10.2\% (28 million people)\cite{DM-prevalence2}-\cite{DM-prevalence3}. Assuming a similar prevalence of diabetes mellitus, the projected prevalence of individuals with any DR in the United States by the year 2020 is 6 million persons, and 1.34 million persons will have VTDR.\\

\subsection{Stages}
\begin{figure}[htb]
\begin{subfigure}{0.6\textwidth}
\captionsetup{width=0.6\textwidth}
\centering\includegraphics[width=0.6\linewidth]{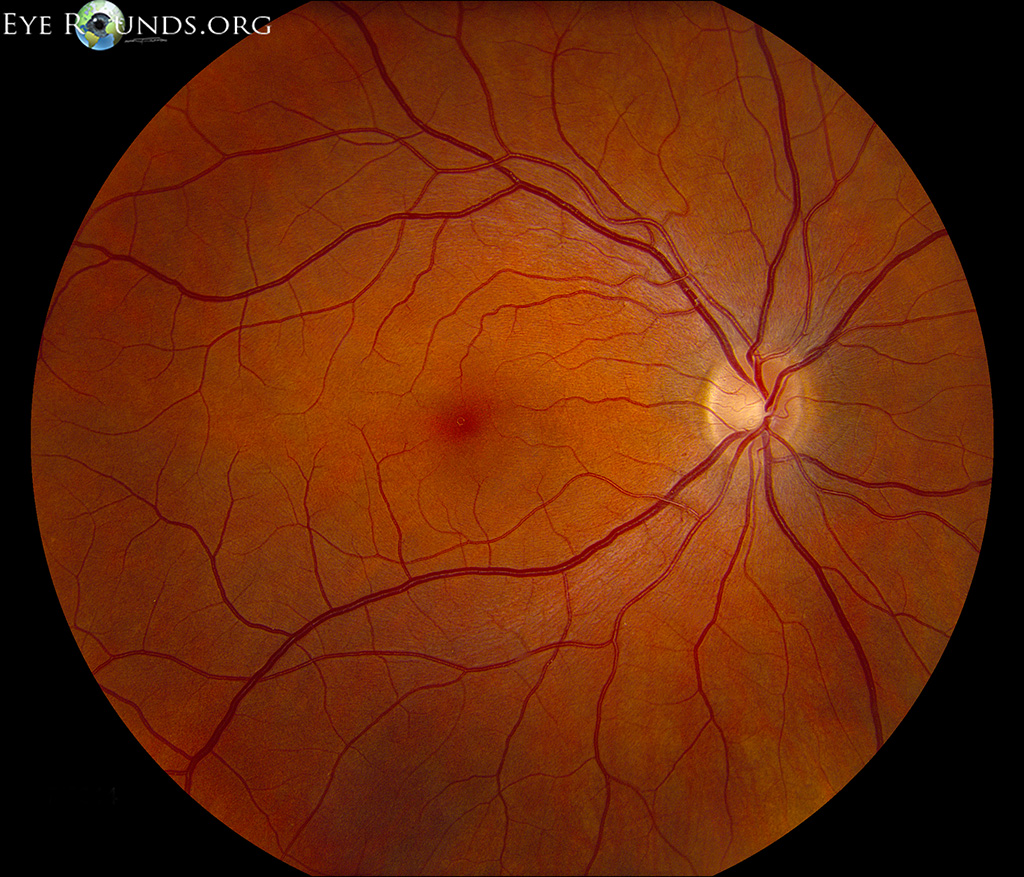}
  \caption{Normal fundus, no DR.}
  \label{fig:normal}
\end{subfigure}
\begin{subfigure}{0.6\textwidth}
\captionsetup{width=0.6\textwidth}
  \centering\includegraphics[width=0.6\linewidth]{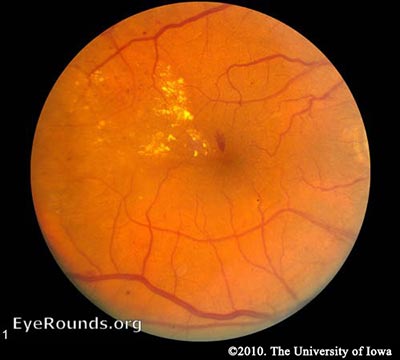}
 \caption{Mild non-proliferative DR. Shown are exudates and microaneurysm.}
  \label{fig:mild-npdr}
\end{subfigure}
\begin{subfigure}{0.6\textwidth}
\captionsetup{width=0.6\textwidth}
  \centering\includegraphics[width=0.6\linewidth]{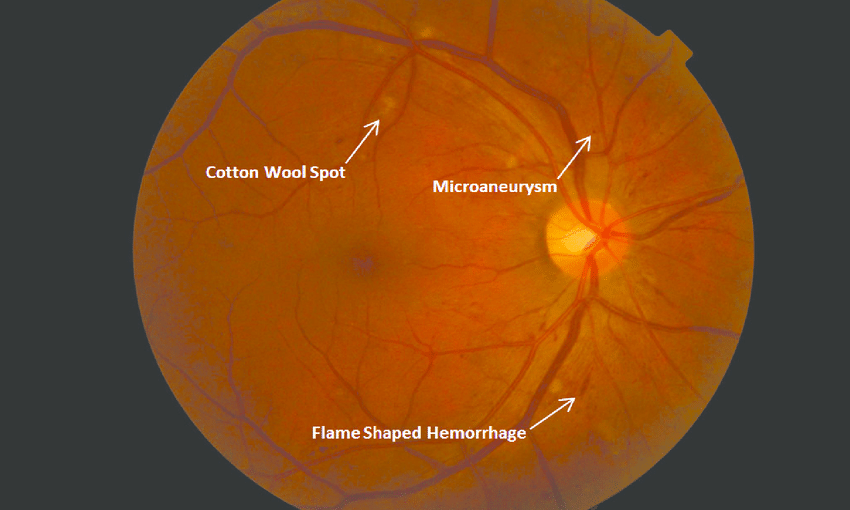}
  \caption{Moderate non-proliferative DR.}
  \label{fig:moderate-NPDR}
\end{subfigure}
\begin{subfigure}{0.6\textwidth}
\captionsetup{width=0.6\textwidth}
  \centering\includegraphics[width=0.6\linewidth]{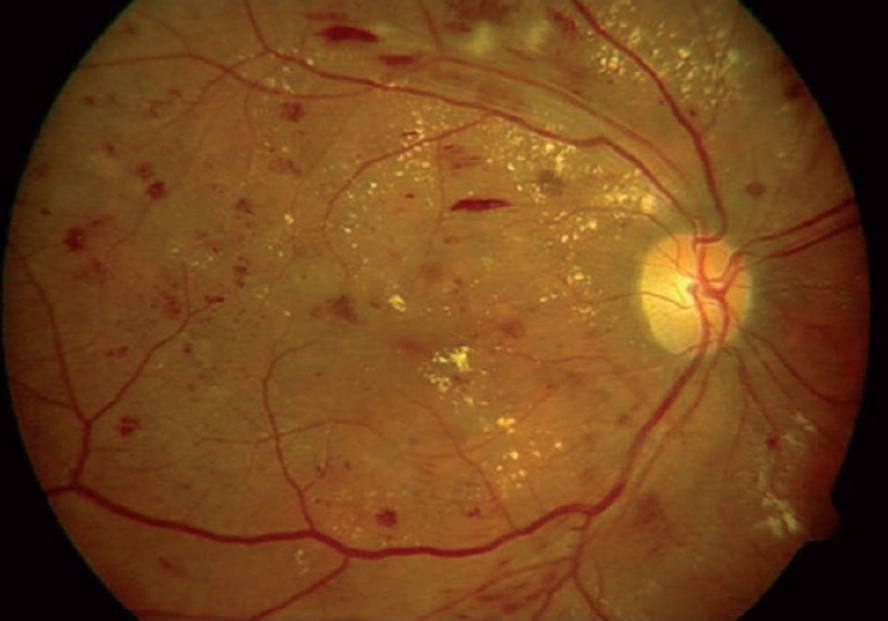}
  \caption{Severe non-proliferative DR. Shown are intra-retinal hemorrhages in 4 quadrants and exudates.}
  \label{fig:severe-NPDR}
\end{subfigure}
\begin{subfigure}{0.6\textwidth}
\captionsetup{width=0.6\textwidth}
  \centering\includegraphics[width=0.6\linewidth]{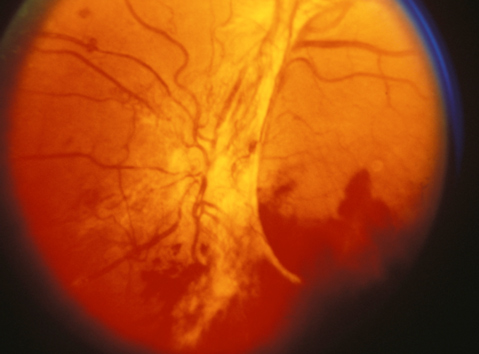}
  \caption{Proliferative DR. Shown are neovascularization of the optic nerve and pre-retinal hemorrhage.}
  \label{fig:PDR}
\end{subfigure}
\begin{subfigure}{0.6\textwidth}
\captionsetup{width=0.6\textwidth}
  \centering\includegraphics[width=0.6\linewidth]{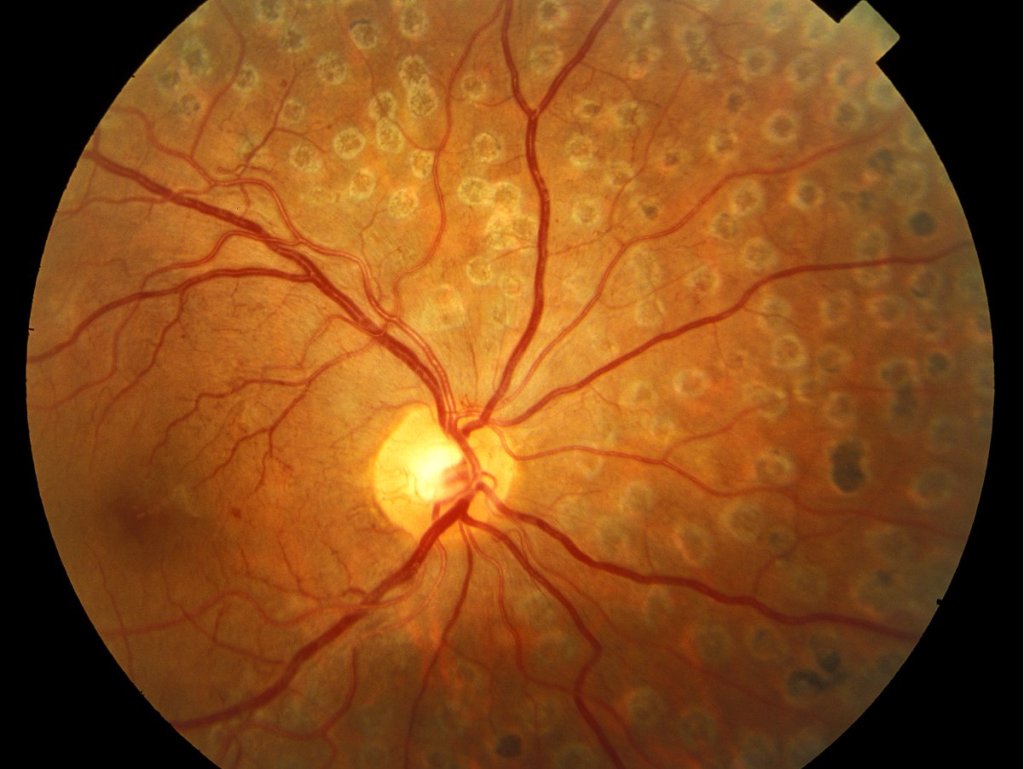}
  \caption{Proliferative DR treated with laser. Pigmented laser scars are seen.}
  \label{fig:PDR-PRP}
\end{subfigure}
\caption[Representative images of Diabetic Retinopathy.]{Example images showing normal (\ref{fig:normal}) and worsening stages of DR (\ref{fig:mild-npdr}-\ref{fig:PDR-PRP}).}
\label{fig:DR_examples}
\end{figure}

\indent DR progresses in an orderly fashion from mild to severe stages when there is not appropriate intervention. It is important to recognize the stages when referal for treatment may be most beneficial. DR is classified into two catagories: non-proliferative diabetic retinopathy (NPDR) and proliferative diabetic retinopathy (PDR).\\
\indent The nonproliferative stages of DR are characterized by retinal vascular related abnormalities, such as microaneurysms, intraretinal hemorrhages, venous dilation, and cotton-wool spots. NPDR is further divided into mild, moderate, and severe stages based on progressively worsening clinical features (see figure \ref{fig:DR_examples}). Managment of NPDR involves close monitoring by an ophthalmologist or optometrist and optimization of the patient's glycemic control by an internist.

\indent The proliferative stages of DR are characterized by abnormal pre-retinal neovascular proliferation, such as on the optic nerve or elsewhere on the retina, which may ultimately lead to vitreous hemorrhage, tractional retinal detachment and irreversible vision loss. Treatment at the proliferative stages involves laser photocoagulation, intravitreal injections of anti-vascular endothelial growth factor (VEGF) agents, or surgery to repair a retinal detachment. Examples of PDR, with and without laser treatment, are shown in figure \ref{fig:DR_examples}.\\

\subsection{Prior work on screening}
\indent Screening for DR is based upon a dilated retinal examination by a trained ophthalmologist or optometrist, or by review of digital retinal images, which may enable early detection of DR along with an appropriate referral \cite{DR-screening1}-\cite{DR-screening9}.\\
\indent A systematic review and meta-analysis of diagnostic accuracy for detection of any level of DR using digital retinal imaging reviewed by certified readers (one field with non-mydriatic imaging) showed 79\% (C.I.\footnote{Confidence Interval (C.I.)} 74-83\%) sensitivity and 96\% (C.I. 95-98\%) specificity.\footnote{See appendix \ref{chap:append_stats} for an overview of sensitivity and specificity.} The authors conclude that screening generates a satisfactory level of sensitivity and specificity, indicating the feasibility of manual DR screening with digital imaging.\\
\indent However, given the increasing prevalence of DR coupled with the fixed number of trained eye care professionals or certified readers, current DR screening programs that rely on expensive labor-intensive manual assessment may fail to address the rising demand for screening, especially in rural or non-developed nations lacking access to qualified screeners. To address this issue, screening with two alternate modalities, crowdsourcing and automated retinal image analysis (ARIA) have been explored.\\

\subsubsection{Crowdsourcing}
One study investigating crowdsourcing with Amazon Mechanical Turk (AMT) found accuracy of 81.3\% for distinguishing between normal and abnormal images \cite{Brady_2014}, while another study investigating AMT found accuracy $\geq$ 90\% for distinguishing between normal vs. severely abnormal images\cite{Mitry_2013}. Both studies reported $>$ 93\% sensitivity, indicating that this screening modality has good potential to address the increasing volume of images that will need to be screened with increasing diabetes prevelance rates. However, as one of the study authors points out, ``crowdsourcing retinal image analysis does have inherent limitations. Almost by definition, control over who is performing the image analysis has been ceded to the abstract entity of 'the crowd,' and efforts to exert choose or credential users of an image grading platform run counter to the spirit of crowdsourcing. Operationally, this means that any crowdsourcing implementation requires meticulous quality control to ensure the results seen in pilot-testing are maintained''\cite{Brady_2017}.\\
\subsubsection{Automated retinal image analysis (ARIA)}
ARIA can be sub-classified based on techniques that rely on traditional computer vision techniques and current state of the art deep learning techniques. A review by Trucco \etal\cite{Trucco} found that 23 studies of various traditional ARIA techniques achieved sensitivity ranging from 45-100\% and specificity ranging from 67.4-99.31\%. These systems were limited by relatively small datasets of images (range 6 to 16,670 images) that were mostly non-public. Moreover, the proprietary nature of these systems makes their general application to other datasets challenging.\\
\subsubsection{Deep Learning efforts} 
\begin{figure}[htb]
\centering\includegraphics[width=0.4\linewidth]{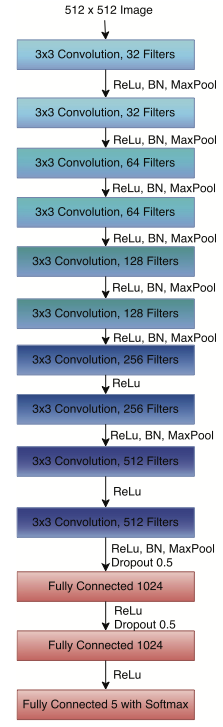}
\caption[CNN architecture of Pratt \etal.]{CNN architecture of Pratt \etal. \cite{Pratt}.}
\label{fig:Pratt_model}
\end{figure}

Gulshan \etal\cite{Gulshan} published a landmark study that used the Inception-v3 neural network trained on 128,175 retinal images and validated on two independent datasets consisting of 9,963 images (EyePACS-1) and 1,748 images (Messidor-2). They found that using the first operating cut point with high specificity, approximating the specificity of ophthalmologists in the development set, on EyePACS-1, the algorithm’s sensitivity was 90.3\% and specificity was 98.1\%. In Messidor-2, the sensitivity was 87.0\% and specificity was 98.5\%.  A second operating point for the algorithm was evaluated, which had a high sensitivity on the development set, reflecting an output that would be used for a screening tool. Using this operating point, on EyePACS-1, the algorithm had a sensitivity of 97.5\% (95\% CI, 95.8\%-98.7\%) and a specificity of 93.4\% (95\% CI, 92.8\%-94.0\%). In Messidor-2, the sensitivity was 96.1\% (95\% CI, 92.4\%-98.3\%) and the specificity was 93.9\% (95\% CI, 92.4\%-95.3\%). The authors conclude that their evaluation of retinal fundus photographs from adults with diabetes, an algorithm based on deep machine learning had high sensitivity and specificity for detecting referable diabetic retinopathy.\\
\indent Gulshan \etal applied transfer learning using the Inception-v3 network, and described their methodology as ``preinitialization using weights from the same network trained to classify objects in the ImageNet data set were used.''  However, it is unclear whether they used the Inception-v3 network as a fixed feature extractor (\ie ``trained the top-layer'') or fine-tuned the network with their training dataset.\\
\indent Moreover, it should be noted that they used a large number (54) of US-licensed ophthalmologists or ophthalmology trainees in their last year of residency, all paid for their work, to grade the training set. Further, US board-certified ophthalmologists with the highest rate of self-consistency were invited to grade the clinical validation sets. Taken together, this methodology for grading the training and validation sets likely provided a highly accurate set of labels, but may not reflect real-world screening conditions where images may be graded by non-ophthalmologists.\\
\indent Pratt \etal\cite{Pratt} applied a custom convolutional neural network (CNN) to classify stages of diabetic retinopathy from the Kaggle Diabetic Retinopathy Detection competition.\footnote{see section \ref{chap:DR_methods} for details of the Kaggle Diabetic Retinopathy Detection competition dataset.} Their CNN architure is reproduced in figure \ref{fig:Pratt_model}. They defined specificity as the number of patients correctly identified as not having DR out of the true total amount not having DR and sensitivity as the number of patients correctly identified as having DR out of the true total amount with DR. Their definition of accuracy was the amount of patients with a correct classification. They reported that their final trained network achieved 95\% specificity, 75\% accuracy and 30\% sensitivity.\\
\indent The high specificity reported in their study indicates that their CNN was able to accurately detect normal healthy eyes. The authors point out (and as indicated in section \ref{chap:DR_methods}) that the majority of images from this Kaggle dataset was made up of normal cases, and thus the CNN may suffer from over-fitting. The authors took steps to address this issue by implementing real-time class weights into their CNN. However, their low sensitivity result indicates issues in detecting truely diseased cases from normals (\ie high false negatives).\\
\indent The authors point out that ``the low sensitivity, mainly from the mild and moderate classes suggests the network struggled to learn deep enough features to detect some of the more intricate aspects of DR. An associated issue identified, which was certified by a clinician, was that by national UK standards around over 10\% of the images in our dataset are deemed ungradable. These images were defined a class on the basis of having at least a certain level of DR. This could have severely hindered our results as the images are misclassified for both training and validation.''\\ 
\indent While the results from Gulshan \etal indicate excellent performance of the Inception-v3 network for classification of diabetic retinopathy, their high sensitivity and specificity results may not reflect real-world performance since their mainly properietary datasets (and public Messidor-2 dataset) were highly validated. Selecting the optimal balance of sensitivity and specificity depends on the purpose for which the test is used. Generally, a screening test should be highly sensitive, whereas a follow-up confirmatory test should be highly specific\cite{Schwartz}.  The results of Pratt \etal suggest that the applicability of deep learning as a screening methodology for real-world images may need further validation.\\
\indent To my knowlege, no prior work has analyzed transfer learning applied to the Kaggle Diabetic Retinopathy Detection competition dataset, which based on the work of Pratt \etal may represent more real-world screening conditions where images are misclassified. Furthermore, it is unclear if transfer learning via the approach of using a pretrained model (e.g. one of the ResNet variants) as a feature extractor (``training the top layer'') vs. fine-tuning the DNN yields higher accuracy. This paper presents the methodology and results of applying transfer learning using several standard DNNs trained on the ImageNet dataset with the Kaggle Diabetic Retinopathy Detection competition dataset, and exploring the accuracy of using these networks as feature extractors vs. fine-tuning them.

\chapter{Methods\label{chap:DR_methods}}

\section{Diabetic Retinopathy Fundus Photographs }

The dataset from the Kaggle Diabetic Retinopathy Detection competition \cite{Kaggle_DR} was used for evlauation of transfer learning for classification of diabetic retinopathy based on digital fundus photography. The task designated by this competition was to create an automated analysis system capable of assigning a score based on a diabetic retinopathy scale (elaborated below).\\
\indent This dataset comprises a large set of high-resolution retina images taken under a variety of imaging conditions. Retinal images for the competition were provided by EyePACS \footnote{Eye Picture Archive Communication System} (\url{http://www.eyepacs.com}), a free platform for retinopathy screening. Briefly, EyePACS allows patients with diabetes to receive retinal evaluations with a digital retinal camera during primary care visits or other eye screening settings. The camera can be operated by a nurse or by other individuals who have been technically trained and certified. The digital images are uploaded to the EyePACS web site where they are interpreted online by trained and certified readers. Recommendations for follow-up and treatment are then made by credentialed doctors, and sent electronically to the patient’s electronic medical record or directly to their primary care physician.\\
\indent This dataset has a total of 88,702 JPEG\footnote{Joint Photographic Experts Group, see \url{https://jpeg.org/}} images. The competition sponsor pre-allocated 31,615 (35.6\%) of these images for training, 3,511 (4.0\%) for validation, and 53,576 (60.4\%) for testing. Class labels were provided by the competition for the training, validation, and test images. All images were rated by a certified reader according to a standard diabetic retinopathy grading scale: 

\begin{itemize}
\item Class 0 - No Diabetic Retinopathy present.
\item Class 1 - Mild Non-proliferative Diabetic Retinopathy present.
\item Class 2 - Moderate Non-proliferative Diabetic Retinopathy present.
\item Class 3 - Severe Non-proliferative Diabetic Retinopathy present.
\item Class 4 - Proliferative Diabetic Retinopathy present.
\end{itemize}

\begin{figure}[htb]
\begin{subfigure}{0.6\textwidth}
\captionsetup{width=0.6\textwidth}
\centering\includegraphics[width=0.6\linewidth]{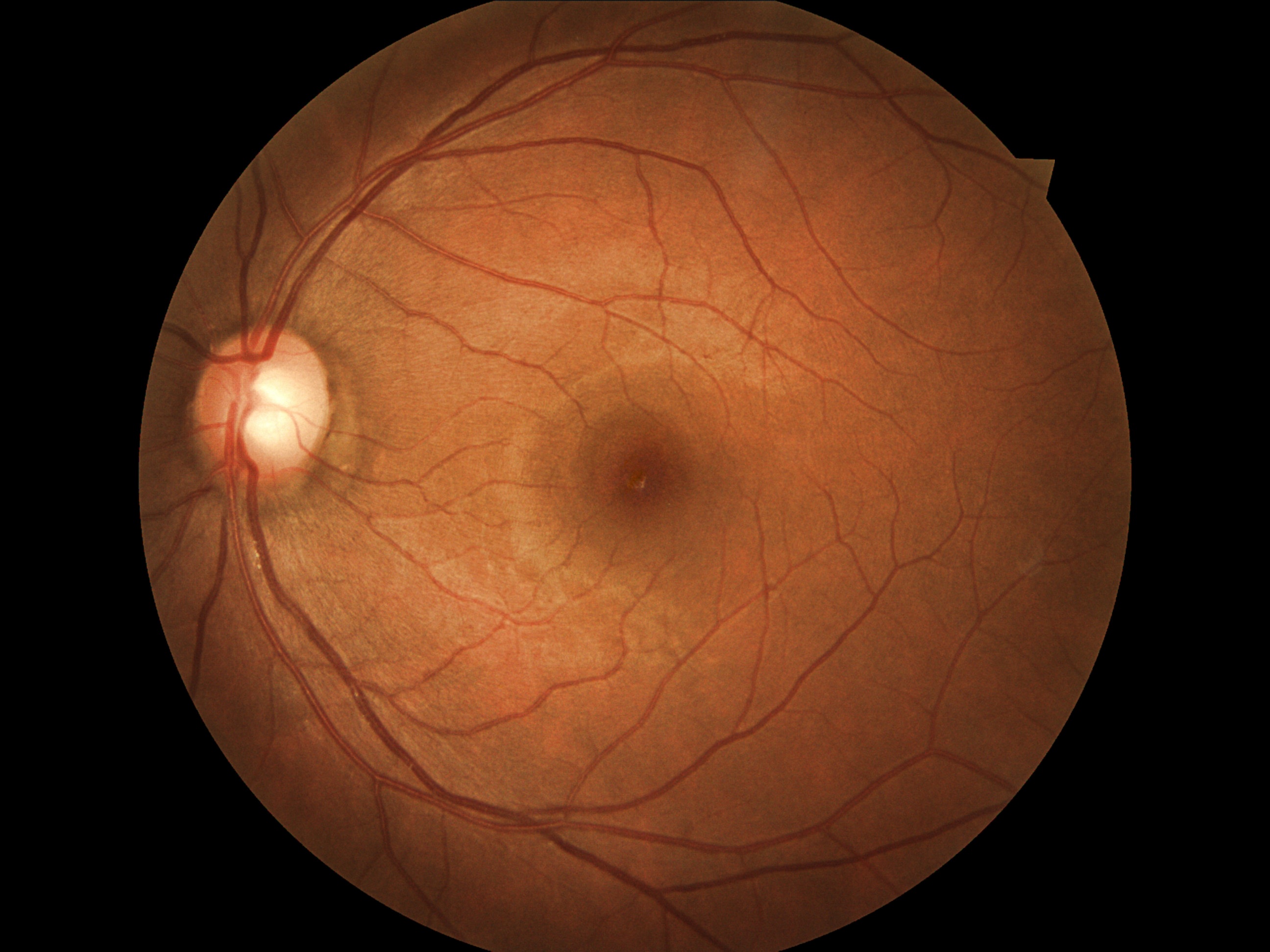}
  \caption{Class 0: no diabetic retinopathy.}
  \label{fig:DR_class0}
\end{subfigure}
\begin{subfigure}{0.6\textwidth}
\captionsetup{width=0.6\textwidth}
  \centering\includegraphics[width=0.6\linewidth]{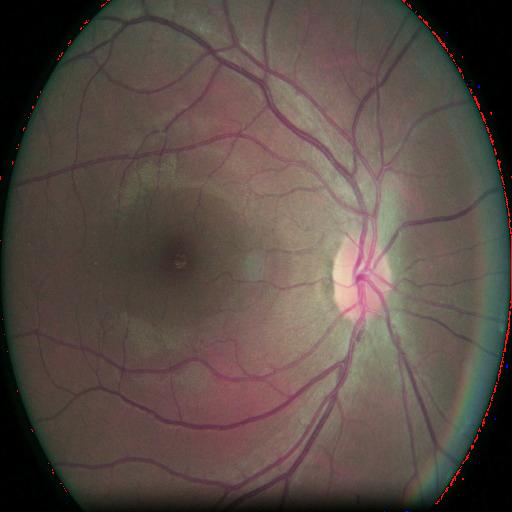}
  \caption{Class 1: mild non-proliferative diabetic retinopathy.}
  \label{fig:DR_class1}
\end{subfigure}
\begin{subfigure}{0.6\textwidth}
\captionsetup{width=0.6\textwidth}
  \centering\includegraphics[width=0.6\linewidth]{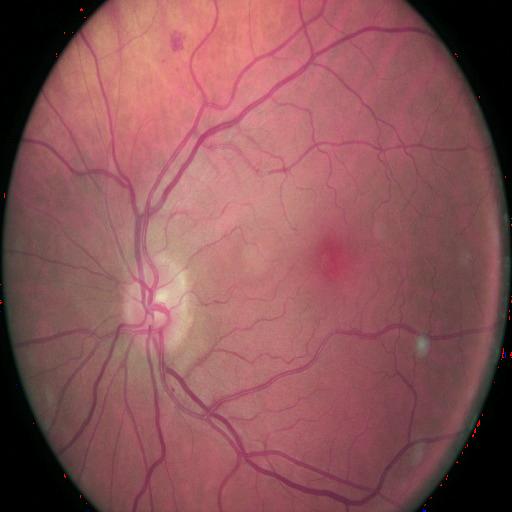}
  \caption{Class 2: moderate non-proliferative diabetic retinopathy.}
  \label{fig:DR_class2}
\end{subfigure}
\begin{subfigure}{0.6\textwidth}
\captionsetup{width=0.6\textwidth}
  \centering\includegraphics[width=0.6\linewidth]{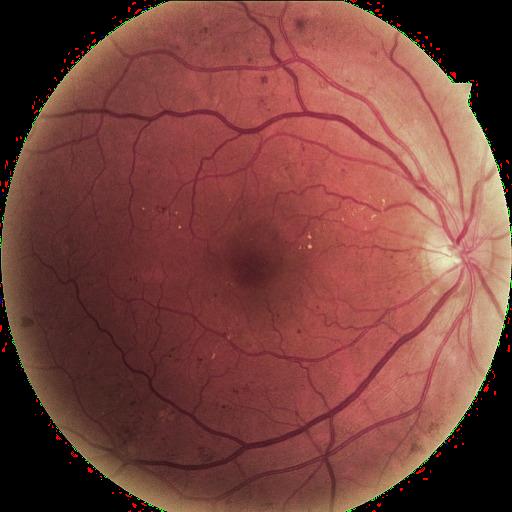}
  \caption{Class 3: severe non-proliferative diabetic retinopathy.}
  \label{fig:DR_class3}
\end{subfigure}
\begin{subfigure}{0.6\textwidth}
\captionsetup{width=0.6\textwidth}
  \centering\includegraphics[width=0.6\linewidth]{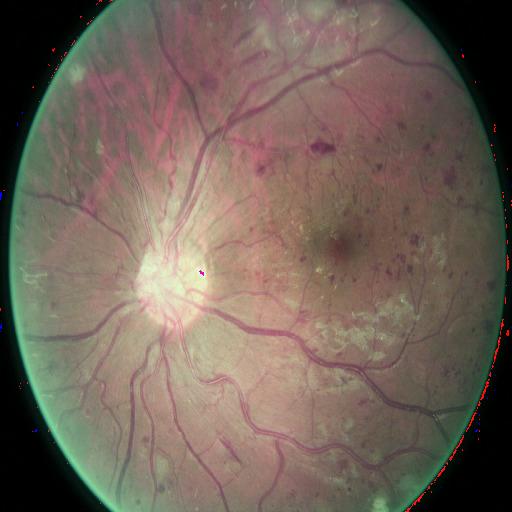}
  \caption{Class 4: proliferative diabetic retinopathy.}
  \label{fig:DR_class4}
\end{subfigure}
\begin{subfigure}{0.6\textwidth}
\captionsetup{width=0.6\textwidth}
  \centering\includegraphics[width=0.6\linewidth]{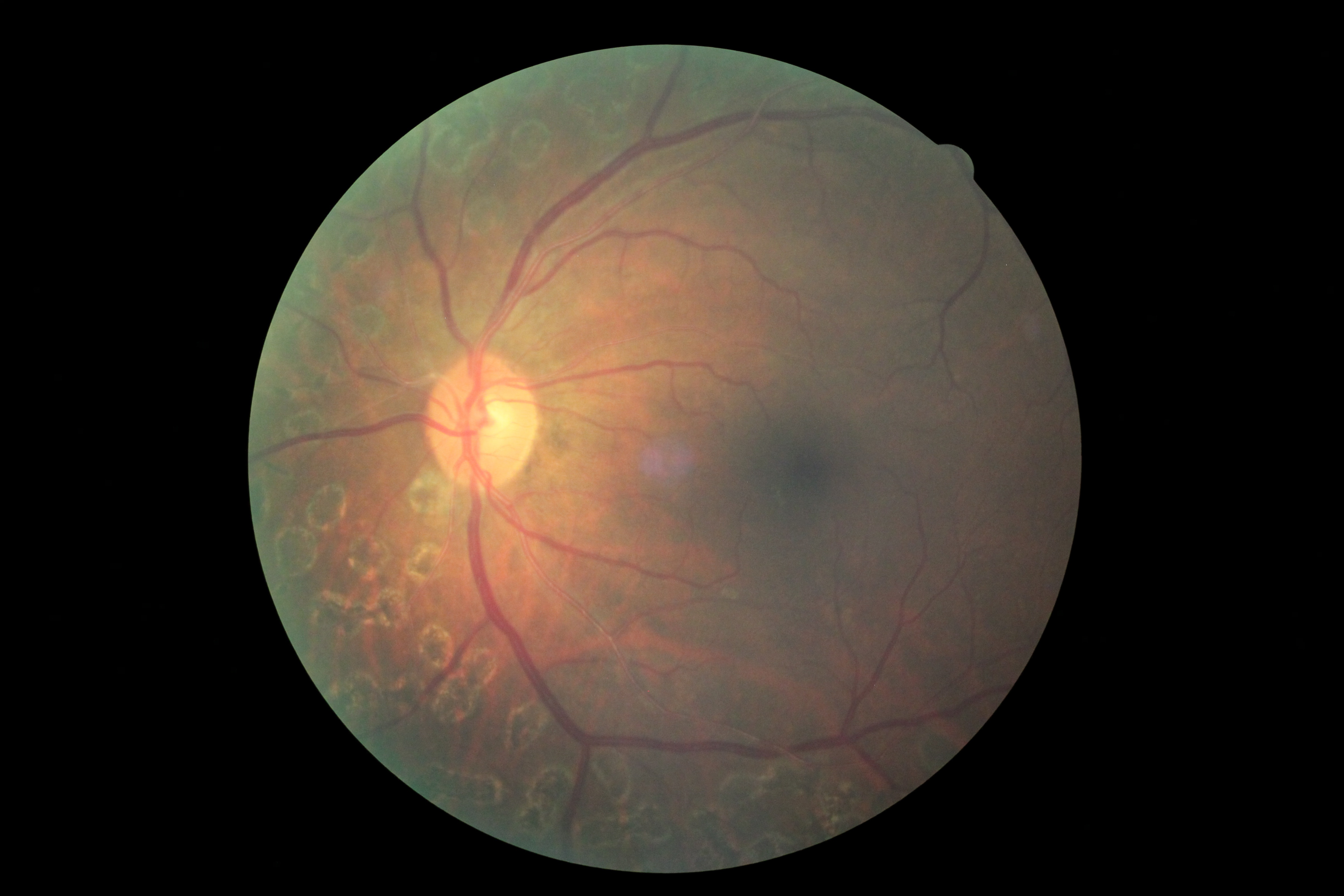}
  \caption{Class 4: proliferative diabetic retinopathy treated with laser.}
  \label{fig:DR_class4_2}
\end{subfigure}
\caption[Sample images from Kaggle Diabetic Retinopathy Detection competition.]{Sample images from Kaggle Diabetic Retinopathy Detection competition.}
\label{fig:DR_class_examples}
\end{figure}

Examples of images from each class are show in figure \ref{fig:DR_class_examples}. The distribution of labels for the training, validation, and test datasets is listed in table \ref{tab:DR_dist}. Note that the relative distribution across classes for the training, validation, and test datasets were kept nearly identical.\\
\indent Rather than accuracy, the quadratic weighted kappa statistic\footnote{See appendix \ref{chap:append_stats} for an overview of quadratic weighted kappa.} was used as the benchmark for evaluation of submissions for the Kaggle Diabetic Retinopathy Detection competition. Briefly, the quadratic weighted kappa is a chance-adjusted index of agreement. In machine learning it can be used to quantify the amount of agreement between an algorithm's predictions and some trusted labels of the same objects. A generally agreed upon scale is: $<$0.20(Poor), 0.21-0.40(Fair), 0.41-0.60(Moderate), 0.61-0.80(Good), and 0.81-1.00(Very good).\\
\indent The top two submissions achieved quadratic weighted kappa scores of 0.84957 and 0.84478, respectively. The system employed by the top score used a pre-processing step to compensate for different lighting conditions, the SparseConvNet\footnote{\url{https://github.com/btgraham/SparseConvNet}}, and Python/Scikit-Learn to train a random forest to combine predictions from the two eyes into a single prediction, and output the final submission\cite{Graham}. The system that scored second place used a custom CNN described in their report\cite{team_o_O}. They utilized a resampling strategy to compensate for the class imbalance of the dataset and a ``blending'' algorithm for both patient eyes to increase performance.

\begin{table}[htb]
\caption[Diabetic retinopathy dataset: distribution of labels across training, validation, and test sets.]{Distribution of labels across training, validation, and test sets for the diabetic retinopathy dataset.}\label{tab:DR_dist}
  \begin{center}
\scalebox{0.8}{
  \begin{tabular}{|l|r|r|r|r|r|r|}
	\hline
   & \textbf{Class 0} & \textbf{Class 1} & \textbf{Class 2} & \textbf{Class 3} & \textbf{Class 4} & \textbf{Total}\\
   \hline
   Training & 23,229(73.5\%) & 2,199(7.0\%) & 4,763(15.0\%) & 786(2.5\%) & 638(2.0\%) & 31,615\\
   Validation & 2,581(73.5\%) & 244(7.0\%) & 529(15.0\%) & 87(2.5\%) & 70(2.0\%) & 3,511\\
   Test & 39,533(73.8\%) & 3,762(7.0\%) & 7,861(14.7\%) & 1,214(2.3\%) & 1206(2.3\%) & 53,576\\
  \hline
\end{tabular}}
  \end{center}
\end{table}

\section{Model selection and runtime configuration\label{sec:models}}

Development and runtime were done on the High Performance Computing (HPC) environment at New York University, specifically the HPC Prince Cluster.\footnote{See \url{https://wikis.nyu.edu/display/NYUHPC/Clusters+-+Prince} for details about this system.} All coding was done with Python 3.6.6. and PyTorch 0.4.1. on a Linux environment.\\
\indent The networks listed in table \ref{tab:networks} were evaluated. This set of 16 networks consists of AlexNet, the 4 variants of VGG, one of the VGG networks with batch normalization, the 5 variants of ResNet, the 4 variants of DenseNet, and the Inception network. A PyTorch transfer learning template\footnote{See \url{https://pytorch.org/tutorials/beginner/transfer_learning_tutorial.html}} was modified to run these pre-built networks.\\
\indent As a ``baseline'' evaluation, each of these 16 networks was evaluated as not pretrained (\ie the pretrained argument at model creation was set to False). The process was then repeated with pretrained networks, \ie the networks were loaded with the weights from their repective training on the ImageNet dataset.\\
\indent Within the pretrained/not pre-trained modes, each network was evaluated twice: once in a configuration ``fine tuning the convnet'' and once as ``fixed feature extractor.'' As described in the PyTorch transfer learning tutorial\cite{pytorch_tutorial}, these configurations are:
\begin{itemize}
\item  Finetuning the convnet: Instead of random initializaion, we initialize the network with a pretrained network, like the one that is trained on imagenet 1000 dataset. Rest of the training looks as usual.
\item Fixed feature extractor: Here, we will freeze the weights for all of the network except that of the final fully connected layer. This last fully connected layer is replaced with a new one with random weights and only this layer is trained. 
\end{itemize}

\indent Rough guidelines for what type of transfer learning to use are discussed in \cite{transfer_learning}. The two main factors are the size of the new dataset (small or big), and its similarity to the original dataset (\eg ImageNet-like in terms of the content of images and the classes, or very different, such as microscope images). Clearly, the retinal images are not similar to the ImageNet dataset, but what value is considered ``big'' or ``small,'' though, is unclear. Hence, the rationale to experiment with both training modes.\\
\indent For the purpose of establishing a ``not pre-trained'' baseline analysis in this report, the finetuning and fixed feature extractor modes were run with not pretrained and pre-trained models, \ie 4 possible combinations:
\begin{enumerate}
\item Not-pretrained network, fine-tuning.
\item Not-pretrained network, fixed feature extractor.
\item Pretrained network, fine-tuning.
\item Pretrained network, fixed feature extractor.
\end{enumerate}

\indent Data augmentation was done using \texttt{torchvision.transforms}, specifically with \texttt{transforms.RandomResizedCrop(224)} and \texttt{transforms.RandomHorizontalFlip()} for the training set, and \texttt{transforms.Resize(256)} and \texttt{transforms.CenterCrop(224)} for the validation set. Training, validation, and test images were normalized using \texttt{transforms.Normalize([0.485, 0.456, 0.406], [0.229, 0.224, 0.225])}, where the first list specifies the mean and the second list specifies the standard deviation of the 3 respective color channels. These values were as recommended\cite{pytorch_tutorial} from the ImageNet dataset. Unless otherwise noted in chapter \ref{chap:DR_results}, 
\begin{itemize}
\item The networks were trained and validated over 10 epochs.
\item Stochastic Gradient Descent (SGD) was used as the optimizer, with initial learning rate of 0.001 and momentum of 0.9.
\item The \texttt{StepLR} learning rate scheduler was used, such that the learning rate decayed by a factor $\gamma=0.1$ every 7 epochs.
\item Cross Entropy Loss was used to train the models with equal weighting across the classes, unless noted otherwise in chapter \ref{chap:DR_results}.
\end{itemize}

\begin{figure}[htb]
\centering\includegraphics[width=0.9\linewidth]{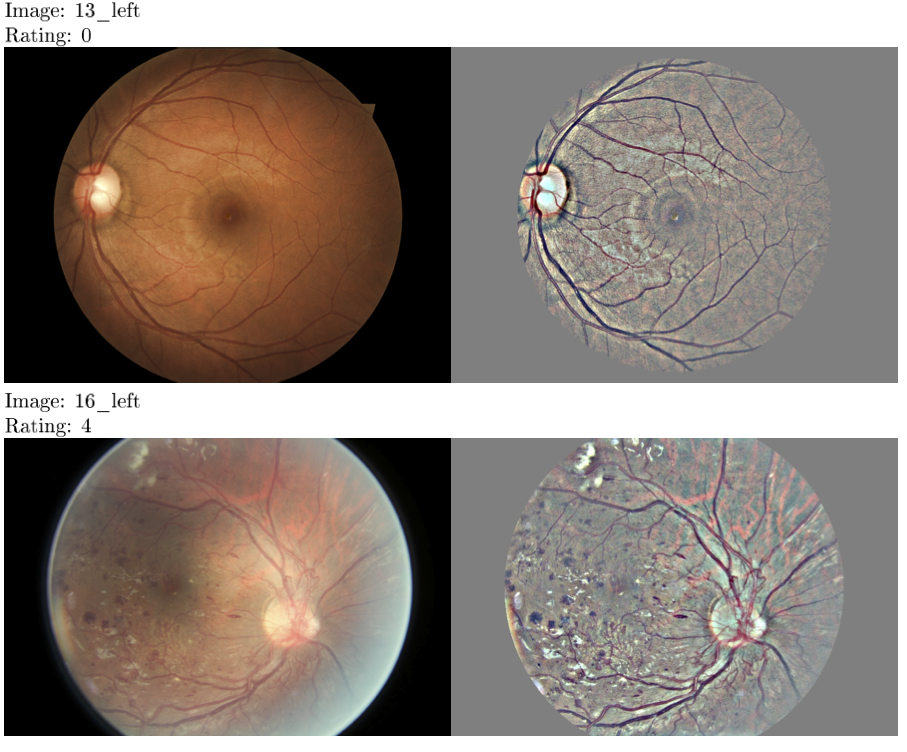}
\caption[Diabetic retinopathy: example pre-processed images]{Diabetic retinopathy: example pre-processed images. Two images from the training set. Original images on the left and preprocessed images on the right. Reproduced from Graham\cite{Graham}.}
\label{fig:preprocessed_examples}
\end{figure}

\indent The PyTorch code was cuda enabled and was submitted via the slurm workload manager\footnote{See \url{https://slurm.schedmd.com/}} with one GPU. All statistical analysis was done with Python. Specifically, the scipy package was used to compute mean, median, standard deviation, and the Wilcoxon signed-rank or Mann-Whitney U test (see appendix \ref{chap:append_stats}). Accuracy, sensitivity, specifcity, and quadratic weighted kappa (see appendix \ref{chap:append_stats}) were calculated for a selected subset of networks.\footnote{Quadratic weighted kappa was calculated using the routine available at \url{https://github.com/benhamner/Metrics/blob/master/Python/ml_metrics/quadratic_weighted_kappa.py}.}\\
\indent To assess if performance could be further improved, 3 modifications were evaluated for selected subsets of the networks: (1) increase the number of epochs to 50; (2) use an image pre-processing step; and (3) use class-adjusted weightings.\\
\indent A selected subset of networks were run for 50 epochs to assess if the training or validation loss and accuracy values significantly change after 10 epochs. In addition, a selected subset of networks were re-evaluated with an image pre-processing step described by Graham\cite{Graham}. This pre-procssing involved: (1) rescale the images to have the same radius (500 pixels was chosen); (2) subtract the local average color; the local average gets mapped to 50\% gray; and (3) clip the images to 90\% size to remove the boundary effects. Two examples of the image pre-processing are shown in figure \ref{fig:preprocessed_examples}.\\
\indent To assess for underfitting or overfitting of the networks, the training to validation loss ratios were computed for each run and were plotted for comparisons. If the training loss is much lower than validation loss (ratio $<$1) then this may indicate that the network might be overfitting. If, on the other hand, roughly training loss equals validation loss, the network may be underfitting\cite{ratio_source}. Lastly, because of the class imbalance where the majority (about 73\%) of training, validation, and test images are from Class 0 cases, class-adjusted weighting were evaluated (specific weights passed to the Cross Entropy Loss function).

\chapter{Results\label{chap:DR_results}}

There was a total of 64 runs (16 networks x [pre-trained vs. not pre-trained] x [fine-tuning vs. fixed feature extractor]). Table \ref{tab:DR_loss_results} shows the loss results for the 16 networks that were evaluated. The results are grouped by not-pretrained and pre-trained catagories, then by training or validation phases, and lastly by the type of implementation (fine-tuning the network or using it as feature extractor). The results of this table are summarized in the series of boxplots in appendix \ref{chap:append_network_plots}, figure \ref{fig:DR_network_loss_boxplots}.

\begin{table}[htb!]
\caption[Diabetic retinopathy: Network loss results]{\textbf{Diabetic retinopathy: Network loss results.} Shown are the loss values for each network at the last epoch. Values are grouped by whether the newtork was pre-trained or not pre-trained, the phase (training or validation), and whether the network was used in fine-tuning or fixed feature extractor mode. All networks were trained for 10 epochs.}\label{tab:DR_loss_results}
  \begin{center}
\scalebox{0.7}{
  \begin{tabular}{|l|r|r|r|r|r|r|r|r|}
\toprule
& \multicolumn{4}{c}{\underline{\textbf{Not-Pretrained}}} & \multicolumn{4}{c}{\underline{\textbf{Pre-Trained}}}\\
& \multicolumn{2}{c}{\underline{\textbf{Training Loss}}} & \multicolumn{2}{c}{\underline{\textbf{Validation Loss}}} & \multicolumn{2}{c}{\underline{\textbf{Training Loss}}} & \multicolumn{2}{c}{\underline{\textbf{Validation Loss}}}\\
& \textbf{Fine} & \textbf{Feature} & \textbf{Fine} & \textbf{Feature} & \textbf{Fine} & \textbf{Feature} & \textbf{Fine} & \textbf{Feature}\\
\textbf{Network} & \textbf{Tuning} & \textbf{Extractor} & \textbf{Tuning} & \textbf{Extractor} & \textbf{Tuning} & \textbf{Extractor} & \textbf{Tuning} & \textbf{Extractor}\\
\midrule
AlexNet & 0.8548 & 0.8676 & 0.8538 & 0.8662 & 0.7475 & 1.1955 & 0.7431 & 0.9628\\
DenseNet-121 & 0.8597 & 0.8630 & 1.1321 & 0.8885 & 0.6261 & 0.8122 & 0.6333 & 0.7697\\
DenseNet-161 & 0.8607 & 0.8632 & 1.2018 & 0.9268 & 0.6067 & 0.7917 & 0.6162 & 0.7861\\
DenseNet-169 & 0.8607 & 0.8632 & 1.2018 & 0.9268 & 0.6217 & 0.8011 & 0.6461 & 0.7615\\
DenseNet-201 & 0.8616 & 0.8649 & 1.0251 & 0.9524 & 0.6177 & 0.8037 & 0.6839 & 0.7838\\
Inception-v3 & 0.8724 & 0.9367 & 0.8940 & 0.9249 & 0.6368 & 0.8719 & 0.6923 & 0.8061\\
ResNet-18 & 0.8602 & 0.8700 & 0.8644 & 0.8695 & 0.6467 & 0.8267 & 0.6616 & 0.8120\\
ResNet-34 & 0.8645 & 0.8792 & 0.8673 & 0.8705 & 0.6262 & 0.8172 & 0.6422 & 0.8312\\
ResNet-50 & 0.8631 & 0.9259 & 0.8934 & 0.8958 & 0.6361 & 0.7999 & 0.6277 & 0.7966\\
ResNet-101 & 0.8636 & 0.9273 & 1.1319 & 0.8891 & 0.6818 & 0.8055 & 0.6691 & 2.1481\\
ResNet-152 & 0.8675 &  0.9289 & 1.4395 & 0.9680 & 0.6493 & 0.7971 & 1.6231 & 0.8171\\
VGG-11 & 0.6198 & 0.9812 & 0.6193 & 0.8189 & 0.8540 & 0.8709 & 0.8510 & 0.8644\\
VGG-11-BN & 0.8649 & 2.2726 & 0.8635 & 1.0197 & 0.6185 & 0.9871 & 0.6200 & 0.7945\\
VGG-13 & 0.8542 & 0.8781 & 0.8529 & 0.8654 & 0.6119 & 1.0039 & 0.5899 & 0.8230\\
VGG-16 & 0.8530 & 0.8714 & 0.8491 & 0.8639 & 0.6045 & 1.0107 & 0.5943 & 0.8451\\
VGG-16 & 0.8530 & 0.8714 & 0.8491 & 0.8639 & 0.6045 & 1.0107 & 0.5943 & 0.8451\\
VGG-19 & 0.8546 & 0.8746 & 0.8542 & 0.8651 &  0.6055 & 0.9577 & 0.5985 & 0.8153\\
\bottomrule
\end{tabular}}
\end{center}
\end{table}

\indent Similarly, table \ref{tab:DR_acc_results} shows the accuracy results for the networks that were evaluated. The results in that table are also grouped by not-pretrained and pre-trained catagories, then by training or validation phases, and lastly by the type of implementation (fine-tuning the network or using it as feature extractor). The results of this table are summarized in the series of boxplots in appendix \ref{chap:append_network_plots}, figure \ref{fig:DR_network_acc_boxplots}.\\
\indent Detailed loss and accuracy plots for each run are plotted in appendix \ref{chap:append_network_plots}, specifically figures \ref{fig:AlexNet_acc_plt}-\ref{fig:Diabetic_Retinopathy_VGG-19_acc_plt}. The plots show loss or accuracy data over 10 epochs, for training and validation. Each plot corresponds to one of the 64 network/pre-trained or not pre-trained/fine-tuning or fixed feature extractor combinations.

\begin{table}[htb!]
\caption[Diabetic retinopathy: Network accuracy results]{\textbf{Diabetic retinopathy: Network accuracy results.} Shown are the accuracy values for each network at the last epoch for training phase and the best accuracy for validation phase. Values are grouped by whether the newtork was pre-trained or not pre-trained, the phase (training or validation), and whether the network was used in fine-tuning or fixed feature extractor mode. All networks were trained for 10 epochs.}\label{tab:DR_acc_results}
  \begin{center}
\scalebox{0.7}{
  \begin{tabular}{|l|r|r|r|r|r|r|r|r|}
\toprule
& \multicolumn{4}{c}{\underline{\textbf{Not-Pretrained}}} & \multicolumn{4}{c}{\underline{\textbf{Pre-Trained}}}\\
& \multicolumn{2}{c}{\underline{\textbf{Training Accuracy}}} & \multicolumn{2}{c}{\underline{\textbf{Validation Accuracy}}} & \multicolumn{2}{c}{\underline{\textbf{Training Accuracy}}} & \multicolumn{2}{c}{\underline{\textbf{Validation Accuracy}}}\\
& \textbf{Fine} & \textbf{Feature} & \textbf{Fine} & \textbf{Feature} & \textbf{Fine} & \textbf{Feature} & \textbf{Fine} & \textbf{Feature}\\
\textbf{Network} & \textbf{Tuning} & \textbf{Extractor} & \textbf{Tuning} & \textbf{Extractor} & \textbf{Tuning} & \textbf{Extractor} & \textbf{Tuning} & \textbf{Extractor}\\
\midrule
AlexNet & 0.7347 & 0.7347 & 0.7351 & 0.7351 & 0.7520 & 0.6558 & 0.7579 & 0.7285\\
DenseNet-121 & 0.7347 & 0.7347 & 0.7345  & 0.7351 & 0.7921 & 0.7342 & 0.7960 & 0.7414\\
DenseNet-161 & 0.7347 & 0.7346 & 0.7345 & 0.7340 & 0.8003 & 0.7396 & 0.8038 & 0.7422\\
DenseNet-169 & 0.7347 & 0.7346 & 0.7345 & 0.7340 & 0.7952 & 0.7378 & 0.7921 & 0.7428\\
DenseNet-201 & 0.7347 & 0.7347 & 0.7351 & 0.7323 & 0.7957 & 0.7365 & 0.7938 & 0.7411\\
Inception-v3 & 0.7347 & 0.7241 & 0.7351 & 0.7354 & 0.7896 & 0.7299 & 0.7884 & 0.7368\\
ResNet-18 & 0.7347 & 0.7347 & 0.7351 & 0.7351 & 0.7847 & 0.7337 & 0.7952 & 0.7382\\
ResNet-34 & 0.7347 & 0.7346 & 0.7351 & 0.7351 & 0.7929 & 0.7339 & 0.7983 & 0.7410\\
ResNet-50 & 0.7347 & 0.7266 & 0.7351 & 0.7351 & 0.7894 & 0.7390 & 0.7975 & 0.7462\\
ResNet-101 & 0.7347 & 0.7265 & 0.7351 & 0.7351 & 0.7743 & 0.7359 & 0.7841 & 0.7448\\
ResNet-152 & 0.7347 & 0.7247 & 0.7351 & 0.7351 & 0.7851 & 0.7379 & 0.7739 & 0.7479\\
VGG-11 & 0.7961 & 0.6937 & 0.7986 & 0.7391 & 0.7347 & 0.7347 & 0.7351 & 0.7351\\
VGG-11-BN & 0.7347 & 0.5982 & 0.7351 & 0.7351 & 0.7930 & 0.6912 & 0.8003 & 0.7380\\
VGG-13 & 0.7347 & 0.7345 & 0.7351 & 0.7351 & 0.7991 & 0.6877 & 0.8083 & 0.7417\\
VGG-16 & 0.7347 & 0.7347 & 0.7351 & 0.7351 & 0.7997 & 0.6865 & 0.8112 & 0.7365\\
VGG-19 & 0.7347 & 0.7347 & 0.7351 & 0.7351 & 0.7990 & 0.6979 & 0.8106 & 0.7383\\
\bottomrule
\end{tabular}}
\end{center}
\end{table}

Table \ref{tab:DR_loss_stats} summarizes the mean, standard deviation, and median values for loss across the models for the various combinations pre-trained vs. not pretrained, training vs. validation phases, and fine-tuning vs. feature-extractor. Table \ref{tab:DR_acc_stats} provides these statistics for the accuracy.

\begin{table}[htb!]
\caption[Diabetic retinopathy: Network loss statistics]{\textbf{Diabetic retinopathy: Network loss statistics.} Shown are the mean, standard deviation, and median values for the loss data at the last epoch. Results are across the 16 networks evaluated and are grouped by pre-trained vs. not pre-trained networks, training or validation phase, and whether the networks were used as fine-tuning or as fixed feature extractors.}\label{tab:DR_loss_stats}
  \begin{center}
\scalebox{0.7}{
  \begin{tabular}{|l|r|r|r|r|r|r|r|r|}
\toprule
& \multicolumn{4}{c}{\underline{\textbf{Not-Pretrained}}} & \multicolumn{4}{c}{\underline{\textbf{Pre-Trained}}}\\
& \multicolumn{2}{c}{\underline{\textbf{Training}}} & \multicolumn{2}{c}{\underline{\textbf{Validation}}} & \multicolumn{2}{c}{\underline{\textbf{Training}}} & \multicolumn{2}{c}{\underline{\textbf{Validation}}}\\
& \textbf{Fine} & \textbf{Feature} & \textbf{Fine} & \textbf{Feature} & \textbf{Fine} & \textbf{Feature} & \textbf{Fine} & \textbf{Feature}\\
& \textbf{Tuning} & \textbf{Extractor} & \textbf{Tuning} & \textbf{Extractor} & \textbf{Tuning} & \textbf{Extractor} & \textbf{Tuning} & \textbf{Extractor}\\
\midrule
Mean & 0.846 & 0.979 & 0.972 & 0.901 & 0.649 & 0.885 & 0.718 & 0.901\\
Stdv & 0.059 & 0.336 & 0.195 & 0.048 & 0.063 & 0.112 & 0.242 & 0.325\\
Median & 0.861 & 0.876 & 0.880 & 0.889 & 0.626 & 0.822 & 0.644 & 0.814\\
\bottomrule
\end{tabular}}
\end{center}
\end{table}

\begin{table}[htb!]
\caption[Diabetic retinopathy: Network accuracy statistics]{\textbf{Diabetic retinopathy: Network accuracy stastistics}. Shown are the mean, standard deviation, and median values for the accuracy data (best accuracy over the epochs). Results are across the 16 networks evaluated and are grouped by pre-trained vs. not pre-trained networks, training or validation phase, and whether the networks were used as fine-tuning or as fixed feature extractors.}\label{tab:DR_acc_stats}
  \begin{center}
\scalebox{0.7}{
  \begin{tabular}{|l|r|r|r|r|r|r|r|r|}
\toprule
& \multicolumn{4}{c}{\underline{\textbf{Not-Pretrained}}} & \multicolumn{4}{c}{\underline{\textbf{Pre-Trained}}}\\
& \multicolumn{2}{c}{\underline{\textbf{Training}}} & \multicolumn{2}{c}{\underline{\textbf{Validation}}} & \multicolumn{2}{c}{\underline{\textbf{Training}}} & \multicolumn{2}{c}{\underline{\textbf{Validation}}}\\
& \textbf{Fine} & \textbf{Feature} & \textbf{Fine} & \textbf{Feature} & \textbf{Fine} & \textbf{Feature} & \textbf{Fine} & \textbf{Feature}\\
& \textbf{Tuning} & \textbf{Extractor} & \textbf{Tuning} & \textbf{Extractor} & \textbf{Tuning} & \textbf{Extractor} & \textbf{Tuning} & \textbf{Extractor}\\
\midrule
Mean & 0.739 & 0.721 & 0.739 & 0.735 & 0.786 & 0.720 & 0.790 & 0.740\\
Stdv & 0.015 & 0.033 & 0.015 & 0.001 & 0.018 & 0.026 & 0.019 & 0.005\\
Median & 0.735 & 0.735 & 0.735 & 0.735 & 0.792 & 0.734 & 0.796 & 0.741\\
\bottomrule
\end{tabular}}
\end{center}
\end{table}

\indent From tables \ref{tab:DR_loss_results}-\ref{tab:DR_acc_stats} and the boxplots \ref{fig:DR_network_loss_boxplots} and \ref{fig:DR_network_acc_boxplots} in appendix \ref{chap:append_network_plots}, several observations are made:\\
\\
(1) The loss was generally lower in the pre-trained models compared to the not pre-trained models when comparing fixed pairs of fine-tuning or feature extractor and training or validation phases. For instance, the losses in the pre-trained validation phase for fine-tuning were lower than that of not pre-trained validation phase for fine-tuning. These trends were graphically observed in the boxplots in appendix \ref{chap:append_network_plots}, in particular figure \ref{fig:DR_loss_boxplots_a}-\ref{fig:DR_loss_boxplots_d} shows that the pre-trained group has lower loss than the not pre-trained group. This was confirmed in table \ref{tab:DR_loss_acc_comparisons}, which showed that there were statistically significant differences for each of the pairwise pre-trained vs. not pre-trained comparisons, except for the pretrained vs. not-pretrained training phase feature-extractor comparison (corresponds to figure \ref{fig:DR_loss_boxplots_b} [see table \ref{tab:DR_loss_acc_comparisons} test no. 2]). The accuracy results showed the inverse pattern as the loss results, where the accuracy was higher in 3 of 4 cases for pre-trained networks (see table \ref{tab:DR_acc_results}, table \ref{tab:DR_acc_stats}, table \ref{tab:DR_loss_acc_comparisons} [test no. 1-4], and the boxplots in figure \ref{fig:DR_acc_boxplots_a}-\ref{fig:DR_acc_boxplots_d}).\\
\begin{table}[htb!]
\caption[Diabetic retinopathy: Network loss and accuracy comparisons]{\textbf{Diabetic retinopathy: Network loss and accuracy comparisons}. Shown are pair-wise Wilcoxon signed-rank (Test no. 1-8) or Mann-Whitney (Test no. 9-12) comparison tests between two catagories, where 2 of 3 variables (pre-trained vs. not pretrained, training or validation phase, fine-tuning vs. fixed feature extractor) were held fixed and the other varied. p-values are indicated, where p $<$ 0.05 (marked in bold) indicates a statistically significant difference between the two categories.}\label{tab:DR_loss_acc_comparisons}
  \begin{center}
\scalebox{0.6}{
  \begin{tabular}{|l|l|l|r|r|}
\toprule
& \multicolumn{2}{c}{\textbf{Category}} & \multicolumn{2}{c}{\textbf{p-value}}\\
\textbf{Test no.} & \textbf{A} & \textbf{B} & \textbf{Loss} & \textbf{Accuracy}\\
\midrule
1 & Not pretrained network/training phase/fine-tuning & Pretrained network/training phase/fine-tuning & \textbf{0.002} & \textbf{0.004}\\
2 & Not pretrained network/training phase/feature-extractor & Pretrained network/training phase/feature extractor & 0.301 & 0.642\\
3 & Not pretrained network/validation phase/fine-tuning & Pretrained network/validation phase/fine-tuning & \textbf{0.002} & \textbf{0.003}\\
4 & Not pretrained network/validation phase/feature-extractor & Pretrained network/validation phase/feature extractor & \textbf{0.034} & \textbf{0.006}\\
& & & \\
5 & Not pretrained network/training phase/feature extractor & Not pretrained network/training phase/fine-tuning & \textbf{$<$0.001} & \textbf{0.005}\\
6 & Not pretrained network/validation phase/feature extractor & Not pretrained network/validation phase/fine-tuning & 0.569 & 0.248\\
7 & Pretrained network/training phase/feature extractor & Pretrained network/training phase/fine-tuning & \textbf{$<$0.001} & \textbf{0.001}\\
8 & Pretrained network/validation phase/feature extractor & Pretrained network/validation phase/fine-tuning & \textbf{0.006} & \textbf{0.001}\\
& & &\\
9 & Not pretrained/training phase/fine-tuning & Not pretrained/validation phase/fine-tuning & 0.053 & \textbf{0.003}\\
10 & Not pretrained network/training phase/feature extractor & Not pretrained network/validation phase/feature extractor & 0.462 & \textbf{$<$0.001}\\
11 & Pretrained network/training phase/fine-tuning & Pretrained network/vaidation phase/fune-tuning & 0.220 & 0.146\\
12 & Pretrained network/training phase/feature extractor & Pretrained network/validation phase/feature extractor & 0.084 & \textbf{$<$0.001}\\
\bottomrule
\end{tabular}}
\end{center}
\end{table}

\noindent (2) Holding the pre-trained or not pre-trained variable fixed, the loss for fine-tuning was generally lower than that for feature-extractor, in both the training and validation phases (see boxplots \ref{fig:DR_loss_boxplots_e}-\ref{fig:DR_loss_boxplots_h}). These differenes were statistically significant except for the not pre-trained fine-tuning vs. feature extractor comparison in the validation phase (see figure \ref{fig:DR_loss_boxplots_f} and table \ref{tab:DR_loss_acc_comparisons} test no. 6). The accuracy results showed the inverse pattern as the loss results, where the accuracy was higher in 3 of 4 cases for fine-tuning (see table \ref{tab:DR_acc_results}, table \ref{tab:DR_acc_stats}, table \ref{tab:DR_loss_acc_comparisons} [test no. 5-8], and the boxplots in figure \ref{fig:DR_acc_boxplots_e}-\ref{fig:DR_acc_boxplots_h}).\\
\\
(3) The last group of comparisons held the pre-training vs. not pre-training variable fixed and the fine-tuning vs. feature extractor variable fixed, while evaluating for a difference between the training and validation phases. As indicated in table \ref{tab:DR_loss_acc_comparisons} there were no statistically differences for pair-wise comparisons between training and validation losses while holding the pre-trained/not pre-trained and fine-tuning/feature extractor variables fixed. Moreover, the values in tables \ref{tab:DR_acc_results}[test no. 9-12] and \ref{tab:DR_acc_stats}, and the boxplots in figure \ref{fig:DR_loss_boxplots_i}-\ref{fig:DR_loss_boxplots_l} indicate that the losses were very similar between the pairs. Table \ref{tab:DR_loss_acc_comparisons} indicates that there was a statistically significant difference in accuracy between the training and validation groups for 3 of the 4 cases, with the exception (see table \ref{tab:DR_loss_acc_comparisons} test no. 11) being when the models were pre-trained and used as fine-tuning (see figure \ref{fig:DR_acc_boxplots_i}-\ref{fig:DR_acc_boxplots_l}).\\
\\
(4) Batch normalization(BN), applied to VGG-11, had mixed results. In the not pre-trained cases, the loss increased and the accuracy decreased compared to VGG withouth BN. In the pre-trained cases, the loss decreased and the accuracy increased in 3 of 4 cases (the exception being training accuracy as fixed feature extractor). In the pretrained, fine-tuning case validation accuracy substantially increased by 6.53\%, and was comparable in accuracy to VGG-19 without BN.\\

\indent To evaluate for under- or over-fitting of the networks, the training to validation loss ratios were tabulated (table \ref{tab:DR_ratio_stats}) and plotted (figure \ref{fig:DR_ratio_boxplots}). The following patterns were observed:
\begin{itemize}
\item AlexNet had a ratio near 1, across the type of training and fine-tuning vs. fixed feature extractor mode.
\item The DenseNet variants had ratios $<$1 when not pretrained, indicating these networks have higher validation loss than training loss and may be overfitting. However, when pretrained their ratios increased to $>$1, implying that overfitting is not likely present.
\item Inception\_v3 showed the same pattern whether it was pretrained or not pretrained: the fine-tuning ratios were $<$1, implying over-fitting, while the fixed feature extractor ratios were $>$1.
\item The ResNet variants showed differing ratios: not pre-trained fine-tuning ratios were all $<$1, while pretrained fine-tuning were $>$1, execept for ResNet-101 and ResNet-152. The ResNet variants showed mixed ratios (table \ref{tab:DR_ratio_stats}) as fixed feature extractors. ResNet-101 suffered a high validation loss when pre-trained as a fixed feature extractor, leading to an outlier ratio of 0.274.
\item The VGG variants had ratios $>$1 across the combinations of pretrained or not pretrained and fine-tuning or fixed feature extractor, implying these networks were not over-fitting.
\item When networks were not-pretrained and in fine-tuning mode, 10/16 (62.5\%) had ratios $<$1.
\item When networks were not-pretrained and in fixed feature extractor mode, 6/16 (37.5\%) had ratios $<$1.
\item When networks were pretrained and in fine-tuning mode, 3/16 (18.75\%) had ratios $<$1.
\item When networks were pretrained and in fixed feature extrator mode, 4/16 (25\%) had ratios $<$1.
\end{itemize}

\begin{figure}[htb]
\begin{subfigure}{0.8\textwidth}
\captionsetup{width=0.8\textwidth}
\centering\includegraphics[width=0.8\linewidth]{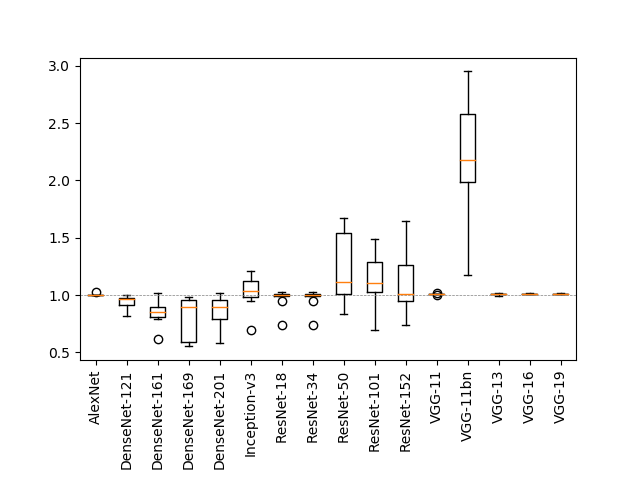}
\caption{Not pre-trained, fixed feature extractor.}
\label{fig:DR_ratio_boxplots_a}
\end{subfigure}
\begin{subfigure}{0.8\textwidth}
\captionsetup{width=0.8\textwidth}
\centering\includegraphics[width=0.8\linewidth]{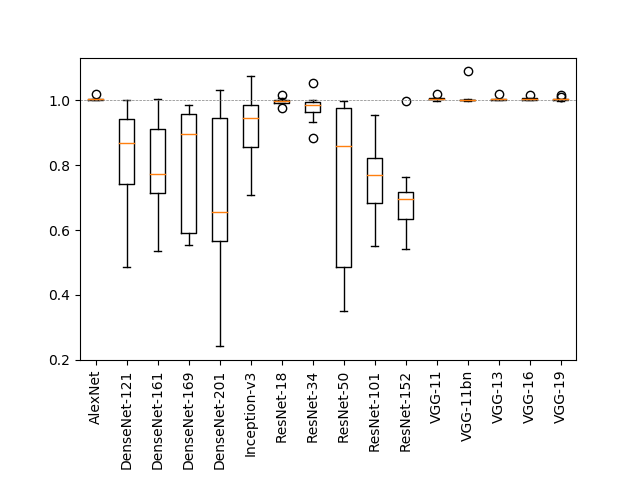}
\caption{Not pre-trained, fine-tuning.}
\label{fig:DR_ratio_boxplots_b}
\end{subfigure}
\label{fig:DR_ratio_boxplots}
\caption[Diabetic Retinopathy: Network training to validation ratio boxplots.]{\textbf{Diabetic Retinopathy: Network training to validation ratio boxplots.} Plotted are the training to validation ratios for each network. A ratio $<$1 implies over-fitting of the network.}\label{fig:DR_ratio_boxplots}
\end{figure}

\begin{figure}
\ContinuedFloat
\begin{subfigure}{0.8\textwidth}
\captionsetup{width=0.8\textwidth}
\centering\includegraphics[width=0.8\linewidth]{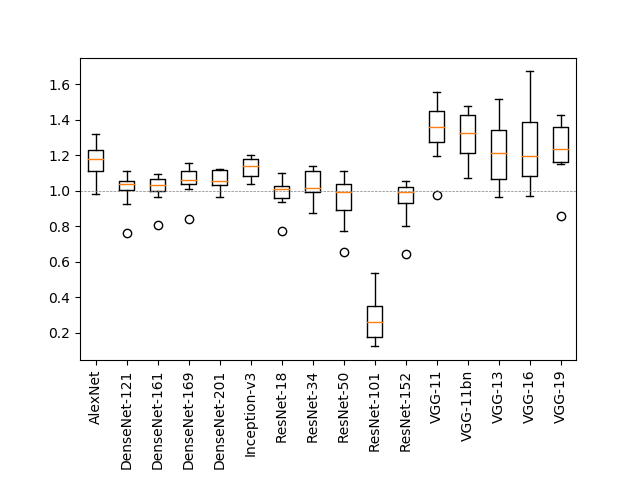}
\caption{Pre-trained, fixed feature extractor.}
\label{fig:DR_ratio_boxplots_c}
\end{subfigure}
\begin{subfigure}{0.8\textwidth}
\captionsetup{width=0.8\textwidth}
\centering\includegraphics[width=0.8\linewidth]{boxplot_not_pretrained_fine_tune.png}
\caption{Pre-trained, fine-tuning.}
\label{fig:DR_ratio_boxplots_d}
\end{subfigure}
\caption[Continued - Diabetic Retinopathy: Network training to validation ratio boxplots.]{\textbf{Continued - Diabetic Retinopathy to Network training to validation ratio boxplots.} A ratio $<$1 implies over-fitting of the network.}
\label{fig:DR_ratio_boxplots2}
\end{figure}

\begin{table}[htb!]
\caption[Diabetic retinopathy: Network training to validation ratio statistics]{\textbf{Diabetic retinopathy: Network training to validation ratio statistics.} Shown are the training to validation ratios for each network, grouped by pre-trained vs. not-pretrained network, training vs. validation phase, and whether the network was used in fine-tuning or fixed feature extrator mode.}\label{tab:DR_ratio_stats}
  \begin{center}
\scalebox{0.7}{
  \begin{tabular}{|l|r|r|r|r|}
\toprule
& \multicolumn{2}{c}{\underline{\textbf{Not-Pretrained}}} & \multicolumn{2}{c}{\underline{\textbf{Pre-Trained}}}\\ 
\textbf{Network} & \textbf{Fine} & \textbf{Feature} & \textbf{Fine} & \textbf{Feature}\\
& \textbf{Tuning} & \textbf{Extractor} & \textbf{Tuning} & \textbf{Extractor}\\
\midrule
AlexNet & 1.004(0.005) & 1.004(0.007) & 1.019(0.011) & 1.168(0.093)\\
DenseNet-121 & 0.823(0.156) & 0.938(0.058) & 1.019(1.019) & 1.009(0.095)\\
DenseNet-161 & 0.789(0.150) & 0.845(0.101) & 1.024(0.060) & 1.013(0.080)\\
DenseNet-169 & 0.802(0.180) & 0.802(0.180) & 1.015(0.065) & 1.055(0.083)\\
DenseNet-201 & 0.691(0.261) & 0.857(0.125) & 1.017(0.059) & 1.066(0.050)\\
Inception-v3 & 0.922(0.102) & 1.032(0.138) & 0.962(0.055) & 1.130(0.054)\\
ResNet-18 & 0.998(0.010) & 0.974(0.082) & 1.016(0.047) & 0.988(0.088)\\
ResNet-34 & 0.976(0.043) & 0.974(0.082) & 1.005(0.062) & 1.034(0.079)\\
ResNet-50 & 0.743(0.260) & 1.244(0.301) & 1.016(0.044) & 0.947(0.134)\\
ResNet-101 & 0.759(0.123) & 1.119(0.255) & 0.912(0.174) & 0.274(0.119)\\
ResNet-152 & 0.701(0.117) & 1.096(0.253) & 0.836(0.216) & 0.943(0.122)\\
VGG-11 & 1.006(0.005) & 1.011(0.005) & 1.025(0.032) & 1.341(0.159)\\
VGG-11bn & 1.010(0.027) & 2.220(0.483) & 1.031(0.023) & 1.305(0.137)\\
VGG-13 & 1.005(0.005) & 1.010(0.007) & 1.058(0.023) & 1.223(0.190)\\
VGG-16 & 1.005(0.005) & 1.010(0.003) & 1.032(0.032) & 1.252(0.218)\\
VGG-19 & 1.005(0.005) & 1.011(0.003) & 1.026(0.030) & 1.230(0.156)\\
\bottomrule
\end{tabular}}
\end{center}
\end{table}

\indent Five of the 16 networks were selected for further evaluation with the test dataset. AlexNet and Inception\_v3 were selected along with one each of the ``best'' variants of ResNet, DenseNet, and VGG. ResNet-18 was selected because it had comparable accuracy (while minimizing complexity) to the other ResNet variants (table \ref{tab:DR_acc_results}) and a ratio near 1 when pretrained in fine-tuning mode. DenseNet-161 had highest accuracy when pretrained and in fine-tuning mode (table \ref{tab:DR_acc_results}) and had a ratio $>$1 with these parameters. Lastly, VGG-19 was near the top in validation accuracy when pretrained and in fine-tuning mode (table \ref{tab:DR_acc_results}) and had a ratio $>$1.\\
\indent Accuracy of these 5 networks (pretrained) as fixed feature extractors or in fine-tuning mode when run with the test dataset is shown in table \ref{tab:DR_test_accuracy}. With fixed feature extractor, 4 of 5 networks had accuracy around 72-73\%, while DenseNet-161 had lower accuracy around 68\%. With fine-tuning, AlexNet and Inception\_v3 had about 3-4\% lower accuracy than the other 3 networks. DenseNet-161, ResNet-18, and VGG-19 had comparable accuracy around 76-77\%.

\begin{table}[htb!]
\caption[Diabetic retinopathy: Network Test Accuracy]{\textbf{Diabetic retinopathy: Network Test Accuracy}. Accuracy(\%) is indicated for the specified networks (all pre-trained) for fixed feature extractor and fine tuning modes.}\label{tab:DR_test_accuracy}
  \begin{center}
\scalebox{0.8}{
  \begin{tabular}{|l|r|r|}
\hline
\textbf{Network} & \textbf{Fixed Feature Extractor} & \textbf{Fine-Tuning}\\
\hline
AlexNet & 72.2 & 73.6\\
DenseNet-161 & 68.3 & 77.5\\
Inception-v3 & 73.7 & 73.8\\
ResNet-18 & 72.8 & 76.1\\
VGG-19 & 73.2 & 77.4\\
\hline
\end{tabular}}
\end{center}
\end{table}

\indent Quadratic weighted kappa was tabulated for each of the 5 networks (table \ref{tab:DR_test_kappa}). In fixed feature extractor mode, Inception\_v3 and VGG-19 showed poor level of agreement, while AlexNet, DenseNet-161, and ResNet-18 showed fair agreement. In fine-tuning mode, AlexNet and Inception\_v3 showed poor level of agreement, while DenseNet-161, ResNet-18, and VGG-19 showed moderate level of agreement.

\begin{table}[htb!]
\caption[Diabetic retinopathy: Network Test Quadratic Weighted Kappa]{\textbf{Diabetic retinopathy: Network Test Quadratic Weighted Kappa.} Quadratic weighted kappa values are indicated for the specified networks (all pre-trained) for fixed feature extractor and fine tuning modes. A generally agreed upon scale is: $<$0.20(Poor), 0.21-0.40(Fair), 0.41-0.60(Moderate), 0.61-0.80(Good), and 0.81-1.00(Very good).}\label{tab:DR_test_kappa}
  \begin{center}
\scalebox{0.8}{
  \begin{tabular}{|l|r|r|}
\hline
\textbf{Network} & \textbf{Fixed Feature Extractor} & \textbf{Fine-Tuning}\\
\hline
AlexNet & 0.21 & 0.18\\
DenseNet-161 & 0.32 & 0.55\\
Inception-v3 & 0.005 & 0.15\\
ResNet-18 & 0.23 & 0.47\\
VGG-19 & 0.16 & 0.56\\
\hline
\end{tabular}}
\end{center}
\end{table}

\indent Sensitivity and specificity were tabulated for each of the 5 networks (table \ref{tab:DR_s_and_s}). Since sensitivity and specificity are based on a binary classification, they were calculated in two ways: (1) for ``any DR,'' where the binary classification was Class 0 vs. any level of DR (Class 1, 2, 3, or 4); and (2) ``referable DR,''\footnote{See \cite{DR-screening9} for definition of ``referable DR.''} where the binary classification was non-referable (Class 0 or 1) vs. referable (Class 2, 3, or 4). Table \ref{tab:DR_s_and_s} shows that:
\begin{itemize}
\item Sensitivity was generally low for all the networks whether fine-tuning vs. fixed feature extractor.
\item Sensitivity was in all cases higher for fine-tuning compared to fixed feature extractor. For VGG-19, it was substantially higher (57.2\% vs. 8.3\%).
\item The highest sensitivity was achieved by VGG-19 (fine-tuning).
\item Inception\_v3 showed poor sensitivity.
\item Specificity was generally high ($>$94\%) for all the networks (fine-tuning).
\end{itemize}

\begin{table}[htb!]
\caption[Diabetic retinopathy: Network Test Sensitivity and Specificity]{\textbf{Diabetic retinopathy: Network Test Sensitivity and Specificity.} Sensitivity(\%) and specificity(\%) values are indicated for the specified networks (all pre-trained) for fixed feature extractor and fine tuning modes. Any DR indicates any level of diabetic retinopathy, referable DR indicates referable diabetic retinopathy, which is class 2 or higher (i.e., moderate diabetic retinopathy or worse).}\label{tab:DR_s_and_s}
  \begin{center}
\scalebox{0.7}{
  \begin{tabular}{|l|r|r|r|r|r|r|r|r|r|}
\toprule
& \multicolumn{4}{c}{\textbf{\underline{Fixed Feature Extractor}}} & \multicolumn{4}{c}{\textbf{\underline{Fine-tuning}}}\\
& \multicolumn{2}{c}{\textbf{\underline{Any DR}}} & \multicolumn{2}{c}{\textbf{\underline{Referable DR}}} & \multicolumn{2}{c}{\textbf{\underline{Any DR}}} & \multicolumn{2}{c}{\textbf{\underline{Referable DR}}}\\
\textbf{Network} & \textbf{Sensitivity} & \textbf{Specificity} & \textbf{Sensitivity} & \textbf{Specificity} & \textbf{Sensitivity} & \textbf{Specificity} & \textbf{Sensitivity} & \textbf{Specificity}\\
\midrule
AlexNet & 10.8 & 96.6 & 13.7 & 96.7 & 12.1 & 97.3 & 15.7 & 97.3\\
DenseNet-161 & 34.6 & 85.5 & 41.2 & 85.3 & 41.6 & 95.1 & 54.1 & 94.9\\
Inception-v3 & 3.0 & 99.9 & 4.0 & 99.9 & 10.5  & 97.6 & 13.3 & 97.8\\
ResNet-18 & 13.7 & 96.1 & 17.3 & 96.2 & 33.9 & 95.4 & 44.3 & 95.3\\
VGG-19 & 6.7 & 98.2 & 8.3 & 98.4 & 44.9 & 94.4 & 57.2 & 94.1\\
\bottomrule
\end{tabular}}
\end{center}
\end{table}

\indent A selected subset of networks (DenseNet-161, Inception\_v3, ResNet-18, and VGG-19) were re-run (as pretrained, fine-tuning) for 50 epochs. Their loss and accuracy curves are plotted in figure \ref{fig:DR_50_epochs}. These plots show that the loss and accuracy begin to converge at around 10 epochs for these networks and there is minimal improvement in loss or accuracy past 10 epochs.

\begin{figure}[htb]
\begin{subfigure}{0.8\textwidth}
\captionsetup{width=0.8\textwidth}
\centering\includegraphics[width=0.8\linewidth]{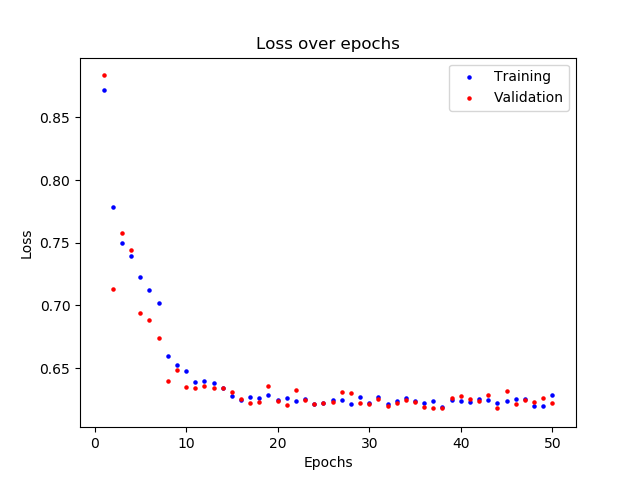}
\caption{Loss for ResNet-18.}
\label{fig:DR_50_epochs_a}
\end{subfigure}
\begin{subfigure}{0.8\textwidth}
\captionsetup{width=0.8\textwidth}
\centering\includegraphics[width=0.8\linewidth]{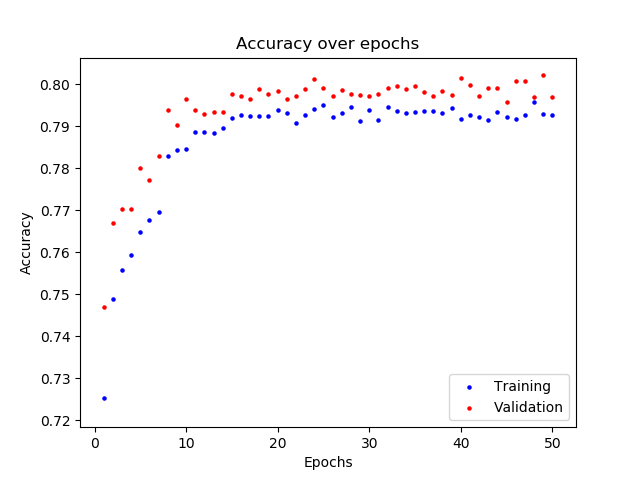}
\caption{Accuracy for ResNet-18.}
\label{fig:DR_50_epochs_b}
\end{subfigure}
\caption[Diabetic Retinopathy: Selected networks loss and accuracy curves over 50 epochs.]{\textbf{Diabetic Retinopathy: Selected networks loss and accuracy curves over 50 epochs.} Mode was set to fine-tuning and networks were pre-trained.}\label{fig:DR_50_epochs}
\end{figure}

\begin{figure}
\ContinuedFloat
\begin{subfigure}{0.8\textwidth}
\captionsetup{width=0.8\textwidth}
\centering\includegraphics[width=0.8\linewidth]{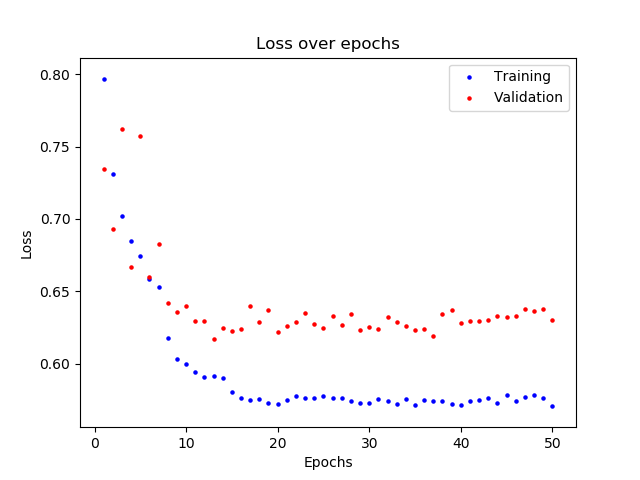}
\caption{Loss for Inception\_v3.}
\label{fig:DR_50_epochs_c}
\end{subfigure}
\begin{subfigure}{0.8\textwidth}
\captionsetup{width=0.8\textwidth}
\centering\includegraphics[width=0.8\linewidth]{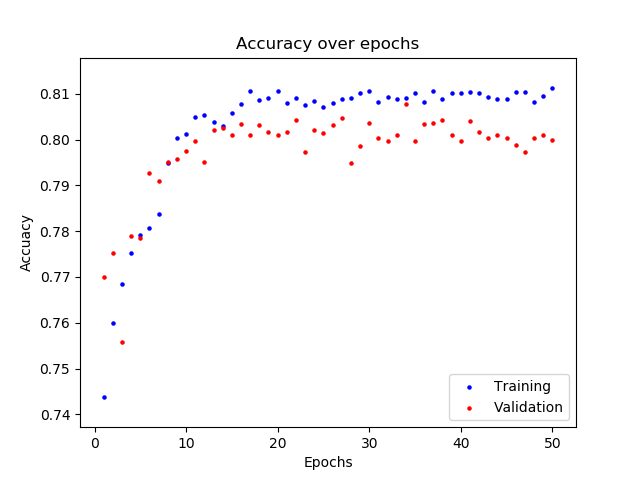}
\caption{Accuracy for Inception\_v3.}
\label{fig:DR_50_epochs_d}
\end{subfigure}
\caption[Continued - Diabetic Retinopathy: Selected networks loss and accuracy curves over 50 epochs.]{\textbf{Continued - Diabetic Retinopathy: Selected networks loss and accuracy curves over 50 epochs.} Mode was set to fine-tuning and networks were pre-trained.}\label{fig:DR_50_epochs2}
\end{figure}

\begin{figure}
\ContinuedFloat
\begin{subfigure}{0.8\textwidth}
\captionsetup{width=0.8\textwidth}
\centering\includegraphics[width=0.8\linewidth]{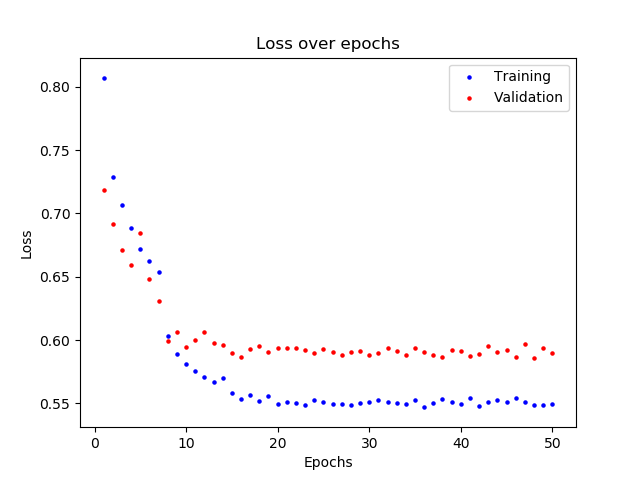}
\caption{Loss for Densenet-161.}
\label{fig:DR_50_epochs_c}
\end{subfigure}
\begin{subfigure}{0.8\textwidth}
\captionsetup{width=0.8\textwidth}
\centering\includegraphics[width=0.8\linewidth]{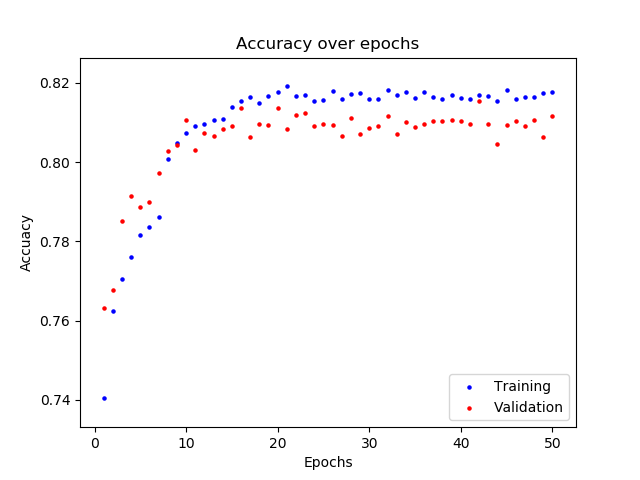}
\caption{Accuracy for Densenet-161.}
\label{fig:DR_50_epochs_d}
\end{subfigure}
\caption[Continued - Diabetic Retinopathy: Selected networks loss and accuracy curves over 50 epochs.]{\textbf{Continued - Diabetic Retinopathy: Selected networks loss and accuracy curves over 50 epochs.} Mode was set to fine-tuning and networks were pre-trained.}\label{fig:DR_50_epochs3}
\end{figure}

\begin{figure}
\ContinuedFloat
\begin{subfigure}{0.8\textwidth}
\captionsetup{width=0.8\textwidth}
\centering\includegraphics[width=0.8\linewidth]{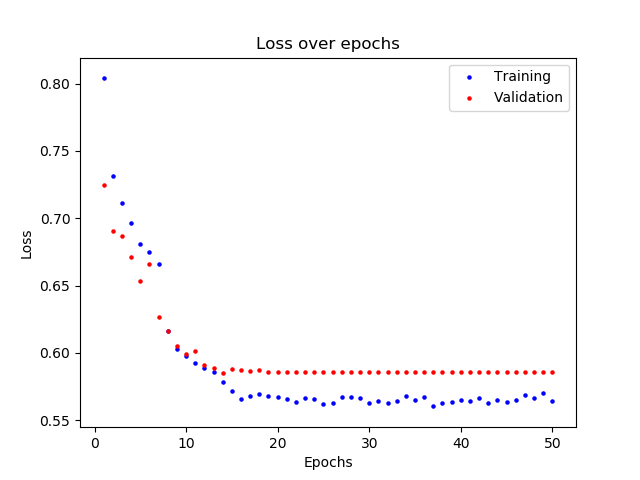}
\caption{Loss for VGG-19.}
\label{fig:DR_50_epochs_c}
\end{subfigure}
\begin{subfigure}{0.8\textwidth}
\captionsetup{width=0.8\textwidth}
\centering\includegraphics[width=0.8\linewidth]{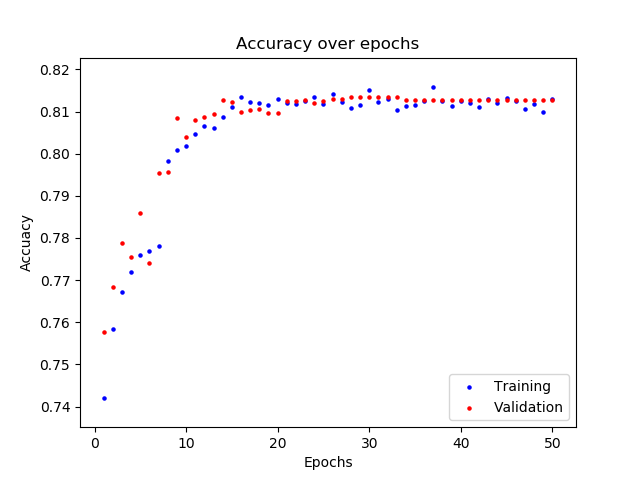}
\caption{Accuracy for VGG-19.}
\label{fig:DR_50_epochs_d}
\end{subfigure}
\caption[Continued - Diabetic Retinopathy: Selected networks loss and accuracy curves over 50 epochs.]{\textbf{Continued - Diabetic Retinopathy: Selected networks loss and accuracy curves over 50 epochs.} Mode was set to fine-tuning and networks were pre-trained.}\label{fig:DR_50_epochs4}
\end{figure}

\indent The effect of image pre-processing (see pre-processing methodology in chapter \ref{chap:DR_methods}) was analyzed for selected networks (AlexNet, DenseNet-161, Inception\_v3, ResNet-18, and VGG-19). Networks were pre-trained and used for fine-tuning and run for 10 epochs. Table \ref{tab:DR_image_preprocessing} shows there was minimal improvement of loss and accuracy for AlexNet, DenseNet-161 or ResNet-18. The validation accuracy for VGG-19 decreased very slightly (0.66\%). For Inception\_v3, there was a significant improvement in validation accuracy (6\%). 

\begin{table}[htb!]
\caption[Diabetic Retinopathy: Loss and Accuracy with image pre-processing.]{\textbf{Diabetic Retinopathy: Loss and Accuracy with image pre-processing.} Shown are the loss and accuracy for selected networks, comparing these values between runs with and without image pre-processng. Networks were pre-trained, run in fine-tuning mode for 10 epochs.}\label{tab:DR_image_preprocessing}
  \begin{center}
\scalebox{0.7}{
  \begin{tabular}{|l|r|r|r|r|}
\toprule
& \multicolumn{2}{c}{\textbf{\underline{Loss}}} & \multicolumn{2}{c}{\textbf{\underline{Accuracy}}}\\
\textbf{Network} & \textbf{Training} & \textbf{Validation} & \textbf{Training} & \textbf{Validation}\\
\midrule
AlexNet, without pre-processing & 0.7475 & 0.7431 & 0.7520 & 0.7579\\
AlexNet, with pre-processing & 0.7475 & 0.7402 & 0.7524 & 0.7528\\
DenseNet-161, without pre-processing & 0.6067 & 0.6162 & 0.8003 & 0.8038 \\
DenseNet-161, with pre-processing & 0.5808 & 0.5946 & 0.8073 & 0.8105\\
Inception\_v3, without pre-processing & 0.6368 & 0.6923 & 0.7896 & 0.7368 \\
Inception\_v3, with pre-processing, & 0.5996 & 0.6401 & 0.8013 & 0.7974\\
ResNet-18, without pre-processing & 0.6467 & 0.6616 & 0.7847 & 0.7952\\
ResNet-18, with pre-processing & 0.6271 & 0.6177 & 0.7927 & 0.7969\\
VGG-19, without pre-processing & 0.6055 & 0.5985 & 0.7990 & 0.8106\\
VGG-19, with pre-processing & 0.5975 & 0.5995 & 0.8018 & 0.8040\\
\bottomrule
\end{tabular}}
\end{center}
\end{table}

\indent The effect of adjusting class weights to account for class imbalance was investigated. VGG-19 (pretrained, fine-tuning) was re-run using various class-adjusted weights specified in table \ref{tab:DR_class_weighted}. The weights were adjusted to weigh Class 0 less than the other classes. One recommendation was to use weights corresponding to the inverse of the class size, which is the second row in table {\ref{tab:DR_class_weighted}. Other class weights were experimented. The results indicate that adjusting the weights did not lower the loss or increase the accuracy over equally weighted classes.

\begin{table}[htb!]
\caption[Diabetic Retinopathy: VGG-19 class-weighted loss and accuracy.]{\textbf{Diabetic Retinopathy: VGG-19 class-weighted loss and accuracy.} Shown are the class-weighted loss and accuracy for VGG-19, run as pre-trained in fine-tuning mode for 10 epochs. For comparison reference, the first row is the result from table \ref{tab:DR_loss_results} where the run was for equal weights across classes.}\label{tab:DR_class_weighted}
  \begin{center}
\scalebox{0.7}{
  \begin{tabular}{|r|r|r|r|r|r|r|r|r|}
\toprule
\multicolumn{5}{c}{\textbf{\underline{Class-adjusted weight}}} & \multicolumn{2}{c}{\textbf{\underline{Loss}}} & \multicolumn{2}{c}{\textbf{\underline{Accuracy}}}\\
\textbf{Class 0} & \textbf{Class 1} & \textbf{Class 2} & \textbf{Class 3} & \textbf{Class 4} & \textbf{Training} & \textbf{Validation} & \textbf{Training} & \textbf{Validation}\\
\midrule
1.0 & 1.0 & 1.0 & 1.0 & 1.0 & 0.6055 & 0.5985 & 0.7990 & 0.8106\\
0.000004 & 0.00045 & 0.0002 & 0.0013 & 0.0016 & 1.0820 & 1.0840 & 0.6651 & 0.7390\\
0.01 & 1 & 1 & 10 & 10 & 1.0917 & 1.0458 & 0.6494 & 0.6917\\
0.25 & 1.0 & 0.85 & 1.0 & 1.0 & 0.8698 & 0.8662 & 0.7809 & 0.7926\\
1 & 100 & 100 & 1000 & 10000 & 1.0913 & 1.0624 & 0.6624 & 0.7567\\
\bottomrule
\end{tabular}}
\end{center}
\end{table}

\indent Lastly, the effect of using bilateral eye data was investigated. The rationale is that if either of a given patient's eye meets criteria for referal from a screening, then the patient is referred for further evaluation. For example, if the right eye is Class 0 and the left eye is Class 3 (severe NPDR) then the patient is referred for further evaluation.\\
\indent The Kaggle Diabetic Retinopathy data set consists of bilateral eye data. The test dataset consists of 53,576 images from 26,788 patients, where each patient contributed a right and left eye image. To analyze if performance could be increased, the test cases were ``blended'' from eye to patient level by assigning a patient level diagnosis based on the eye with worse severity. This analysis was carried out for VGG-19 (pre-trained, fine-tuning), which showed accuracy of 73\%, sensitivity and specificity of 66\% and 90\% (for referable DR), and quadratic weighted kappa of 0.58. This was a decrease of 3.4\% in accuracy, an increase of 8.8\% in sensitivity, a decrease of 4.1\% in specificity, and an increase of 0.02 in quadratic weighted kappa compared to eye-level statistics.

\chapter{Discussion\label{chap:DR_discussion}}

This report detailed the results of a transfer learning evaluation for classification of diabetic retinopathy by digital fundus photography. Prior reports (see chapter \ref{chap:background_DR}) have investigated select DNNs, such as Inception\_v3 or custom CNNs. However, to the best of my knowledge, no prior work has systematically investigated the performance of the full suite of PyTorch torchvision models.\\
\indent In this study, transfer learning techniques were applied to AlexNet, Inception\_v3, and the variants of DenseNet, ResNet and VGG for classification of diabetic retinopathy by digital fundus photographs of the retina. The largest publically available dataset (as of the time of this report) from the Kaggle Diabetic Retinopathy Detection competition was utilized for training, validation, and testing. This dataset comprises just over 88K retinal fundus images and was acquired in a screening setting, where the images were taken under a variety of conditions.\\
\indent The 16 DNNs under evaluation were run in 4 configurations: in a not pre-trained mode to establish a baseline for comparison, and then with the pre-trained configuration (\ie network weights established previously from the ImageNet dataset). Within each mode, the networks were run in configuration of either a fixed-feature extractor (\ie ``training the top layer'') or in fine-tuning mode, whereby the network is retrained with the data from the transfer domain. Which of these modes of training yields higher accuracy is unclear since prior reports do not detail which methodology has been utilized, although prior work likely used the ``training the top layer approach.'' The rationale for this approach is that the torchvision models, such as ResNet, have been trained on millions of images, and thus can be sucessfully transferred to detect lower level features (\eg edge corners) that are agnostic to a particular target domain. Presumably, then, retraining of the top layer to another domain while freezing the lower layers should yield good accuracy. The key findings from this study are:\\
\\
\noindent (1) Pre-trained networks, in general, performed better than not pre-trained (considered baseline models in this study), as indicated by lower loss and higher accuracy. This was confirmed at the level of each particular DNN (tables \ref{tab:DR_loss_results} and \ref{tab:DR_acc_results}, figures \ref{fig:AlexNet_acc_plt}-\ref{fig:Diabetic_Retinopathy_VGG-19_acc_plt}) and when the networks were aggregated for analysis (tables \ref{tab:DR_loss_stats}, \ref{tab:DR_acc_stats}, \ref{tab:DR_loss_acc_comparisons}[Test no. 1-4], figures \ref{fig:DR_loss_boxplots}, \ref{fig:DR_acc_boxplots}).  This finding is expected, since networks that were evaluated as pre-trained on the very large ImageNet dataset (on order of millions of images) should have better ability to detect low level features, and in turn to result in better performance at the higher layers, compared to the same network trained on the relatively smaller Kaggle dataset (on order of tens of thousands of images).\\
\\
\noindent (2) Except for VGG, adding layers to a particular network generally did not improve performance. In particular, the loss (across pretrained vs. not-pretrained and fine-tuning vs. fixed feature extractor combinations) stayed stable for the DenseNet variants as the number of layers increased (table \ref{tab:DR_loss_results}, figure \ref{fig:DR_network_loss_boxplots}). The loss for the ResNet variants mostly stayed stable or slightly increased, except ResNet-101, which showed a significant increase in loss in one combination (pretrained, validation phase, feature extractor). There was a slight increase in loss from VGG-11 to the more complex VGG variants, except for the pretrained, validation phase, fine tuning combination where loss decreased from VGG-11 to VGG-19. The accuracy data showed that adding layers in the networks did not substantially change the accuracy for DenseNet or ResNet (table \ref{tab:DR_acc_results}, figure \ref{fig:DR_network_acc_boxplots}). The (pretrained) validation accuracy for VGG (fine-tuning) improved substantially from VGG-11 (73.5\%) to VGG-13 (80.8\%), with minimal change with VGG-16 or VGG-19. On the other hand, the accuracy for (pretrained) VGG (feature extractor) did not change substantially with more layers.\\
\\
\noindent (3) The network training to validation ratios showed that when not pre-trained, most of the networks had ratios near or $>$1, except for the DenseNet variants, and ResNet-50, -101, and -152 when evaluated in fine-tuning mode (table \ref{tab:DR_ratio_stats}, figure \ref{fig:DR_ratio_boxplots}). This may imply that DenseNet and those particular ResNet variants were overfitting. When the networks were configured as pre-trained, the ratios were all near 1 or $>$1 for all the networks, except ResNet-152. There was a deviation in validation loss for ResNet-101 as a feature extractor (table \ref{tab:DR_loss_results}), which caused an outlier ratio. In general, however, the ratio data suggests that most of the models were not overfitting when pre-trained (either fixed feature extractor or fine-tuning) and that those models that may have been overfitting when not pretrained (the DenseNet variants) were not overfitting when pretrained.\\
\\
\noindent With the result of higher performance of pretrained over not pretrained networks established, the remaining analysis explored pretrained networks only, and in particular a subset of candidate networks for further analysis (AlexNet, DenseNet-161, Inception\_v3, ResNet-18, and VGG-19). Further key findings were:\\
\\
\noindent (4) Most of the networks in this subset showed $>$72\% accuracy, with the exception of DenseNet-161 (68.3\%). This is inline with the $>$70\% accuracy (interpreted as 100\% - Top-1 error in table \ref{tab:networks}) of these networks (excluding AlexNet) reported on ImageNet. The accuracy of each network was higher in fine-tuning mode compared to fixed feature extractor (range of difference 0.1-9.2\%), indicating that these networks had comparable (AlexNet, Inception\_v3) or in some cases (DenseNet-161, ResNet-18, and VGG-19) significantly higher accuracy in fine-tuning mode (table \ref{tab:DR_test_accuracy}). DenseNet-161 and VGG-19 achieved the highest accuracies (77.5 and 77.4\%, respectively) amongst the networks when fine-tuned.\\
\\

\noindent (5) The quadratic weighted kappa was poor in 2 cases (Inception\_v3, VGG-19) and fair in 3 cases (AlexNet, DenseNet-161, ResNet-18) in fixed feature extractor mode (table \ref{tab:DR_test_kappa}). In fine-tuning mode, the quadratic weighted kappa was poor in 2 cases (AlexNet, Inception\_v3) and moderate in 3 cases (DenseNet-161, ResNet-18, VGG-19). As in the accuracy results, DenseNet-161 and VGG-19 achieved the highest two kappa scores.\\
\\
\noindent (6) The sensitivity of this subset of networks for DR detection was generally poor or fair (table \ref{tab:DR_s_and_s}). The sensitivity of each network was much higher when in fine-tuning compared to fixed feature extractor mode, in some cases substantially higher (ResNet-18 and VGG-19). These relatively low sensitivity values indicate that there was a high proportion of false negatives. In the case of the ``any DR'' binary classification, this implied a high number of cases were predicted to be Class 0 by the network but identified as diseased by the ground truth. In the case of the ``referable DR'' binary classification, this implied that the network predicted cases to be non-referable when in fact the ground truth labeled them as referable. The issue of poor sensitivity and misclassification is explored further below. Comparing the sensitivity when the fine-tuning or fixed feature extractor variable was kept the same showed that the sensitivity was (as expected) higher for referable DR compared to any DR, but not substantially so.\\
\\
\noindent (7) The specificity of this subset of networks for DR detection was generally high: $>$94\% for all networks (fine-tuning or fixed feature extractor), except for DenseNet-161, which achieved about 85\% specificity with fixed feature extractor (table \ref{tab:DR_s_and_s}). These results indicate that the networks generally were able to correctly identifiy no DR (or non-referable) cases very well, with a small portion of false positives.\\
\\
\indent Taken together, these finding indicate that pretained networks performed better than not-pretrained networks, and that the networks generally performed better when fine-tuned rather than with the fixed feature extractor configuration. Adding layers to the DenseNet and ResNet networks generally did not enhance performance, but there was a significant boost from VGG-11 to the other VGG variants. The ratio data suggests that the networks were generally not overfitting when pretrained.\\
\indent The accuracy of the subset of networks selected to be evaluated with the test data (AlexNet, DenseNet-161, Inception\_v3, ResNet-18, and VGG-19), showed similar accuracy when used in fine-tuning mode (range 73.6-77.5\% accuracy). The accuracy was higher for fine-tuning compared to fixed feature extractor in 4 of 5 cases (AlexNet, DenseNet-161, ResNet-18, and VGG-19), and stayed stable for Inception\_v3.\\
\indent The quadratic weighted kappa results indicate that the networks generally had poor to fair level of agreement in fixed feature extractor mode. Although the kappa values improved with fine-tuning, the highest kappa value (0.56 for VGG-19), still indicated only moderate level of agreement. Of note, the metric used by the Kaggle Diabetic Retinopathy Detection competition was the quadratic weighted kappa rather than accuracy, with the two top submissions receiving scores of 0.84957 and 0.84478, respectively.\\
\indent Both of those submissions used techniques (random forest, resampling) to address the class imbalance issue with this dataset, specifically that a large portion of the dataset is labeled as Class 0. In this analysis, the class weights were adjusted for the VGG-19 network (pretrained, fine-tuning), however no improvement in accuracy was observed (table \ref{tab:DR_class_weighted}). Interestingly, Pratt \etal\cite{Pratt} recognized the risk of class imbalance overfitting, and described their strategy as ``for every batch loaded for back-propagation, the class-weights were updated with a ratio respective to how many images in the training batch were classified as having no signs of DR. This reduced the risk of over-fitting to a certain class to be greatly reduced.''\\
\indent It is unclear how much their technique reduced the risk, since no comparative data without this method is provided in their report. However, they reported that their final trained network achieved 95\% specificity, 75\% accuracy and 30\% sensitivity (based on an ``any DR'' classification scheme). In this study, the best results were achieved by VGG-19 (with no class weight adjustments), which showed 77.4\% accuracy, 44.9\% sensitivity, and 94.4\% specificity (``any DR,'' when pretrained and fine-tuning). Thus, VGG-19's accuracy and specificity (without class weighted adjustment) was comparable to that of the CNN by Pratt \etal while the sensitivity of VGG-19 was about 15\% higher. It should be noted that Pratt \etal did not report a kappa statistic in their paper. Furthermore, it is unclear how much of an effect the class imbalance strategies employed by the top two Kaggle submissions had since scores were only provided on the final overall models. Further study of the role of class imbalance and strategies to compensate for it are needed.\\
\indent Another strategy that the top Kaggle submission and Pratt \etal used was image pre-processing. The second place Kaggle submission stated that ``we crop away all background and resize the images to squares of 128, 256 and 512 pixel'' but did not use other image pre-processing. Pratt \etal described a ``colour normalisation'' scheme, while the top Kaggle submission used an image pre-processing technique detailed in chapter \ref{chap:DR_methods}. As in the case of class imbalance above, it is unclear how much the image pre-processing increased performance, as results for only the final model are provided. In this report, the effect of image pre-processing (utilizing the same methodology as the top Kaggle submission) was analyzed (table \ref{tab:DR_image_preprocessing}) for a subset of networks and showed that performance improved modestly (6\%) for only 1 of the 5 networks (Inception\_v3).\\
\indent Lastly, an examination of the possible sources of error is instructive. For VGG-19 (pre-trained, fine-tuning) the 53,576 predictions consisted of 6,311 (11.8\%) true positives, 37,306 (69.6\%) true negatives, 2,227 (4.2\%) false positives, and 7,732 (14.4\%) false negatives. As indicated by the high specificity, we observe that there was a relatively small percentage of false positives. The confusion matrix\footnote{See \url{https://en.wikipedia.org/wiki/Confusion_matrix}} in figure \ref{fig:confusion_matrix} shows that amongst the 2,227 false positives, 2,129 (95.6\%) occured when VGG-19 predicted class 2 (moderate NPDR) when ground truth was labeled as class 0.\\

\begin{figure}[htb!]
\centering\includegraphics[width=0.8\linewidth]{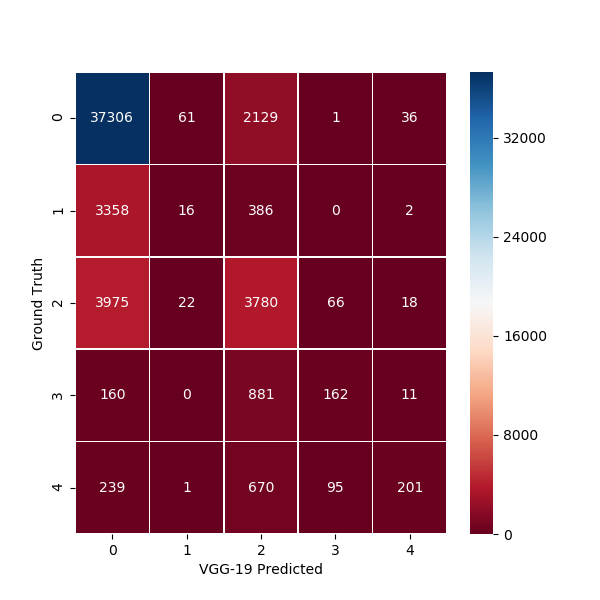}
\caption[Confusion Matrix for VGG-19]{Confusion Matrix for VGG-19.}
\label{fig:confusion_matrix}
\end{figure}

\indent The author of this report, a board-certified ophthalmologist, reviewed a random sample of 50 cases among the 2,129 class 2/class 0 misclassifications. In that random sample, 18 (36\%) were false positives because the eye had drusen (deposits typically found in age-related macular degeneration, not associated with diabetic retinopathy), but no diabetic retinopathy. Presumably, the neural network detected the drusen incorrectly as retinal hemorrhages or exudates and misclassified the diagnosis. Another 18 (36\%) cases were due to ``smudge'' artifacts from the camera, which the network likely identified as diabetic findings, but were correctly labeled by ground truth as class 0. Six (12\%) cases had poor lighting/artifact, and presumably the neural network identified dark regions as areas of disease. Three (6\%) cases had retinal scars (not due to diabetes) that were likely identified as diabetic findings by the network. Three (6\%) cases had a true retinal hemorrhage, and thus were misclassified by ground truth as normal. Two cases (4\%) had sufficient quality (no artifact or lighting issues) and no retinal pathology, and were thus true errors by the network. Figure \ref{fig:DR_FP_examples} shows representative examples from each type of misclassification.

\begin{figure}[htb]
\begin{subfigure}{0.8\textwidth}
\captionsetup{width=0.8\textwidth}
\centering\includegraphics[width=0.8\linewidth]{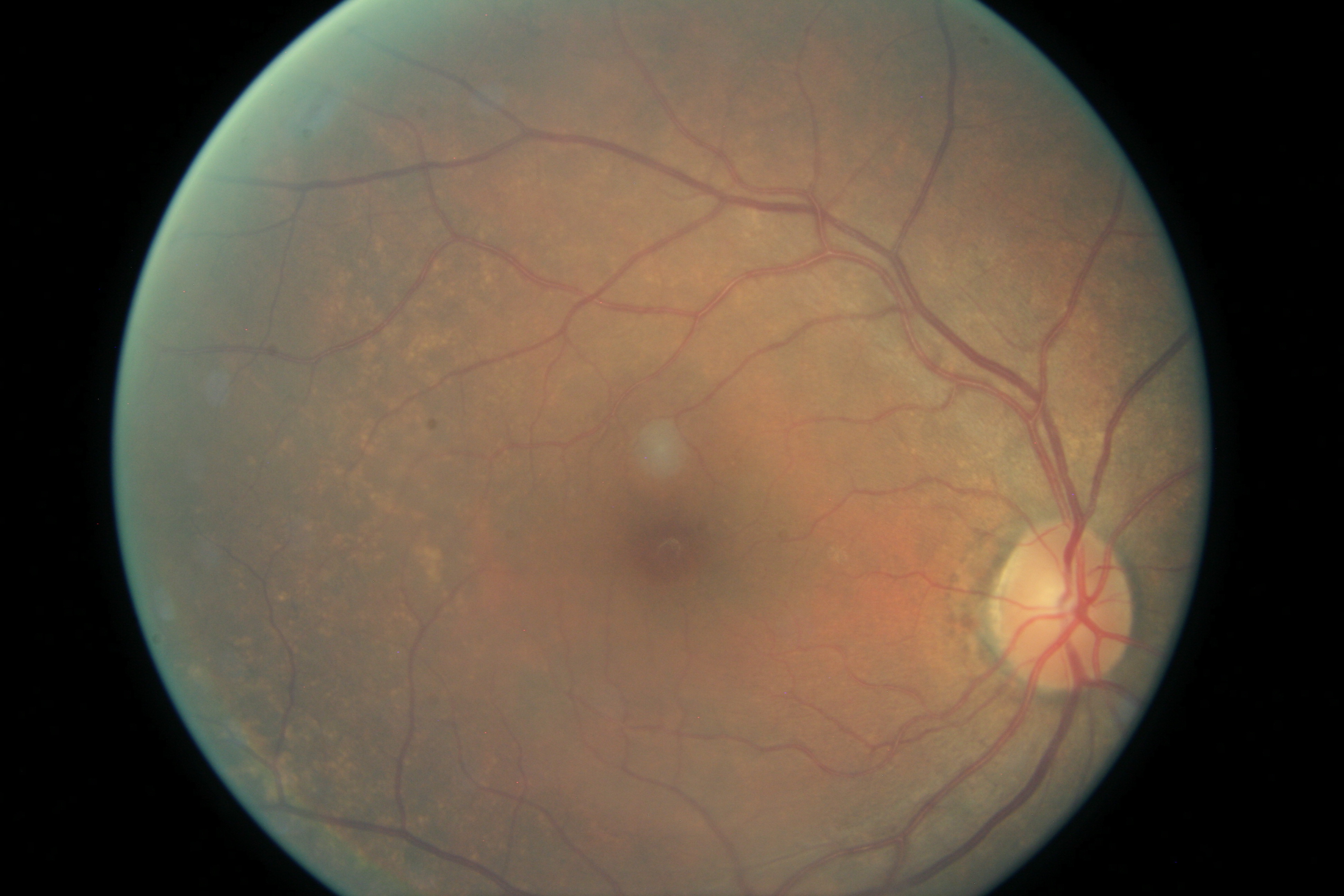}
  \caption{Drusen (yellow deposits on the left side of image) may have been misclassified as DR feature.}
  \label{fig:FP_a}
\end{subfigure}
\begin{subfigure}{0.8\textwidth}
\captionsetup{width=0.8\textwidth}
\centering\includegraphics[width=0.8\linewidth]{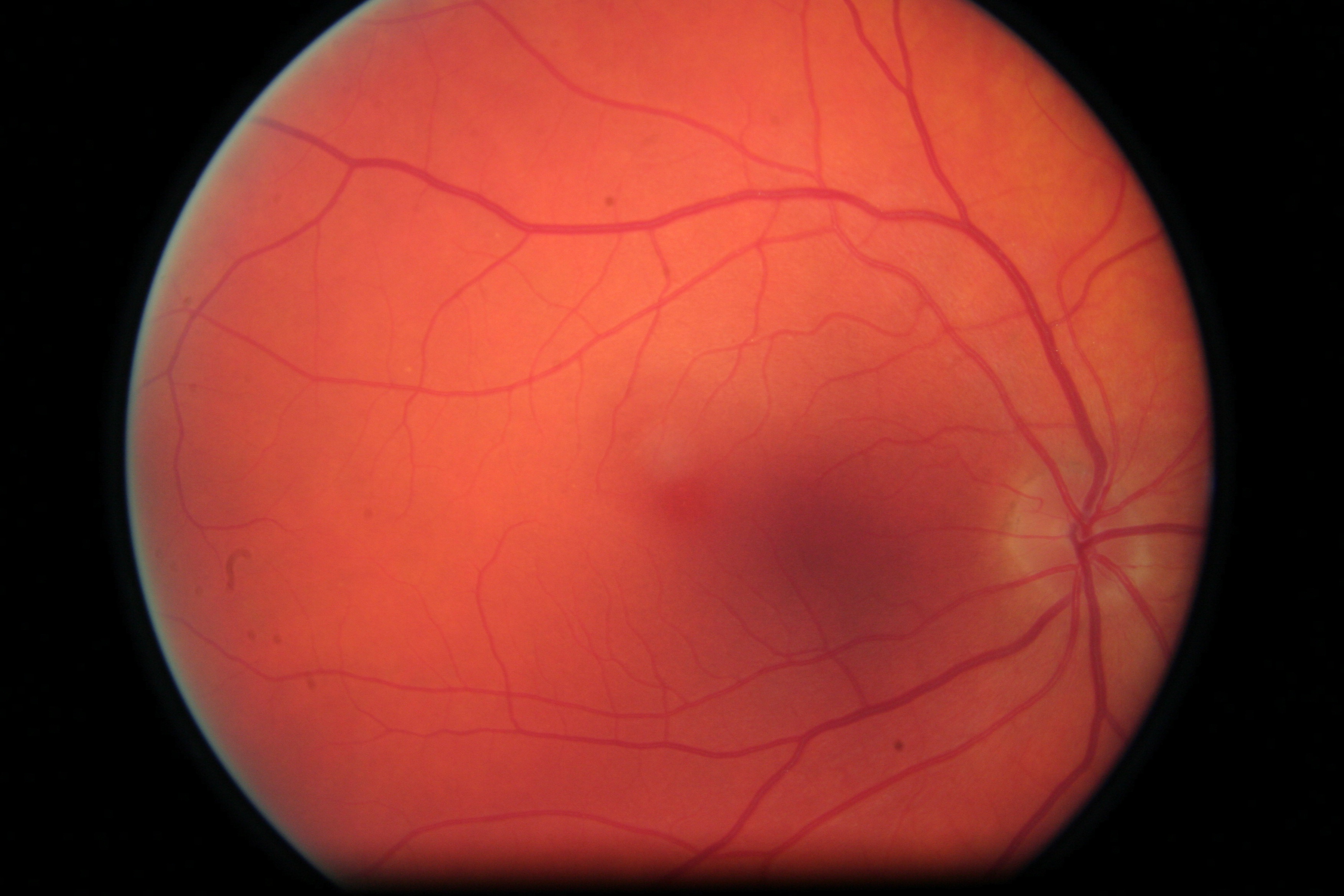}
  \caption{``Smudges'' due to camera artifact likely may have been misclassified as DR feature.}
  \label{fig:FP_b}
\end{subfigure}
\caption[Diabetic Retinopathy: examples of false positives.]{Diabetic Retinopathy: examples of false positives.}
\label{fig:DR_FP_examples}
\end{figure}

\begin{figure}
\ContinuedFloat
\begin{subfigure}{0.8\textwidth}
\captionsetup{width=0.8\textwidth}
\centering\includegraphics[width=0.8\linewidth]{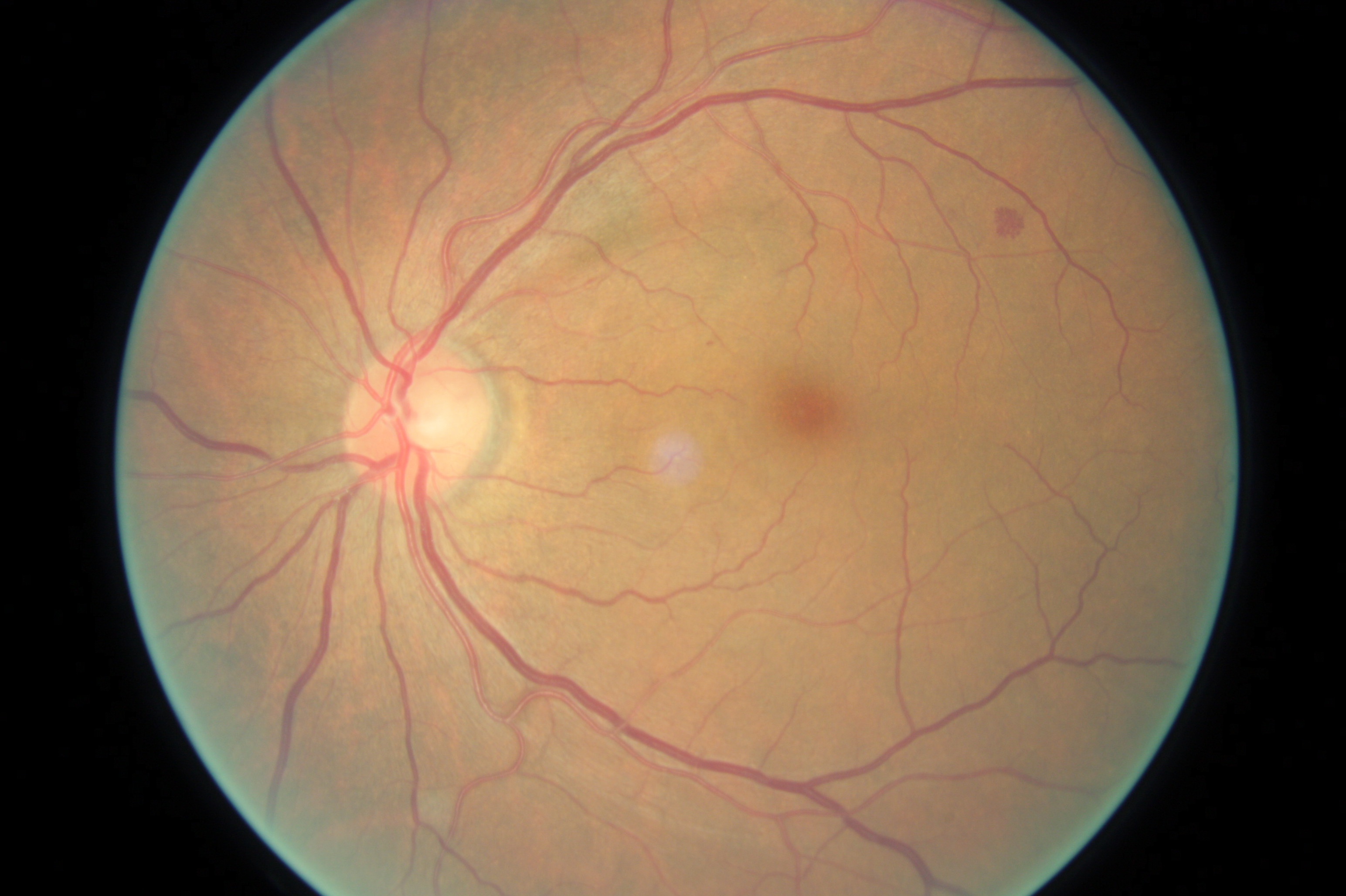}
  \caption{Image was mis-labeled by ground truth. Has a retinal hemorrhage, indicating presence of DR.}
  \label{fig:FP_c}
\end{subfigure}
\begin{subfigure}{0.8\textwidth}
\captionsetup{width=0.8\textwidth}
\centering\includegraphics[width=0.8\linewidth]{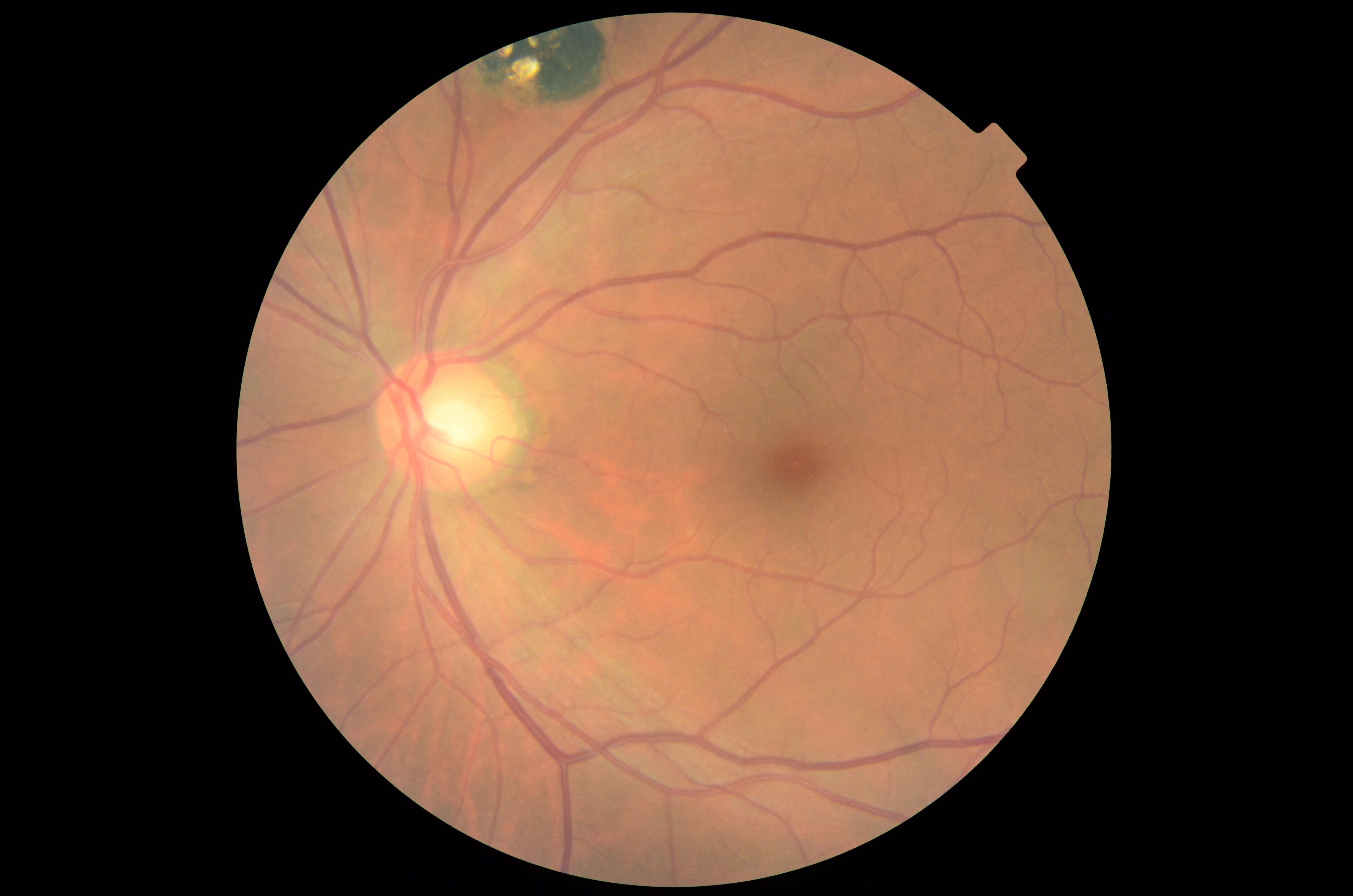}
  \caption{Image has a retinal scar at the top, unrelated to diabetes, and was likely misclassified as a DR feature.}
  \label{fig:FP_d}
\end{subfigure}
\caption[Continued - Diabetic Retinopathy: examples of false positives.]{Continued - Diabetic Retinopathy: examples of false positives.}
\label{fig:DR_FP_examples2}
\end{figure}

\begin{figure}
\ContinuedFloat
\begin{subfigure}{0.8\textwidth}
\captionsetup{width=0.8\textwidth}
\centering\includegraphics[width=0.8\linewidth]{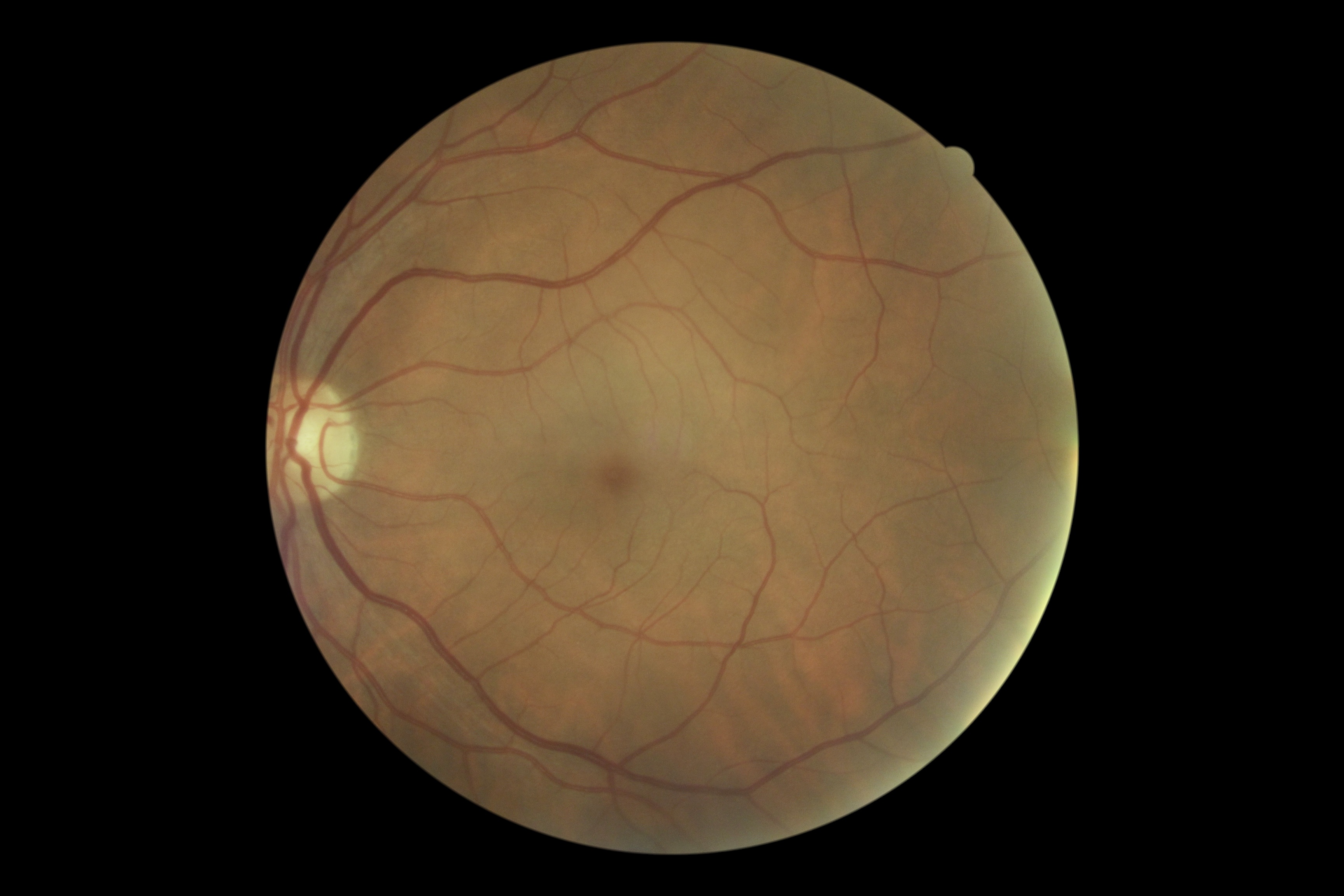}
  \caption{Image was labled correctly as class 0 by ground truth. The network misclassified this image as class 2 (unclear why).}
  \label{fig:FP_e}
\end{subfigure}
\caption[Continued - Diabetic Retinopathy: examples of false positives.]{Continued - Diabetic Retinopathy: examples of false positives.}
\label{fig:DR_FP_examples3}
\end{figure}

\indent It is unclear what triggered the network to misclassify the last two particular cases (true errors), but this type of true error represented a small (4\%) percentage of the random sample. Cases where other types of retinal pathology (drusen, retinal scars) would lead to referal for evaluation would be indicated anyway if the photo was reviewed manually. Poor lighting and camera artifact are troublesome issues and represented a non-trival percentage (48\%) of the random sample. A strategy to deal with these issues is identify the poor quality photo at the time of aquisition, either manually or by deep learning techniques\cite{zago} and attempt to re-image the patient or refer if still indequate. While such sources of false positives are troublesome in terms of adding extra cases to manually screen, from a clinical standpoint it is better to refer potentially normal cases for further evaluation rather than miss potential disease.\\
\indent The confusion matrix in figure \ref{fig:confusion_matrix} shows that amongst the 7,732 false negatives, the majority occured when the ground truth label was class 1 (3,358 cases, 43.4\%) or class 2 (3,975 cases, 51.4\%), with a smaller proportion when ground truth was class 3 (160 cases, 2.1\%) or class 4 (239 cases, 3.1\%). A random sample of 100 false negative cases were selected for further review, 25 images from each class.\\
\indent Table \ref{tab:VGG_FN} indicates that in total, 42\% of the random sample were false negatives where VGG predicted Class 0, but the image did show features of diabetic retinopathy. 27\% of images were of poor quality, and likely should have not been labeled by ground truth (\ie excluded). 31\% of images were labeled incorrectly, and either were normal or had other retinal features (\eg drusen) but not diabetic retinopathy. Examples of misclassifications are shown in figure \ref{fig:DR_FN_examples}.

\begin{table}[htb!]
\caption[Diabetic retinopathy: VGG-19 False Negatives]{\textbf{Diabetic retinopathy: VGG-19 False Negatives} Listed is a breakdown of the cause of a false negative classification for 100 random images classified as false negatives. The VGG-19 predictions were all Class 0. Each table entry listed the number of cases.}
\label{tab:VGG_FN}
  \begin{center}
\scalebox{0.8}{
  \begin{tabular}{|l|l|l|l|}
\hline
\textbf{Ground Truth Label} & \textbf{Actual FN} & \textbf{Image Poor Quality} & \textbf{Ground Truth Label Error}\\
\hline
Class 1 & 8 & 6 & 11\\
Class 2 & 7 & 7 & 11\\
Class 3 & 12 & 7 & 6\\
Class 4 & 15 & 7 & 3\\
\hline
Totals & 42(42\%) & 27(27\%) & 31(31\%)\\
\hline
\end{tabular}}
\end{center}
\end{table}

\begin{figure}[htb]
\begin{subfigure}{0.8\textwidth}
\captionsetup{width=0.8\textwidth}
\centering\includegraphics[width=0.8\linewidth]{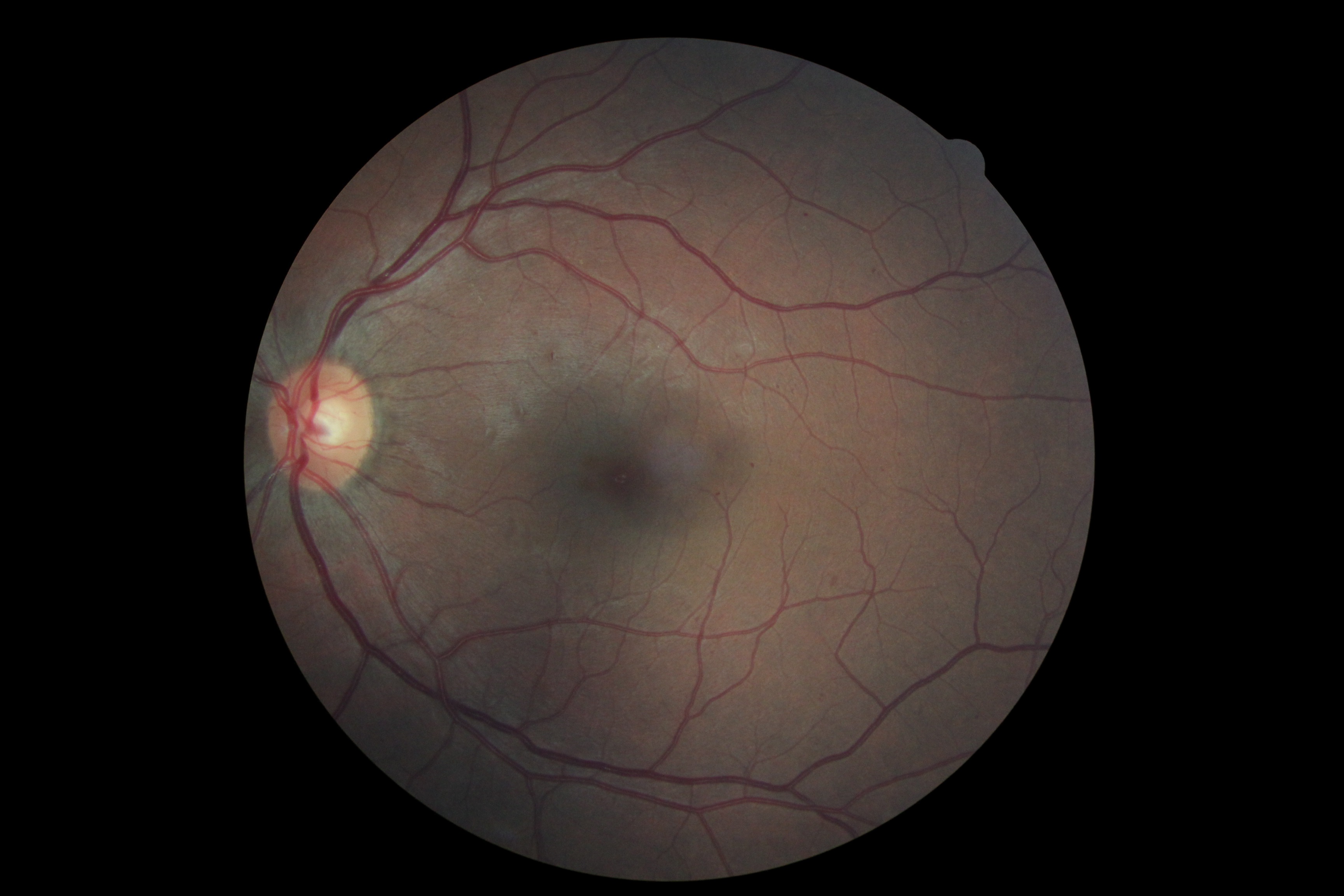}
  \caption{The network missed very small microaneurysms (small red dots) present in the image, indicating presence of DR.}
  \label{fig:FN_a}
\end{subfigure}
\begin{subfigure}{0.8\textwidth}
\captionsetup{width=0.8\textwidth}
\centering\includegraphics[width=0.8\linewidth]{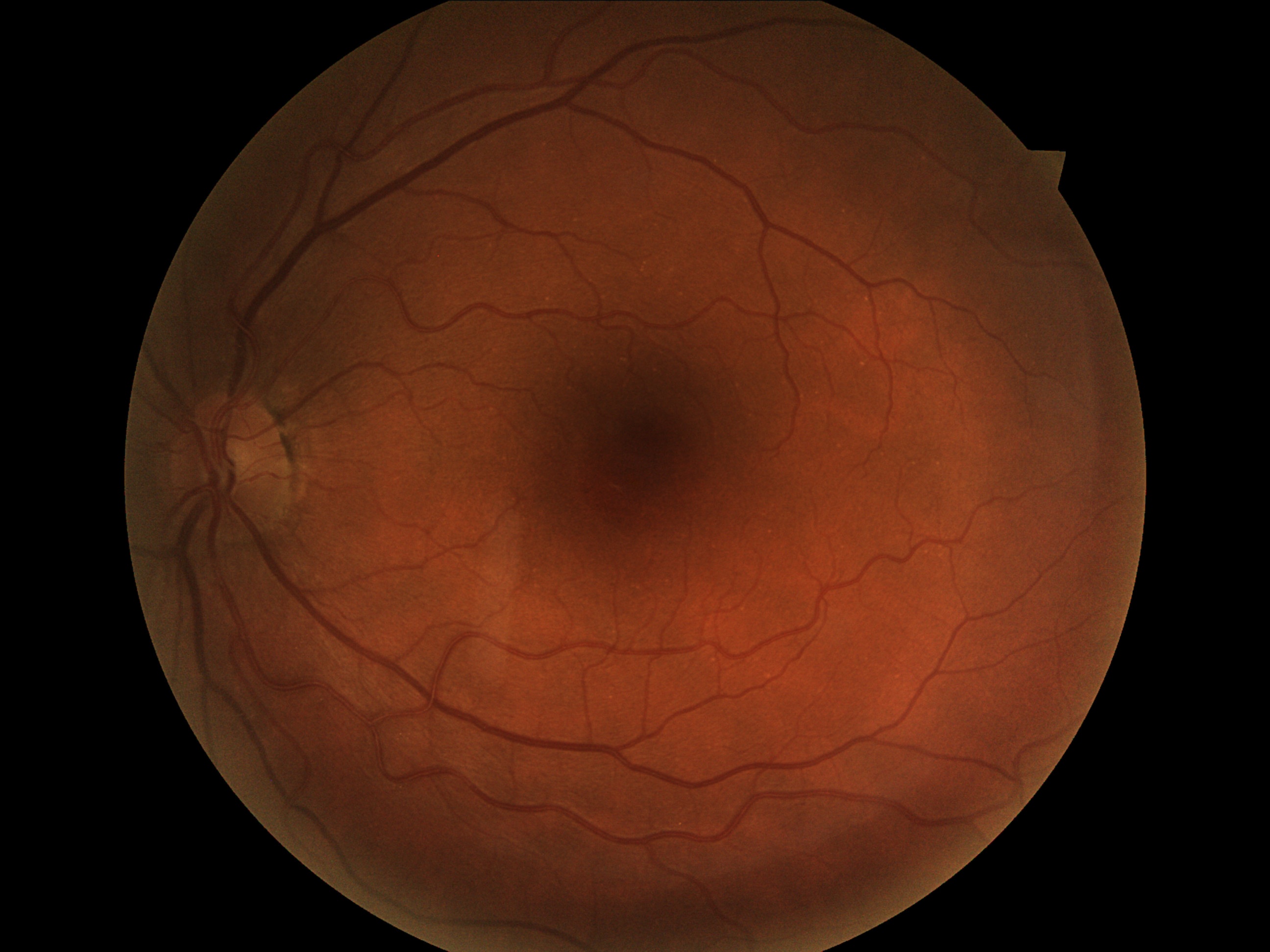}
  \caption{Likely a mislabel by ground truth. Drusen (small yellow dots) are present, but no features of DR are evident.}
  \label{fig:FP_b}
\end{subfigure}
\caption[Diabetic Retinopathy: examples of false negatives.]{Diabetic Retinopathy: examples of false negatives.}
\label{fig:DR_FN_examples}
\end{figure}

\begin{figure}
\ContinuedFloat
\begin{subfigure}{0.8\textwidth}
\captionsetup{width=0.8\textwidth}
\centering\includegraphics[width=0.8\linewidth]{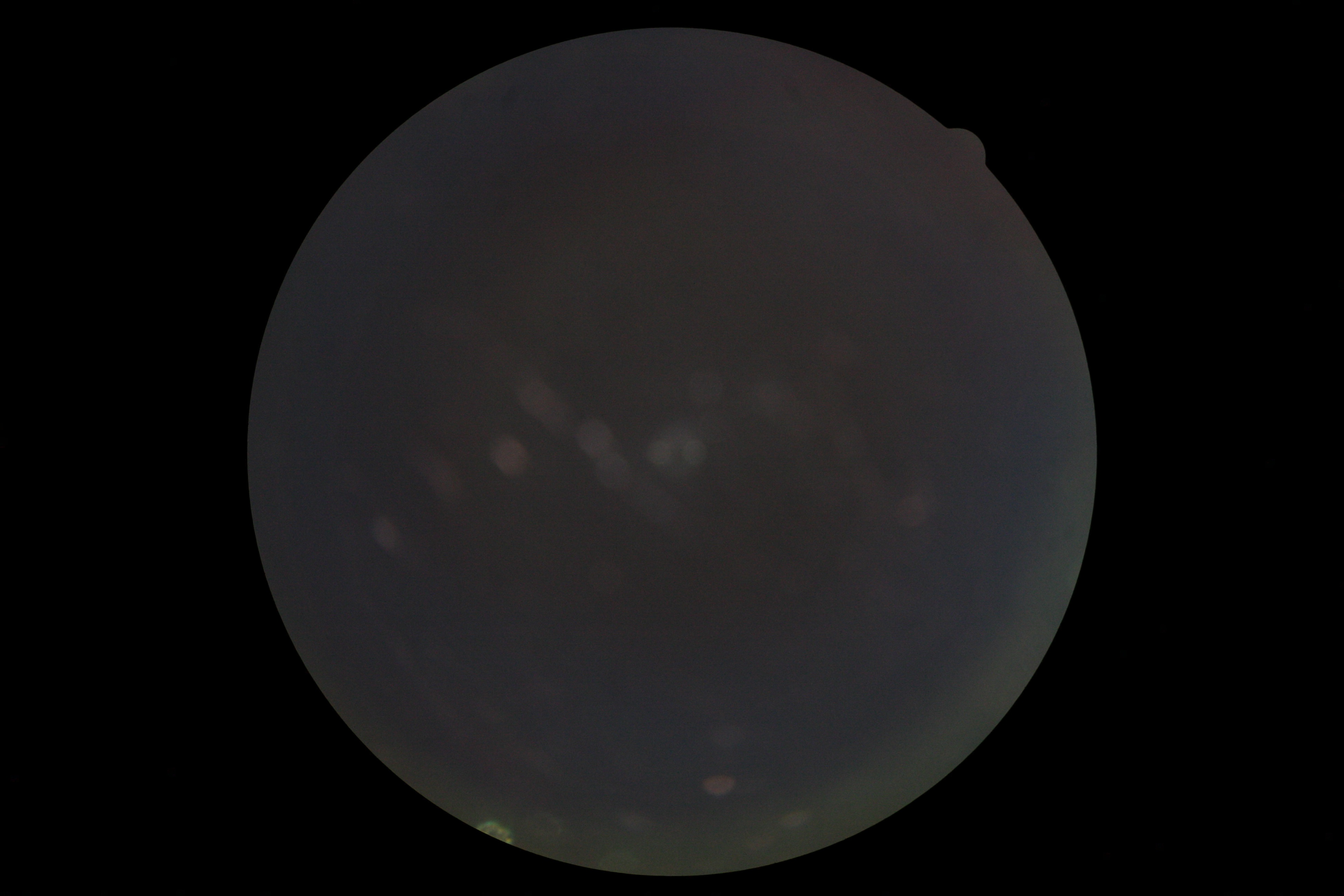}
  \caption{An example of a poor quality image, labeled as having DR by ground truth, likely should have been excluded.}
  \label{fig:FN_c}
\end{subfigure}
\begin{subfigure}{0.8\textwidth}
\captionsetup{width=0.8\textwidth}
\centering\includegraphics[width=0.8\linewidth]{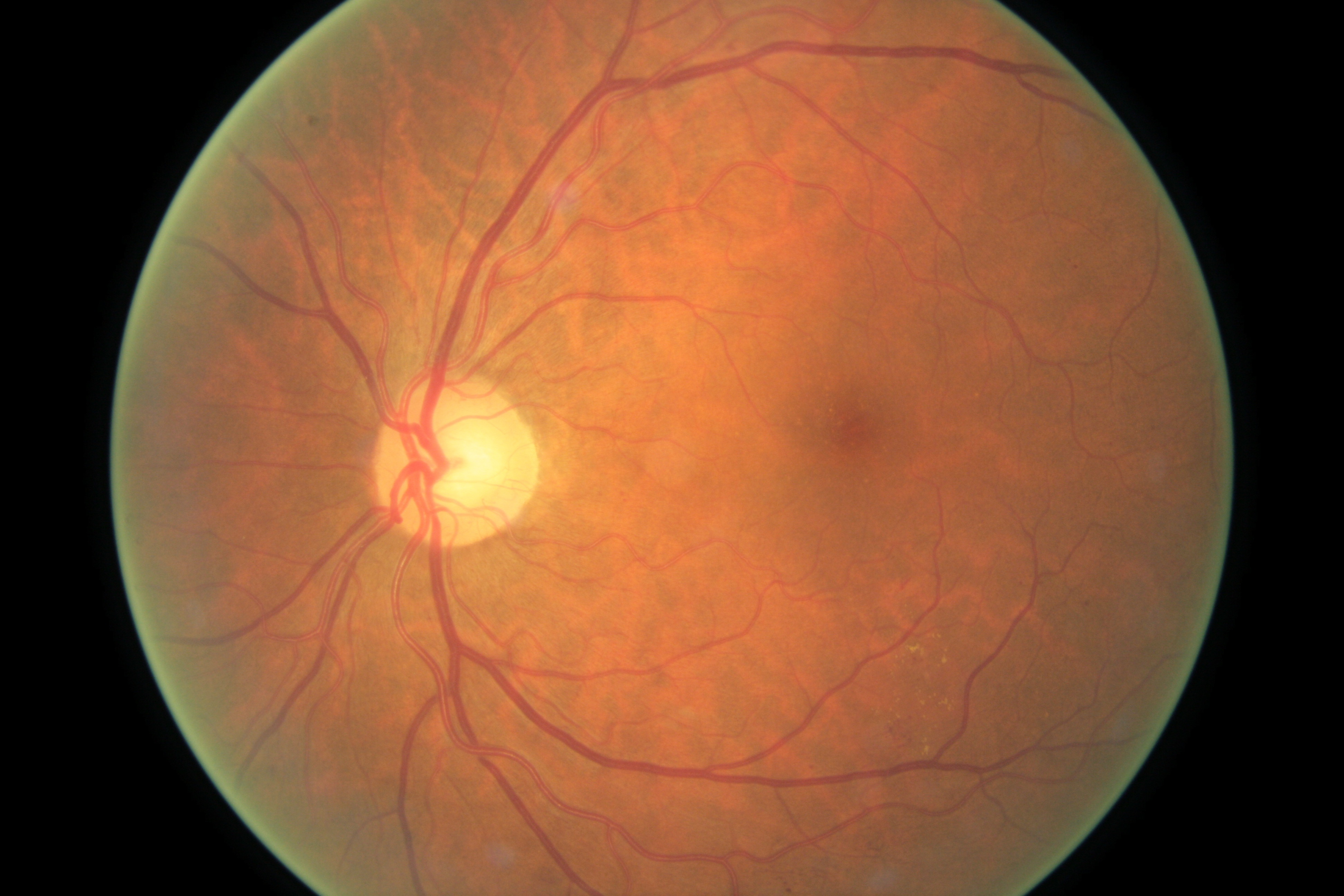}
  \caption{The network missed subtle microaneurysms and exudates present, indicated DR.}
  \label{fig:FN_d}
\end{subfigure}
\caption[Continued - Diabetic Retinopathy: examples of false negatives.]{Continued - Diabetic Retinopathy: examples of false negatives.}
\label{fig:DR_FN_examples2}
\end{figure}

\begin{figure}
\ContinuedFloat
\begin{subfigure}{0.8\textwidth}
\captionsetup{width=0.8\textwidth}
\centering\includegraphics[width=0.8\linewidth]{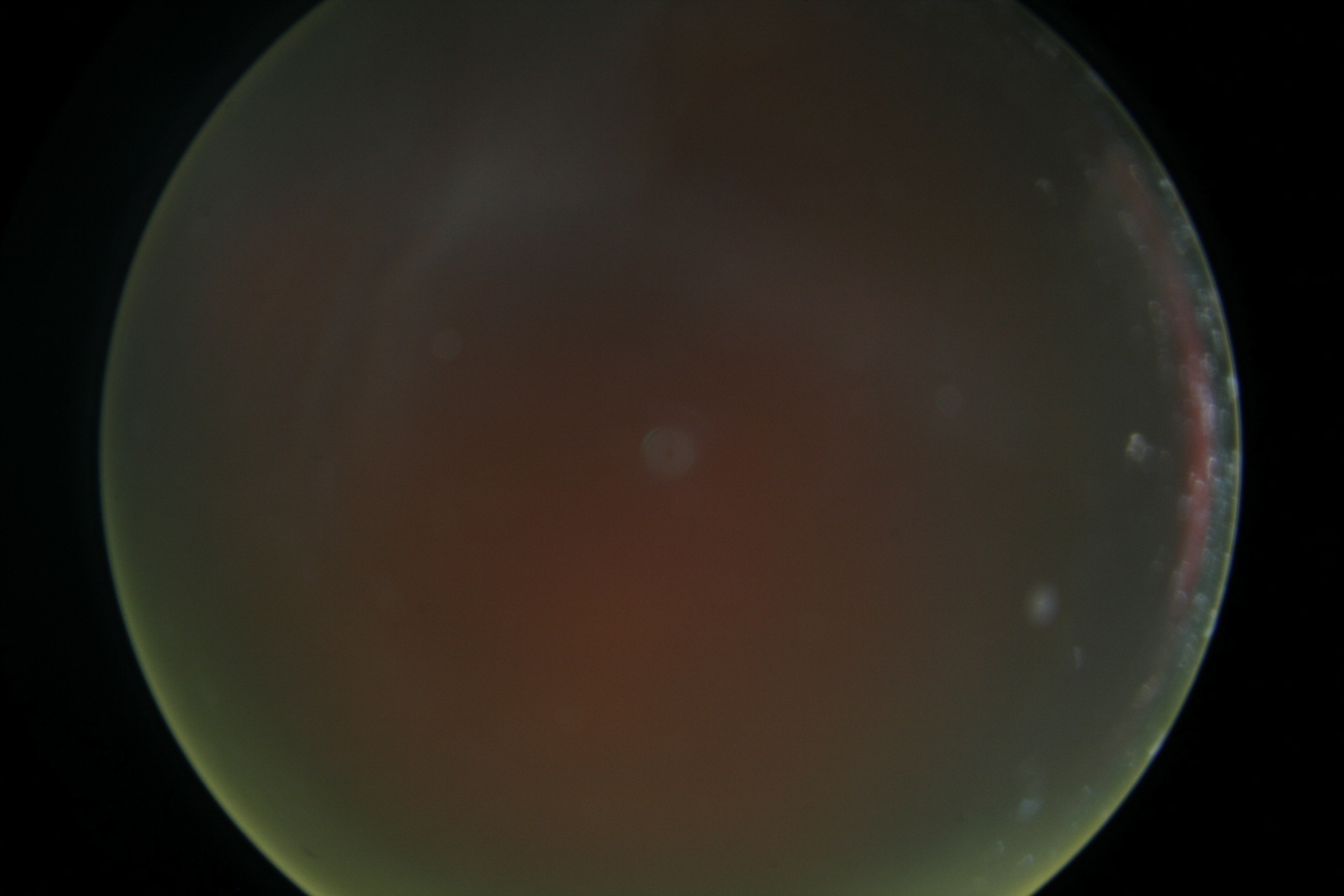}
  \caption{The image has a vitreous hemorrhage, indicating PDR.}
  \label{fig:FN_e}
\end{subfigure}
\begin{subfigure}{0.8\textwidth}
\captionsetup{width=0.8\textwidth}
\centering\includegraphics[width=0.8\linewidth]{6876_left.jpeg}
  \caption{The network missed the laser scars, indicating PDR.}
  \label{fig:FN_f}
\end{subfigure}
\caption[Continued - Diabetic Retinopathy: examples of false negatives.]{Continued - Diabetic Retinopathy: examples of false negatives.}
\label{fig:DR_FN_examples3}
\end{figure}

\indent Upon detailed review, it appears that cause of acutal false negatives were due to the network missing subtle features, such as very small retinal hemorrhages or exudates, especially when they are located in areas of the photo with low light or contrast. Additionally, the network may have not been trained on enough images of PDR showing vitreous hemorrhage or PDR treated with laser, and thus misclassified these cases as Class 0. Further study with a larger training and validation dataset of such relatively uncommon cases would be beneficial. It also appears that image pre-processing may have not helped to ``enhance'' very small features such as microaneurysms in low contrast regions of an image. This may represent an inherit limitation of DNNs in diabetic retinopathy detection, and a particular scenario where a human's ability to detect such subtle features still outperforms the network's. Further study of image pre-processing techniques or other network enhancements to detect such small features is warranted.\\
\indent The finding that a non-trivial percentage (31\%) of the random images classified as false negatives were due to ground truth label errors may imply that the accuracy and sensitivity were underestimated. Possible sources of error could be that ground truth readers defaulted to label a case as having DR when image quality was inadequate for diagnosis, or since the ground truth labels were human generated, there always will exist some level of error.\\
\indent Moreover, Pratt \etal\cite{Pratt} noted ``an associated issue identified, which was certified by a clinician, was that by national UK standards around over 10\% of the images in our dataset[Kaggle] are deemed ungradable. These images were defined a class on the basis of having at least a certain level of DR. This could have severely hindered our results as the images are misclassified for both training and validation." Gulshan \etal\cite{Gulshan} achieved a mich higher sensitivity using their dataset with the Inception\_v3 network, but it should be noted that a large number of ophthalmologists or ophthalmology trainees graded their datasets, highlighting that an accurately labeled ground truth dataset is critical and may shift accuracy and sensitivity results of the DNN.\\
\indent A limitation of this study was access to a large dataset of labeled retinal images (the Kaggle dataset being the largest in the public domain at this time). Further study with the smaller public Messidor-2 dataset may be informative. As a greater number of large datasets become available in the public domain, further research in the performance of transfer learning for diabetic retinopathy using retinal fundus photography will be of practical importance to validate this methodology as a screening technique.

\part{Transfer Learning for Classification of Diabetic Macular Edema, Choroidal Neovascularization and Drusen by Optical Coherence Tomography\label{part:three}}
\chapter{Background}\label{chap:part3_background}}

This part of the report analyzes the inter-disease classification performance of a suite of DNNs. The input images were JPEG photos of optical coherence tomography (OCT) scans through the macula. The DNNs were trained on photos showing 4 types of classes: (1) no disease (normal); (2) diabetic macular edema (DME); (3) choroidal neovascularization (CNV); or (4) drusen. CNV and drusen are two features associated with age-related macular degeneration (ARMD). OCT imaging is briefly reviewed next and in the following sections each of these clinical entities is briefly reviewed.

\section{Optical Coherence Tomography}

Optical Coherence Tomography (OCT) is a non-invasive imaging technique that provides high resolution cross sections of the retina with an axial resolution in the 5-7 $\mu$m range (see figure \ref{fig:AAO_ppp} for an example of a normal OCT)\cite{OCT_review}. OCT provides histological level detail of the retina. OCT employs light from a broadband light source, which is divided into a reference and a sample beam, to obtain a reflectivity versus depth profile of the retina. The light waves that are backscattered from the retina, interfere with the reference beam, and this interference pattern is used to measure the light echoes versus the depth profile of the tissue in vivo\cite{OCT_paper1}\cite{OCT_paper2}. For technical details of OCT, the reader is referred to \cite{OCT_book}.\\
\indent OCT is used extensively for imaging the macula and has become pivotal for diagnosis of retinal diseases such as DME and ARMD. It is estimated that 30 million OCT studies are done worldwide per year\cite{OCT_numbers}, and that figure is projected to continue growing with increasing prevalence of disease such as diabeties and macular degeneration. With rising demand for OCT imaging, there will likely be challenges keeping up with this demand for human based interpretation of images. Deep learning methods, discussed below, could be a promising technology to address this need.

\begin{figure}[htb]
\centering\includegraphics[width=0.8\linewidth]{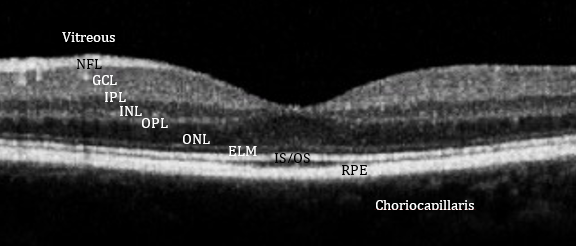}
\caption[Example of a Normal Optical Coherence Tomography Image.]{Example of a Normal Optical Coherence Tomography Image. Normal anatomy of the central macula. NFL, nerve fiber layer; GCL, ganglion cell layer; IPL, inner plexiform layer; INL, inner nuclear layer; OPL, outer plexiform layer; ONL, outer nuclear layer; ELM, external limiting membrane; IS/OS, inner segment/outer segment junction; RPE, retinal pigment epithelium. Reproduced from \url{https://www.aao.org/young-ophthalmologists/yo-info/article/4-tips-assessing-macular-oct-scan}.}
\label{fig:AAO_ppp}
\end{figure}

\section{Diabetic Macular Edema}

An overview of diabetic retinopathy (DR) is provided in chapter \ref{chap:background_DR}. Diabetic macular edema (DME) is a form of DR that involves the macula, the center of the retina.\footnote{Portions of this section courtesy of Diabetic Retinopathy Preferred Practice Pattern (2017)\cite{DR-PPP}.} Depending on the location of the DME and its chronicity, permanent vision loss can ensue. DME is a clinical diagnosis based upon a steroscopic dilated retinal fundus examination. The clinical diagnosis recognizes two forms of DME: clinically significant macular edema (CSME) and non-CSME. Criteria for CSME,  described in \cite{DR-PPP}, include any of the following:
\begin{itemize}
\item Thickening of the retina at or within 500 $\mu$m of the center of the macula.
\item Hard exudates at or within 500 $\mu$m of the center of the macula, when associated with adjacent retinal thickening. (This criteria does not apply to residual hard exudates that remain after successful treatment of prior retinal thickening.)
\item A zone or zones of retinal thickening one disc area or larger, where any portion of the thickening is within one disc diameter of the center of the macula
\end{itemize}

\begin{figure}[htb]
\centering\includegraphics[width=0.6\linewidth]{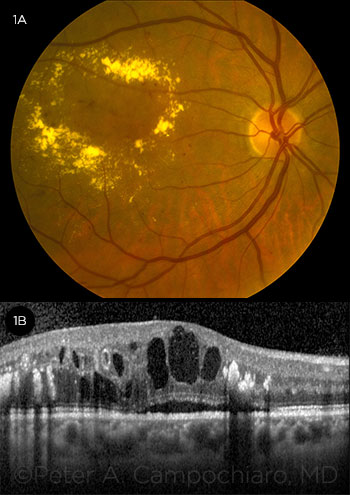}
\caption[Representative photo and OCT of Diabetic Macular Edema.]{Representative color fundus photo (top) and OCT (bottom) of Diabetic Macular Edema, reproduced from \url{https://www.aao.org/eyenet/article/diabetic-macular-edema-personalizing-treatment}. The photo shows a circinate ring of hard exudates (yellow deposits), wich may be associated with macular edema.}
\label{fig:DME_example}
\end{figure}

With the advent of OCT and intravitreal anti-vascular endothelial growth factor (VEGF) treatment, it has now become more appropriate to subdivide DME according to involvement at the center of the macula, because the risk of visual loss and the need for treatment is greater when the center is involved.\\
\indent OCT has become a pivotal imaging technology for ophthalmic imaging. However, the American Academy of Ophthalmology's preferred practice pattern does not recommend OCT to screen a patient with no or minimal diabetic retinopathy (figure \ref{fig:AAO_ppp}). As such, OCT, unlike digital fundus photography, is not currently recommended for population wide screening. Rather, OCT's current role in DME is to quantify the progression of DME or response to treatment, in patients with an established history of DME. Stereoscopic fundus photography is useful to document DME, but is not readily used for screening purposes, and because of the readier availability of OCT and ability to quantify DME over steroscopic imaging, OCT is generally utlilized in favor of steroscopic imaging. Non-stereoscopic fundus imaging, taken in a screening setting, can possibily detect DME based upon the presence of hard exudates, which are sometimes associated with macular edema (figure \ref{fig:DME_example}).

\begin{figure}[htb]
\centering\includegraphics[width=0.8\linewidth]{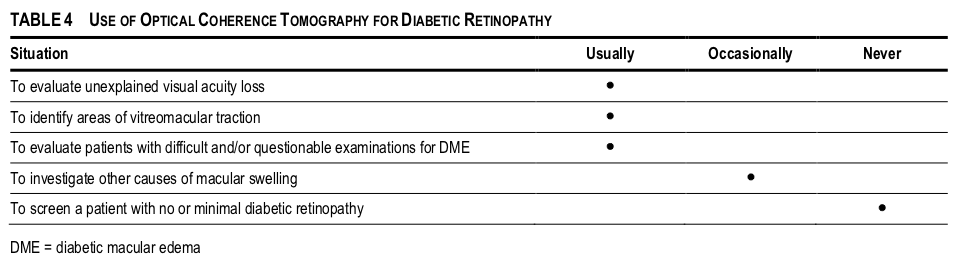}
\caption[Use of Optical Coherence Tomography for Diabetic Retinopathy.]{American Academy of Ophthalmology guidelines for use of optical coherence tomography for diabetic retinopathy, reproduced from \cite{DR-PPP}.}
\label{fig:AAO_ppp}
\end{figure}

\section{Age-related Macular Degeneration}

Age-related macular degeneration (ARMD) is a disorder of the macula, the center of the retina responsible for central visual acuity. Patient are typically 50 years of age or older. ARMD is a leading cause of severe, irreversible vision impairment in developed countries. The prevalence of ARMD varies across ethnicity and age, with increased prevalence in Caucasian and older patients. Estimates suggest that the 1.75 million individuals affected by advanced ARMD in at least one eye are expected to increase to nearly 3 million by year 2020. ARMD is divided into a ``dry'' form (about 80\% of cases) and a ``wet'' form (about 20\% of cases), which is responsible for nearly 90\% of the severe visual acuity loss. Effective treatments currently exist for the ``wet'' form using anti-VEGF therapy, as well as anti-oxidant therapy to slow the progression of the ``dry'' form.\\
\indent A rigorous classification system based on the Age-Related Eye Disease Study was created. It is based on several morphological features and the reader is referred to \cite{AMD-PPP} for details. This study investigated two particular morphological features associated with ARMD, drusen and choroidal neovascularization, discussed in further detail below.

\subsection{Choroidal Neovascularization}

\begin{figure}[htb]
\centering\includegraphics[width=0.8\linewidth]{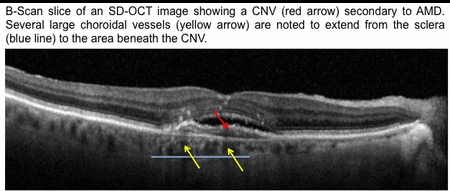}
\caption[Representative OCT of CNV.]{Representative OCT of CNV, reproduced from \url{https://iovs.arvojournals.org/article.aspx?articleid=2150855}}
\label{fig:OCT_CNV}
\end{figure}

Choroidal neovascularization (CNV) occurs when blood vessels from the choriocapillaris (the layer below the retina) perforate through Bruch's membrane (a barrier between the retinal pigment epithelium and the choriocapillaris), leading to hemorrhage underneath the retinal pigment epithelium or the retina. The hemorrhage can lead to profound vision loss that may be irreversible without prompt treatment.\\
\indent CNV can be detected by clinical examination and its presence confirmed by fluorescein angiography. It can also be detected on OCT (figure \ref{fig:OCT_CNV}). CNV is associated with the ``wet'' type of macular degeneration, but it not exclusively found in this disease as it could be seen in other retinal diseases such as high myopia and histoplasmosis. Recognition of CNV, such as with OCT, is important for diagnostic and therapeutic purposes.

\subsection{Drusen}

\begin{figure}[htb]
\centering\includegraphics[width=0.8\linewidth]{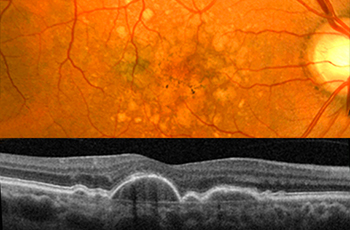}
\caption[Representative OCT of Drusen.]{Representative OCT of Drusen. Shown are a color photo (top) of the fundus (back of the eye) and optical coherence tomography (OCT) scan (bottom) in a patient with ARMD, showing multiple large drusen (lipid and fatty protein deposits). Reproduced from \url{https://www.masseyeandear.org/news/press-releases/2016/02/patients-with-high-risk-macular-degeneration-show-improvement-with-high-dose-statin-treatment}}
\label{fig:OCT_drusen}
\end{figure}

Drusen are yellow lesions at the level of the basement membrane of the retinal pigment epithelium. Though drusen are the hallmark of ARMD, they can be seen in other retinal conditions such as dominant drusen. Drusen are diagnosed by clinical examination and can be confirmed on OCT imaging (figure \ref{fig:OCT_drusen}). It is important to recognize drusen for diagnostic purposes and then to characterize them by size (small, medium, or large) and morphology (soft or hard) for prognostic purposes.

\section{Prior work on OCT-based classification}

Deep learning (DL) is a promising technology to automate classification of OCT images. Prior techniques that relied on pre-DL methodologies, such as using hand crafted segmentation, will not be reviewed here. This section briefly reviews prior DL-based work on OCT-based classification for DME, CNV and drusen (in the context of ARMD).\\
\indent Alsaih \etal evaluated machine learning techniques for DME classification on OCT images\cite{Machine_learning_DME_paper}. They used extraction of histogram of oriented gradients and local binary pattern (LBP) features within a multiresolution approach as well as principal component analysis (PCA) and bag of words (BoW) representations, and found that their best results led to a sensitivity and specificity of 87.5 and 87.5\%, respectively.\\
\indent Kamble \etal\cite{Automated_DME_analysis} finetuned Inception-Resnet-v2\cite{inception-resnet} on the publicly available data set of the Singapore Eye Research Institute (SERI). They reported that they achieved 100\% classification accuracy (on DME images from OCT) with the Inception-Resnet-v2 model using a leave-one-out crossvalidation strategy.\\
\indent Venhuizen \etal evaluated a machine learning algorithm that automatically grades ARMD severity stages from OCT scans\cite{automated_staging}. Their algorithm was built around the Bag of Words (BoW) approach. They report that their system achieved an area under the ROC curve of 0.980 with a sensitivity of 98.2\% at a specificity of 91.2\%. They state that this compares favorably with the performance of human observers who achieved sensitivities of 97.0\% and 99.4\% at specificities of 89.7\% and 87.2\%, respectively. Moreover, they report a good level of agreement with the reference was obtained ($\kappa=$ 0.713) and was in concordance with the human observers ($\kappa=$ 0.775 and $\kappa=$ 0.755, respectively).\\
\indent A landmark study by Kermany \etal\cite{OCT_paper} utilized transfer learning applied to the largest ($\sim$100K images) publically available dataset for classification of DME, CNV, and drusen. They report using Tensorflow adapted with an Inception V3 architecture pretrained on the ImageNet dataset, and the convolutional layers were frozen and used as fixed feature extractors.  In a multi-class comparison between CNV, DME, drusen, and normal, they report that they achieved an accuracy of 96.6\%, with a sensitivity of 97.8\%, a specificity of 97.4\%, and a weighted error of 6.6\%.

\chapter{Methods\label{chap:part3_methods}}

\begin{figure}[htb]
  \centering\includegraphics[width=1.0\linewidth]{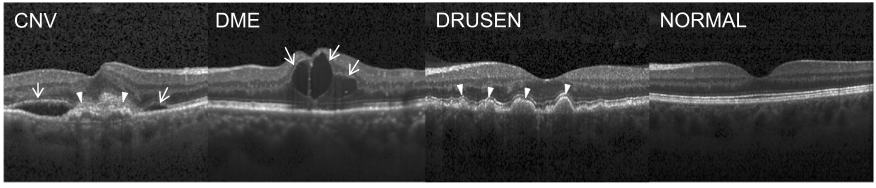}
\caption[Representative images from the OCT classification dataset.]{Representative images from the OCT classification dataset. (Far left) choroidal neovascularization (CNV) with neovascular membrane (white arrowheads) and associated subretinal fluid (arrows). (Middle left) Diabetic macular edema (DME) with retinal-thickening-associated intraretinal fluid (arrows). (Middle right) Multiple drusen (arrowheads) present in early ARMD. (Far right) Normal retina with preserved foveal contour and absence of any retinal fluid/edema. Reproduced from \cite{OCT_paper}.}
\label{fig:OCT_class_examples}
\end{figure}

Methodology similar to that in part \ref{part:two} was used in this part of the report. The dataset used in this part of the report was obtained from \cite{OCT_paper}.\footnote{Available at \url{https://data.mendeley.com/datasets/rscbjbr9sj/2}} This dataset consists of 109,309 OCT JPEG images, taken through the macular center, from patients with: no retinal disease (Class 0), DME (Class 1), CNV (Class 2), and drusen (Class 3). The dataset provider partitioned the images into 108,309 for training and 1,000 for testing. In this report, the author randomly selected 10,831(10\%) of the 108,309 images for a validation dataset, leaving 97,478(90\%) for training. The distribution of the dataset, broken down by training, validation, and test data, and by classification category is listed in table \ref{tab:OCT_dist}. Examples of each class are shown in figure \ref{fig:OCT_class_examples}. 

\begin{table}[htb]
\caption[OCT classification dataset: distribution of labels across training, validation, and test sets.]{Distribution of labels across training, validation, and test sets for the OCT classification dataset.}\label{tab:OCT_dist}
  \begin{center}
\scalebox{0.8}{
  \begin{tabular}{|l|r|r|r|r|r|}
        \hline
   & \textbf{Class 0} & \textbf{Class 1} & \textbf{Class 2} & \textbf{Class 3}  & \textbf{Total}\\
   \hline
   Training & 46,026(47.2\%) & 10,213(10.5\%) & 33,485(34.3\%) & 7,754(8.0\%) & 97,478 \\
   Validation & 5,114(47.2\%) & 1,135(10.5\%) & 3,720(34.3\%) & 862(8.0) & 10,831\\
   Test & 250(25\%) & 250(25\%) & 250(25\%) & 250(25\%) & 1,000\\ 
  \hline
\end{tabular}}
  \end{center}
\end{table}

\indent In this part of the report, the following networks were selected for baseline analysis: AlexNet, DensetNet-121, Inception\_v3, Resnet-18, and VGG-11. Each network was run pre-trained, in fixed-feature extractor and fine-tuning mode (see \ref{sec:models} for details about these configurations). The same settings as in \ref{sec:models} were used as far as data augmentation, image normalization, SGD learning rate and momentum, learning rate scheduler, and cross entropy loss. No image preprocessing was done other than resizing images to the expected input size required for each respective model. The networks were all run for 25 epochs. To investigate the effect of adding layers, DenseNet-161, ResNet-34, and VGG-19 were run with the same settings, in fixed-feature extractor and fine-tuning modes, for 25 epochs. The networks were run on the same previously described computing environment on a GPU.\\
\indent Network loss and accuracy over the epochs were tabulated and plotted. Each model was evaluated with the test data using weights established at the last epoch. As in part \ref{part:two}, accuracy and quadratic weighted kappa were reported.

\chapter{Results\label{chap:part3_results}}

Tables \ref{tab:OCT_loss_results} and \ref{tab:OCT_accuracy_results} summarize the training and validation loss and accuracy for each network, run in fixed-feature extractor and fine-tuning modes. Loss and accuracy plots for 5 networks (AlexNet, DenseNet-121, Inception\_v3, ResNet-18, and VGG-11) are plotted in figures \ref{fig:networks_loss_plots} and \ref{fig:networks_acc_plots}, respectively. Several observations are made:\\
\\
\noindent (1) The training and valiation losses were much lower in fine-tuning vs. fixed feature mode. In particular, the validation loss across the networks was on order 10 times less for fine-tuning compared to fixed-feature (mean loss 0.0682 vs. 0.4453)\\
\\
\noindent (2) Within the fixed-feature extractor and fine-tuning modes, the loss was generally similar across the networks (table \ref{tab:OCT_loss_results}, standard deviation results).\\
\\
\noindent (3) The training and validation accuracies were signficantly higher for fine-tuning mode vs. fixed-feature extractor. In particular the validation accuracy across the networks was much higher for fine-tuning (0.9778) vs. fixed-feature extractor (0.8509).\\
\\
\noindent (4) Within the fixed-feature extractor and fine-tuning modes, the accuracy was generally similar across the networks (table \ref{tab:OCT_accuracy_results}, standard deviation results). In particular, for fine-tuning mode, all the networks had $>$97\% validation accuracy.\\
\\
\noindent (5) Adding layers to the DenseNet, ResNet, and VGG networks (via DenseNet-161, ResNet-34, and VGG-19, respectively) lowered the loss slightly in all cases. This also increased the accuracies in all cases by very small amounts.

\begin{table}[htb!]
\caption[OCT-based Classification Network loss results]{\textbf{OCT-based Classification Network loss results.} Shown are the loss values for each network at the last epoch. Values are grouped by whether the network was used in fine-tuning or fixed feature extractor mode. All networks were trained for 25 epochs.}\label{tab:OCT_loss_results}
  \begin{center}
\scalebox{0.8}{
  \begin{tabular}{|l|r|r|r|r|}
\toprule
& \multicolumn{2}{c}{\underline{\textbf{Fixed-Feature}}} & \multicolumn{2}{c}{\underline{\textbf{Fine-Tuning}}}\\
\textbf{Network} & \textbf{Training} & \textbf{Validation} & \textbf{Training} & \textbf{Validation}\\
\midrule
AlexNet & 0.6513 & 0.4168 & 0.2084 & 0.0921\\
DenseNet-121 & 0.6666 & 0.4937 & 0.1514 & 0.0665\\
DenseNet-161 & 0.6308 & 0.3513 & 0.1382 & 0.0599\\
Inception & 0.7957 & 0.5926 & 0.1387 & 0.0595\\
ResNet-18 & 0.7212 & 0.4403 & 0.1666 & 0.0698\\
ResNet-34 & 0.7034 & 0.4386 & 0.1532 & 0.0665\\
VGG-11 & 0.6348 & 0.4156 & 0.1572 & 0.0679\\
VGG-19 & 0.6283 & 0.4134 & 0.1504 & 0.0630\\
\hline
\hline
Mean & 0.6790 & 0.4453 & 0.1508 & 0.0682\\ 
Stddev & 0.0545 & 0.0668 & 0.0093 & 0.0097\\
\bottomrule
\end{tabular}}
\end{center}
\end{table}

\begin{table}[htb!]
\caption[OCT-based Classification Network accuracy results]{\textbf{OCT-based Classification Network accuracy results.} Shown are the accuracy values for each network at the last epoch. Values are grouped by whether the network was used in fine-tuning or fixed feature extractor mode. All networks were trained for 25 epochs.}\label{tab:OCT_accuracy_results}
  \begin{center}
\scalebox{0.8}{
  \begin{tabular}{|l|r|r|r|r|}
\toprule
& \multicolumn{2}{c}{\underline{\textbf{Fixed-Feature}}} & \multicolumn{2}{c}{\underline{\textbf{Fine-Tuning}}}\\
\textbf{Network} & \textbf{Training} & \textbf{Validation} & \textbf{Training} & \textbf{Validation}\\
\midrule
AlexNet & 0.7716 & 0.8554 & 0.9271 & 0.9716\\
DenseNet-121 & 0.7529 & 0.8624 & 0.9471 & 0.9788\\
DenseNet-161 & 0.7671 & 0.8875 & 0.9518 & 0.9804\\
Inception & 0.6995 & 0.8003 & 0.9516 & 0.9797\\
ResNet-18 & 0.7318 & 0.8464 & 0.9416 & 0.9765\\
ResNet-34 & 0.7393 & 0.8519 & 0.9459 & 0.9779\\
VGG-11 & 0.7706 & 0.8491 & 0.9447 & 0.9776\\
VGG-19 & 0.7752 & 0.8545 & 0.9475 & 0.9795\\
\hline
\hline
Mean & 0.7510 & 0.8509 & 0.9447 & 0.9778\\
Stddev & 0.245 & 0.0226 & 0.0074 & 0.0026\\
\bottomrule
\end{tabular}}
\end{center}
\end{table}

\begin{figure}[htb]
\begin{subfigure}{0.8\textwidth}
\captionsetup{width=0.8\textwidth}
\centering\includegraphics[width=0.8\linewidth]{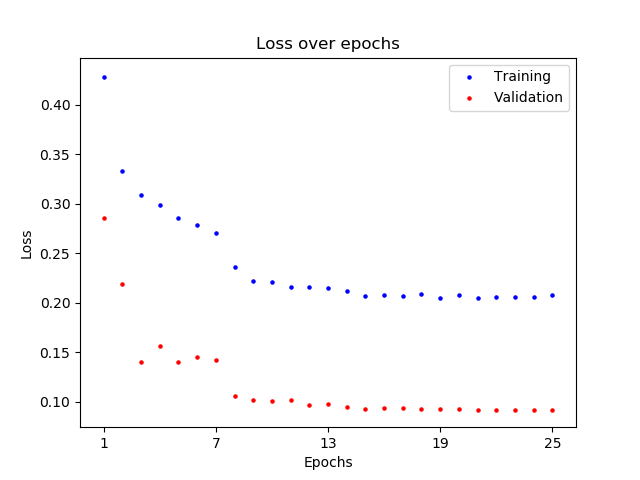}
\caption{AlexNet, fine-tuning.}
\label{fig:alexnet_loss1}
\end{subfigure}
\begin{subfigure}{0.8\textwidth}
\captionsetup{width=0.8\textwidth}
\centering\includegraphics[width=0.8\linewidth]{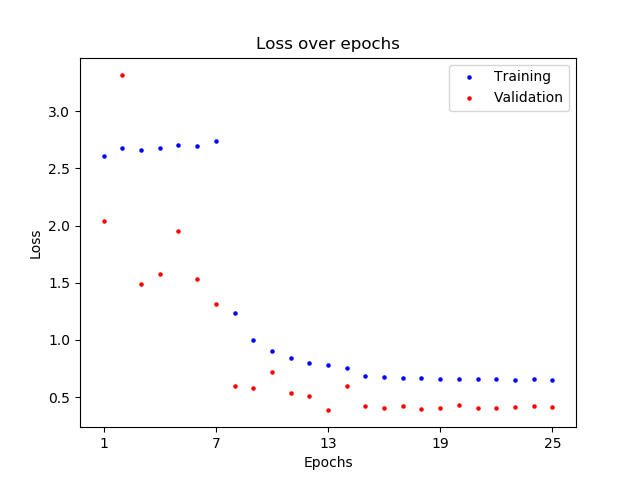}
\caption{AlexNet, feature-extractor.}
\label{fig:alexnet_loss2}
\end{subfigure}
\caption[Network loss plots.]{Plotted are the network training and validation loss over the epochs. All networks were run pre-trained for 25 epochs.}\label{fig:networks_loss_plots}
\end{figure}

\begin{figure}
\ContinuedFloat
\begin{subfigure}{0.8\textwidth}
\captionsetup{width=0.8\textwidth}
\centering\includegraphics[width=0.8\linewidth]{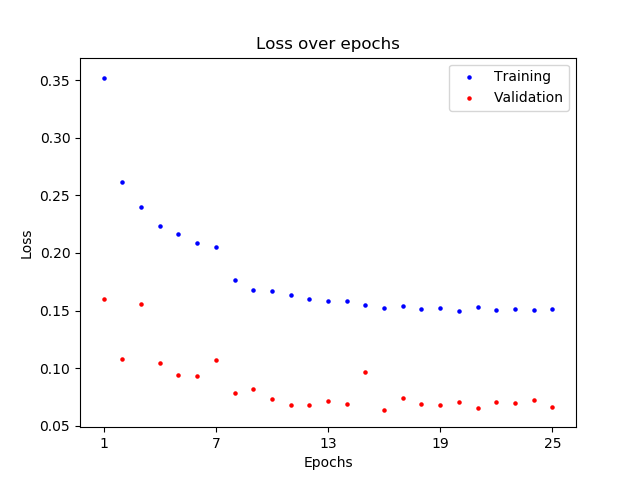}
\caption{DenseNet121, fine-tuning.}
\label{fig:densenet121_loss1}
\end{subfigure}
\begin{subfigure}{0.8\textwidth}
\captionsetup{width=0.8\textwidth}
\centering\includegraphics[width=0.8\linewidth]{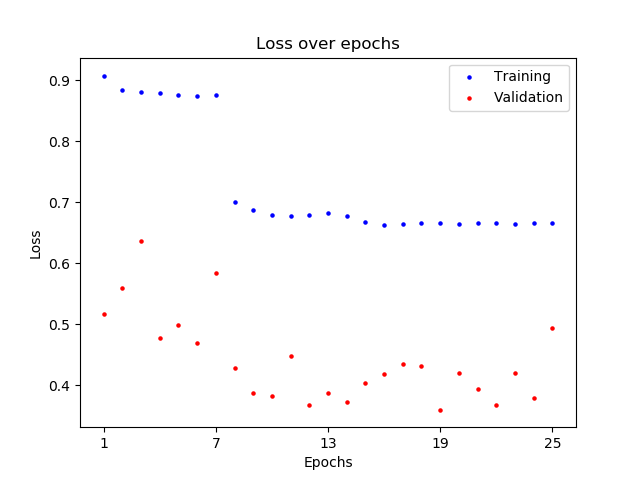}
\caption{DenseNet121, feature-extractor.}
\label{fig:alexnet_loss2}
\end{subfigure}
\caption[Network loss plots - Continued.]{Plotted are the network training and validation loss over the epochs. All networks were run pre-trained for 25 epochs.}\label{fig:networks_loss_plots2}
\end{figure}

\begin{figure}
\ContinuedFloat
\begin{subfigure}{0.8\textwidth}
\captionsetup{width=0.8\textwidth}
\centering\includegraphics[width=0.8\linewidth]{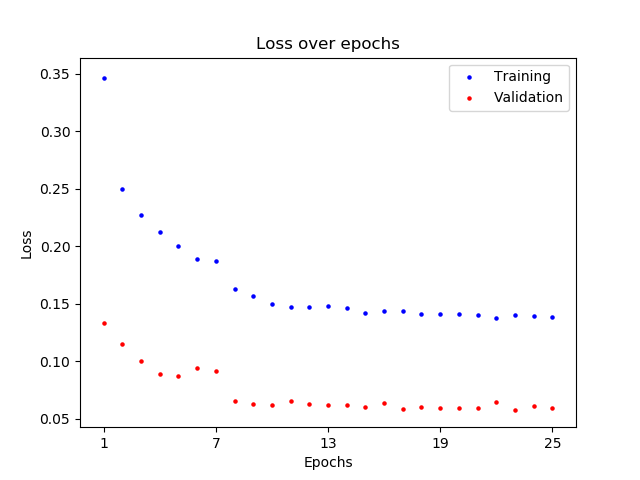}
\caption{Inception\_v3, fine-tuning.}
\label{fig:inception_loss1}
\end{subfigure}
\begin{subfigure}{0.8\textwidth}
\captionsetup{width=0.8\textwidth}
\centering\includegraphics[width=0.8\linewidth]{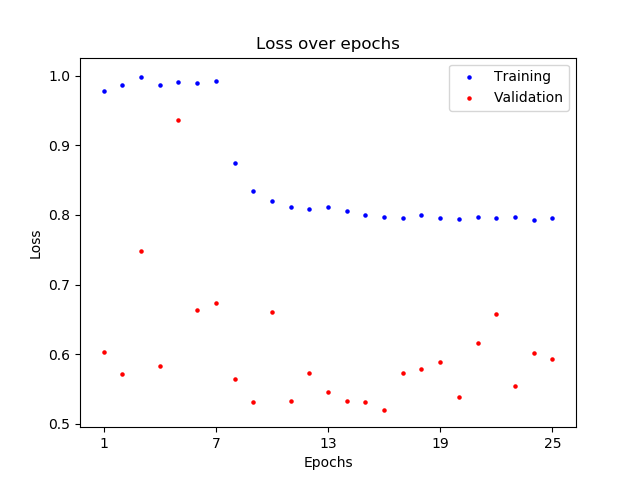}
\caption{Inception\_v3, feature-extractor.}
\label{fig:inception_loss2}
\end{subfigure}
\caption[Network loss plots - Continued.]{Plotted are the network training and validation loss over the epochs. All networks were run pre-trained for 25 epochs.}\label{fig:networks_loss_plots3}
\end{figure}

\begin{figure}
\ContinuedFloat
\begin{subfigure}{0.8\textwidth}
\captionsetup{width=0.8\textwidth}
\centering\includegraphics[width=0.8\linewidth]{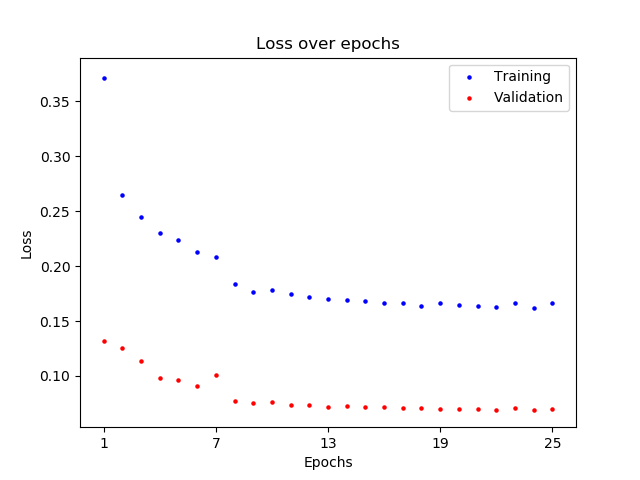}
\caption{ResNet-18, fine-tuning.}
\label{fig:resnet18_loss1}
\end{subfigure}
\begin{subfigure}{0.8\textwidth}
\captionsetup{width=0.8\textwidth}
\centering\includegraphics[width=0.8\linewidth]{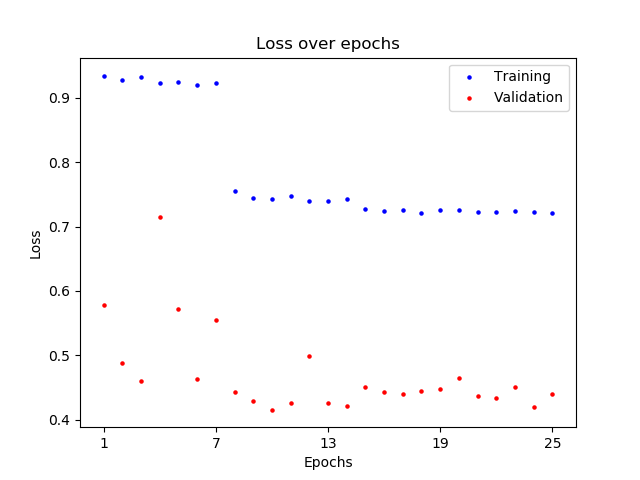}
\caption{ResNet-18, feature-extractor.}
\label{fig:resnet18_loss2}
\end{subfigure}
\caption[Network loss plots - Continued.]{Plotted are the network training and validation loss over the epochs. All networks were run pre-trained for 25 epochs.}\label{fig:networks_loss_plots3}
\end{figure}

\begin{figure}
\ContinuedFloat
\begin{subfigure}{0.8\textwidth}
\captionsetup{width=0.8\textwidth}
\centering\includegraphics[width=0.8\linewidth]{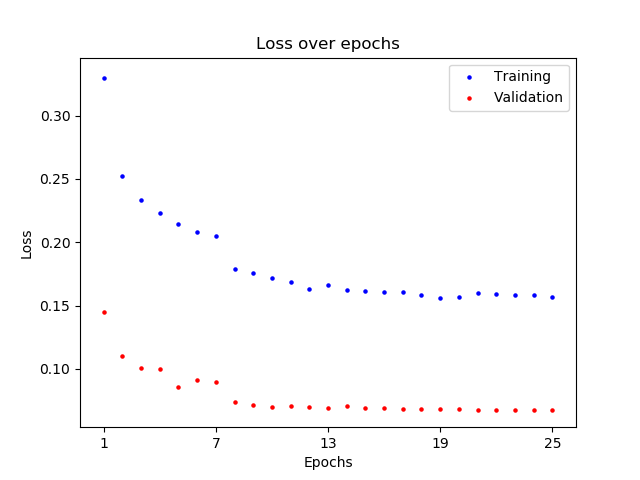}
\caption{VGG-11, fine-tuning.}
\label{fig:vgg11_loss1}
\end{subfigure}
\begin{subfigure}{0.8\textwidth}
\captionsetup{width=0.8\textwidth}
\centering\includegraphics[width=0.8\linewidth]{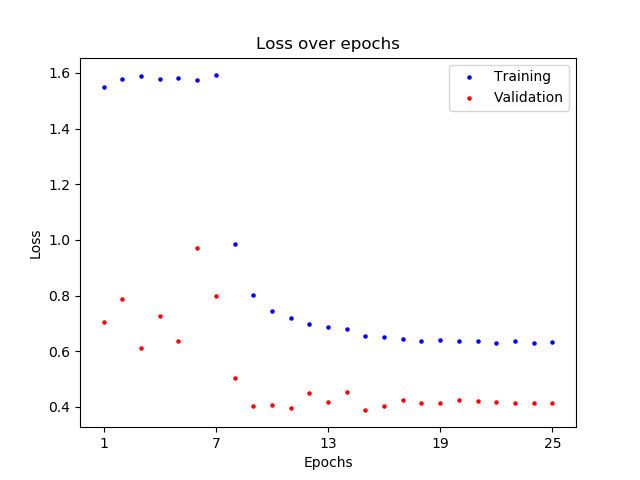}
\caption{VGG-11, feature-extractor.}
\label{fig:vgg11_loss2}
\end{subfigure}
\caption[Network loss plots - Continued.]{Plotted are the network training and validation loss over the epochs. All networks were run pre-trained for 25 epochs.}\label{fig:networks_loss_plots4}
\end{figure}

\begin{figure}[htb]
\begin{subfigure}{0.8\textwidth}
\captionsetup{width=0.8\textwidth}
\centering\includegraphics[width=0.8\linewidth]{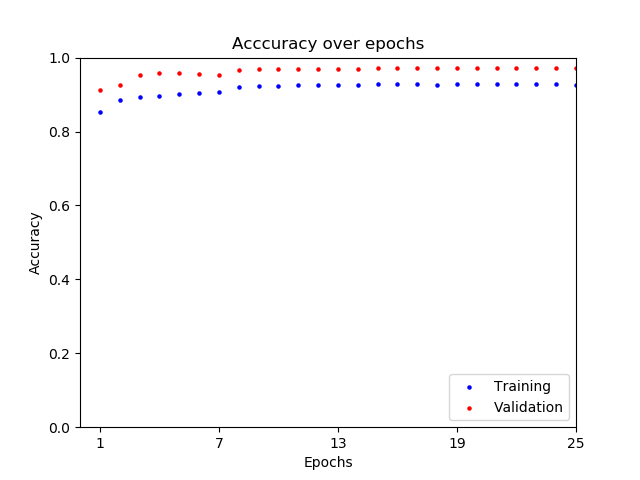}
\caption{AlexNet, fine-tuning.}
\label{fig:alexnet_acc1}
\end{subfigure}
\begin{subfigure}{0.8\textwidth}
\captionsetup{width=0.8\textwidth}
\centering\includegraphics[width=0.8\linewidth]{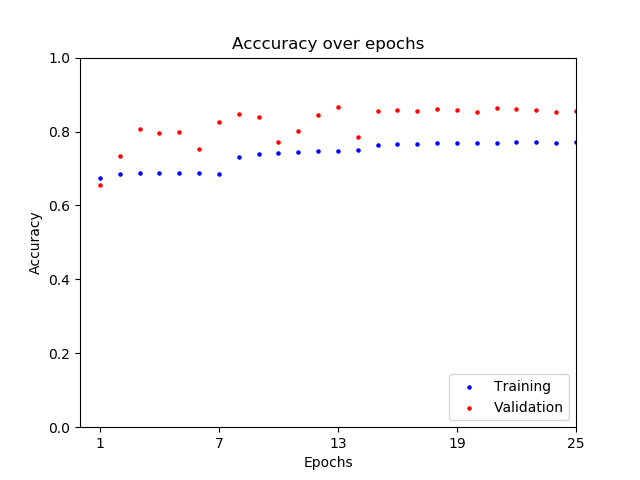}
\caption{AlexNet, feature-extractor.}
\label{fig:alexnet_acc2}
\end{subfigure}
\caption[Network accuracy plots.]{Plotted are the network training and validation accuracy over the epochs. All networks were run pre-trained for 25 epochs.}\label{fig:networks_acc_plots}
\end{figure}

\begin{figure}
\ContinuedFloat
\begin{subfigure}{0.8\textwidth}
\captionsetup{width=0.8\textwidth}
\centering\includegraphics[width=0.8\linewidth]{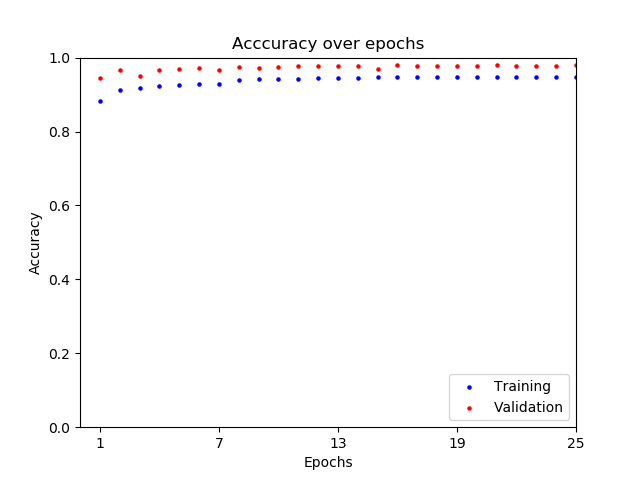}
\caption{DenseNet121, fine-tuning.}
\label{fig:densenet121_acc1}
\end{subfigure}
\begin{subfigure}{0.8\textwidth}
\captionsetup{width=0.8\textwidth}
\centering\includegraphics[width=0.8\linewidth]{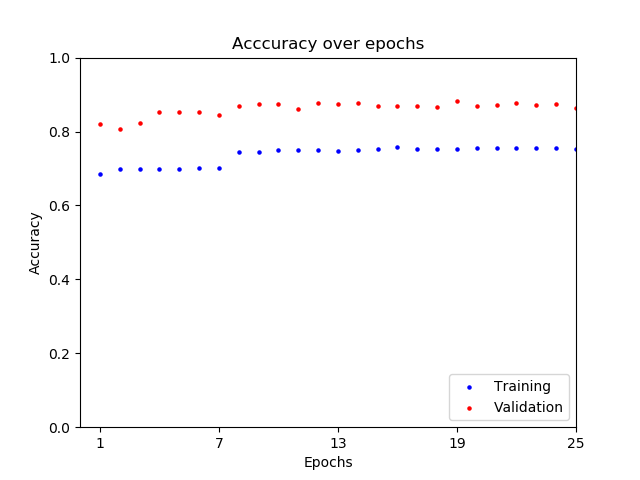}
\caption{DenseNet121, feature-extractor.}
\label{fig:densenet121_acc2}
\end{subfigure}
\caption[Network accuracy plots - Continued.]{Plotted are the network training and validation accuracy over the epochs. All networks were run pre-trained for 25 epochs.}\label{fig:networks_acc_plots2}
\end{figure}

\begin{figure}
\ContinuedFloat
\begin{subfigure}{0.8\textwidth}
\captionsetup{width=0.8\textwidth}
\centering\includegraphics[width=0.8\linewidth]{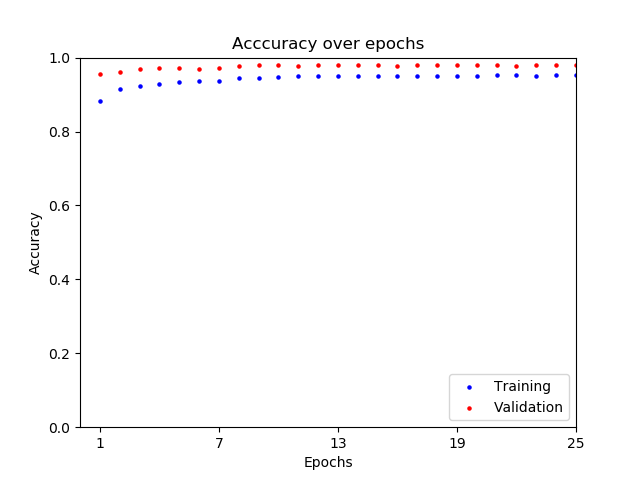}
\caption{Inception\_v3, fine-tuning.}
\label{fig:inception_acc1}
\end{subfigure}
\begin{subfigure}{0.8\textwidth}
\captionsetup{width=0.8\textwidth}
\centering\includegraphics[width=0.8\linewidth]{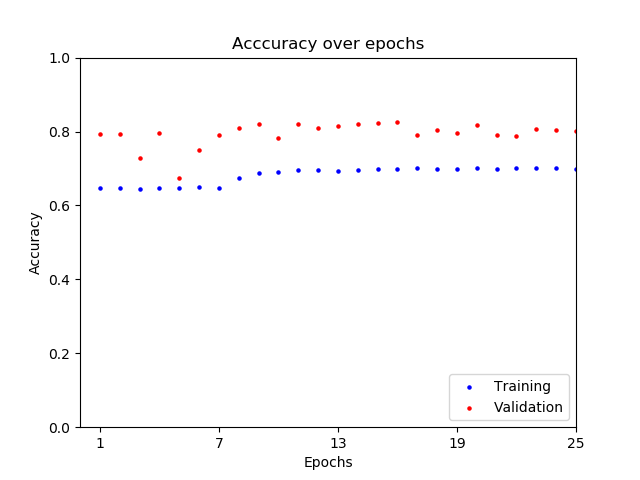}
\caption{Inception\_v3, feature-extractor.}
\label{fig:inception_acc2}
\end{subfigure}
\caption[Network accuracy plots - Continued.]{Plotted are the network training and validation accuracy over the epochs. All networks were run pre-trained for 25 epochs.}\label{fig:networks_acc_plots3}
\end{figure}

\begin{figure}
\ContinuedFloat
\begin{subfigure}{0.8\textwidth}
\captionsetup{width=0.8\textwidth}
\centering\includegraphics[width=0.8\linewidth]{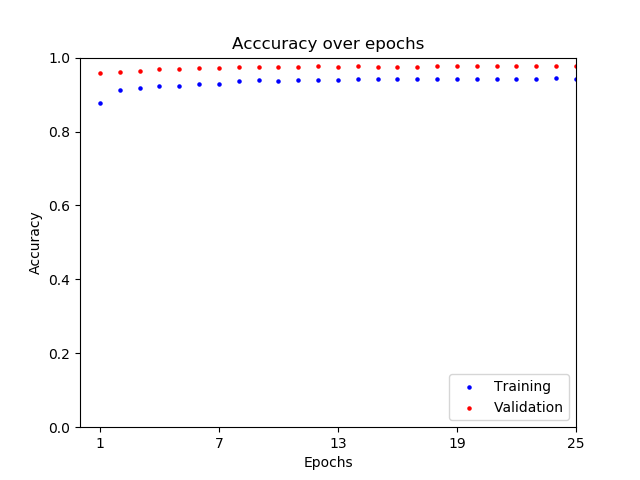}
\caption{ResNet-18, fine-tuning.}
\label{fig:resnet18_acc1}
\end{subfigure}
\begin{subfigure}{0.8\textwidth}
\captionsetup{width=0.8\textwidth}
\centering\includegraphics[width=0.8\linewidth]{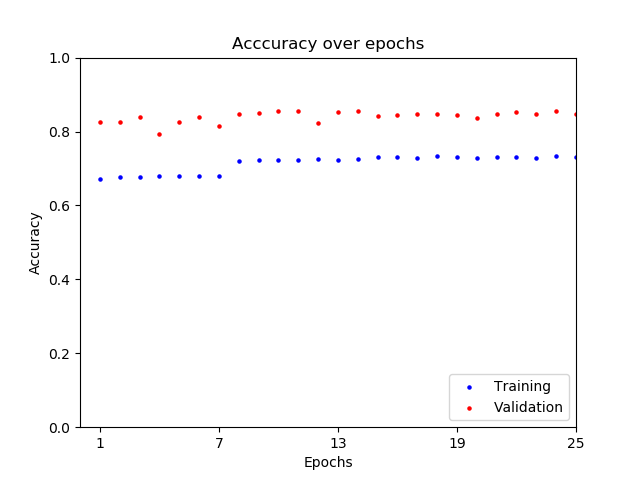}
\caption{ResNet-18, feature-extractor.}
\label{fig:vgg11_acc2}
\end{subfigure}
\caption[Network accuracy plots - Continued.]{Plotted are the network training and validation accuracy over the epochs. All networks were run pre-trained for 25 epochs.}\label{fig:networks_acc_plots3}
\end{figure}
\begin{figure}

\ContinuedFloat
\begin{subfigure}{0.8\textwidth}
\captionsetup{width=0.8\textwidth}
\centering\includegraphics[width=0.8\linewidth]{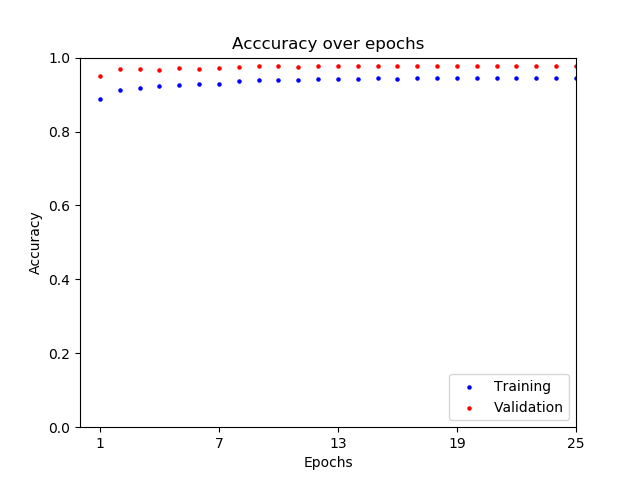}
\caption{VGG-11, fine-tuning.}
\label{fig:vgg11_acc1}
\end{subfigure}
\begin{subfigure}{0.8\textwidth}
\captionsetup{width=0.8\textwidth}
\centering\includegraphics[width=0.8\linewidth]{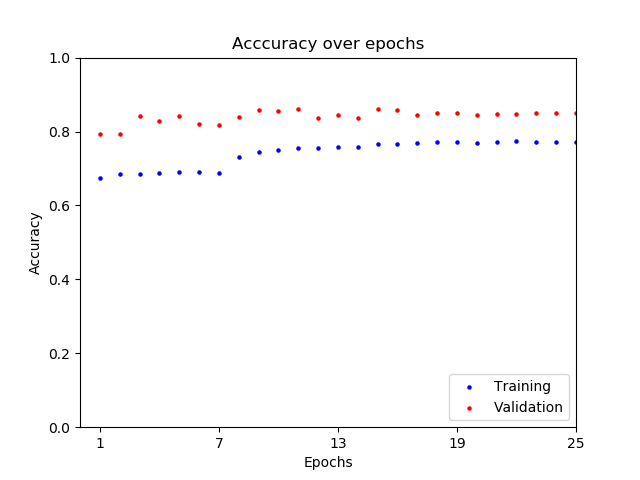}
\caption{VGG-11, feature-extractor.}
\label{fig:vgg11_acc2}
\end{subfigure}
\caption[Network accuracy plots - Continued.]{Plotted are the network training and validation accuracy over the epochs. All networks were run pre-trained for 25 epochs.}\label{fig:networks_acc_plots4}
\end{figure}

\begin{figure}[htb]
\begin{subfigure}{0.6\textwidth}
\captionsetup{width=0.6\textwidth}
\centering\includegraphics[width=0.6\linewidth]{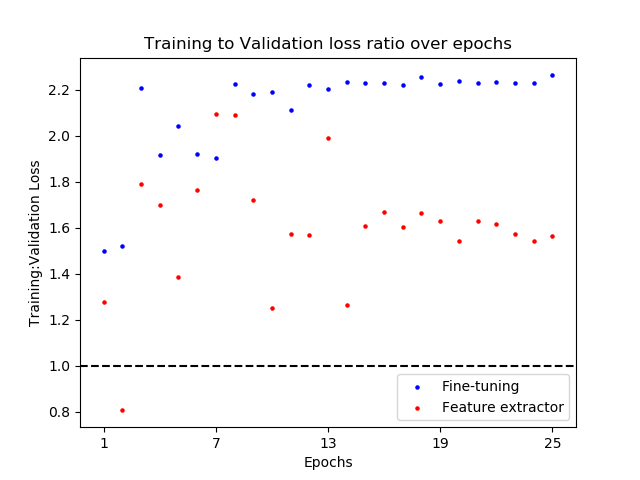}
\caption{AlexNet.}
\label{fig:alexnet_ratio}
\end{subfigure}
\begin{subfigure}{0.6\textwidth}
\captionsetup{width=0.6\textwidth}
\centering\includegraphics[width=0.6\linewidth]{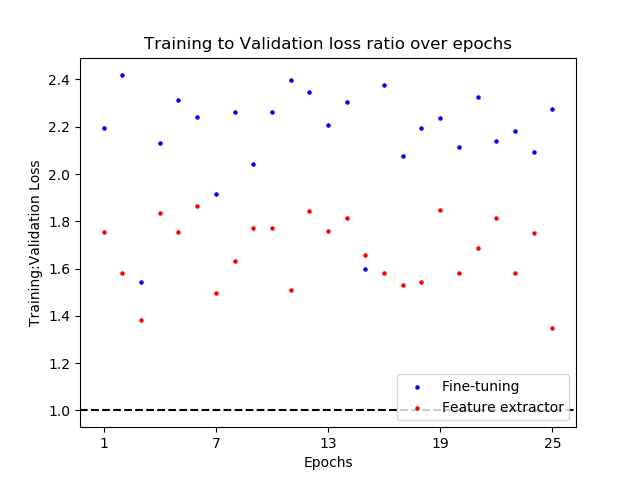}
\caption{DenseNet121.}
\label{fig:densenet121_ratio}
\end{subfigure}
\begin{subfigure}{0.6\textwidth}
\captionsetup{width=0.6\textwidth}
\centering\includegraphics[width=0.6\linewidth]{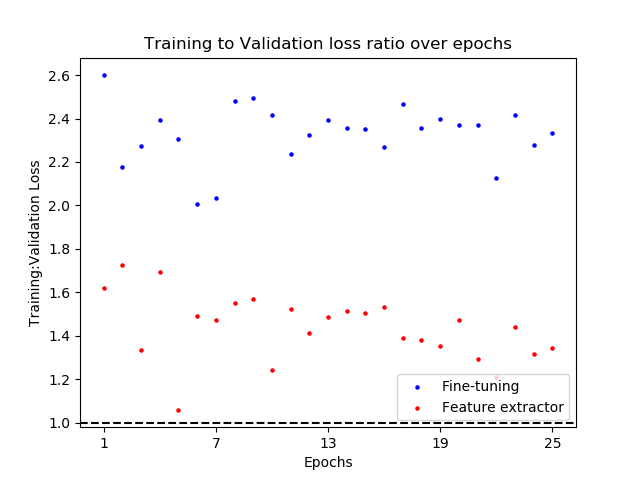}
\caption{Inception\_v3.}
\label{fig:inception_ratio}
\end{subfigure}
\begin{subfigure}{0.6\textwidth}
\captionsetup{width=0.6\textwidth}
\centering\includegraphics[width=0.6\linewidth]{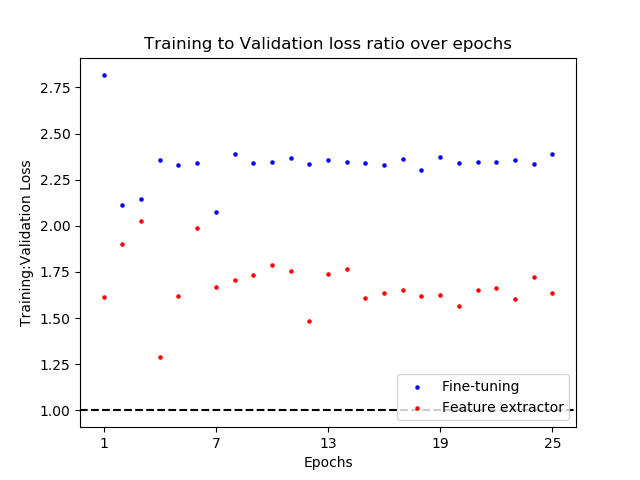}
\caption{ResNet-18.}
\label{fig:resnet18_ratio}
\end{subfigure}
\begin{subfigure}{0.6\textwidth}
\captionsetup{width=0.6\textwidth}
\centering\includegraphics[width=0.6\linewidth]{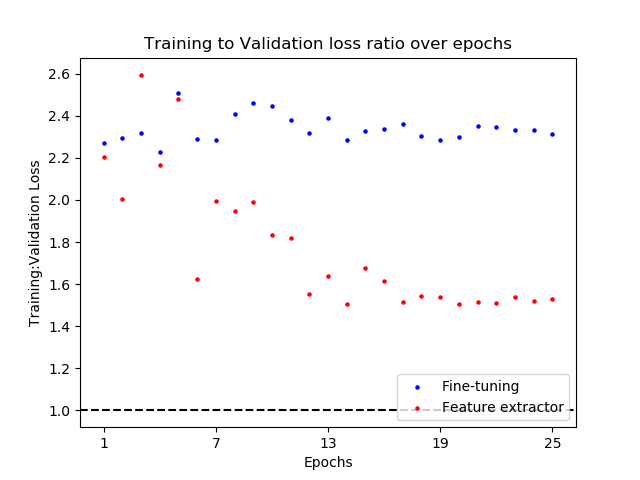}
\caption{VGG-11.}
\label{fig:vgg11_ratio}
\end{subfigure}
\caption[Network training to validation ratio plots.]{Plotted are the network training to validation ratios. All networks were run pre-trained for 25 epochs.}\label{fig:networks_ratio_boxplots}
\end{figure}

\indent Figure \ref{fig:networks_ratio_boxplots} shows the network training to validation ratios. Except for a single outlier for AlexNet early in training, all ratios were $>$1. In general, the ratio for fine-tuning was greater than feature-extractor as the number of epochs progressed. This implies that the networks were not over-fitting and the fitting profiles for fine-tuning mode were better.

\indent Lastly, the test dataset accuracy and quadratic weighted kappa results are shown in tables \ref{tab:oct_test_accuracy} and \ref{tab:oct_test_kappa}, respectively. The accuracy for fine-tuning mode was much higher than fixed-feature extractor (mean accuracy 95.1\% for fine-tuning vs. 68.4\% for fixed-feature extractor). Similarly, the quadratic weighted kappa was much higher for fine-tuning compared to fixed-feature extractor (mean quadratic weighted kappa 0.969 vs. 0.534).

\begin{table}[htb!]
\caption[OCT Classification Network Test Accuracy]{\textbf{OCT Classification Network Test Accuracy}. Accuracy(\%) is indicated for the specified networks (all pre-trained) for fixed feature extractor and fine tuning modes.}\label{tab:oct_test_accuracy}
  \begin{center}
\scalebox{0.8}{
  \begin{tabular}{|l|r|r|}
\hline
\textbf{Network} & \textbf{Fixed Feature Extractor} & \textbf{Fine-Tuning}\\
\hline
AlexNet & 73.4 & 92.2\\
DenseNet-121 & 76.8 & 96.2\\
DenseNet-161 & 79.1 & 96.4\\
Inception-v3 & 25.0 & 92.4\\
ResNet-18 & 73.8 & 95.3\\
ResNet-34 & 74.5 & 95.8\\
VGG-11 & 69.6 & 95.5\\
VGG-19 & 75.1 & 96.9\\
\hline\hline
Mean & 68.4 & 95.1\\
Stdev & 16.6 & 1.7\\
\hline
\end{tabular}}
\end{center}
\end{table}

\begin{table}[htb!]
\caption[OCT Classification Network Test Quadratic Weighted Kappa]{\textbf{OCT Classification Network Test Quadratic Weighted Kappa}. Quadratic weighted kappa is indicated for the specified networks (all pre-trained) for fixed feature extractor and fine tuning modes. A generally agreed upon scale is: $<$0.20(Poor), 0.21-0.40(Fair), 0.41-0.60(Moderate), 0.61-0.80(Good), and 0.81-1.00(Very good).}\label{tab:oct_test_kappa}
  \begin{center}
\scalebox{0.8}{
  \begin{tabular}{|l|r|r|}
\hline
\textbf{Network} & \textbf{Fixed Feature Extractor} & \textbf{Fine-Tuning}\\
\hline
AlexNet & 0.517 & 0.944\\
DenseNet-121 & 0.647 & 0.983\\
DenseNet-161 & 0.676 & 0.976\\
Inception-v3 & 0.000 & 0.951\\
ResNet-18 & 0.607 & 0.968\\
ResNet-34 & 0.591 & 0.976\\
VGG-11 & 0.582 & 0.969\\
VGG-19 & 0.655 & 0.984\\
\hline\hline
Mean & 0.534 & 0.969\\
Stdev & 0.207 & 0.014\\
\hline
\end{tabular}}
\end{center}
\end{table}

\chapter{Discussion\label{chap:part3_discussion}}

This part of the report analyzed the performance of transfer learning for an inter-disease classification scheme of normal (no retinal disease), DME, CNV, and drusen. In contrast, part \ref{part:two} analyzed transfer learning for an intra-disease classification scheme, specifically cases of no diabetic retinopathy vs. increasing severity of diabetic retinopathy.\\
\indent While a previous report \cite{OCT_paper} analyzed a single network (Inception\_v3), to the best of my knowledge, no prior work has analyzed a suite of networks. Several key findings were observed in this report:\\
\\
\noindent (1) The validation accuracies at the end of training were significantly higher when the networks were configured in fine-tuning compared to fixed-feature extractor mode. On average, across the networks evaluated (table \ref{tab:OCT_accuracy_results}), the validation accuracy for fine-tuning was 0.9778 compared to 0.8509 for fixed-feature extractor.\\
\\
\noindent (2) All the networks did very well, with validation accuracies $>$97\% for fine-tuning mode. Adding layers to networks (specifically DenseNet, ResNet, and VGG) did not significantly improve performance.\\
\\
\noindent (3) The mean accuracy for all the networks using the test dataset was much higher (95.1\%) in fine-tuning mode compared to fixed-feature extractor (68.4\%). A similar finding occured with the quadratic weighted kappa, where fine-tuning had a kappa of 0.969 (very good) while fixed-feature extractor had a kappa of 0.534 (moderate). The accuracy and quadratic weighted kappa for fixed-feature extractor were brought down by low values for Inception\_v3, however, similar findings were observed when this network was excluded.\\

\indent Interestingly, the best test accuracy (fine-tuning, 96.9\%) of this inter-disease classification scheme was much higher than the accuracy found in the intra-disease classification scheme (fine-tuning, 77.5\%) for diabetic retinopathy detection in part \ref{part:two}. This is expected, since the features that distinguish the classes in the inter-disease scheme (normal, DME, CNV, and drusen) are much different and richer than the more subtle features that distinguish cases of no diabetic retinopathy compared to mild diabetic retinopathy, or mild diabetic retinopathy compared to moderate diabetic retinopathy.\\
\indent To investigate sources of error, confusion matrices were constructed (appendix \ref{chap:append_CM}). Certain patterns were observed:\\
\\
\noindent (1) In fine-tuning mode, the networks generally did very well classifying normal, DME, and CNV test cases. The sources of error mostly came from misclassifcation of drusen test cases. The majority of these errors occured when the the network misclassified the test case as CNV and ground truth was drusen. Presumably, there were morphologic characteristics that confused the networks into classifying these cases as CNV and further research is beneficial.\\
\\
\noindent (2) In general, the majority of ground truth test cases of DME and CNV that were misclassified by the networks in fixed-feature extractor mode were correctly classified in fine-tuning mode. Although the misclassification of drusen substantially improved from fixed-feature extractor to fine-tuning, there was still a small portion of drusen test cases that were misclassified. For example, consider VGG-11: in fixed-feature extractor mode 182 (72.8\%) of 250 drusen test cases were misclassified, which improved to 33 (13.2\%) misclassifications with fine-tuning.\\
\\
\noindent (3) Adding layers did not significantly improve the misclassification rates, except for VGG-11 to -19 with fixed-feature extractor, where there was a significant decrease in misclassifications of DME and drusen.\\

\indent Inception\_v3 was examined in more detail (figure \ref{fig:inception_CM_a} and \ref{fig:inception_CM_b}) because it showed behaviour in fixed-feature extractor that deviated from the other networks. The findings showed that: (1) for fixed-feature extractor, the network classification broke down, whereby all the test cases were classified as CNV (Class 2). It is unclear why this occured only for this particular network; (2) for fine-tuning, this network correctly identified all normal and CNV cases, but misclassified 8\% of DME cases as CNV, 2\% of drusen cases as normal, and 20.4\% of drusen cases as CNV. The network was re-run in fixed-feature mode and showed consistent results of 24.5\% test accuracy  and 0.00 quadratic weighted kappa on the second run.\\
\indent Kermany \etal\cite{OCT_paper} described ``results based adapting Inception\_v3 pretrained on the ImageNet dataset. Retraining consisted of initializing the convolutional layers with loaded pretrained weights and retraining the final, softmax layer to recognize our classes from scratch. In this study, the convolutional layers were frozen and used as fixed feature extractors.'' Moreover, they report that ``attempts at ‘‘fine-tuning’’ the convolutional layers by unfreezing and updating the pretrained weights on our medical images using backpropagation tended to decrease model performance due to overfitting.'' However, the results described in this report support the opposite conclusion, in that fine-tuning Inception\_v3 yielded much higher accuracies, while the training to validation ratio plots did not imply overfitting. It should be noted that Kermany \etal used Tensorflow while this study used PyTorch. SGD with the same learning rate were used in both studies.\\
\indent It should be noted as well that Kamble \etal\cite{Automated_DME_analysis} used Inception-Resnet-v2 fine-tuned in their study and reported 100\% classification accuracy. They note that their developed model was also compared to other fine tuned models, such as Resnet-50 and Inception-v3. It is unclear why the implications of fine-tuning and fixed-feature extractor differed between this thesis and Kermany \etal Further study is warranted.\\
\indent Lastly, several study limitations should be noted:\\
\\
\noindent (1) The dataset provider allocated a relatively small number of test cases (1,000 in total, with 250 cases from each class). Further evaluation with larger test datasets, especially with varying degrees of image quality, would be insightful.\\
\\
\noindent (2) The images were derived from one type of instrument (Spectralis OCT, Heidelberg Engineering, Germany). Further study with other OCT vendor cameras would be beneficial.\\
\\
\noindent (3) The objective of the second part of this study was to evaluate transfer learning for classification of DME, CNV, and drusen by OCT, rather than classification of DME and ARMD by OCT. The rationale for this was two-fold: (a) CNV and drusen, although associated with ARMD, can be seen in other retinal conditions besides ARMD and it unclear from the methodology of Kermany \etal\cite{OCT_paper} if all cases were from ARMD; and (b) the AREDS classification of ARMD recognizes other morphologic features besides drusen and CNV, such as geographic atrophy, which may signify ARMD and were not evaluated in the dataset used by this report.

\appendix
\chapter{List of abbreviations\label{chap&append_abbr}}

\begin{tabular}{ll}\\
\small
AMT& Amazon mechanical Turk\\
ARIA& Automated retinal image analysis\\
ARMD & Age-Related Macular Degeneration\\
BN& Batch normalization\\
CNV& choroidal neovascularization\\
DL& Deep Learning\\
DME& diabetic macular edema\\
DNN& Deep neural network\\
DR& diabetic retinopathy\\
JPEG& Joint Photographic Experts Group\\
NPDR& Non-proliferative diabetic retinopathy\\
OCT& Optical coherence tomography\\
PDR& Proliferative diabetic retinopathy\\
ResNet& Residual Neural Network\\
VGG& Visual Geometry Group\\
VTDR& Vision-threatening diabetic retinopathy\\
\normalsize
\end{tabular}

\chapter{Review of Statistical Topics Used in this Thesis\label{chap:append_stats}}

\section{Sensitivity and Specificity}

In assessing the performance of a diagnostic test, there are 4 cases to consider in comparing the label predicted by a system (\eg a deep learning network) versus the true label assigned by a grader to that test label:
\begin{enumerate}
\item \textbf{True positive (TP)} - the predicted and true labels agree and the test case has the disease.
\item \textbf{True negative (TP)} - the predicted and true labels agree and the test case does not have the disease.
\item \textbf{False Negative (FN)} - the predicted label is that the test case does not have the disease, while the true label is that the test case has the disease.
\item \textbf{False positive (FP)} - the predicted label is that the test case has the disease, while the true label is that the test case does not have the disease.
\end{enumerate}

\textbf{Sensitivity} (also called the true positive rate, the recall, or probability of detection \cite{stats-book} in some fields) measures the proportion of actual positives that are correctly identified as such (\eg the percentage of sick people who are correctly identified as having the condition) and is defined by:
\begin{equation}
Sensitivity = \frac{TP}{TP + FN}
\end{equation}
\indent \textbf{Specificity} (also called the true negative rate) measures the proportion of actual negatives that are correctly identified as such (\eg the percentage of healthy people who are correctly identified as not having the condition) and is defined by:
\begin{equation}
Specificity = \frac{TN}{TN + FP}
\end{equation}

\indent \textbf{Accuracy} is the percentage of cases where the predicted and true labels agree. In a binary classification scheme, accuracy can by defined by:
\begin{equation}
Accuracy = \frac{TP + TN}{TP + TN + FP + FN}
\end{equation}

\section{Wilcoxon signed-rank and Mann-Whitney U Test}
\indent The \textbf{Wilcoxon signed-rank} test is a non-parametric statistical hypothesis test used to compare two related samples, matched samples, or repeated measurements on a single sample to assess whether their population mean ranks differ (\ie it is a paired difference test). It can be used as an alternative to the paired Student's t-test, t-test for matched pairs, or the t-test for dependent samples when the population cannot be assumed to be normally distributed. A Wilcoxon signed-rank test is a nonparametric test that can be used to determine whether two dependent samples were selected from populations having the same distribution\cite{wilcoxon}. See \cite{wilcoxon} for details of the test procedure.\\
\indent The \textbf{Mann-Whitney U} test is a nonparametric test of the null hypothesis that it is equally likely that a randomly selected value from one sample will be less than or greater than a randomly selected value from a second sample\cite{mann-whitney}. See \cite{mann-whitney} for details of the test calculation.  

\section{Quadratic weighted kappa}
\indent \textbf{Cohen's kappa coefficient ($\kappa$)} is a statistic which measures inter-rater agreement for qualitative (categorical) items\cite{kappa}. Kappa takes into account the possibility of agreement occuring by chance (\ie guessing) and is a more robust measure than simple percent agreement. For details of the kappa calculation, the reader is referred to \cite{wiki_kappa}. Kappa does not take into account the degree of disagreement between observers and all disagreement is treated equally as total disagreement.  Therefore, a weighting scheme has been devised, with either linear or quadratic weightings.\footnote{For example see the medcalc implementation at \url{https://www.medcalc.org/manual/kappa.php}, which advised: Use linear weights when the difference between the first and second category has the same importance as a difference between the second and third category, etc. If the difference between the first and second category is less important than a difference between the second and third category, etc., use quadratic weights.} A generally agreed upon scale for level of agreement is shown in table \ref{tab:kappa}.

\begin{table}[htb!]
\caption[Kappa categories.]{Kappa catagories.}\label{tab:kappa}
  \begin{center}
\scalebox{0.9}{
\begin{tabular}{|l|l|}
\hline
$\kappa$ & Level of Agreement\\
\hline
$<$0.20 & Poor\\
0.21-0.40 & Fair\\
0.41-0.60 & Moderate\\
0.61-0.80 & Good\\
0.81-1.00 & Very good\\
\hline
\end{tabular}}
\end{center}
\end{table}

\chapter{Network loss and accuracy plots\label{chap:append_network_plots}}

\begin{figure}[htb]
\begin{subfigure}{0.6\textwidth}
\captionsetup{width=0.8\textwidth}
\centering
\includegraphics[width=0.6\linewidth]{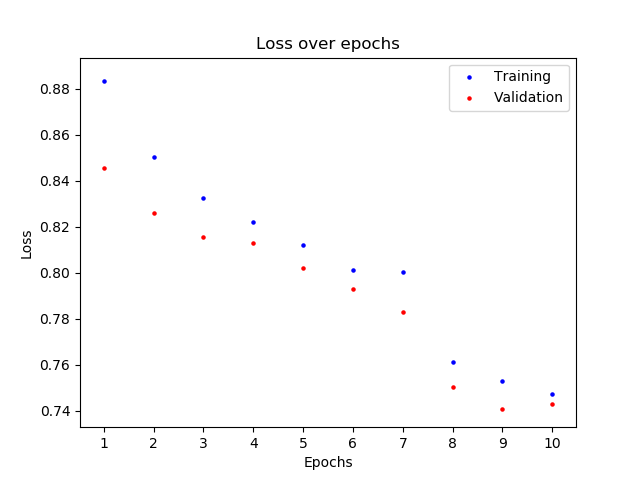}
\caption{Loss; Pretrained; Fine-tuning}
\label{fig:Diabetic_Retinopathy_AlexNet_loss_plt_a}
\end{subfigure}
\begin{subfigure}{0.6\textwidth}
\captionsetup{width=0.8\textwidth}
\centering
\includegraphics[width=0.6\linewidth]{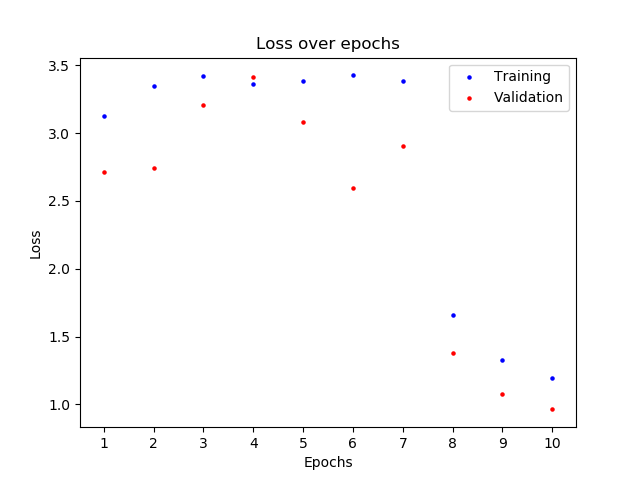}
\caption{Pretrained; Feature extractor}
\label{fig:Diabetic_Retinopathy_AlexNet_loss_plt_b}
\end{subfigure}
\begin{subfigure}{0.6\textwidth}
\captionsetup{width=0.8\textwidth}
\centering
\includegraphics[width=0.6\linewidth]{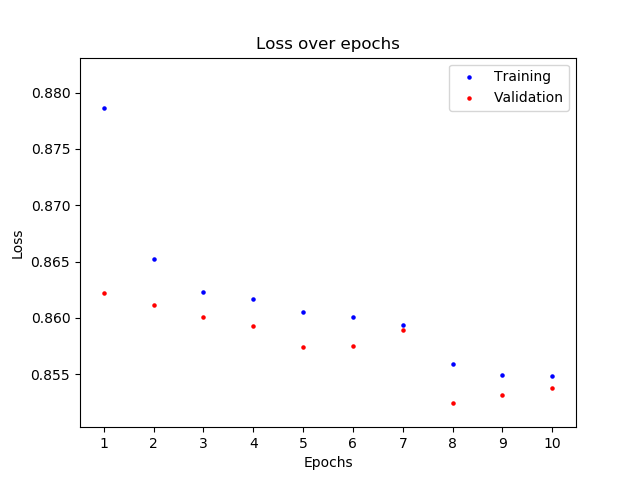}
\caption{Loss; Not Pretrained; Fine-tuning}
\label{fig:Diabetic_Retinopathy_AlexNet_loss_plt_c}
\end{subfigure}
\begin{subfigure}{0.6\textwidth}
\captionsetup{width=0.8\textwidth}
\centering
\includegraphics[width=0.6\linewidth]{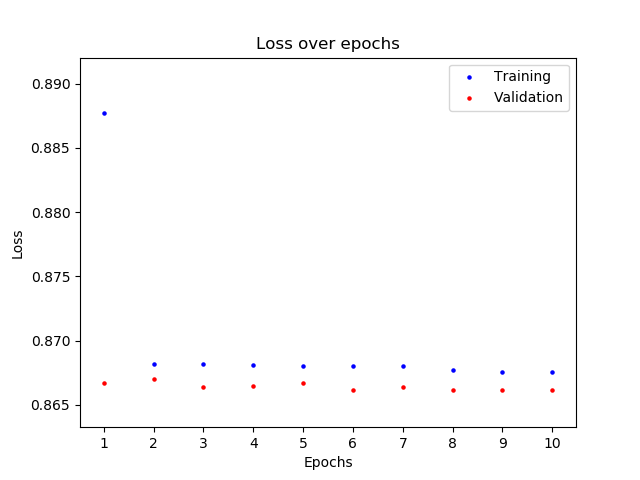}
\caption{Loss; Not Pretrained; Feature extractor}
\label{fig:Diabetic_Retinopathy_AlexNet_loss_plt_d}
\end{subfigure}
\begin{subfigure}{0.6\textwidth}
\captionsetup{width=0.8\textwidth}
\centering\includegraphics[width=0.6\linewidth]{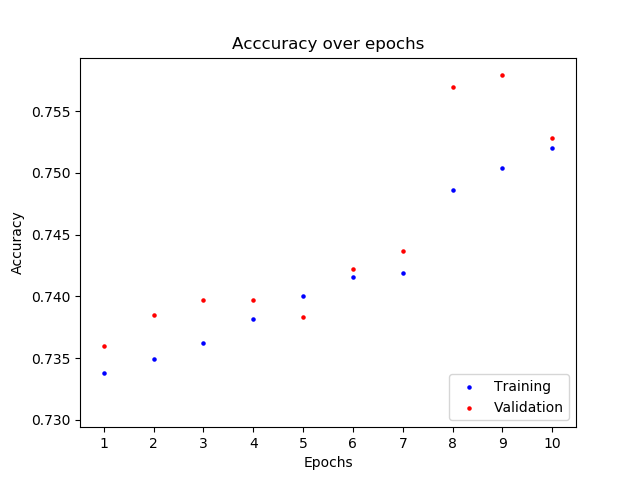}
\caption{Accuracy; Pretrained; Fine-tuning}
\label{fig:Diabetic_Retinopathy_AlexNet_acc_plt_e}
\end{subfigure}
\begin{subfigure}{0.6\textwidth}
\captionsetup{width=0.8\textwidth}
\centering\includegraphics[width=0.6\linewidth]{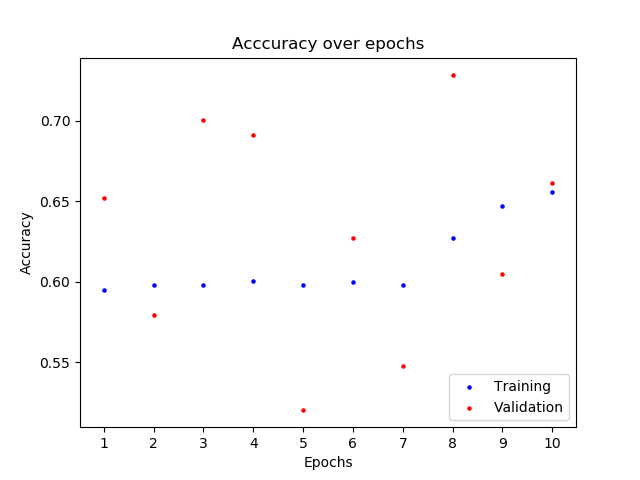}
\caption{Accuracy; Pretrained; Feature extractor}
\label{fig:Diabetic_Retinopathy_AlexNet_acc_plt_f}
\end{subfigure}
\begin{subfigure}{0.6\textwidth}
\captionsetup{width=0.8\textwidth}
\centering\includegraphics[width=0.6\linewidth]{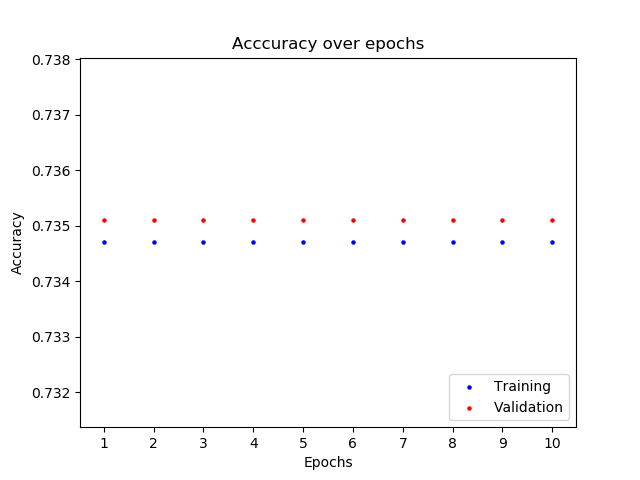}
\caption{Accuracy; Not Pretrained; Fine-tuning}
\label{fig:Diabetic_Retinopathy_AlexNet_acc_plt_g}
\end{subfigure}
\begin{subfigure}{0.6\textwidth}
\captionsetup{width=0.8\textwidth}
\centering\includegraphics[width=0.6\linewidth]{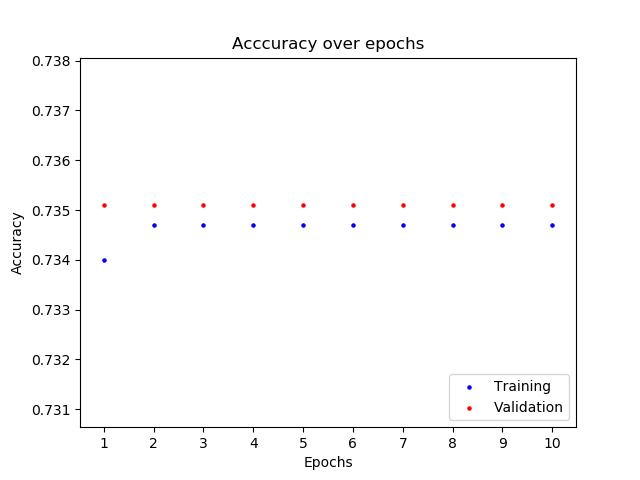}
\caption{Accuracy; Not Pretrained; Feature extractor}
\label{fig:Diabetic_Retinopathy_AlexNet_acc_plt_h}
\end{subfigure}
\caption[Diabetic Retinopathy: AlexNet loss and accuracy plots.]{Diabetic Retinopathy: AlexNet loss and accuracy plots.}
\label{fig:AlexNet_acc_plt}
\end{figure}

\begin{figure}[htb]
\begin{subfigure}{0.6\textwidth}
\captionsetup{width=0.8\textwidth}
\centering
\includegraphics[width=0.6\linewidth]{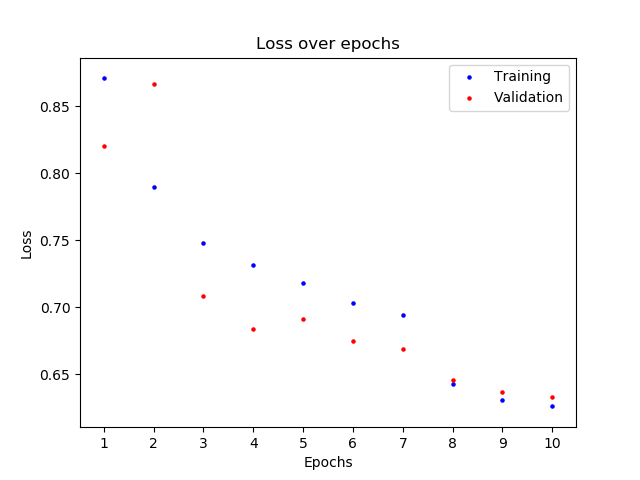}
\caption{Loss; Pretrained; Fine-tuning}
\label{fig:Diabetic Retinopathy: DenseNet-121_loss_plt_a}
\end{subfigure}
\begin{subfigure}{0.6\textwidth}
\captionsetup{width=0.8\textwidth}
\centering
\includegraphics[width=0.6\linewidth]{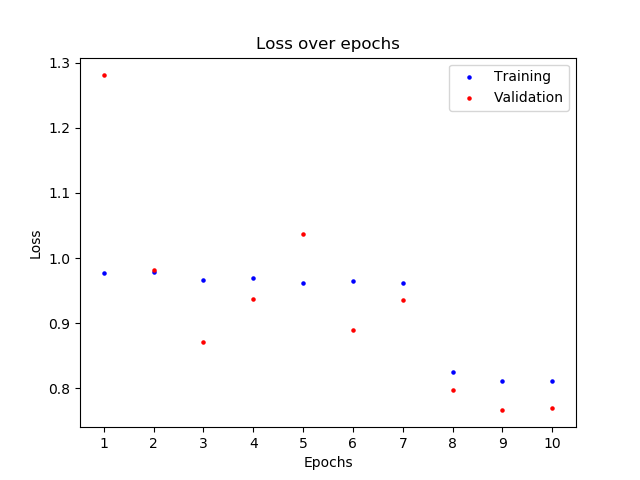}
\caption{Loss; Pretrained; Feature extractor}
\label{fig:Diabetic Retinopathy: DenseNet-121_loss_plt_b}
\end{subfigure}
\begin{subfigure}{0.6\textwidth}
\captionsetup{width=0.8\textwidth}
\centering
\includegraphics[width=0.6\linewidth]{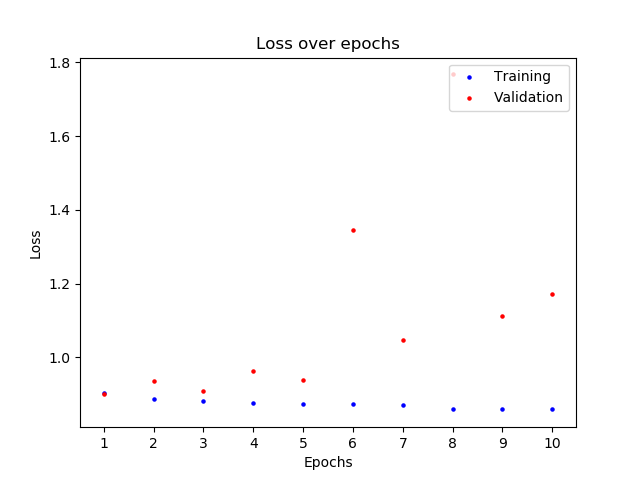}
\caption{Loss, Not Pretrained; Fine-tuning}
\label{fig:Diabetic Retinopathy: DenseNet-121_loss_plt_c}
\end{subfigure}
\begin{subfigure}{0.6\textwidth}
\captionsetup{width=0.8\textwidth}
\centering
\includegraphics[width=0.6\linewidth]{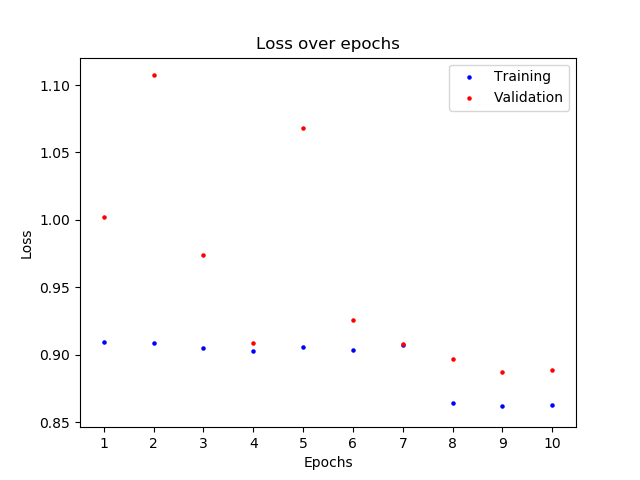}
\caption{Loss; Not Pretrained; Feature extractor}
\label{fig:Diabetic Retinopathy: DenseNet-121_loss_plt_d}
\end{subfigure}
\begin{subfigure}{0.6\textwidth}
\captionsetup{width=0.8\textwidth}
\centering\includegraphics[width=0.6\linewidth]{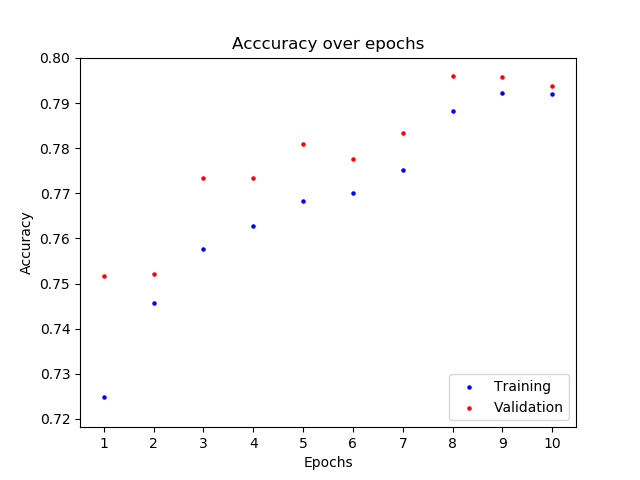}
\caption{Accuracy; Pretrained; Fine-tuning}
\label{fig:Diabetic Retinopathy: DenseNet-121_acc_plt_e}
\end{subfigure}
\begin{subfigure}{0.6\textwidth}
\captionsetup{width=0.8\textwidth}
\centering\includegraphics[width=0.6\linewidth]{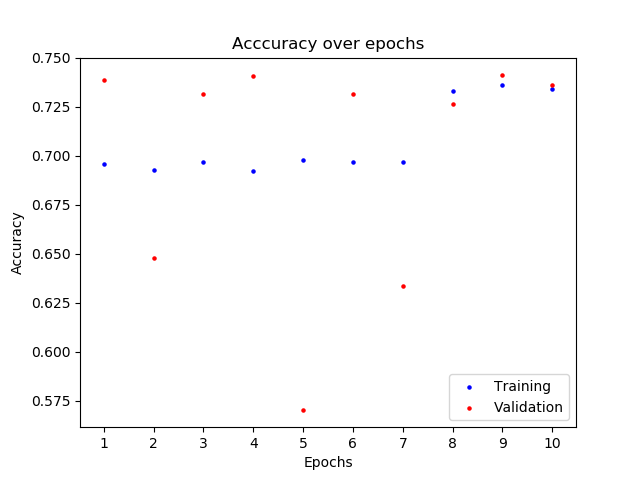}
\caption{Accuracy; Pretrained; Feature extractor}
\label{fig:Diabetic Retinopathy: DenseNet-121_acc_plt_f}
\end{subfigure}
\begin{subfigure}{0.6\textwidth}
\captionsetup{width=0.8\textwidth}
\centering\includegraphics[width=0.6\linewidth]{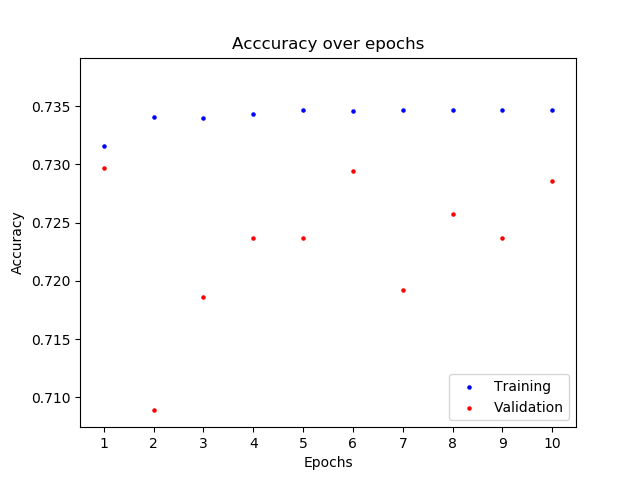}
\caption{Accuracy; Not Pretrained; Fine-tuning}
\label{fig:Diabetic Retinopathy: DenseNet-121_acc_plt_g}
\end{subfigure}
\begin{subfigure}{0.6\textwidth}
\captionsetup{width=0.8\textwidth}
\centering\includegraphics[width=0.6\linewidth]{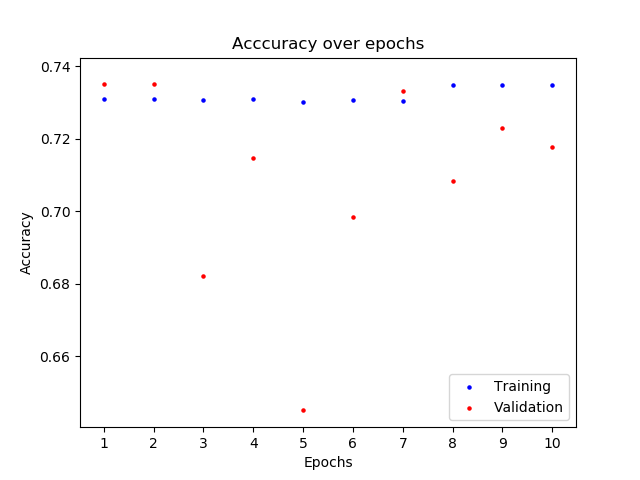}
\caption{Accuracy; Not Pretrained; Feature extractor}
\label{fig:Diabetic Retinopathy: DenseNet-121_acc_plt_h}
\end{subfigure}
\caption[Diabetic Retinopathy: DenseNet-121 loss and accuracy plots.]{Diabetic Retinopathy: DenseNet-121 loss and accuracy plots.}
\label{fig:Diabetic Retinopathy: DenseNet-121_acc_plt}
\end{figure}

\begin{figure}[htb]
\begin{subfigure}{0.6\textwidth}
\captionsetup{width=0.8\textwidth}
\centering
\includegraphics[width=0.6\linewidth]{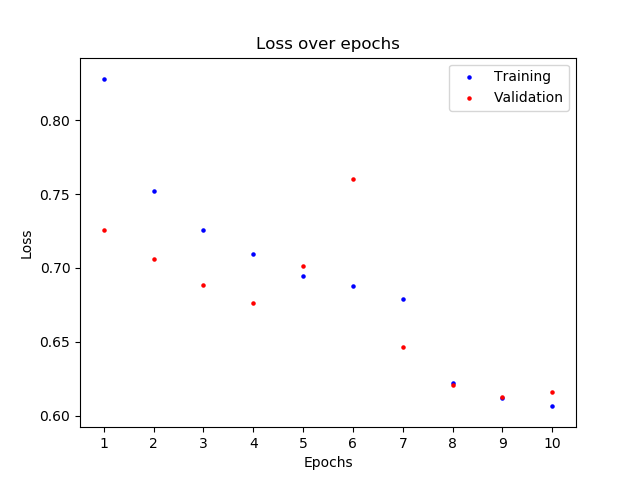}
\caption{Loss; Pretrained; Fine-tuning}
\label{fig:Diabetic Retinopathy: DenseNet-161_loss_plt_a}
\end{subfigure}
\begin{subfigure}{0.6\textwidth}
\captionsetup{width=0.8\textwidth}
\centering
\includegraphics[width=0.6\linewidth]{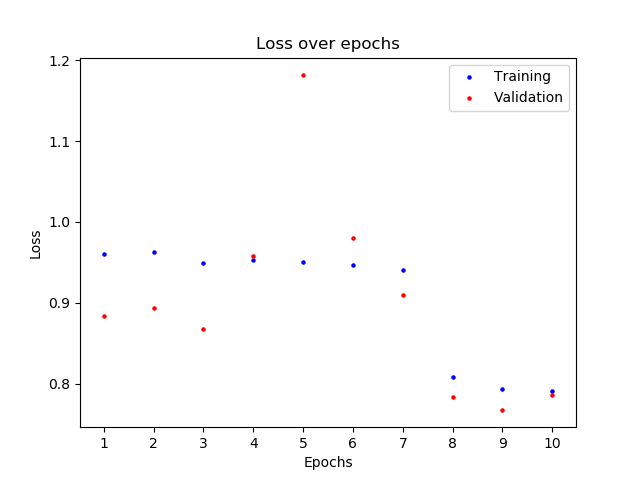}
\caption{Loss; Pretrained; Feature extractor}
\label{fig:Diabetic Retinopathy: DenseNet-161_loss_plt_b}
\end{subfigure}
\begin{subfigure}{0.6\textwidth}
\captionsetup{width=0.8\textwidth}
\centering
\includegraphics[width=0.6\linewidth]{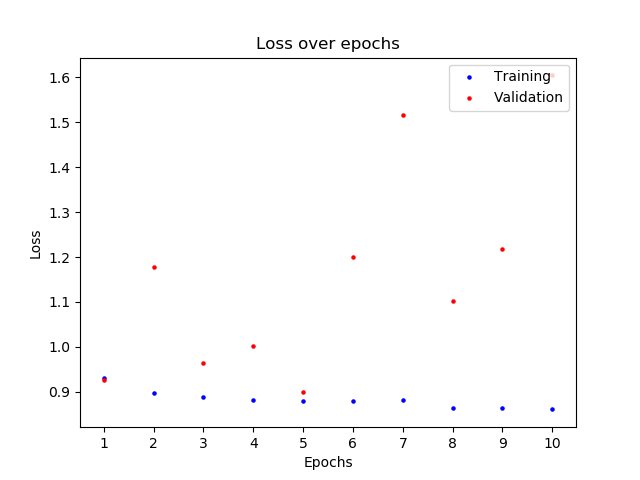}
\caption{Loss; Not Pretrained; Fine-tuning}
\label{fig:Diabetic Retinopathy: DenseNet-161_loss_plt_c}
\end{subfigure}
\begin{subfigure}{0.6\textwidth}
\captionsetup{width=0.8\textwidth}
\centering
\includegraphics[width=0.6\linewidth]{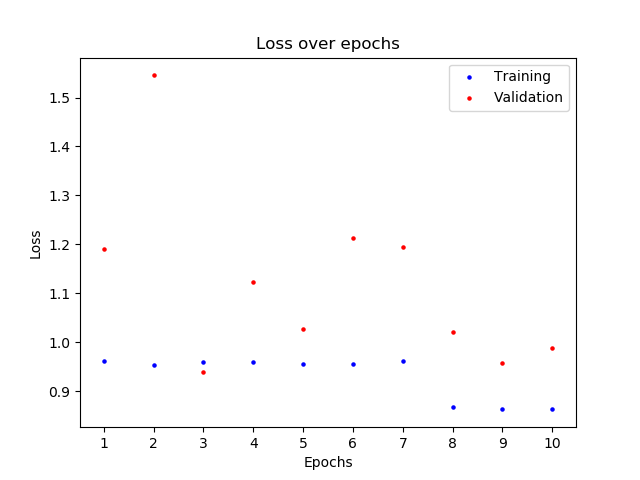}
\caption{Loss; Not Pretrained; Feature extractor}
\label{fig:Diabetic Retinopathy: DenseNet-161_loss_plt_d}
\end{subfigure}
\begin{subfigure}{0.6\textwidth}
\captionsetup{width=0.8\textwidth}
\centering\includegraphics[width=0.6\linewidth]{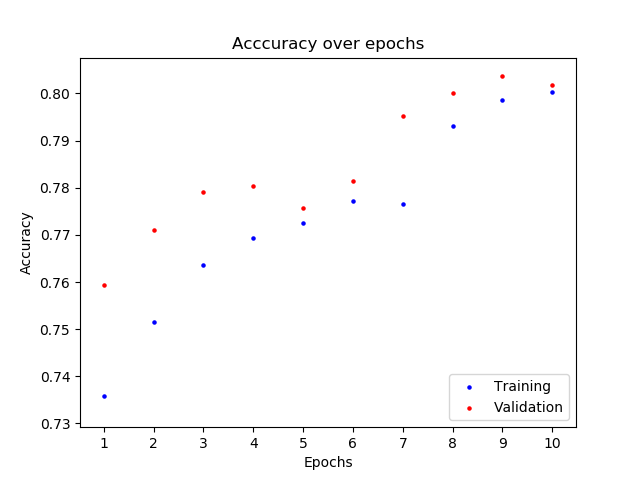}
\caption{Accuracy; Pretrained; Fine-tuning}
\label{fig:Diabetic Retinopathy: DenseNet-161_acc_plt_e}
\end{subfigure}
\begin{subfigure}{0.6\textwidth}
\captionsetup{width=0.8\textwidth}
\centering\includegraphics[width=0.6\linewidth]{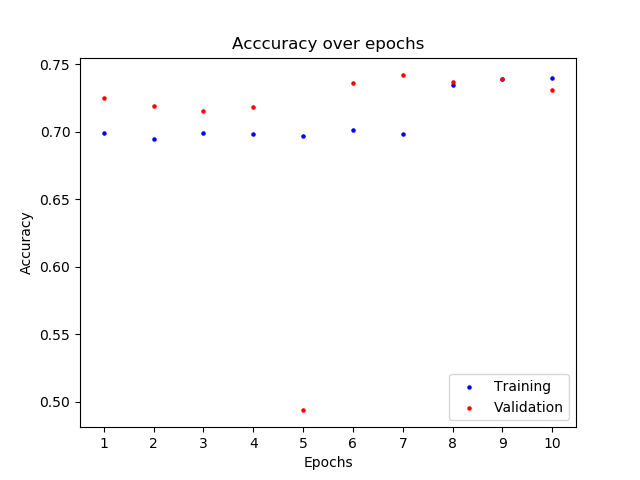}
\caption{Accuracy Pretrained; Feature extractor}
\label{fig:Diabetic Retinopathy: DenseNet-161_acc_plt_f}
\end{subfigure}
\begin{subfigure}{0.6\textwidth}
\captionsetup{width=0.8\textwidth}
\centering\includegraphics[width=0.6\linewidth]{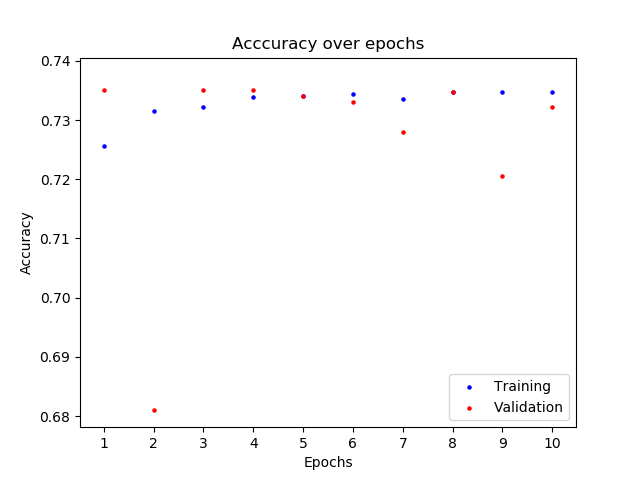}
\caption{Accuracy; Not Pretrained; Fine-tuning}
\label{fig:Diabetic Retinopathy: DenseNet-161_acc_plt_g}
\end{subfigure}
\begin{subfigure}{0.6\textwidth}
\captionsetup{width=0.8\textwidth}
\centering\includegraphics[width=0.6\linewidth]{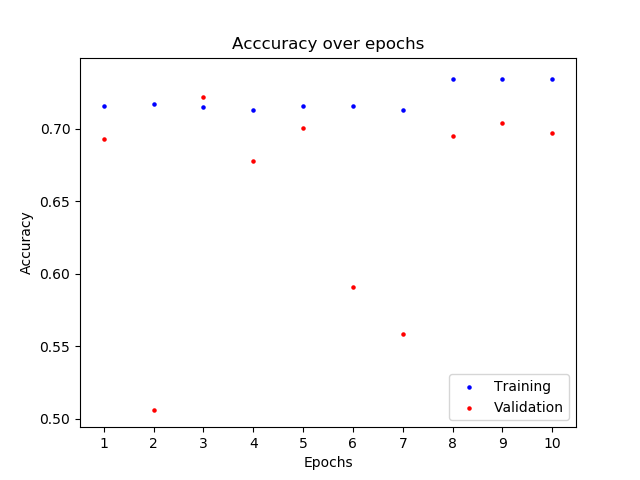}
\caption{Accuracy; Not Pretrained; Feature extractor}
\label{fig:Diabetic Retinopathy: DenseNet-161_acc_plt_h}
\end{subfigure}
\caption[Diabetic Retinopathy: DenseNet-161 loss and accuracy plots.]{Diabetic Retinopathy: DenseNet-161 loss and accuracy plots.}
\label{fig:Diabetic Retinopathy: DenseNet-161_acc_plt}
\end{figure}

\begin{figure}[htb]
\begin{subfigure}{0.6\textwidth}
\captionsetup{width=0.8\textwidth}
\centering
\includegraphics[width=0.6\linewidth]{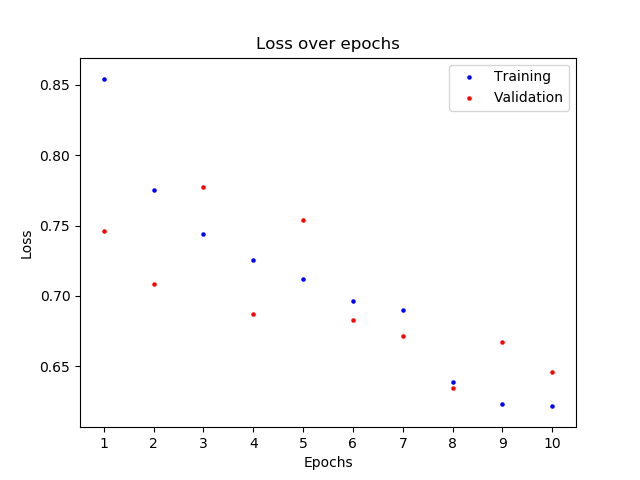}
\caption{Loss; Pretrained; Fine-tuning}
\label{fig:Diabetic Retinopathy: DenseNet-169_loss_plt_a}
\end{subfigure}
\begin{subfigure}{0.6\textwidth}
\captionsetup{width=0.8\textwidth}
\centering
\includegraphics[width=0.6\linewidth]{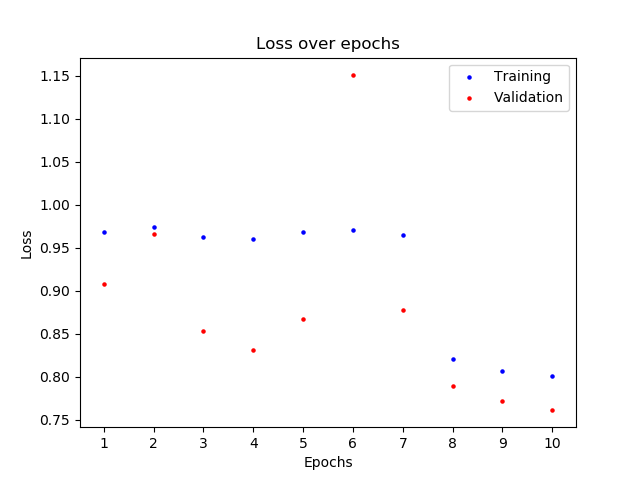}
\caption{Loss; Pretrained; Feature extractor}
\label{fig:Diabetic Retinopathy: DenseNet-169_loss_plt_b}
\end{subfigure}
\begin{subfigure}{0.6\textwidth}
\captionsetup{width=0.8\textwidth}
\centering
\includegraphics[width=0.6\linewidth]{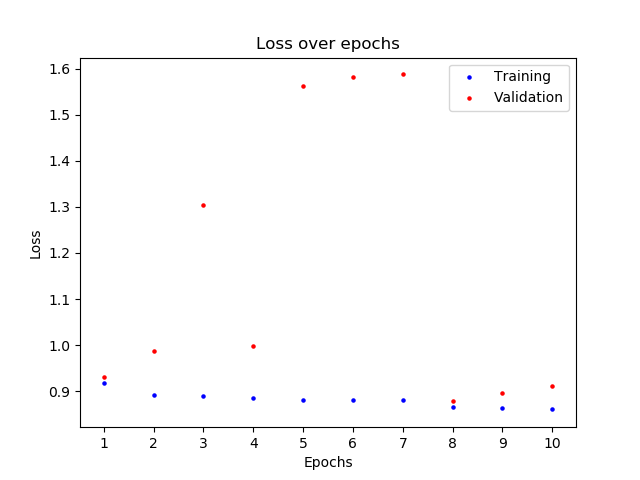}
\caption{Loss; Not Pretrained; Fine-tuning}
\label{fig:Diabetic Retinopathy: DenseNet-169_loss_plt_c}
\end{subfigure}
\begin{subfigure}{0.6\textwidth}
\captionsetup{width=0.8\textwidth}
\centering
\includegraphics[width=0.6\linewidth]{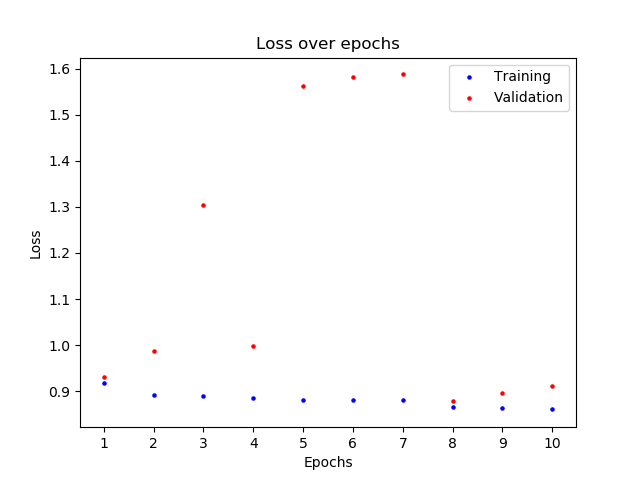}
\caption{Loss; Not Pretrained; Feature extractor}
\label{fig:Diabetic Retinopathy: DenseNet-169_loss_plt_d}
\end{subfigure}
\begin{subfigure}{0.6\textwidth}
\captionsetup{width=0.8\textwidth}
\centering\includegraphics[width=0.6\linewidth]{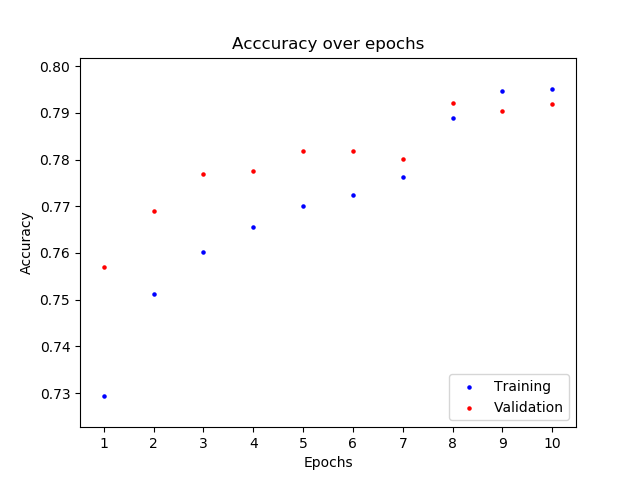}
\caption{Accuracy; Pretrained; Fine-tuning}
\label{fig:Diabetic Retinopathy: DenseNet-169_acc_plt_e}
\end{subfigure}
\begin{subfigure}{0.6\textwidth}
\captionsetup{width=0.8\textwidth}
\centering\includegraphics[width=0.6\linewidth]{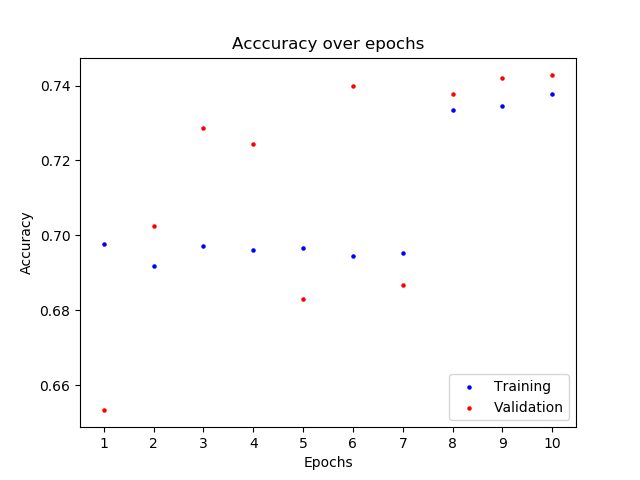}
\caption{Accuracy; Pretrained; Feature extractor}
\label{fig:Diabetic Retinopathy: DenseNet-169_acc_plt_f}
\end{subfigure}
\begin{subfigure}{0.6\textwidth}
\captionsetup{width=0.8\textwidth}
\centering\includegraphics[width=0.6\linewidth]{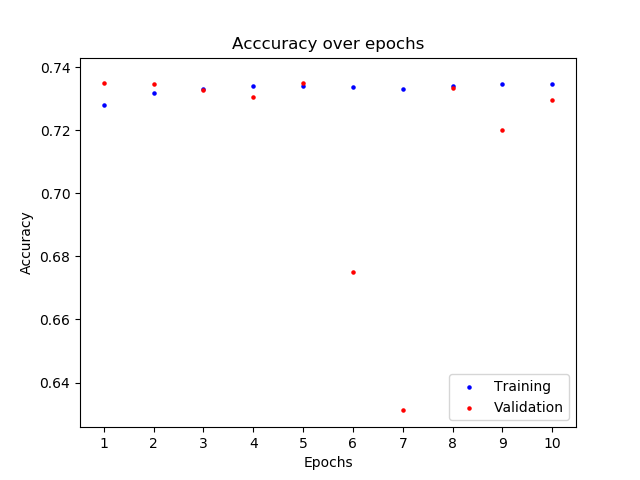}
\caption{Accuracy; Not Pretrained; Fine-tuning}
\label{fig:Diabetic Retinopathy: DenseNet-169_acc_plt_g}
\end{subfigure}
\begin{subfigure}{0.6\textwidth}
\captionsetup{width=0.8\textwidth}
\centering\includegraphics[width=0.6\linewidth]{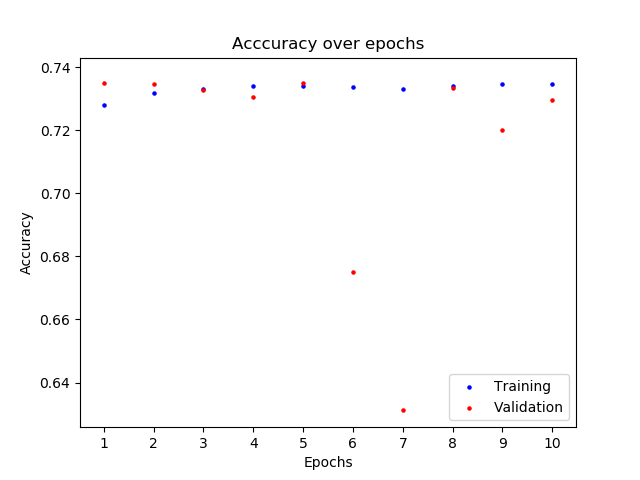}
\caption{Accuracy; Not Pretrained; Feature extractor}
\label{fig:Diabetic Retinopathy: DenseNet-169_acc_plt_h}
\end{subfigure}
\caption[Diabetic Retinopathy: DenseNet-169 loss and accuracy plots.]{Diabetic Retinopathy: DenseNet-169 loss and accuracy plots.}
\label{fig:Diabetic Retinopathy: DenseNet-169_acc_plt}
\end{figure}

\begin{figure}[htb]
\begin{subfigure}{0.6\textwidth}
\captionsetup{width=0.8\textwidth}
\centering
\includegraphics[width=0.6\linewidth]{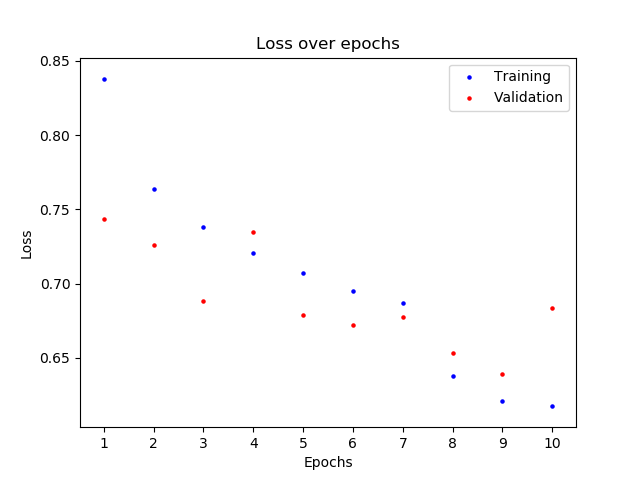}
\caption{Loss; Pretrained; Fine-tuning}
\label{fig:Diabetic Retinopathy: DenseNet-201_loss_plt_a}
\end{subfigure}
\begin{subfigure}{0.6\textwidth}
\captionsetup{width=0.8\textwidth}
\centering
\includegraphics[width=0.6\linewidth]{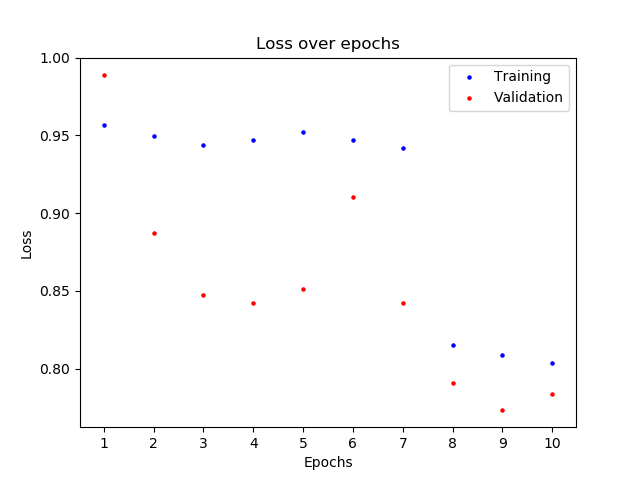}
\caption{Loss; Pretrained; Feature extractor}
\label{fig:Diabetic Retinopathy: DenseNet-201_loss_plt_b}
\end{subfigure}
\begin{subfigure}{0.6\textwidth}
\captionsetup{width=0.8\textwidth}
\centering
\includegraphics[width=0.6\linewidth]{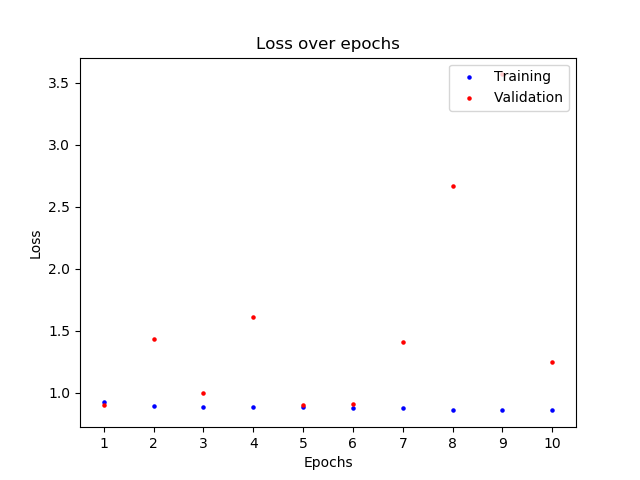}
\caption{Loss; Not Pretrained; Fine-tuning}
\label{fig:Diabetic Retinopathy: DenseNet-201_loss_plt_c}
\end{subfigure}
\begin{subfigure}{0.6\textwidth}
\captionsetup{width=0.8\textwidth}
\centering
\includegraphics[width=0.6\linewidth]{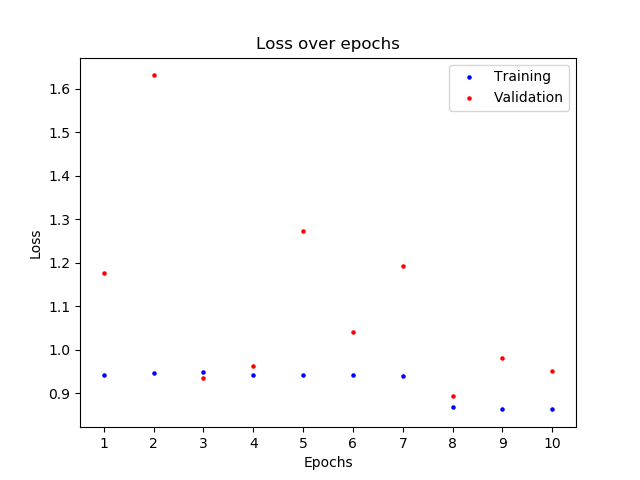}
\caption{Loss; Not Pretrained; Feature extractor}
\label{fig:Diabetic Retinopathy: DenseNet-201_loss_plt_d}
\end{subfigure}
\begin{subfigure}{0.6\textwidth}
\captionsetup{width=0.8\textwidth}
\centering\includegraphics[width=0.6\linewidth]{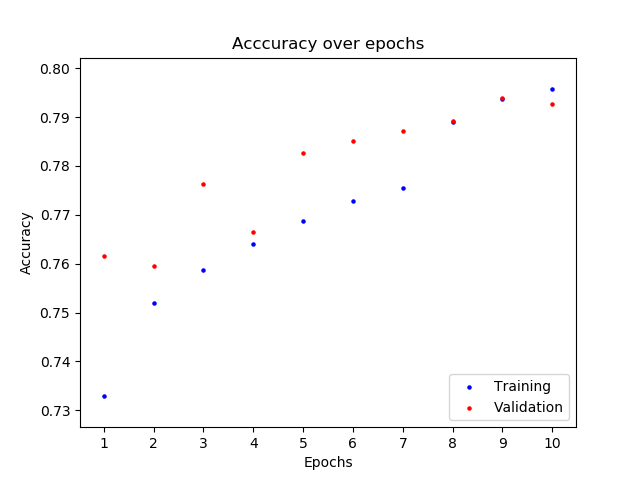}
\caption{Accuracy Pretrained; Fine-tuning}
\label{fig:Diabetic Retinopathy: DenseNet-201_acc_plt_e}
\end{subfigure}
\begin{subfigure}{0.6\textwidth}
\captionsetup{width=0.8\textwidth}
\centering\includegraphics[width=0.6\linewidth]{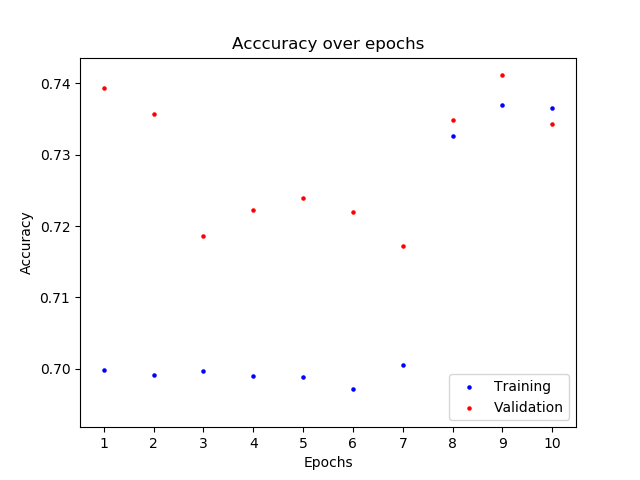}
\caption{Accuracy; Pretrained; Feature extractor}
\label{fig:Diabetic Retinopathy: DenseNet-201_acc_plt_f}
\end{subfigure}
\begin{subfigure}{0.6\textwidth}
\captionsetup{width=0.8\textwidth}
\centering\includegraphics[width=0.6\linewidth]{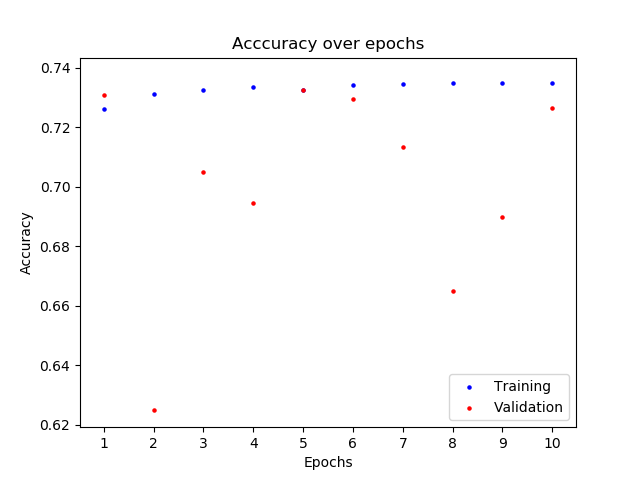}
\caption{Accuracy; Not Pretrained; Fine-tuning}
\label{fig:Diabetic Retinopathy: DenseNet-201_acc_plt_g}
\end{subfigure}
\begin{subfigure}{0.6\textwidth}
\captionsetup{width=0.8\textwidth}
\centering\includegraphics[width=0.6\linewidth]{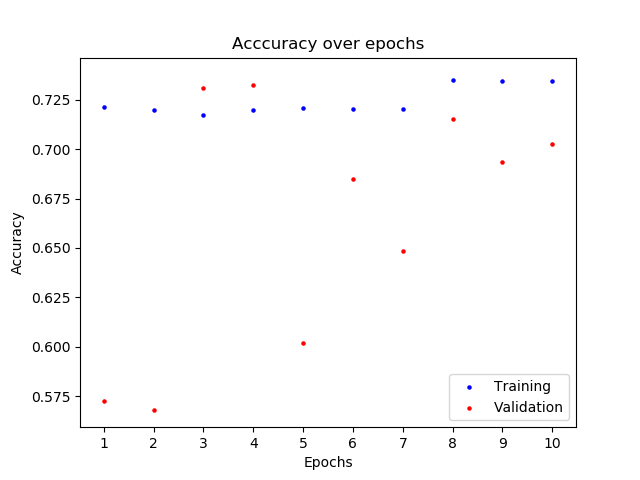}
\caption{Accuracy; Not Pretrained; Feature extractor}
\label{fig:Diabetic Retinopathy: DenseNet-201_acc_plt_h}
\end{subfigure}
\caption[Diabetic Retinopathy: DenseNet-201 loss and accuracy plots.]{Diabetic Retinopathy: DenseNet-201 loss and accuracy plots.}
\label{fig:Diabetic Retinopathy: DenseNet-201_acc_plt}
\end{figure}

\begin{figure}[htb]
\begin{subfigure}{0.6\textwidth}
\captionsetup{width=0.8\textwidth}
\centering
\includegraphics[width=0.6\linewidth]{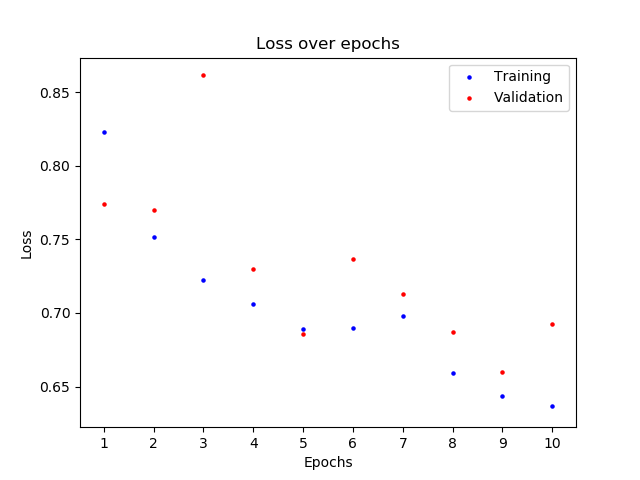}
\caption{Loss; Pretrained; Fine-tuning}
\label{fig:Diabetic Retinopathy: Inception-v3_loss_plt_a}
\end{subfigure}
\begin{subfigure}{0.6\textwidth}
\captionsetup{width=0.8\textwidth}
\centering
\includegraphics[width=0.6\linewidth]{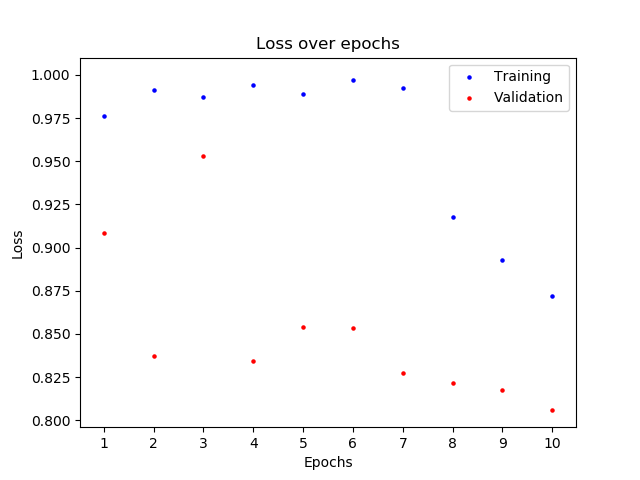}
\caption{Loss; Pretrained; Feature extractor}
\label{fig:Diabetic Retinopathy: Inception-v3_loss_plt_b}
\end{subfigure}
\begin{subfigure}{0.6\textwidth}
\captionsetup{width=0.8\textwidth}
\centering
\includegraphics[width=0.6\linewidth]{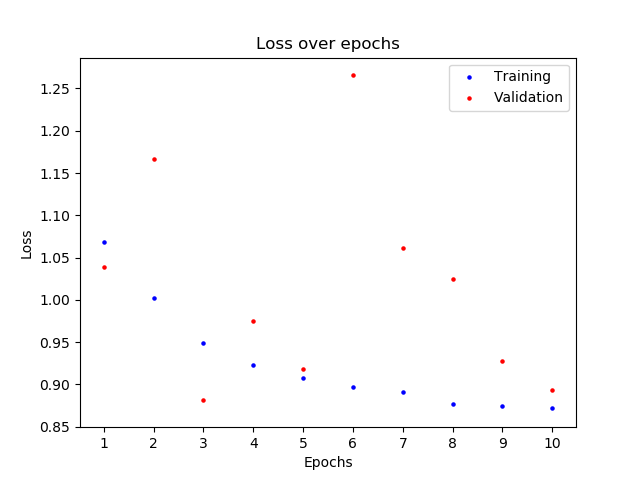}
\caption{Loss; Not Pretrained; Fine-tuning}
\label{fig:Diabetic Retinopathy: Inception-v3_loss_plt_c}
\end{subfigure}
\begin{subfigure}{0.6\textwidth}
\captionsetup{width=0.8\textwidth}
\centering
\includegraphics[width=0.6\linewidth]{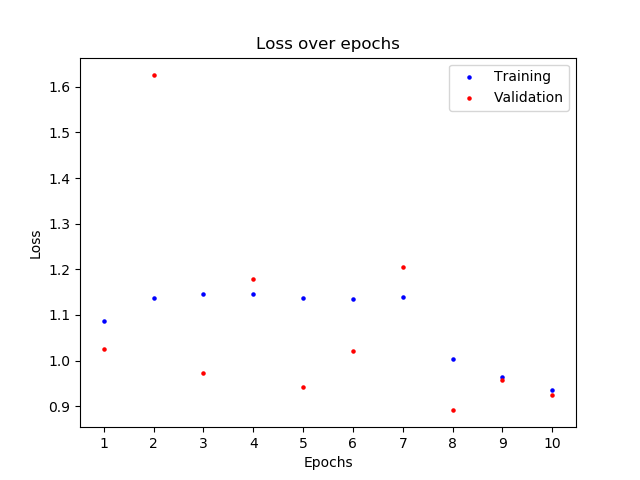}
\caption{Loss; Not Pretrained; Feature extractor}
\label{fig:Diabetic Retinopathy: Inception-v3_loss_plt_d}
\end{subfigure}
\begin{subfigure}{0.6\textwidth}
\captionsetup{width=0.8\textwidth}
\centering\includegraphics[width=0.6\linewidth]{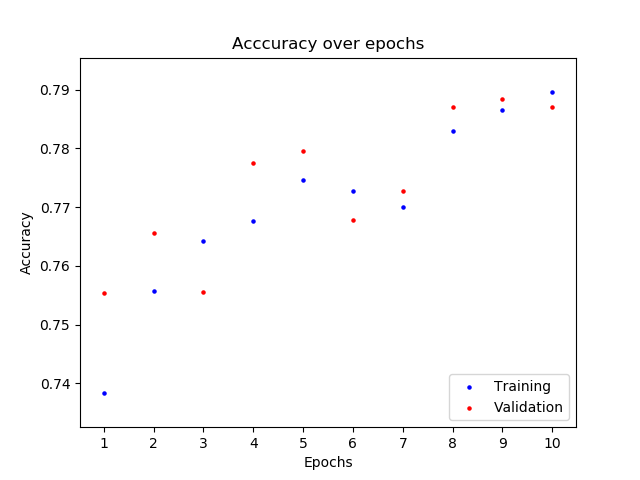}
\caption{Accuracy; Pretrained; Fine-tuning}
\label{fig:Diabetic Retinopathy: Inception-v3_acc_plt_e}
\end{subfigure}
\begin{subfigure}{0.6\textwidth}
\captionsetup{width=0.8\textwidth}
\centering\includegraphics[width=0.6\linewidth]{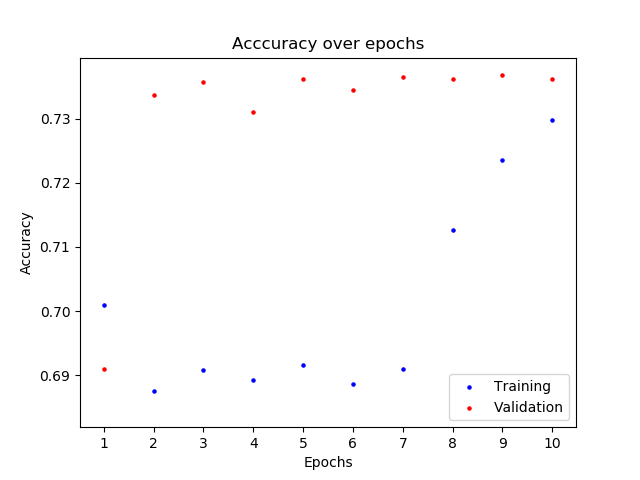}
\caption{Accuacy; Pretrained; Feature extractor}
\label{fig:Diabetic Retinopathy: Inception-v3_acc_plt_f}
\end{subfigure}
\begin{subfigure}{0.6\textwidth}
\captionsetup{width=0.8\textwidth}
\centering\includegraphics[width=0.6\linewidth]{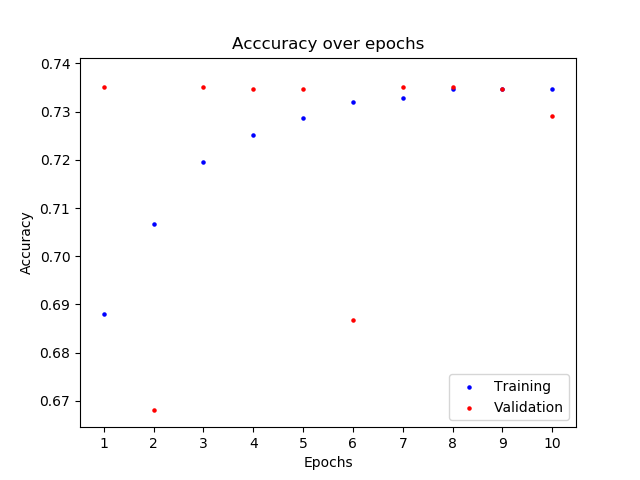}
\caption{Accuracy; Not Pretrained; Fine-tuning}
\label{fig:Diabetic Retinopathy: Inception-v3_acc_plt_g}
\end{subfigure}
\begin{subfigure}{0.6\textwidth}
\captionsetup{width=0.8\textwidth}
\centering\includegraphics[width=0.6\linewidth]{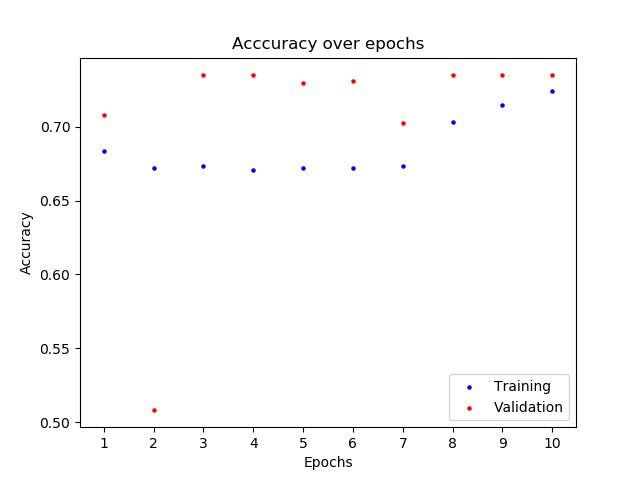}
\caption{Accuracy; Not Pretrained; Feature extractor}
\label{fig:Diabetic Retinopathy: Inception-v3_acc_plt_h}
\end{subfigure}
\caption[Diabetic Retinopathy: Inception-v3 loss and accuracy plots.]{Diabetic Retinopathy: Inception-v3 loss and accuracy plots.}
\label{fig:Diabetic Retinopathy: Inception-v3_acc_plt}
\end{figure}

\begin{figure}[htb]
\begin{subfigure}{0.6\textwidth}
\captionsetup{width=0.8\textwidth}
\centering
\includegraphics[width=0.6\linewidth]{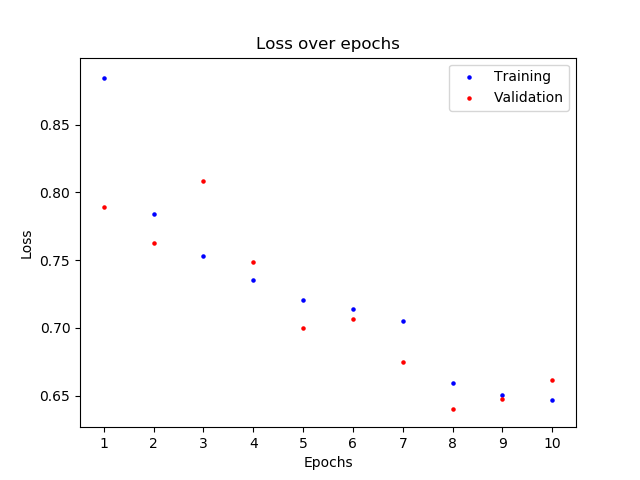}
\caption{Loss; Pretrained; Fine-tuning}
\label{fig:Diabetic Retinopathy: ResNet-18_loss_plt_a}
\end{subfigure}
\begin{subfigure}{0.6\textwidth}
\captionsetup{width=0.8\textwidth}
\centering
\includegraphics[width=0.6\linewidth]{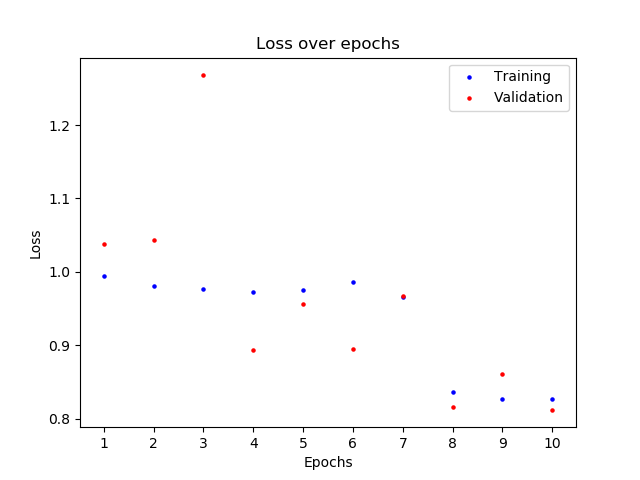}
\caption{Loss; Pretrained; Feature extractor}
\label{fig:Diabetic Retinopathy: ResNet-18_loss_plt_b}
\end{subfigure}
\begin{subfigure}{0.6\textwidth}
\captionsetup{width=0.8\textwidth}
\centering
\includegraphics[width=0.6\linewidth]{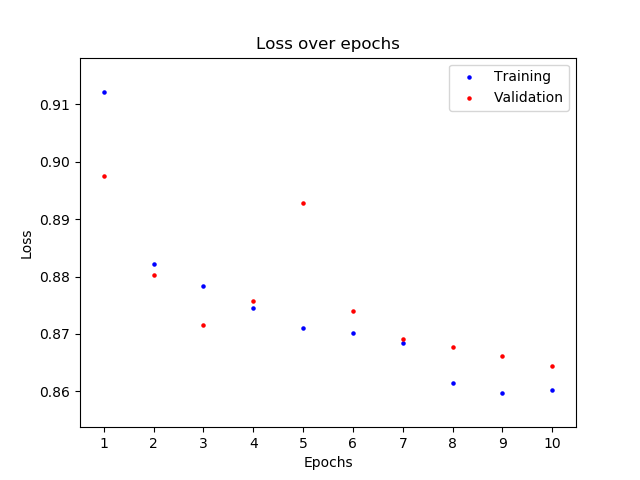}
\caption{Loss; Not Pretrained; Fine-tuning}
\label{fig:Diabetic Retinopathy: ResNet-18_loss_plt_c}
\end{subfigure}
\begin{subfigure}{0.6\textwidth}
\captionsetup{width=0.8\textwidth}
\centering
\includegraphics[width=0.6\linewidth]{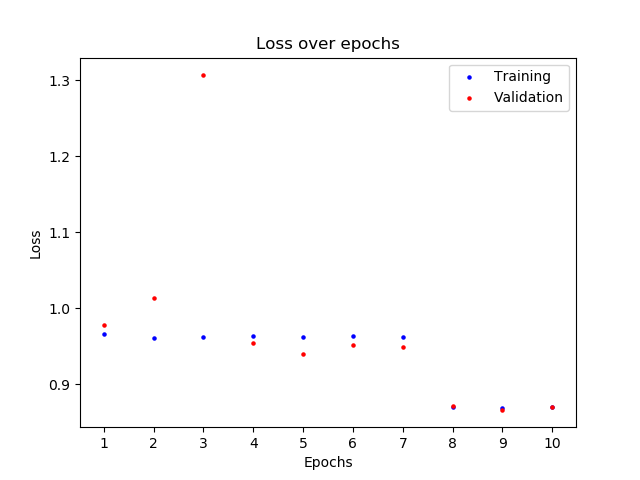}
\caption{Loss; Not Pretrained; Feature extractor}
\label{fig:Diabetic Retinopathy: ResNet-18_loss_plt_d}
\end{subfigure}
\begin{subfigure}{0.6\textwidth}
\captionsetup{width=0.8\textwidth}
\centering\includegraphics[width=0.6\linewidth]{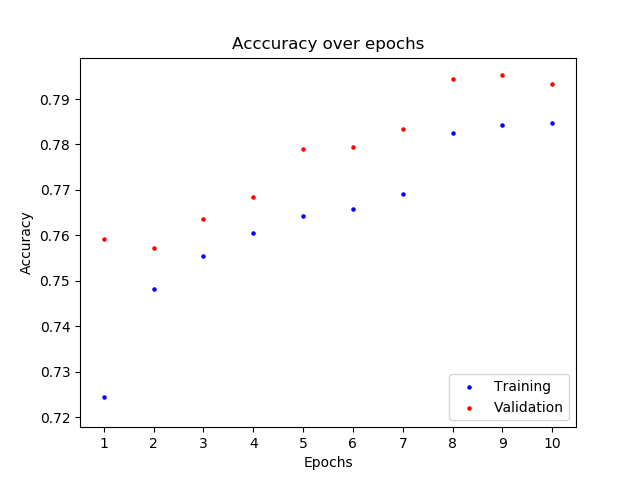}
\caption{Accuracy; Pretrained; Fine-tuning}
\label{fig:Diabetic Retinopathy: ResNet-18_acc_plt_e}
\end{subfigure}
\begin{subfigure}{0.6\textwidth}
\captionsetup{width=0.8\textwidth}
\centering\includegraphics[width=0.6\linewidth]{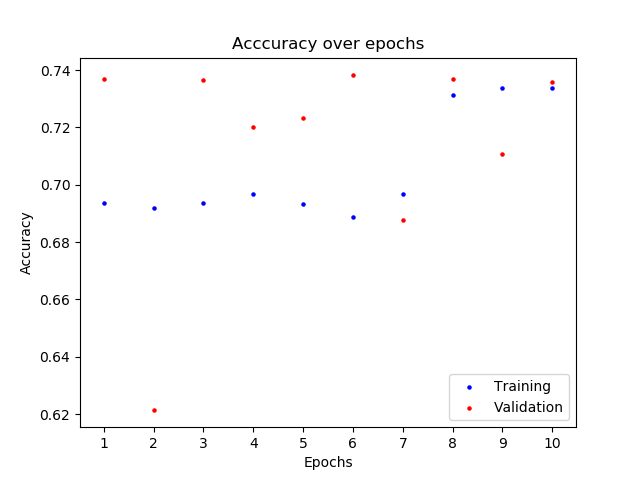}
\caption{Accuracy; Pretrained; Feature extractor}
\label{fig:Diabetic Retinopathy: ResNet-18_acc_plt_f}
\end{subfigure}
\begin{subfigure}{0.6\textwidth}
\captionsetup{width=0.8\textwidth}
\centering\includegraphics[width=0.6\linewidth]{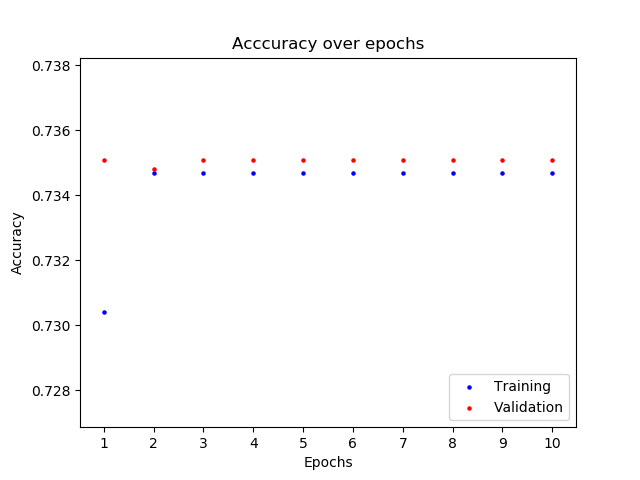}
\caption{Accuracy; Not Pretrained; Fine-tuning}
\label{fig:Diabetic Retinopathy: ResNet-18_acc_plt_g}
\end{subfigure}
\begin{subfigure}{0.6\textwidth}
\captionsetup{width=0.8\textwidth}
\centering\includegraphics[width=0.6\linewidth]{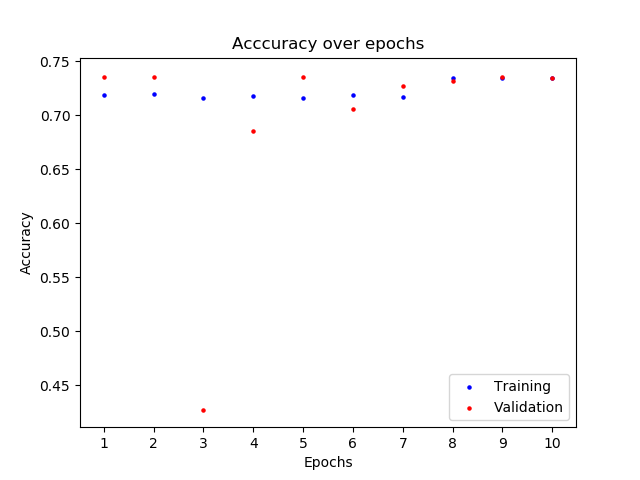}
\caption{Accuracy; Not Pretrained; Feature extractor}
\label{fig:Diabetic Retinopathy: ResNet-18_acc_plt_h}
\end{subfigure}
\caption[Diabetic Retinopathy: ResNet-18 loss and accuracy plots.]{Diabetic Retinopathy: ResNet-18 loss and accuracy plots.}
\label{fig:Diabetic Retinopathy: ResNet-18_acc_plt}
\end{figure}

\begin{figure}[htb]
\begin{subfigure}{0.6\textwidth}
\captionsetup{width=0.8\textwidth}
\centering\includegraphics[width=0.6\linewidth]{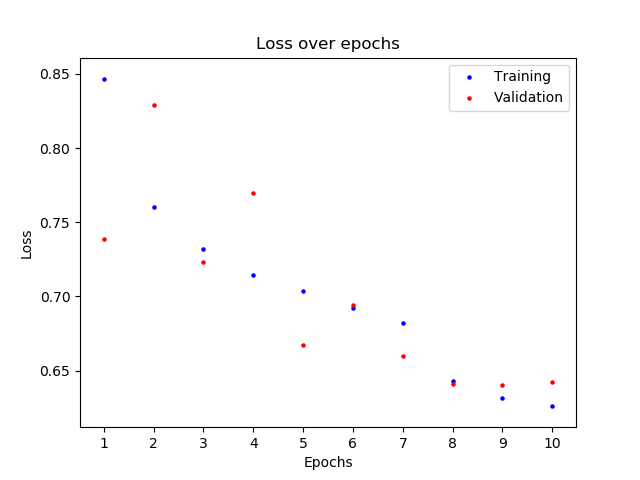}
\caption{Loss; Pretrained; Fine-tuning}
\label{fig:Diabetic Retinopathy: ResNet-34_loss_plt_a}
\end{subfigure}
\begin{subfigure}{0.6\textwidth}
\captionsetup{width=0.8\textwidth}
\centering\includegraphics[width=0.6\linewidth]{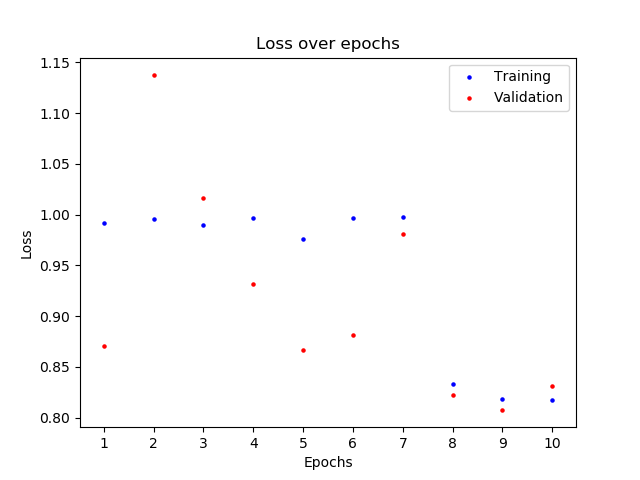}
\caption{Loss; Pretrained; Feature extractor}
\label{fig:Diabetic Retinopathy: ResNet-34_loss_plt_b}
\end{subfigure}
\begin{subfigure}{0.6\textwidth}
\captionsetup{width=0.8\textwidth}
\centering\includegraphics[width=0.6\linewidth]{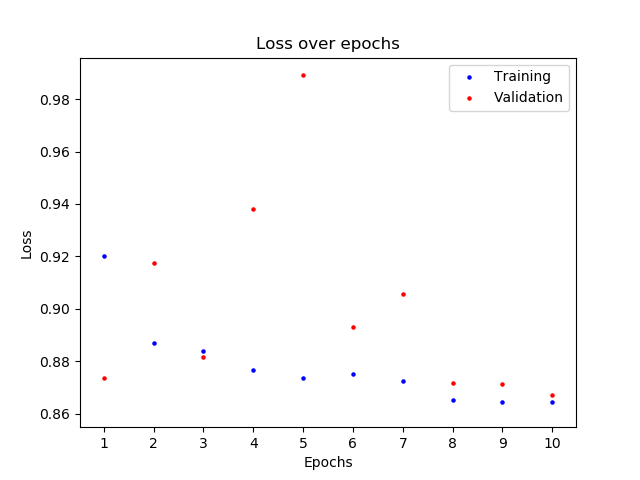}
\caption{Loss; Not Pretrained; Fine-tuning}
\label{fig:Diabetic Retinopathy: ResNet-34_loss_plt_c}
\end{subfigure}
\begin{subfigure}{0.6\textwidth}
\captionsetup{width=0.8\textwidth}
\centering\includegraphics[width=0.6\linewidth]{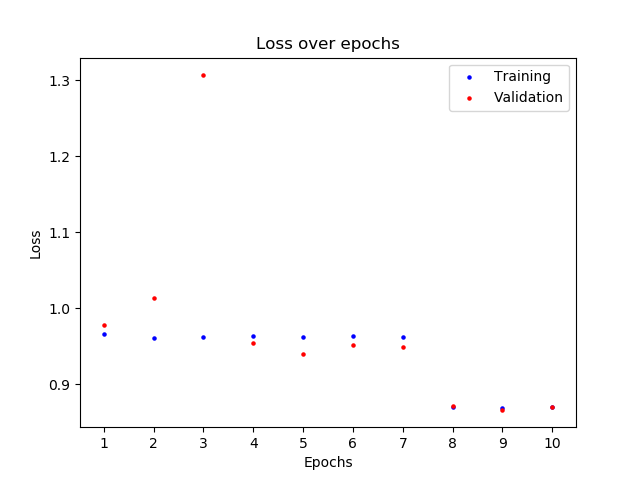}
\caption{Loss; Not Pretrained; Feature extractor}
\label{fig:Diabetic Retinopathy: ResNet-34_loss_plt_d}
\end{subfigure}
\begin{subfigure}{0.6\textwidth}
\captionsetup{width=0.8\textwidth}
\centering\includegraphics[width=0.6\linewidth]{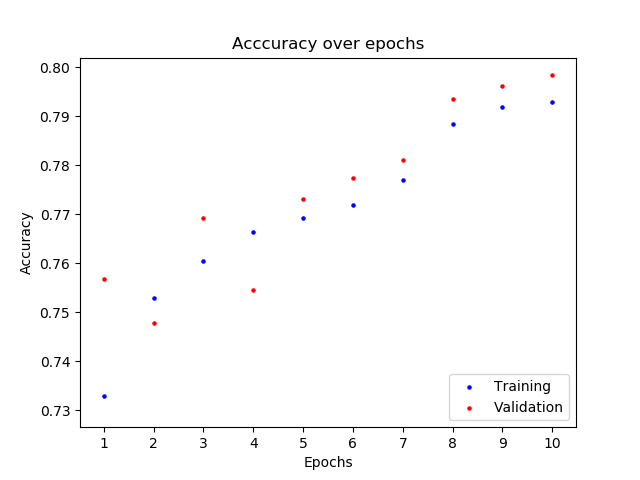}
\caption{Accuracy; Pretrained; Fine-tuning}
\label{fig:Diabetic Retinopathy: ResNet-34_acc_plt_e}
\end{subfigure}
\begin{subfigure}{0.6\textwidth}
\captionsetup{width=0.8\textwidth}
\centering\includegraphics[width=0.6\linewidth]{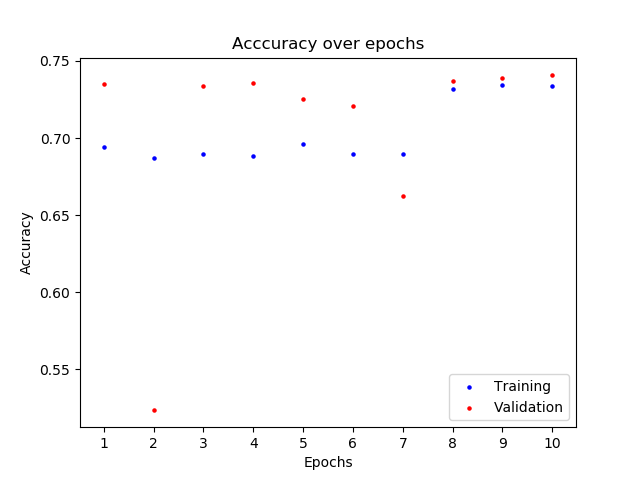}
\caption{Accuracy; Pretrained; Feature extractor}
\label{fig:Diabetic Retinopathy: ResNet-34_acc_plt_f}
\end{subfigure}
\begin{subfigure}{0.6\textwidth}
\captionsetup{width=0.8\textwidth}
\centering\includegraphics[width=0.6\linewidth]{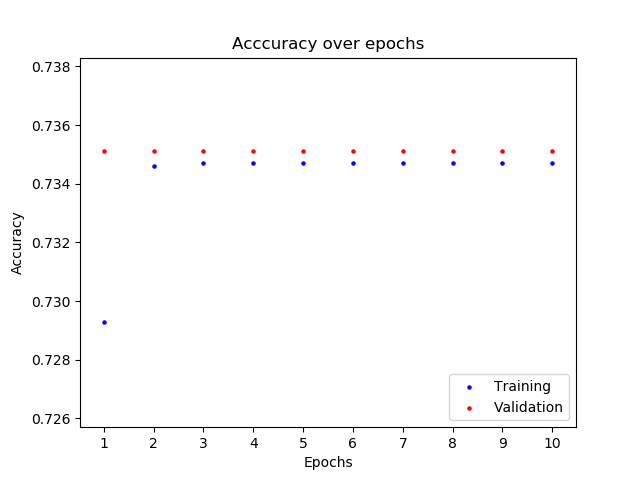}
\caption{Accuracy; Not Pretrained; Fine-tuning}
\label{fig:Diabetic Retinopathy: ResNet-34_acc_plt_g}
\end{subfigure}
\begin{subfigure}{0.6\textwidth}
\captionsetup{width=0.8\textwidth}
\centering\includegraphics[width=0.6\linewidth]{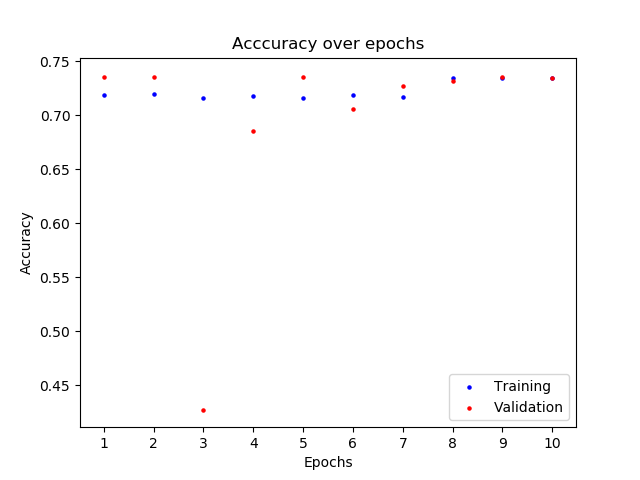}
\caption{Accuracy; Not Pretrained; Feature extractor}
\label{fig:Diabetic Retinopathy: ResNet-34_acc_plt_h}
\end{subfigure}
\caption[Diabetic Retinopathy: ResNet-34 loss and accuracy plots.]{Diabetic Retinopathy: ResNet-34 loss and accuracy plots.}
\label{fig:Diabetic Retinopathy: ResNet-34_acc_plt}
\end{figure}

\begin{figure}[htb]
\begin{subfigure}{0.6\textwidth}
\captionsetup{width=0.8\textwidth}
\centering\includegraphics[width=0.6\linewidth]{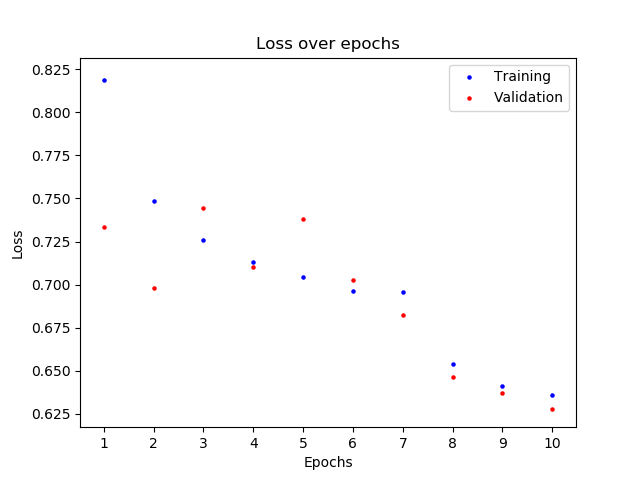}
\caption{Loss; Pretrained; Fine-tuning}
\label{fig:Diabetic Retinopathy: ResNet-50_loss_plt_a}
\end{subfigure}
\begin{subfigure}{0.6\textwidth}
\captionsetup{width=0.8\textwidth}
\centering\includegraphics[width=0.6\linewidth]{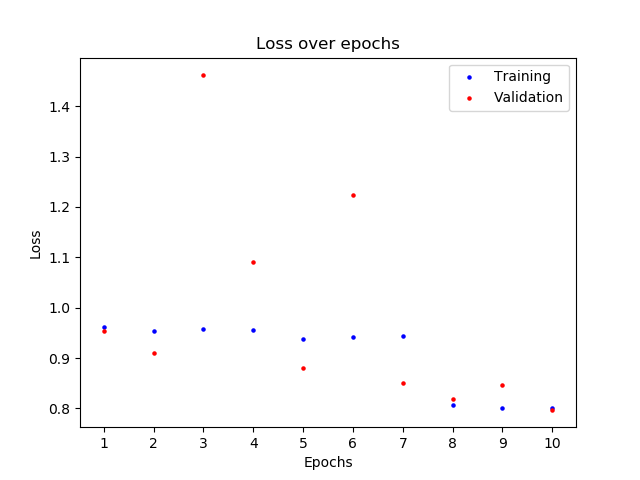}
\caption{Loss; Pretrained; Feature extractor}
\label{fig:Diabetic Retinopathy: ResNet-50_loss_plt_b}
\end{subfigure}
\begin{subfigure}{0.6\textwidth}
\captionsetup{width=0.8\textwidth}
\centering\includegraphics[width=0.6\linewidth]{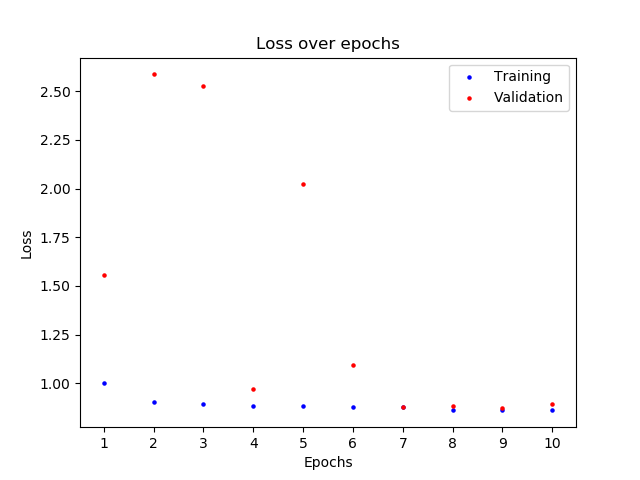}
\caption{Loss; Not Pretrained; Fine-tuning}
\label{fig:Diabetic Retinopathy: ResNet-50_loss_plt_c}
\end{subfigure}
\begin{subfigure}{0.6\textwidth}
\captionsetup{width=0.8\textwidth}
\centering\includegraphics[width=0.6\linewidth]{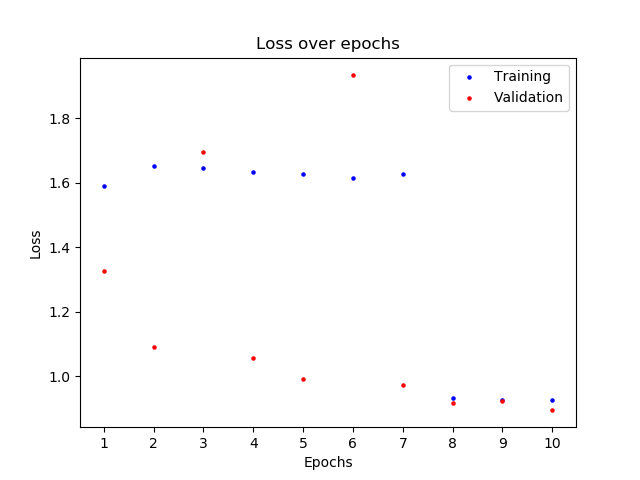}
\caption{Loss; Not Pretrained; Feature extractor}
\label{fig:Diabetic Retinopathy: ResNet-50_loss_plt_d}
\end{subfigure}
\begin{subfigure}{0.6\textwidth}
\captionsetup{width=0.8\textwidth}
\centering\includegraphics[width=0.6\linewidth]{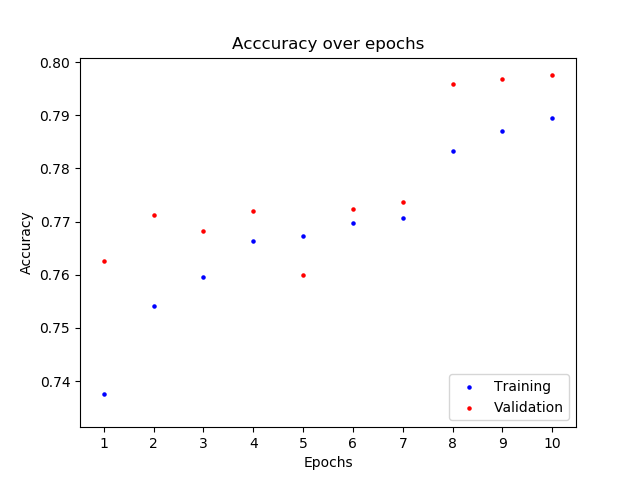}
\caption{Accuracy; Pretrained; Fine-tuning}
\label{fig:Diabetic Retinopathy: ResNet-50_acc_plt_e}
\end{subfigure}
\begin{subfigure}{0.6\textwidth}
\captionsetup{width=0.8\textwidth}
\centering\includegraphics[width=0.6\linewidth]{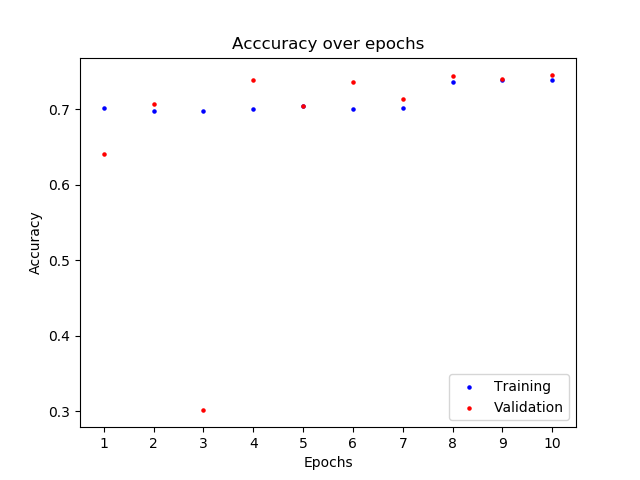}
\caption{Accuracy; Pretrained; Feature extractor}
\label{fig:Diabetic Retinopathy: ResNet-50_acc_plt_f}
\end{subfigure}
\begin{subfigure}{0.6\textwidth}
\captionsetup{width=0.8\textwidth}
\centering\includegraphics[width=0.6\linewidth]{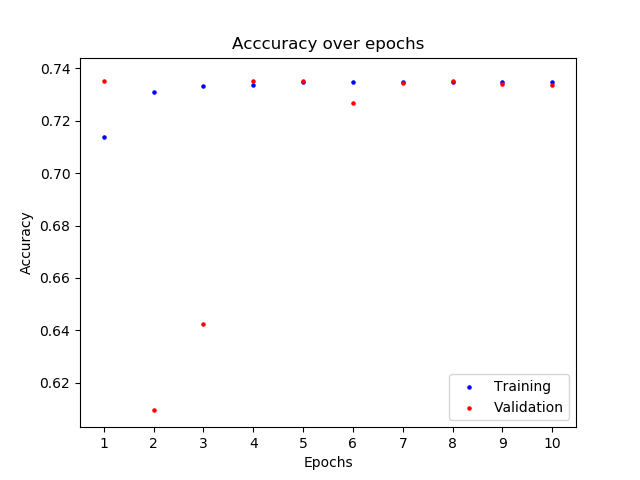}
\caption{Accuracy; Not Pretrained; Fine-tuning}
\label{fig:Diabetic Retinopathy: ResNet-50_acc_plt_g}
\end{subfigure}
\begin{subfigure}{0.6\textwidth}
\captionsetup{width=0.8\textwidth}
\centering\includegraphics[width=0.6\linewidth]{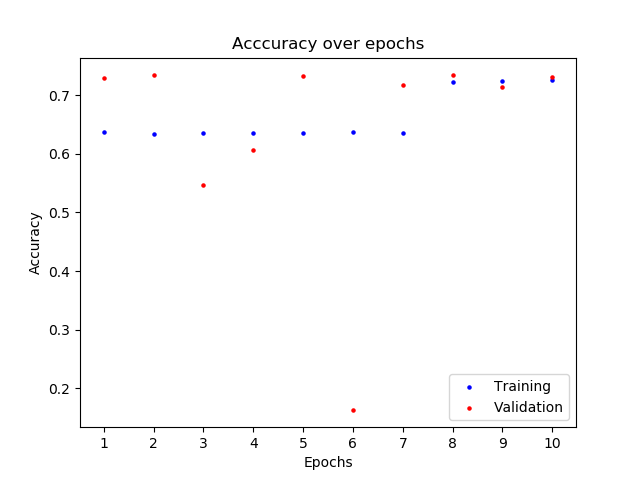}
\caption{Accuracy; Not Pretrained; Feature extractor}
\label{fig:Diabetic Retinopathy: ResNet-50_acc_plt_h}
\end{subfigure}
\caption[Diabetic Retinopathy: ResNet-50 loss and accuracy plots.]{Diabetic Retinopathy: ResNet-50 loss and accuracy plots.}
\label{fig:Diabetic Retinopathy: ResNet-50_acc_plt}
\end{figure}

\begin{figure}[htb]
\begin{subfigure}{0.6\textwidth}
\captionsetup{width=0.8\textwidth}
\centering\includegraphics[width=0.6\linewidth]{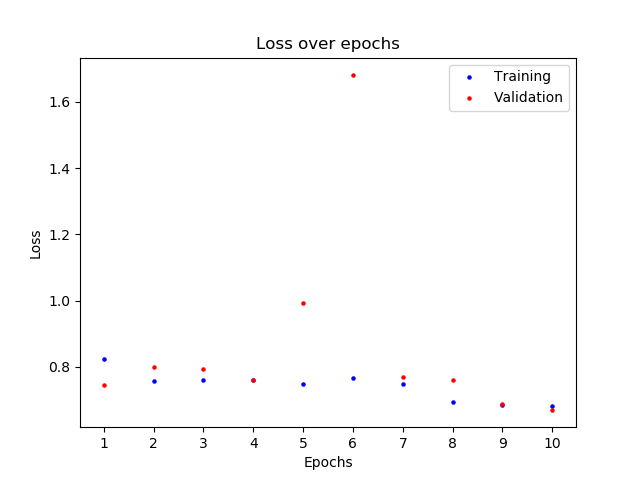}
\caption{Loss; Pretrained; Fine-tuning}
\label{fig:Diabetic Retinopathy: ResNet-101_loss_plt_a}
\end{subfigure}
\begin{subfigure}{0.6\textwidth}
\captionsetup{width=0.8\textwidth}
\centering\includegraphics[width=0.6\linewidth]{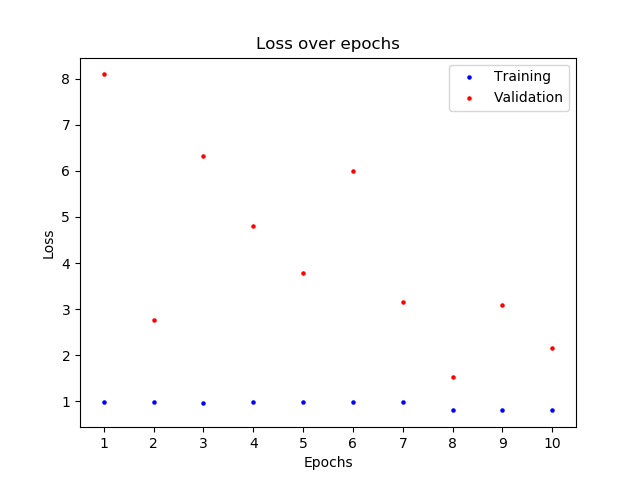}
\caption{Loss; Pretrained; Feature extractor}
\label{fig:Diabetic Retinopathy: ResNet-101_loss_plt_b}
\end{subfigure}
\begin{subfigure}{0.6\textwidth}
\captionsetup{width=0.8\textwidth}
\centering\includegraphics[width=0.6\linewidth]{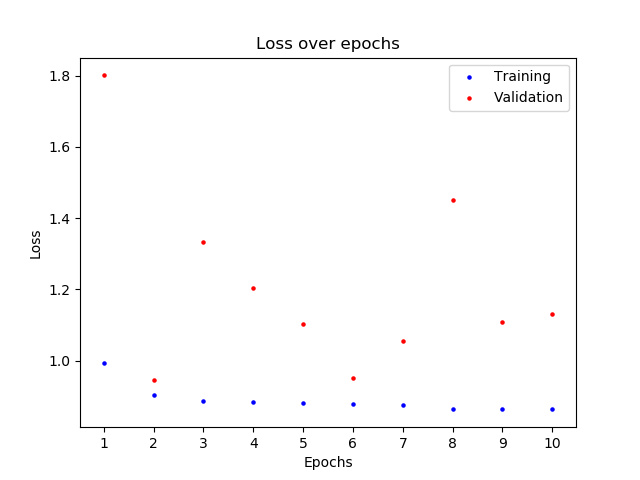}
\caption{Loss; Not Pretrained; Fine-tuning}
\label{fig:Diabetic Retinopathy: ResNet-101_loss_plt_c}
\end{subfigure}
\begin{subfigure}{0.6\textwidth}
\captionsetup{width=0.8\textwidth}
\centering\includegraphics[width=0.6\linewidth]{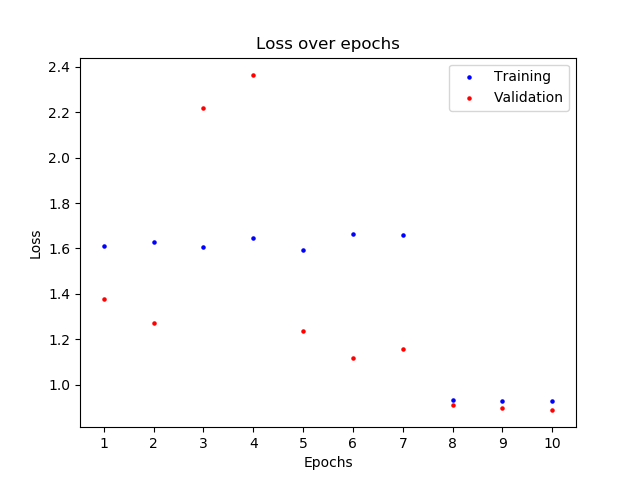}
\caption{Loss; Not Pretrained; Feature extractor}
\label{fig:Diabetic Retinopathy: ResNet-101_loss_plt_d}
\end{subfigure}
\begin{subfigure}{0.6\textwidth}
\captionsetup{width=0.8\textwidth}
\centering\includegraphics[width=0.6\linewidth]{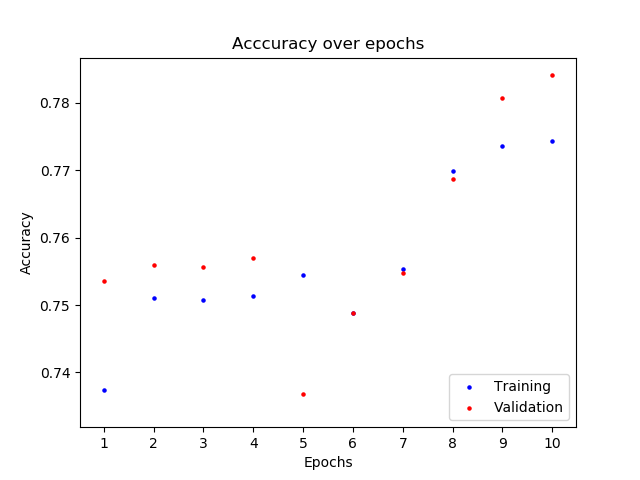}
\caption{Accuracy; Pretrained; Fine-tuning}
\label{fig:Diabetic Retinopathy: ResNet-101_acc_plt_e}
\end{subfigure}
\begin{subfigure}{0.6\textwidth}
\captionsetup{width=0.8\textwidth}
\centering\includegraphics[width=0.6\linewidth]{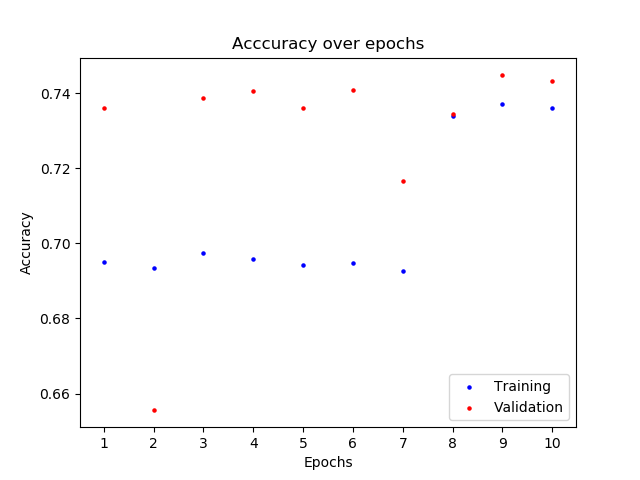}
\caption{Accuracy; Pretrained; Feature extractor}
\label{fig:Diabetic Retinopathy: ResNet-101_acc_plt_f}
\end{subfigure}
\begin{subfigure}{0.6\textwidth}
\captionsetup{width=0.8\textwidth}
\centering\includegraphics[width=0.6\linewidth]{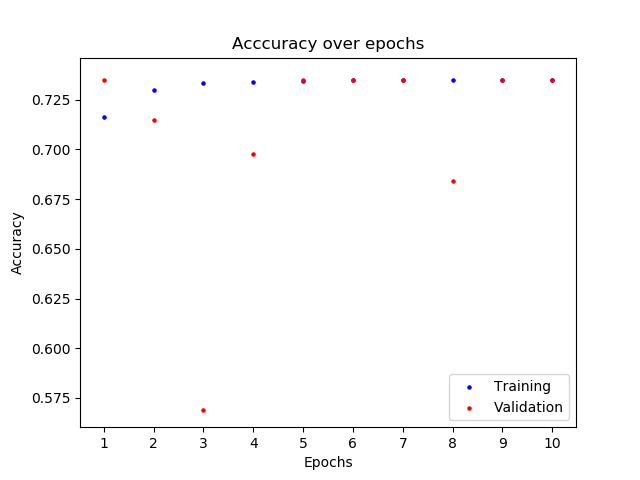}
\caption{Accuracy; Not Pretrained; Fine-tuning}
\label{fig:Diabetic Retinopathy: ResNet-101_acc_plt_g}
\end{subfigure}
\begin{subfigure}{0.6\textwidth}
\captionsetup{width=0.8\textwidth}
\centering\includegraphics[width=0.6\linewidth]{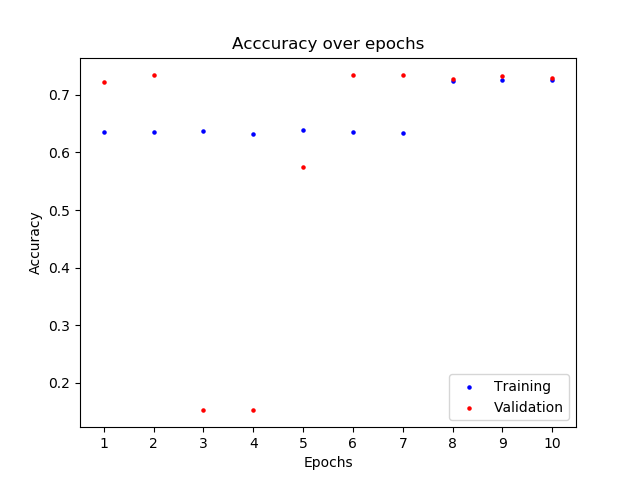}
\caption{Accuracy; Not Pretrained; Feature extractor}
\label{fig:Diabetic Retinopathy: ResNet-101_acc_plt_h}
\end{subfigure}
\caption[Diabetic Retinopathy: ResNet-101 loss and accuracy plots.]{Diabetic Retinopathy: ResNet-101 loss and accuracy plots.}
\label{fig:Diabetic Retinopathy: ResNet-101_acc_plt}
\end{figure}

\begin{figure}[htb]
\begin{subfigure}{0.6\textwidth}
\captionsetup{width=0.8\textwidth}
\centering\includegraphics[width=0.6\linewidth]{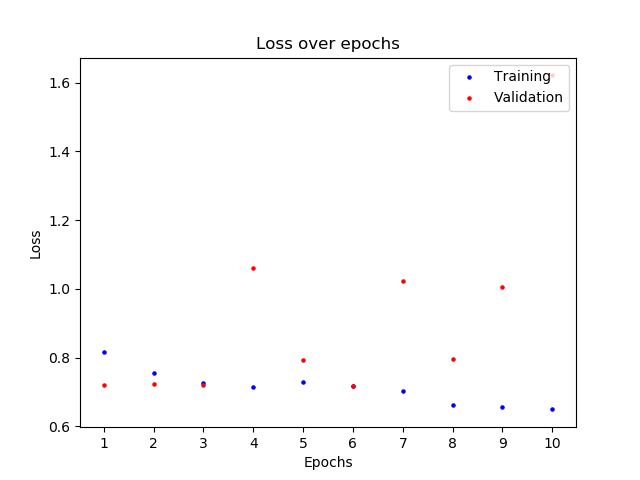}
\caption{Loss; Pretrained; Fine-tuning}
\label{fig:Diabetic Retinopathy: ResNet-152_loss_plt_a}
\end{subfigure}
\begin{subfigure}{0.6\textwidth}
\captionsetup{width=0.8\textwidth}
\centering\includegraphics[width=0.6\linewidth]{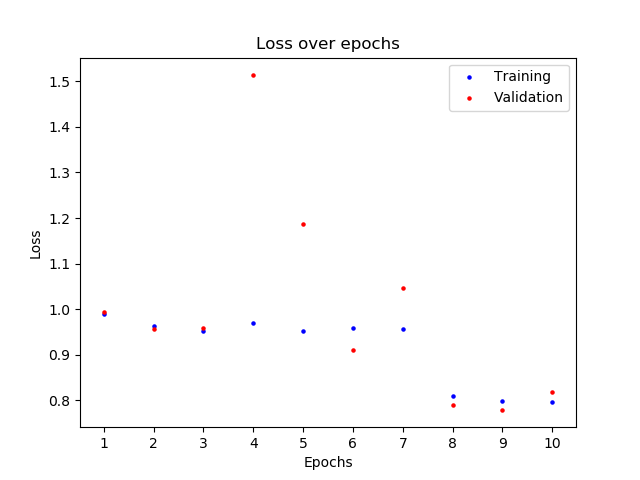}
\caption{Loss; Pretrained; Feature extractor}
\label{fig:Diabetic Retinopathy: ResNet-152_loss_plt_b}
\end{subfigure}
\begin{subfigure}{0.6\textwidth}
\captionsetup{width=0.8\textwidth}
\centering\includegraphics[width=0.6\linewidth]{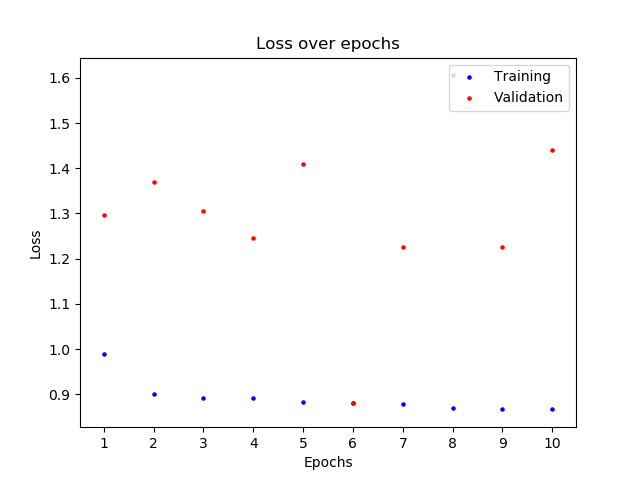}
\caption{Loss; Not Pretrained; Fine-tuning}
\label{fig:Diabetic Retinopathy: ResNet-152_loss_plt_c}
\end{subfigure}
\begin{subfigure}{0.6\textwidth}
\captionsetup{width=0.8\textwidth}
\centering\includegraphics[width=0.6\linewidth]{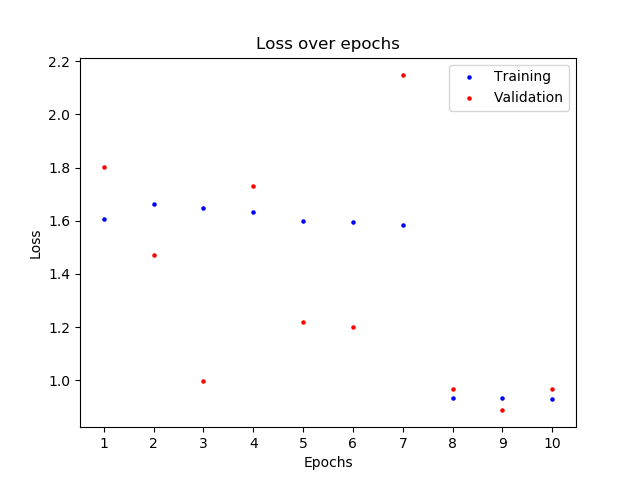}
\caption{Loss; Not Pretrained; Feature extractor}
\label{fig:Diabetic Retinopathy: ResNet-152_loss_plt_d}
\end{subfigure}
\begin{subfigure}{0.6\textwidth}
\captionsetup{width=0.8\textwidth}
\centering\includegraphics[width=0.6\linewidth]{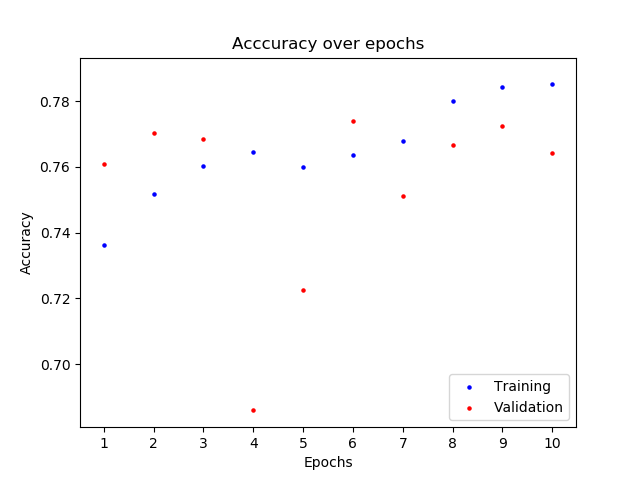}
\caption{Accuracy; Pretrained; Fine-tuning}
\label{fig:Diabetic Retinopathy: ResNet-152_acc_plt_e}
\end{subfigure}
\begin{subfigure}{0.6\textwidth}
\captionsetup{width=0.8\textwidth}
\centering\includegraphics[width=0.6\linewidth]{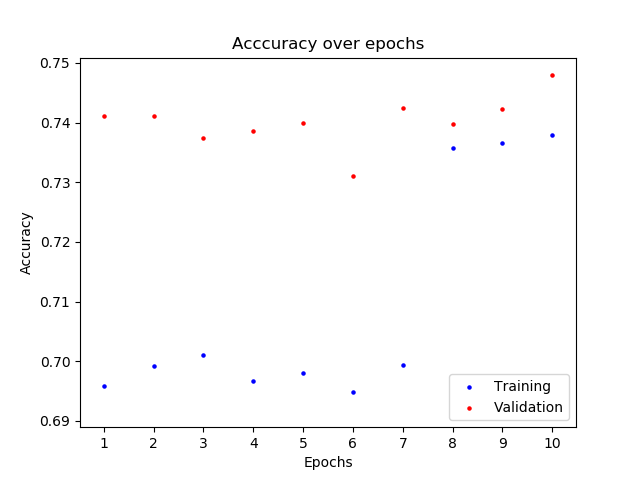}
\caption{Accuracy; Pretrained; Feature extractor}
\label{fig:Diabetic Retinopathy: ResNet-152_acc_plt_f}
\end{subfigure}
\begin{subfigure}{0.6\textwidth}
\captionsetup{width=0.8\textwidth}
\centering\includegraphics[width=0.6\linewidth]{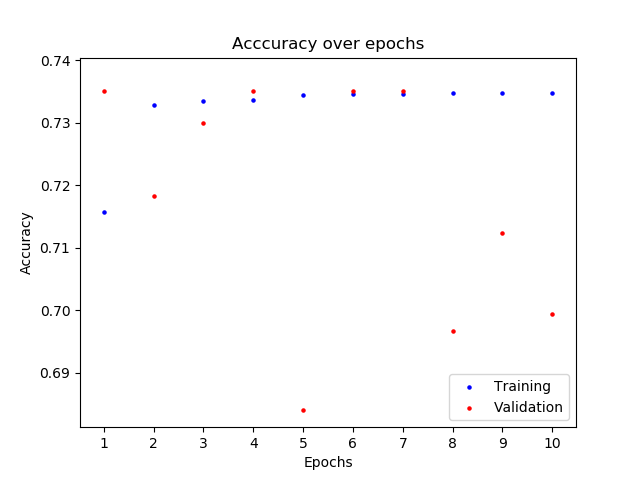}
\caption{Accuracy; Not Pretrained; Fine-tuning}
\label{fig:Diabetic Retinopathy: ResNet-152_acc_plt_g}
\end{subfigure}
\begin{subfigure}{0.6\textwidth}
\captionsetup{width=0.8\textwidth}
\centering\includegraphics[width=0.6\linewidth]{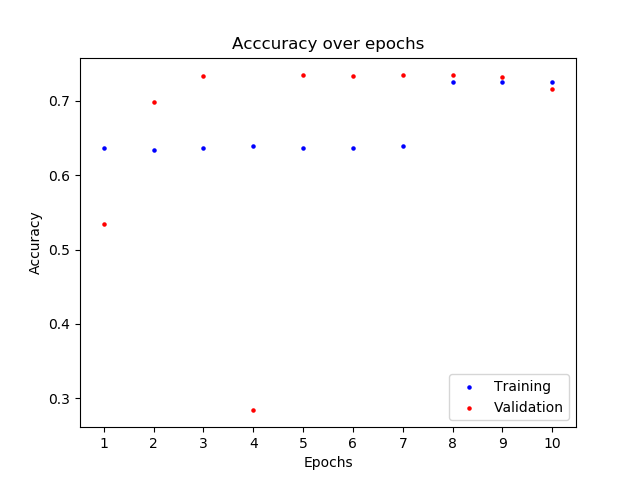}
\caption{Accuracy; Not Pretrained; Feature extractor}
\label{fig:Diabetic Retinopathy: ResNet-152_acc_plt_h}
\end{subfigure}
\caption[Diabetic Retinopathy: ResNet-152 loss and accuracy plots.]{Diabetic Retinopathy: ResNet-152 loss and accuracy plots.}
\label{fig:Diabetic Retinopathy: ResNet-152_acc_plt}
\end{figure}

\begin{figure}[htb]
\begin{subfigure}{0.6\textwidth}
\captionsetup{width=0.8\textwidth}
\centering\includegraphics[width=0.6\linewidth]{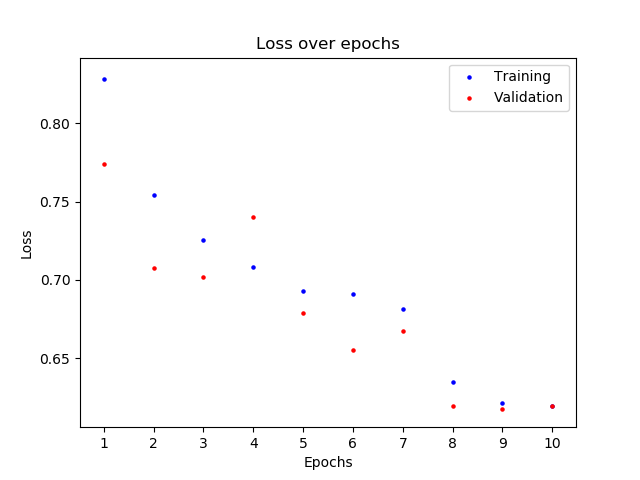}
\caption{Loss; Pretrained; Fine-tuning}
\label{fig:Diabetic Retinopathy: VGG-11_loss_plt_a}
\end{subfigure}
\begin{subfigure}{0.6\textwidth}
\captionsetup{width=0.8\textwidth}
\centering\includegraphics[width=0.6\linewidth]{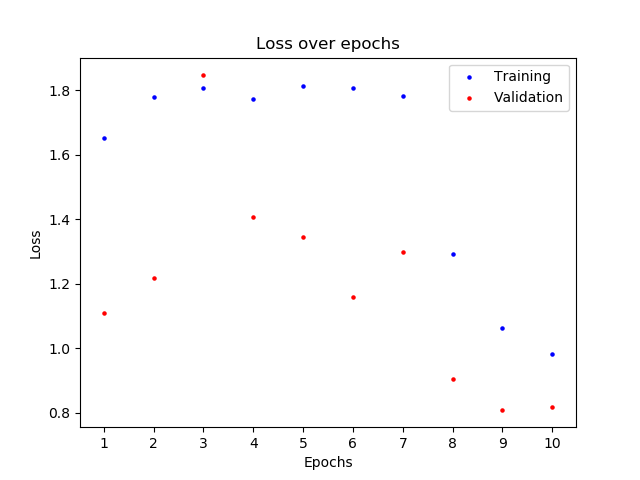}
\caption{Loss; Pretrained; Feature extractor}
\label{fig:Diabetic Retinopathy: VGG-11_loss_plt_b}
\end{subfigure}
\begin{subfigure}{0.6\textwidth}
\captionsetup{width=0.8\textwidth}
\centering\includegraphics[width=0.6\linewidth]{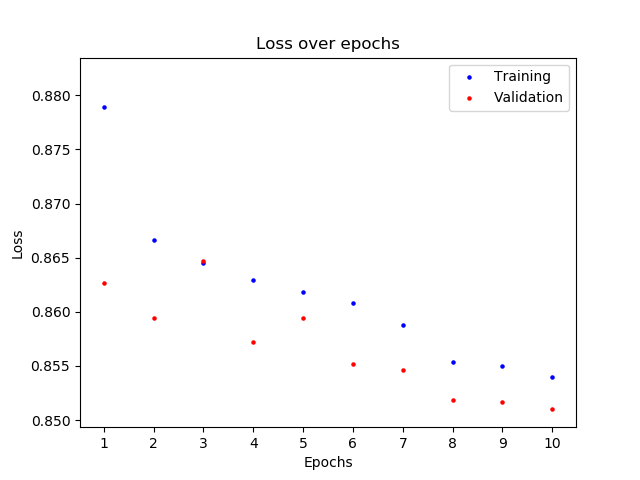}
\caption{Loss; Not Pretrained; Fine-tuning}
\label{fig:Diabetic Retinopathy: VGG-11_loss_plt_c}
\end{subfigure}
\begin{subfigure}{0.6\textwidth}
\captionsetup{width=0.8\textwidth}
\centering\includegraphics[width=0.6\linewidth]{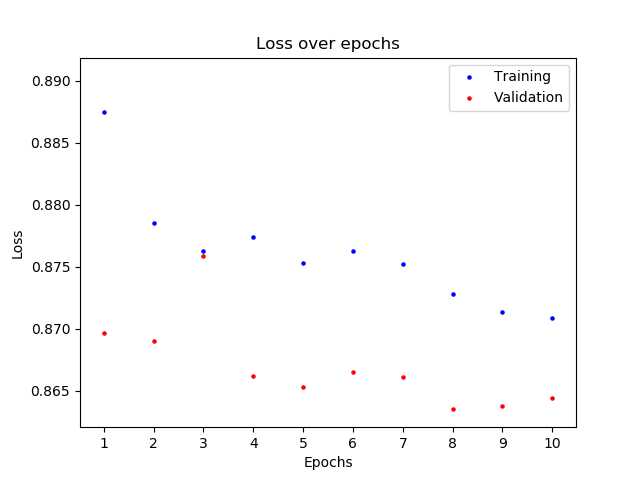}
\caption{Loss; Not Pretrained; Feature extractor}
\label{fig:Diabetic Retinopathy: VGG-11_loss_plt_d}
\end{subfigure}
\begin{subfigure}{0.6\textwidth}
\captionsetup{width=0.8\textwidth}
\centering\includegraphics[width=0.6\linewidth]{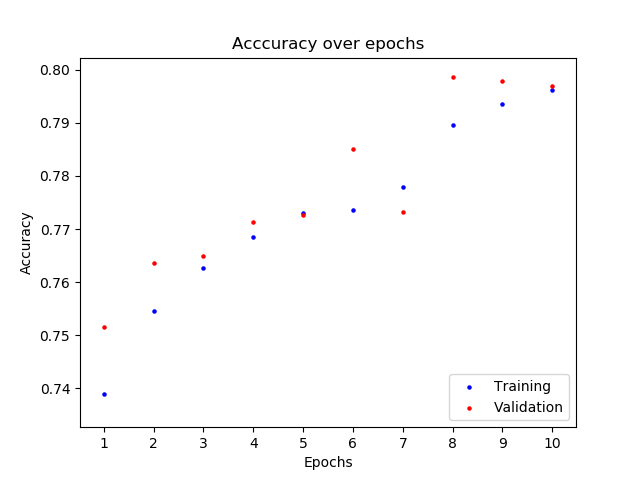}
\caption{Accuracy; Pretrained; Fine-tuning}
\label{fig:Diabetic Retinopathy: VGG-11_acc_plt_e}
\end{subfigure}
\begin{subfigure}{0.6\textwidth}
\captionsetup{width=0.8\textwidth}
\centering\includegraphics[width=0.6\linewidth]{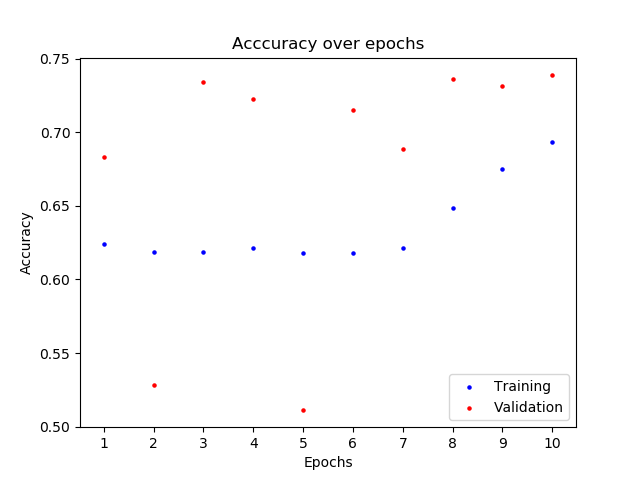}
\caption{Accuracy; Pretrained; Feature extractor}
\label{fig:Diabetic Retinopathy: VGG-11_acc_plt_f}
\end{subfigure}
\begin{subfigure}{0.6\textwidth}
\captionsetup{width=0.8\textwidth}
\centering\includegraphics[width=0.6\linewidth]{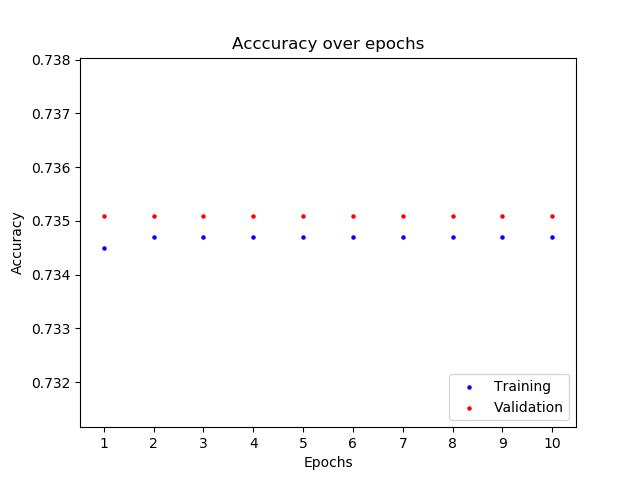}
\caption{Accuracy; Not Pretrained; Fine-tuning}
\label{fig:Diabetic Retinopathy: VGG-11_acc_plt_g}
\end{subfigure}
\begin{subfigure}{0.6\textwidth}
\captionsetup{width=0.8\textwidth}
\centering\includegraphics[width=0.6\linewidth]{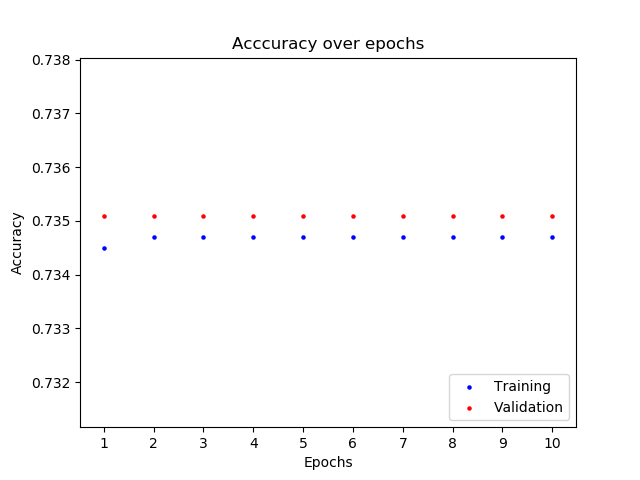}
\caption{Accuracy; Not Pretrained; Feature extractor}
\label{fig:Diabetic Retinopathy: VGG-11_acc_plt_h}
\end{subfigure}
\caption[Diabetic Retinopathy: VGG-11 loss and accuracy plots.]{Diabetic Retinopathy: VGG-11 loss and accuracy plots.}
\label{fig:Diabetic Retinopathy: VGG-11_acc_plt}
\end{figure}

\begin{figure}[htb]
\begin{subfigure}{0.6\textwidth}
\captionsetup{width=0.8\textwidth}
\centering\includegraphics[width=0.6\linewidth]{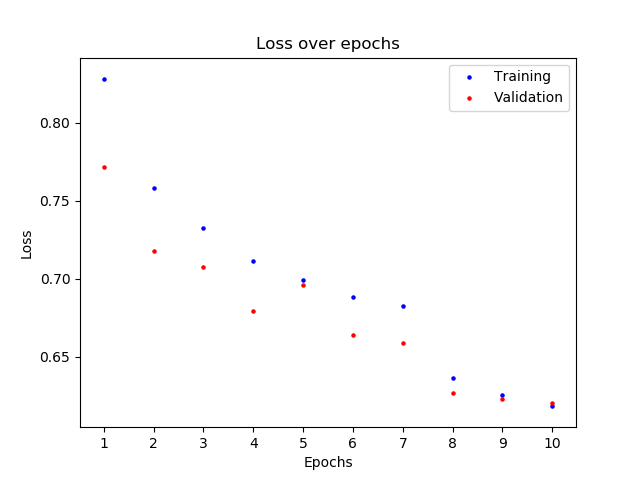}
\caption{Loss; Pretrained; Fine-tuning}
\label{fig:Diabetic Retinopathy: VGG-11bn_loss_plt_a}
\end{subfigure}
\begin{subfigure}{0.6\textwidth}
\captionsetup{width=0.8\textwidth}
\centering\includegraphics[width=0.6\linewidth]{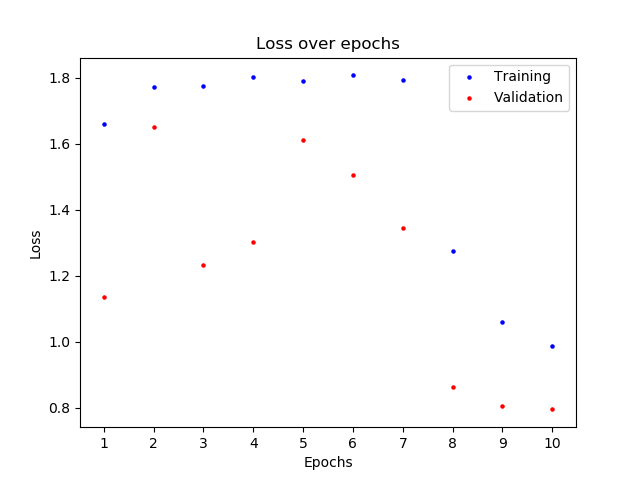}
\caption{Loss; Pretrained; Feature extractor}
\label{fig:Diabetic Retinopathy: VGG-11bn_loss_plt_b}
\end{subfigure}
\begin{subfigure}{0.6\textwidth}
\captionsetup{width=0.8\textwidth}
\centering\includegraphics[width=0.6\linewidth]{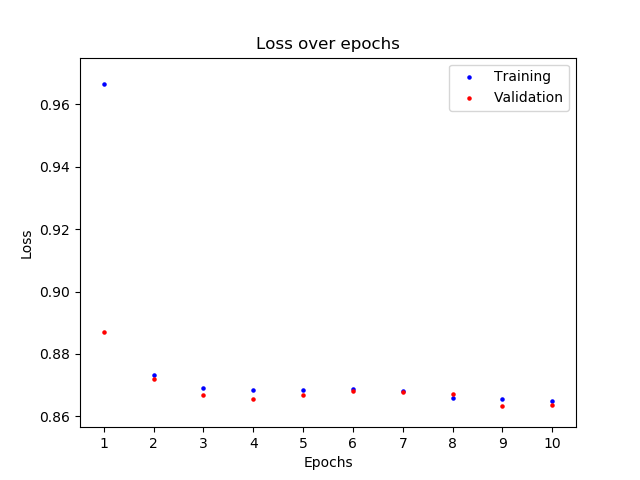}
\caption{Loss; Not Pretrained; Fine-tuning}
\label{fig:Diabetic Retinopathy: VGG-11bn_loss_plt_c}
\end{subfigure}
\begin{subfigure}{0.6\textwidth}
\captionsetup{width=0.8\textwidth}
\centering\includegraphics[width=0.6\linewidth]{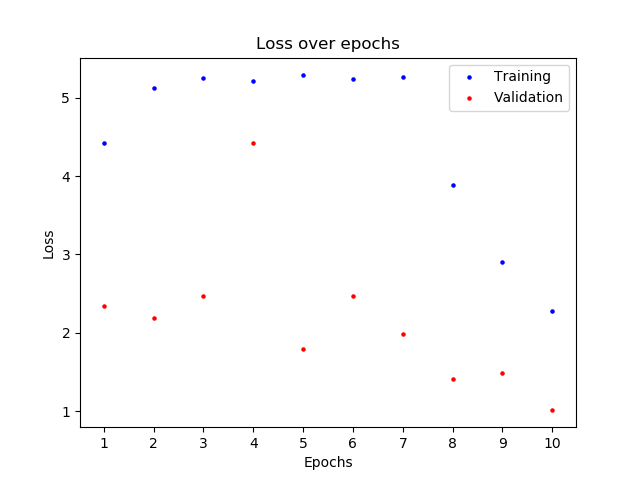}
\caption{Loss; Not Pretrained; Feature extractor}
\label{fig:Diabetic Retinopathy: VGG-11bn_loss_plt_d}
\end{subfigure}
\begin{subfigure}{0.6\textwidth}
\captionsetup{width=0.8\textwidth}
\centering\includegraphics[width=0.6\linewidth]{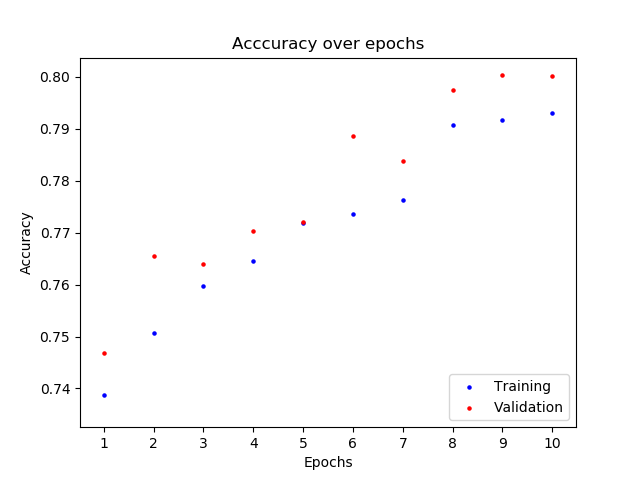}
\caption{Accuracy; Pretrained; Fine-tuning}
\label{fig:Diabetic Retinopathy: VGG-11bn_acc_plt_e}
\end{subfigure}
\begin{subfigure}{0.6\textwidth}
\captionsetup{width=0.8\textwidth}
\centering\includegraphics[width=0.6\linewidth]{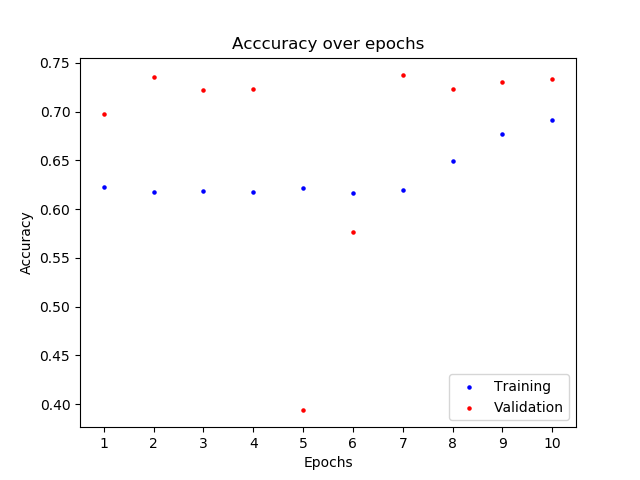}
\caption{Accuracy; Pretrained; Feature extractor}
\label{fig:Diabetic Retinopathy: VGG-11bn_acc_plt_f}
\end{subfigure}
\begin{subfigure}{0.6\textwidth}
\captionsetup{width=0.8\textwidth}
\centering\includegraphics[width=0.6\linewidth]{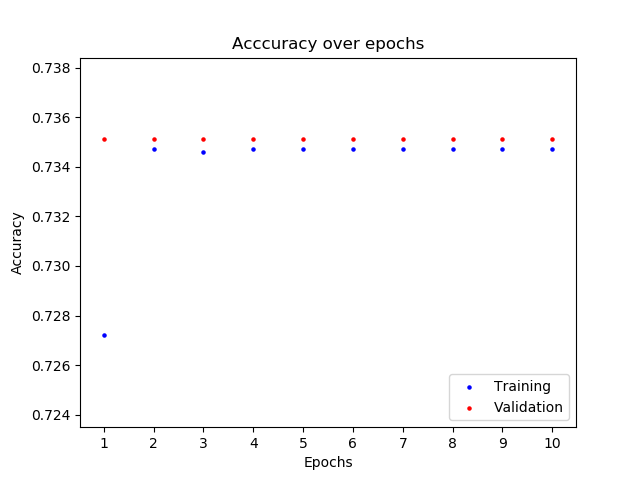}
\caption{Accuracy; Not Pretrained; Fine-tuning}
\label{fig:Diabetic Retinopathy: VGG-11bn_acc_plt_g}
\end{subfigure}
\begin{subfigure}{0.6\textwidth}
\captionsetup{width=0.8\textwidth}
\centering\includegraphics[width=0.6\linewidth]{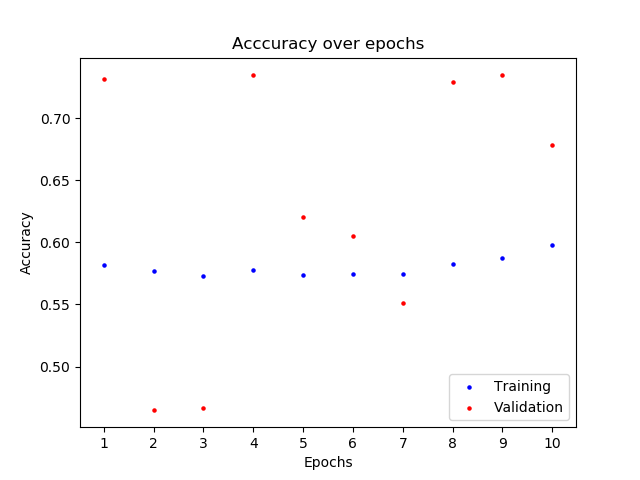}
\caption{Accuracy; Not Pretrained; Feature extractor}
\label{fig:Diabetic Retinopathy: VGG-11bn_acc_plt_h}
\end{subfigure}
\caption[Diabetic Retinopathy: VGG-11bn loss and accuracy plots.]{Diabetic Retinopathy: VGG-11bn loss and accuracy plots.}
\label{fig:Diabetic Retinopathy: VGG-11bn_acc_plt}
\end{figure}

\begin{figure}[htb]
\begin{subfigure}{0.6\textwidth}
\captionsetup{width=0.8\textwidth}
\centering\includegraphics[width=0.6\linewidth]{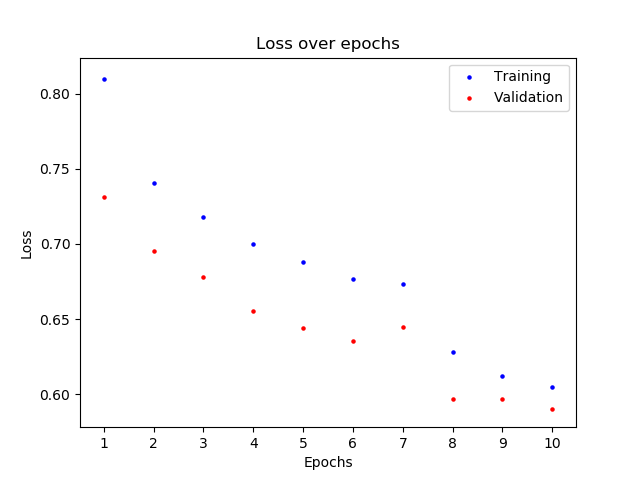}
\caption{Loss; Pretrained; Fine-tuning}
\label{fig:Diabetic Retinopathy: VGG-13_loss_plt_a}
\end{subfigure}
\begin{subfigure}{0.6\textwidth}
\captionsetup{width=0.8\textwidth}
\centering\includegraphics[width=0.6\linewidth]{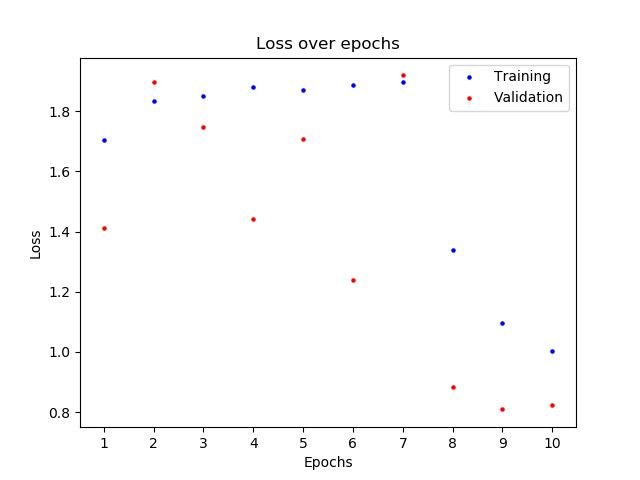}
\caption{Loss; Pretrained; Feature extractor}
\label{fig:Diabetic Retinopathy: VGG-13_loss_plt_b}
\end{subfigure}
\begin{subfigure}{0.6\textwidth}
\captionsetup{width=0.8\textwidth}
\centering\includegraphics[width=0.6\linewidth]{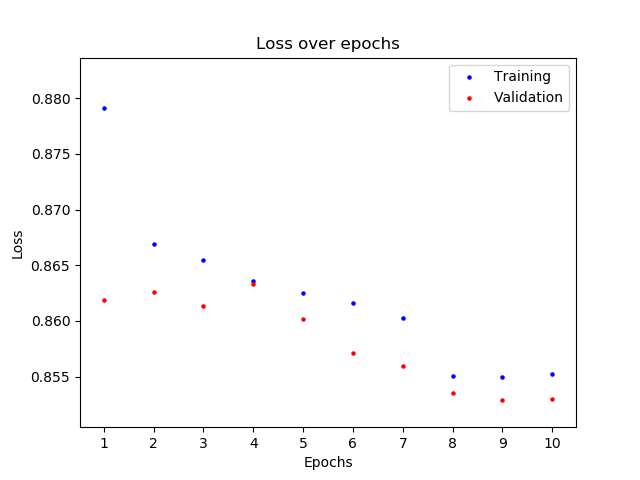}
\caption{Loss; Not Pretrained; Fine-tuning}
\label{fig:Diabetic Retinopathy: VGG-13_loss_plt_c}
\end{subfigure}
\begin{subfigure}{0.6\textwidth}
\captionsetup{width=0.8\textwidth}
\centering\includegraphics[width=0.6\linewidth]{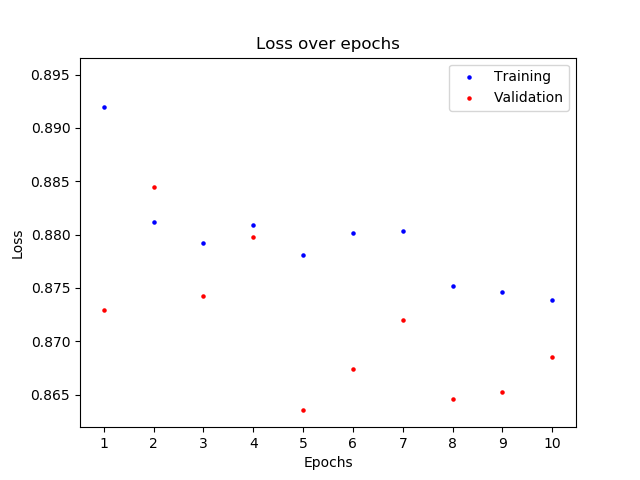}
\caption{Loss; Not Pretrained; Feature extractor}
\label{fig:Diabetic Retinopathy: VGG-13_loss_plt_d}
\end{subfigure}
\begin{subfigure}{0.6\textwidth}
\captionsetup{width=0.8\textwidth}
\centering\includegraphics[width=0.6\linewidth]{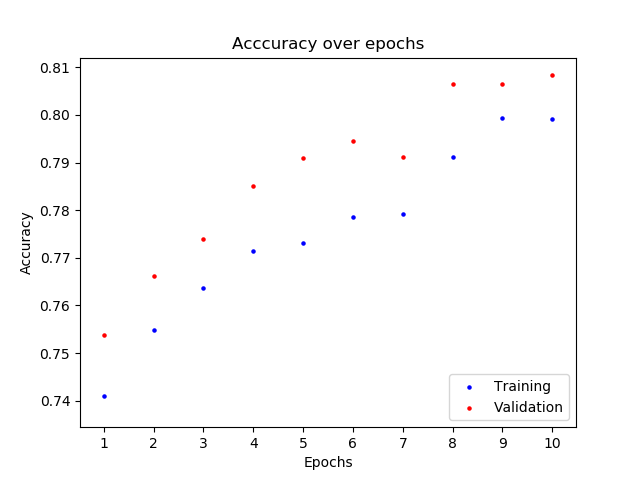}
\caption{Accuracy; Pretrained; Fine-tuning}
\label{fig:Diabetic Retinopathy: VGG-13_acc_plt_e}
\end{subfigure}
\begin{subfigure}{0.6\textwidth}
\captionsetup{width=0.8\textwidth}
\centering\includegraphics[width=0.6\linewidth]{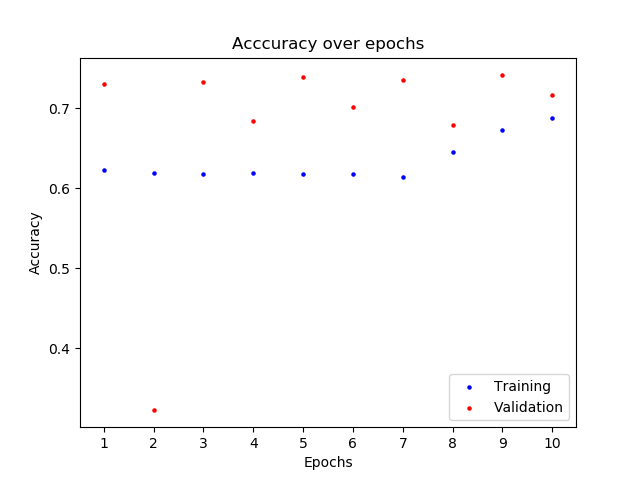}
\caption{Accuracy; Pretrained; Feature extractor}
\label{fig:Diabetic Retinopathy: VGG-13_acc_plt_f}
\end{subfigure}
\begin{subfigure}{0.6\textwidth}
\captionsetup{width=0.8\textwidth}
\centering\includegraphics[width=0.6\linewidth]{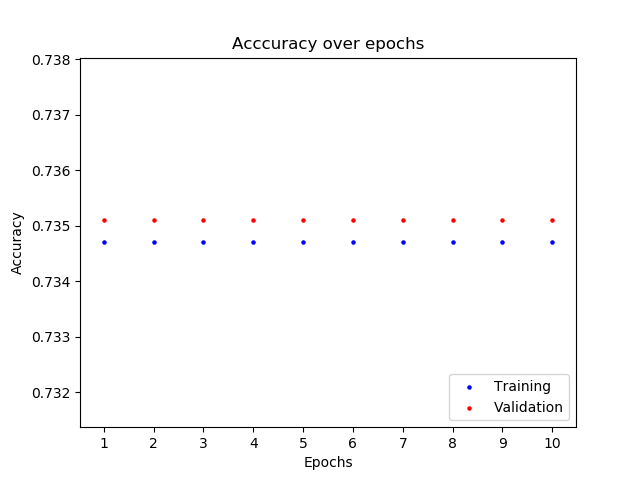}
\caption{Accuracy; Not Pretrained; Fine-tuning}
\label{fig:Diabetic Retinopathy: VGG-13_acc_plt_g}
\end{subfigure}
\begin{subfigure}{0.6\textwidth}
\captionsetup{width=0.8\textwidth}
\centering\includegraphics[width=0.6\linewidth]{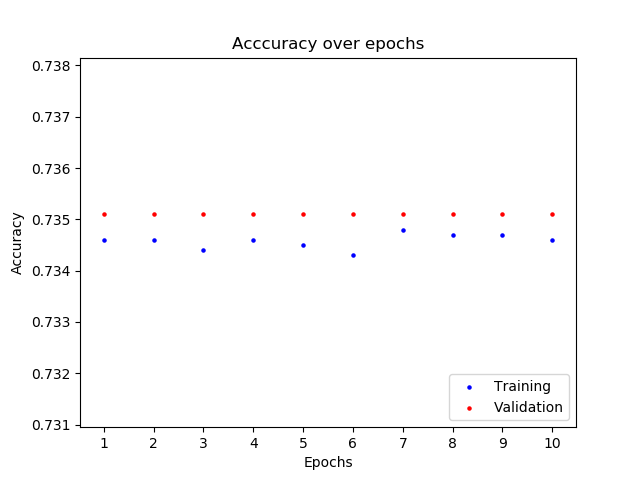}
\caption{Accuracy; Not Pretrained; Feature extractor}
\label{fig:Diabetic Retinopathy: VGG-13_acc_plt_h}
\end{subfigure}
\caption[Diabetic Retinopathy: VGG-13 loss and accuracy plots.]{Diabetic Retinopathy: VGG-13 loss and accuracy plots.}
\label{fig:Diabetic Retinopathy: VGG-13_acc_plt}
\end{figure}

\begin{figure}[htb]
\begin{subfigure}{0.6\textwidth}
\captionsetup{width=0.8\textwidth}
\centering\includegraphics[width=0.6\linewidth]{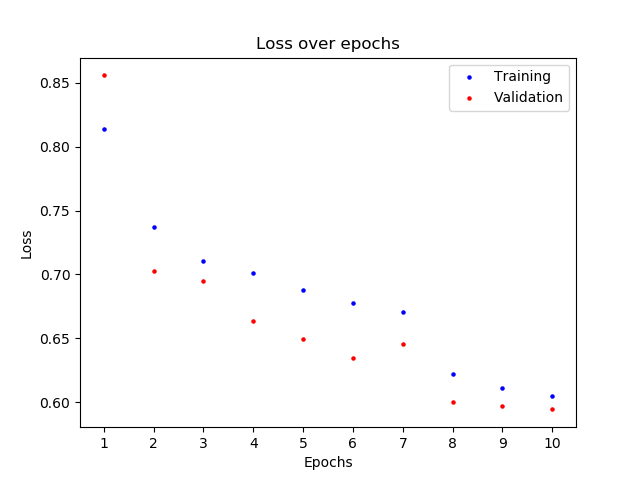}
\caption{Loss; Pretrained; Fine-tuning}
\label{fig:Diabetic Retinopathy: VGG-16_loss_plt_a}
\end{subfigure}
\begin{subfigure}{0.6\textwidth}
\captionsetup{width=0.8\textwidth}
\centering\includegraphics[width=0.6\linewidth]{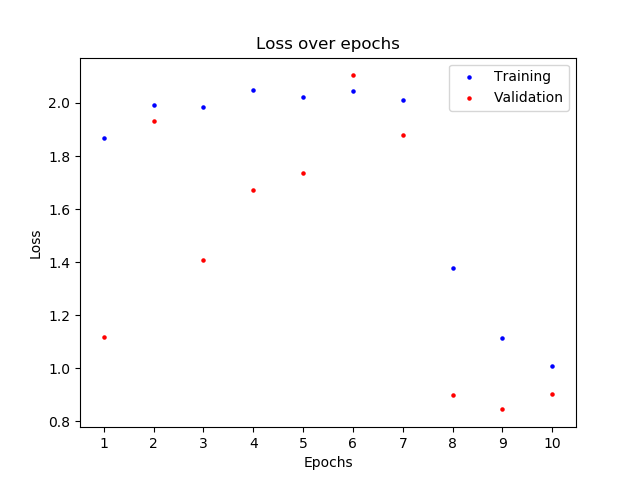}
\caption{Loss; Pretrained; Feature extractor}
\label{fig:Diabetic Retinopathy: VGG-16_loss_plt_b}
\end{subfigure}
\begin{subfigure}{0.6\textwidth}
\captionsetup{width=0.8\textwidth}
\centering\includegraphics[width=0.6\linewidth]{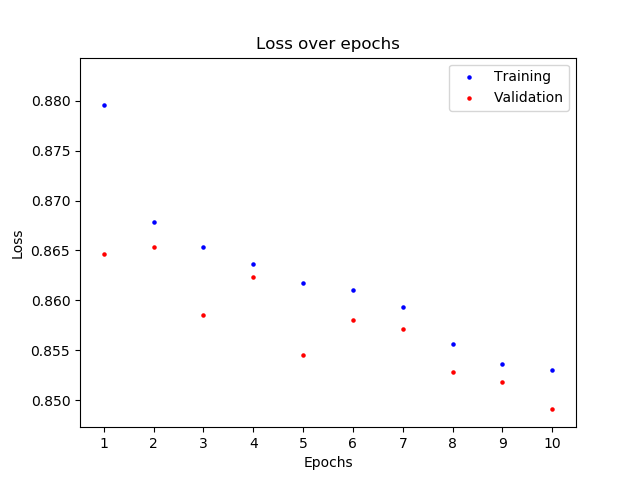}
\caption{Loss; Not Pretrained; Fine-tuning}
\label{fig:Diabetic Retinopathy: VGG-16_loss_plt_c}
\end{subfigure}
\begin{subfigure}{0.6\textwidth}
\captionsetup{width=0.8\textwidth}
\centering\includegraphics[width=0.6\linewidth]{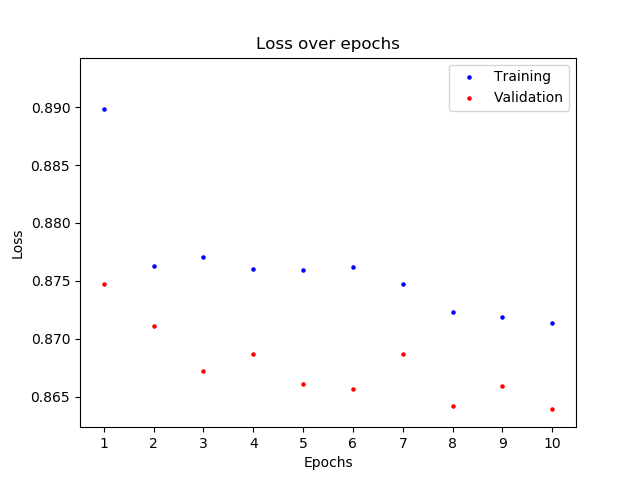}
\caption{Loss; Not Pretrained; Feature extractor}
\label{fig:Diabetic Retinopathy: VGG-16_loss_plt_d}
\end{subfigure}
\begin{subfigure}{0.6\textwidth}
\captionsetup{width=0.8\textwidth}
\centering\includegraphics[width=0.6\linewidth]{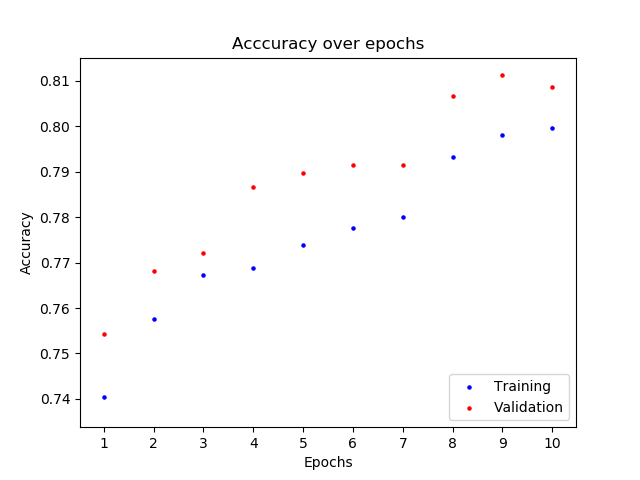}
\caption{Accuracy; Pretrained; Fine-tuning}
\label{fig:Diabetic Retinopathy: VGG-16_acc_plt_e}
\end{subfigure}
\begin{subfigure}{0.6\textwidth}
\captionsetup{width=0.8\textwidth}
\centering\includegraphics[width=0.6\linewidth]{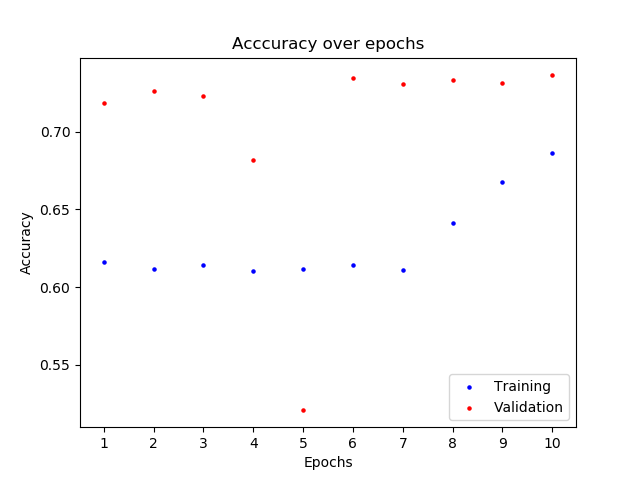}
\caption{Accuracy; Pretrained; Feature extractor}
\label{fig:Diabetic Retinopathy: VGG-16_acc_plt_f}
\end{subfigure}
\begin{subfigure}{0.6\textwidth}
\captionsetup{width=0.8\textwidth}
\centering\includegraphics[width=0.6\linewidth]{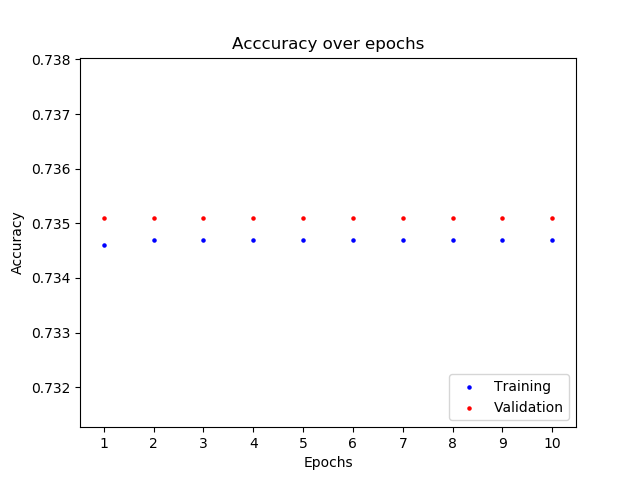}
\caption{Accuracy; Not Pretrained; Fine-tuning}
\label{fig:Diabetic Retinopathy: VGG-16_acc_plt_g}
\end{subfigure}
\begin{subfigure}{0.6\textwidth}
\captionsetup{width=0.8\textwidth}
\centering\includegraphics[width=0.6\linewidth]{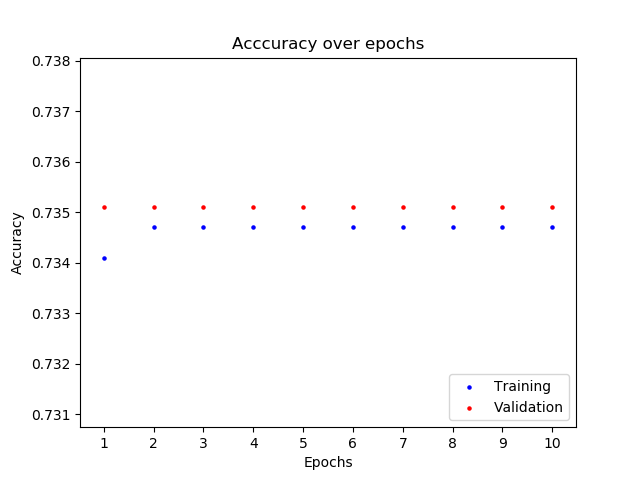}
\caption{Accuracy; Not Pretrained; Feature extractor}
\label{fig:Diabetic Retinopathy: VGG-16_acc_plt_h}
\end{subfigure}
\caption[Diabetic Retinopathy: VGG-16 loss and accuracy plots.]{Diabetic Retinopathy: VGG-16 loss and accuracy plots.}
\label{fig:Diabetic Retinopathy: VGG-16_acc_plt}
\end{figure}

\begin{figure}[htb]
\begin{subfigure}{0.6\textwidth}
\captionsetup{width=0.8\textwidth}
\centering\includegraphics[width=0.6\linewidth]{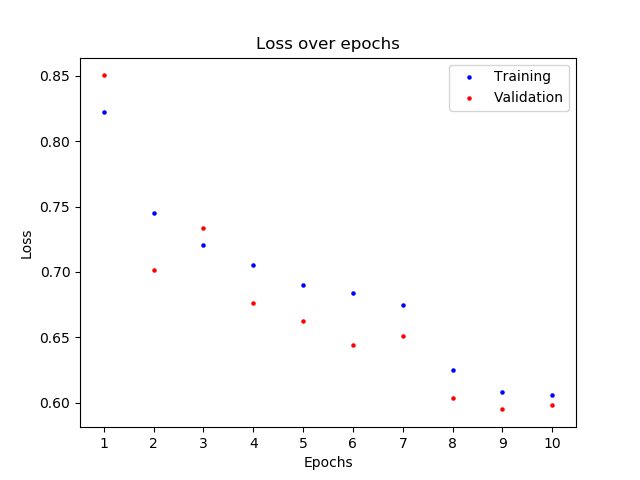}
\caption{Loss; Pretrained; Fine-tuning}
\label{fig:Diabetic Retinopathy: VGG-19_loss_plt_a}
\end{subfigure}
\begin{subfigure}{0.6\textwidth}
\captionsetup{width=0.8\textwidth}
\centering\includegraphics[width=0.6\linewidth]{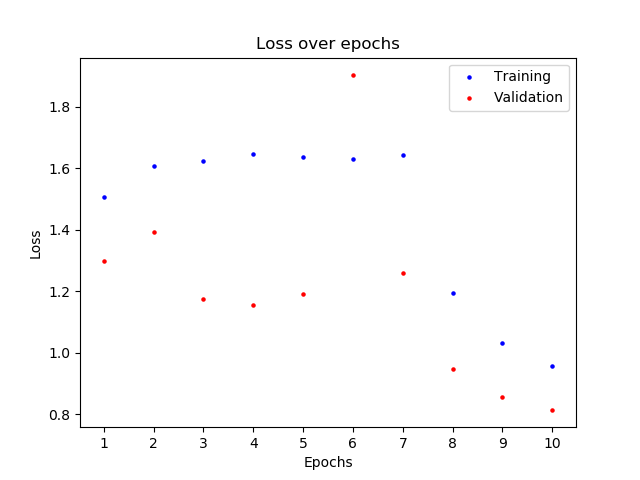}
\caption{Loss; Pretrained; Feature extractor}
\label{fig:Diabetic Retinopathy: VGG-19_loss_plt_b}
\end{subfigure}
\begin{subfigure}{0.6\textwidth}
\captionsetup{width=0.8\textwidth}
\centering\includegraphics[width=0.6\linewidth]{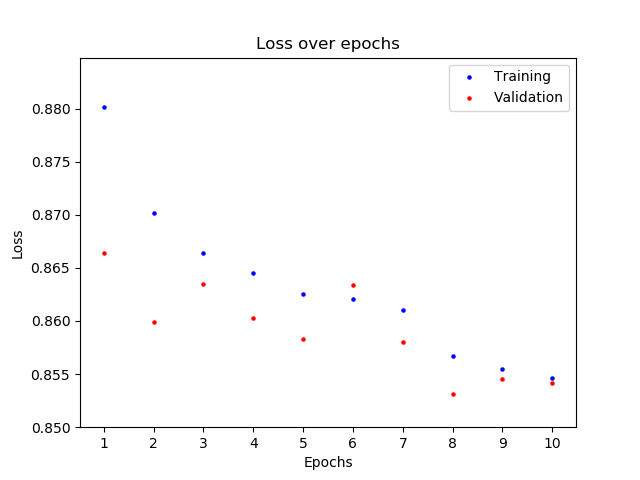}
\caption{Loss; Not Pretrained; Fine-tuning}
\label{fig:Diabetic Retinopathy: VGG-19_loss_plt_c}
\end{subfigure}
\begin{subfigure}{0.6\textwidth}
\captionsetup{width=0.8\textwidth}
\centering\includegraphics[width=0.6\linewidth]{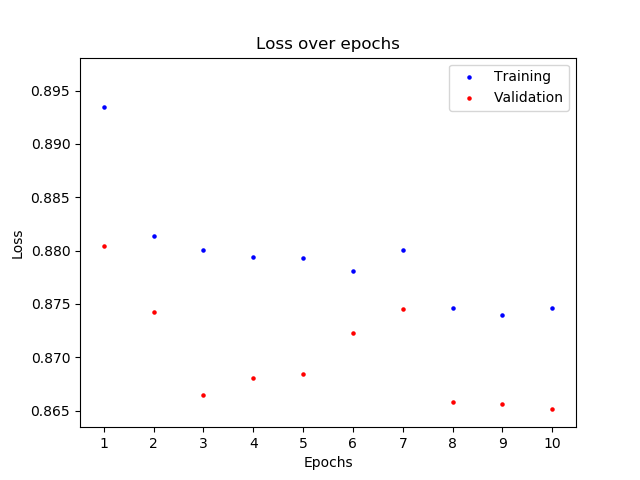}
\caption{Loss; Not Pretrained; Feature extractor}
\label{fig:Diabetic Retinopathy: VGG-19_loss_plt_d}
\end{subfigure}
\begin{subfigure}{0.6\textwidth}
\captionsetup{width=0.8\textwidth}
\centering\includegraphics[width=0.6\linewidth]{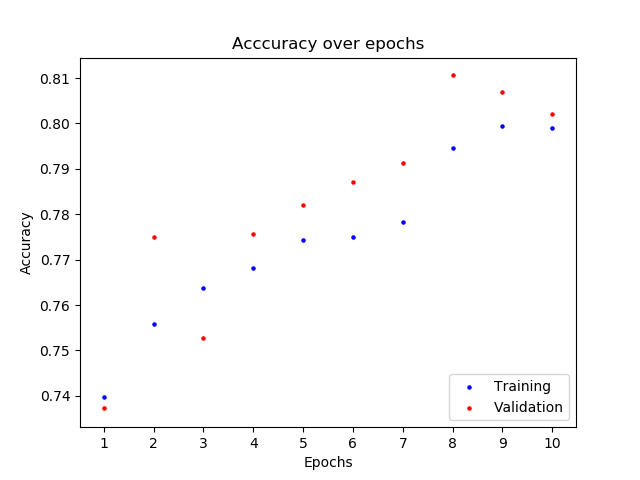}
\caption{Accuracy; Pretrained; Fine-tuning}
\label{fig:Diabetic Retinopathy: VGG-19_acc_plt_e}
\end{subfigure}
\begin{subfigure}{0.6\textwidth}
\captionsetup{width=0.8\textwidth}
\centering\includegraphics[width=0.6\linewidth]{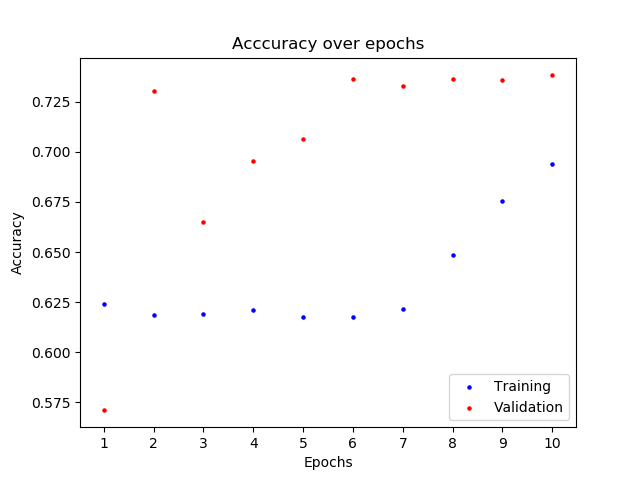}
\caption{Accuracy; Pretrained; Feature extractor}
\label{fig:Diabetic Retinopathy: VGG-19_acc_plt_f}
\end{subfigure}
\begin{subfigure}{0.6\textwidth}
\captionsetup{width=0.8\textwidth}
\centering\includegraphics[width=0.6\linewidth]{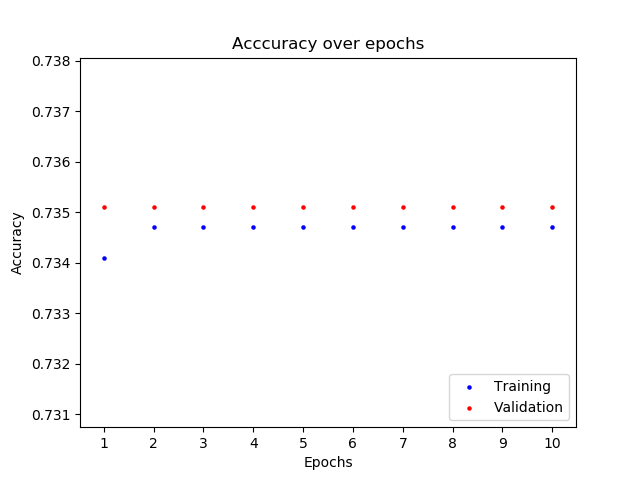}
\caption{Accuracy; Not Pretrained; Fine-tuning}
\label{fig:Diabetic Retinopathy: VGG-19_acc_plt_g}
\end{subfigure}
\begin{subfigure}{0.6\textwidth}
\captionsetup{width=0.8\textwidth}
\centering\includegraphics[width=0.6\linewidth]{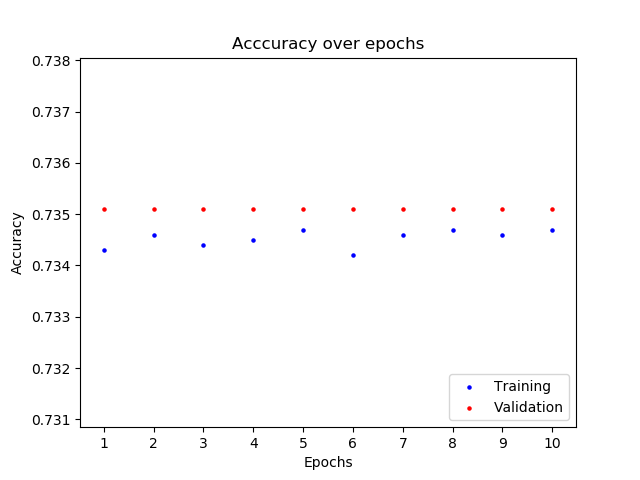}
\caption{Accuracy; Not Pretrained; Feature extractor}
\label{fig:Diabetic Retinopathy: VGG-19_acc_plt_h}
\end{subfigure}
\caption[Diabetic Retinopathy: VGG-19 loss and accuracy plots.]{Diabetic Retinopathy: VGG-19 loss and accuracy plots.}
\label{fig:Diabetic_Retinopathy_VGG-19_acc_plt}
\end{figure}

This appendix is for reference purposes and is organized as so:
\begin{itemize}
\item Figure \ref{fig:DR_network_loss_boxplots} shows boxplots of the loss parameter for the 16 networks that were evaluated. Each plot indicates whether pretrained or not-pretrained networks were used, the phase of training (training or validation), and whether the network was used in fine-tuning or fixed feature extractor mode.
\item Figure \ref{fig:DR_network_acc_boxplots} shows boxplots of the accuracy parameter for the 16 networks that were evaluated. Each plot indicates whether pretrained or not-pretrained networks were used, the phase of training (training or validation), and whether the network was used in fine-tuning or fixed feature extractor mode.
\item Figures \ref{fig:AlexNet_acc_plt}-\ref{fig:Diabetic_Retinopathy_VGG-19_acc_plt} are the detailed plots of loss and accuracy data for each of the 16 networks evaluated (table \ref{tab:networks}). All networks were run for 10 epochs.
\item Figure \ref{fig:DR_loss_boxplots} shows boxplots comparing loss results for two groups (aggregate of the 16 networks), where the groups are defined by a combination of 3 variables: (1) network pretrained or not pretrained; (2) training or validation phase; and (3) used in fine-tuning or fixed feature extractor mode.
\item Figure \ref{fig:DR_acc_boxplots} shows boxplots comparing accuracy results for two groups (aggregate of the 16 networks), where the groups are defined by a combination of 3 variables: (1) network pretrained or not pretrained; (2) training or validation phase; and (3) used in fine-tuning or fixed feature extractor mode.
\end{itemize}

\begin{figure}[htb!]
\begin{subfigure}{0.9\textwidth}
\captionsetup{width=0.9\textwidth}
\centering
\includegraphics[width=0.9\linewidth]{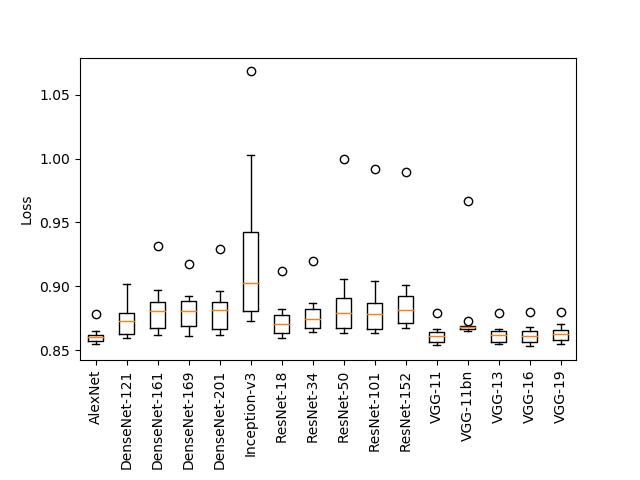}
\caption{Training Loss (Not pretrained, Fine-tuning)}
\label{fig:Diabetic_Retinopathy_loss_boxplot_a}
\end{subfigure}
\begin{subfigure}{0.9\textwidth}
\captionsetup{width=0.9\textwidth}
\centering
\includegraphics[width=0.9\linewidth]{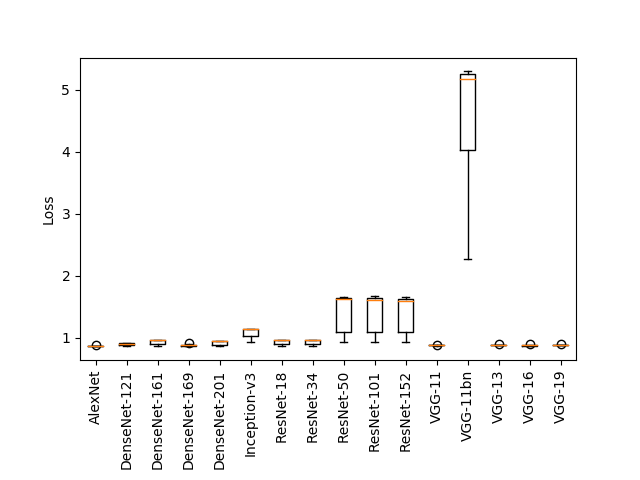}
\caption{Training Loss (Not pretrained, Fixed Feature Extractor)}
\label{fig:Diabetic_Retinopathy_loss_boxplot_b}
\end{subfigure}
\caption[Diabetic Retinopathy: Network loss boxplots.]{Diabetic Retinopathy: Network loss boxplots.}
\label{fig:DR_network_loss_boxplots}
\end{figure}

\begin{figure}[htb!]
\ContinuedFloat
\begin{subfigure}{0.9\textwidth}
\captionsetup{width=0.9\textwidth}
\centering
\includegraphics[width=0.9\linewidth]{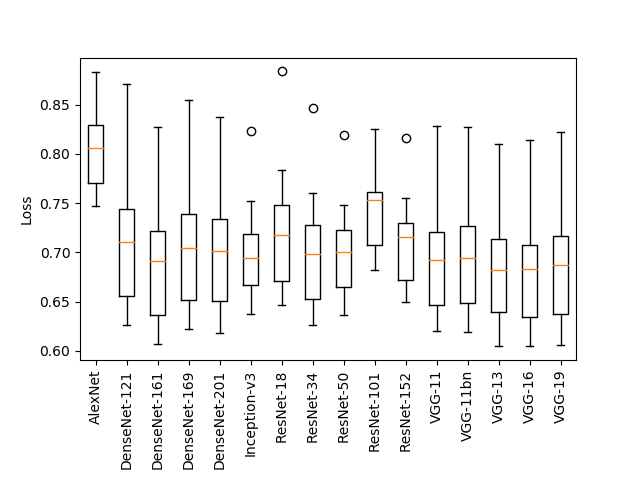}
\caption{Training Loss (Pretrained, Fine-tuning)}
\label{fig:Diabetic_Retinopathy_loss_boxplot_c}
\end{subfigure}
\begin{subfigure}{0.9\textwidth}
\captionsetup{width=0.9\textwidth}
\centering
\includegraphics[width=0.9\linewidth]{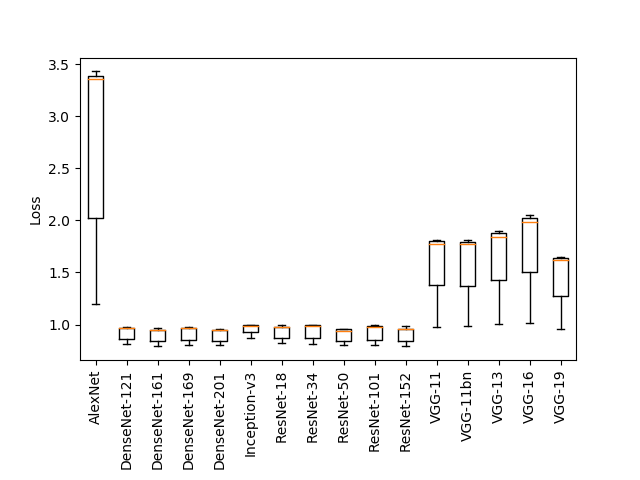}
\caption{Training Loss (Pretrained, Fixed Feature Extractor)}
\label{fig:Diabetic_Retinopathy_loss_boxplot_d}
\end{subfigure}
\caption[Continued - Diabetic Retinopathy: Network loss boxplots.]{Continued - Diabetic Retinopathy: Network loss boxplots.}
\label{fig:DR_network_loss_boxplots2}
\end{figure}

\begin{figure}[htb!]
\ContinuedFloat
\begin{subfigure}{0.9\textwidth}
\captionsetup{width=0.9\textwidth}
\centering
\includegraphics[width=0.9\linewidth]{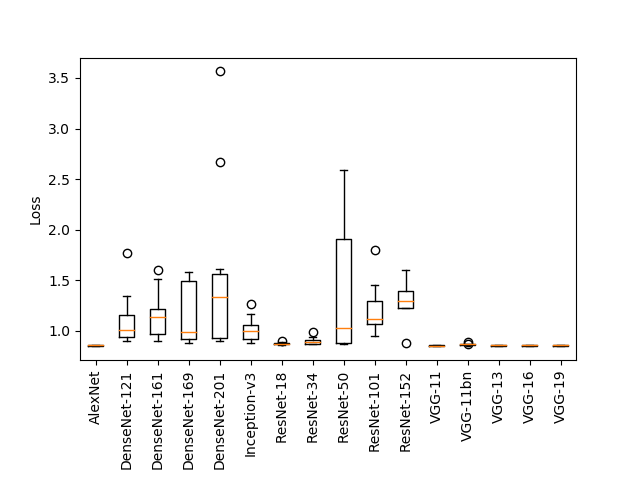}
\caption{Validation Loss (Not pretrained, Fine-tuning)}
\label{fig:Diabetic_Retinopathy_loss_boxplot_e}
\end{subfigure}
\begin{subfigure}{0.9\textwidth}
\captionsetup{width=0.9\textwidth}
\centering
\includegraphics[width=0.9\linewidth]{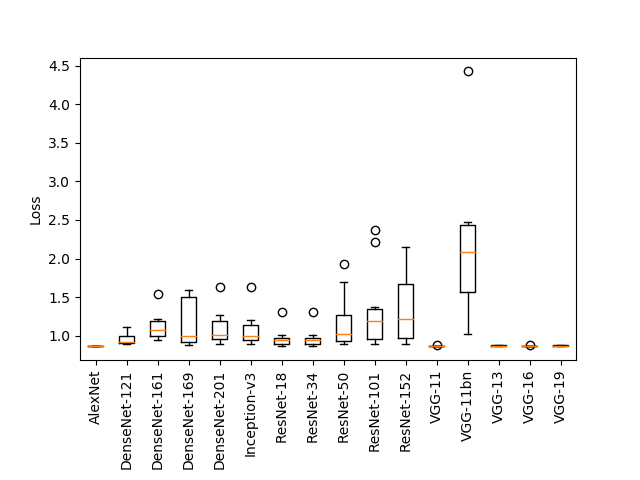}
\caption{Validation Loss (Not pretrained, Fixed Feature Extractor)}
\label{fig:Diabetic_Retinopathy_loss_boxplot_f}
\end{subfigure}
\caption[Continued - Diabetic Retinopathy: Network loss boxplots.]{Continued - Diabetic Retinopathy: Network loss boxplots.}
\label{fig:DR_network_loss_boxplots3}
\end{figure}

\begin{figure}[htb!]
\ContinuedFloat
\begin{subfigure}{0.9\textwidth}
\captionsetup{width=0.9\textwidth}
\centering
\includegraphics[width=0.9\linewidth]{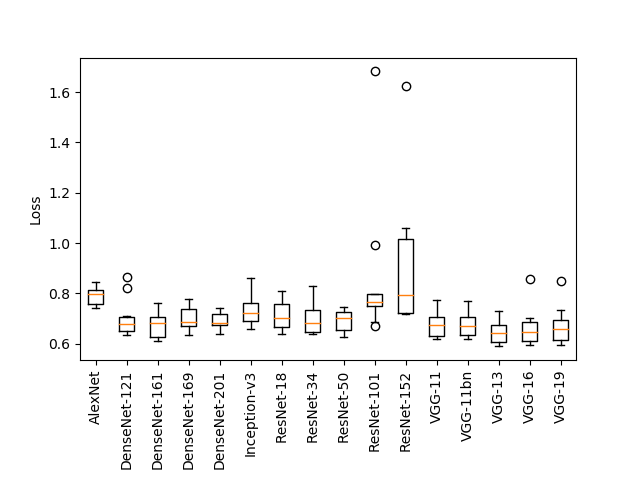}
\caption{Validation Loss (Pretrained, Fine-tuning)}
\label{fig:Diabetic_Retinopathy_loss_boxplot_g}
\end{subfigure}
\begin{subfigure}{0.9\textwidth}
\captionsetup{width=0.9\textwidth}
\centering
\includegraphics[width=0.9\linewidth]{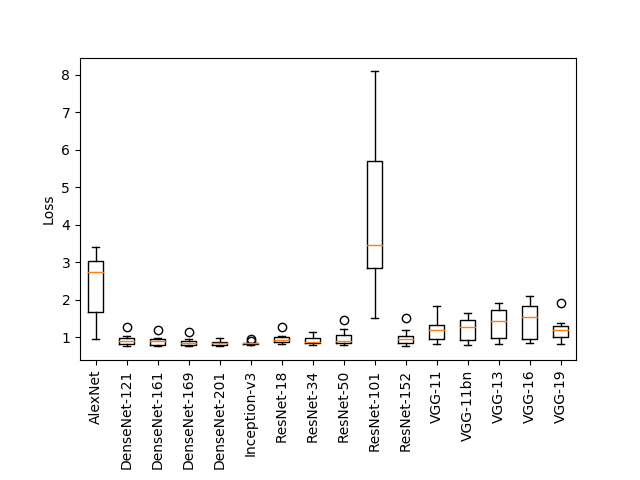}
\caption{Validation Loss (Pretrained, Fixed Feature Extractor)}
\label{fig:Diabetic_Retinopathy_loss_boxplot_h}
\end{subfigure}
\caption[Continued - Diabetic Retinopathy: Network loss boxplots.]{Continued - Diabetic Retinopathy: Network loss boxplots.}
\label{fig:DR_network_loss_boxplots4}
\end{figure}

\begin{figure}[htb!]
\begin{subfigure}{0.9\textwidth}
\captionsetup{width=0.9\textwidth}
\centering
\includegraphics[width=0.9\linewidth]{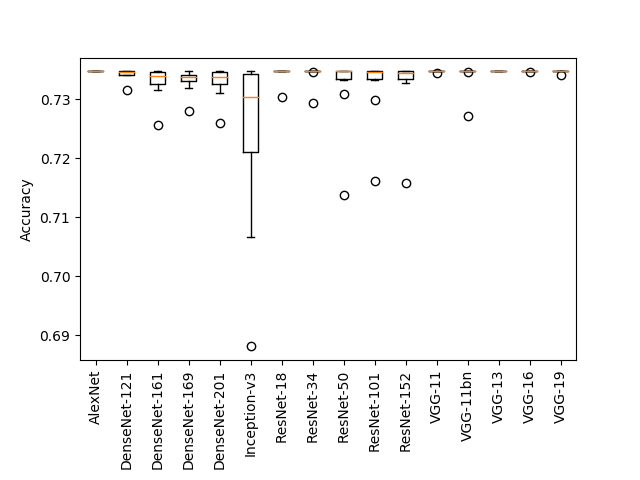}
\caption{Training Accuracy (Not pretrained, Fine-tuning)}
\label{fig:Diabetic_Retinopathy_acc_boxplot_a}
\end{subfigure}
\begin{subfigure}{0.9\textwidth}
\captionsetup{width=0.9\textwidth}
\centering
\includegraphics[width=0.9\linewidth]{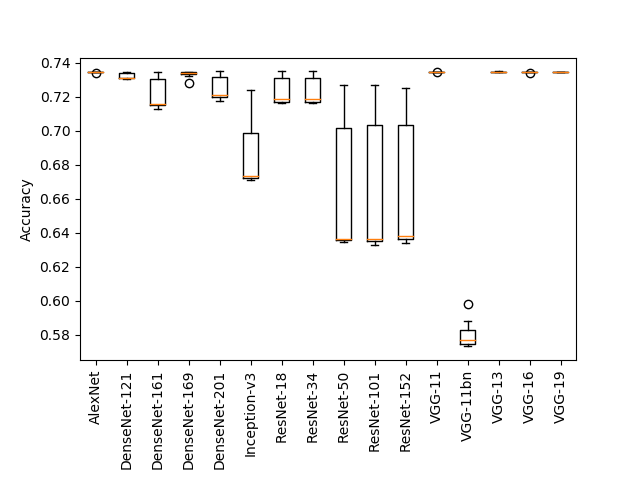}
\caption{Training Accuracy (Not pretrained, Fixed Feature Extractor)}
\label{fig:Diabetic_Retinopathy_acc_boxplot_b}
\end{subfigure}
\caption[Diabetic Retinopathy: Network accuracy boxplots.]{Diabetic Retinopathy: Network accuracy boxplots.}
\label{fig:DR_network_acc_boxplots}
\end{figure}

\begin{figure}[htb!]
\ContinuedFloat
\begin{subfigure}{0.9\textwidth}
\captionsetup{width=0.9\textwidth}
\centering
\includegraphics[width=0.9\linewidth]{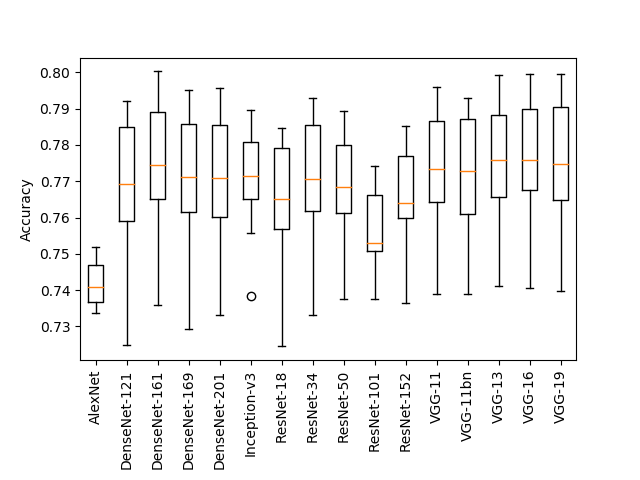}
\caption{Training Accuracy (Pretrained, Fine-tuning)}
\label{fig:Diabetic_Retinopathy_acc_boxplot_c}
\end{subfigure}
\begin{subfigure}{0.9\textwidth}
\captionsetup{width=0.9\textwidth}
\centering
\includegraphics[width=0.9\linewidth]{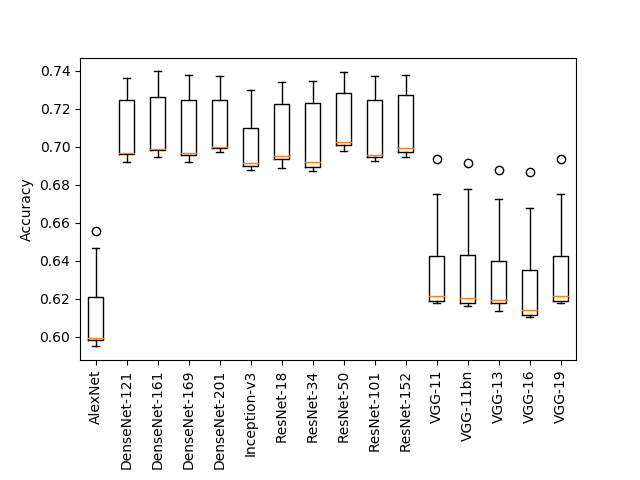}
\caption{Training Accuracy (Pretrained, Fixed Feature Extractor)}
\label{fig:Diabetic_Retinopathy_acc_boxplot_d}
\end{subfigure}
\caption[Continued - Diabetic Retinopathy: Network accuracy boxplots.]{Continued - Diabetic Retinopathy: Network accuracy boxplots.}
\label{fig:DR_network_acc_boxplots2}
\end{figure}

\begin{figure}[htb!]
\ContinuedFloat
\begin{subfigure}{0.9\textwidth}
\captionsetup{width=0.9\textwidth}
\centering
\includegraphics[width=0.9\linewidth]{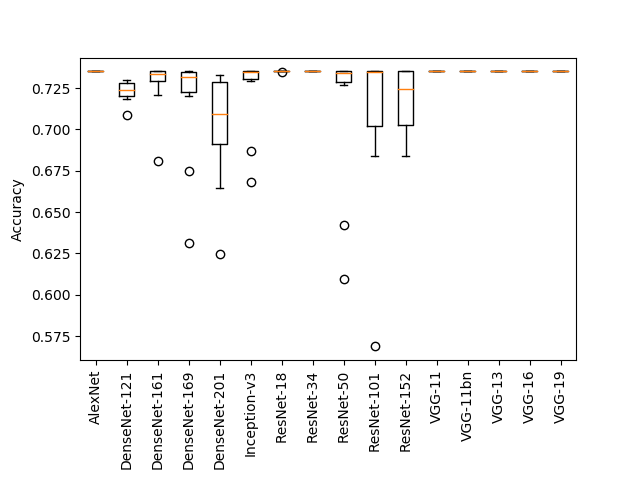}
\caption{Validation Accuracy (Not pretrained, Fine-tuning)}
\label{fig:Diabetic_Retinopathy_acc_boxplot_e}
\end{subfigure}
\begin{subfigure}{0.9\textwidth}
\captionsetup{width=0.9\textwidth}
\centering
\includegraphics[width=0.9\linewidth]{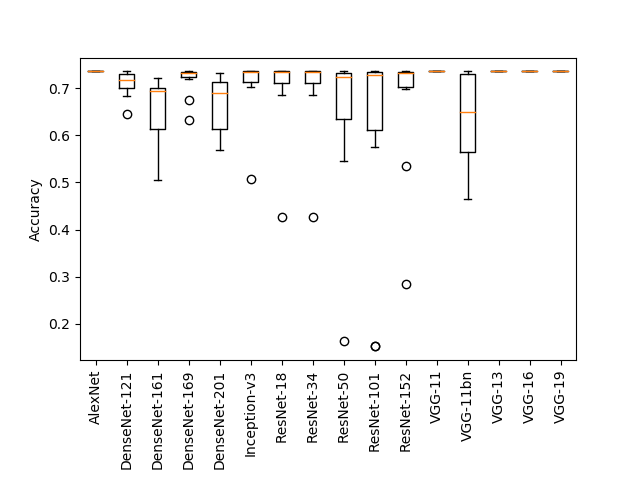}
\caption{Validation Accuracy (Not pretrained, Fixed Feature Extractor)}
\label{fig:Diabetic_Retinopathy_acc_boxplot_f}
\end{subfigure}
\caption[Continued - Diabetic Retinopathy: Network accuracy boxplots.]{Continued - Diabetic Retinopathy: Network accuracy boxplots.}
\label{fig:DR_network_acc_boxplots3}
\end{figure}

\begin{figure}[htb!]
\ContinuedFloat
\begin{subfigure}{0.9\textwidth}
\captionsetup{width=0.9\textwidth}
\centering
\includegraphics[width=0.9\linewidth]{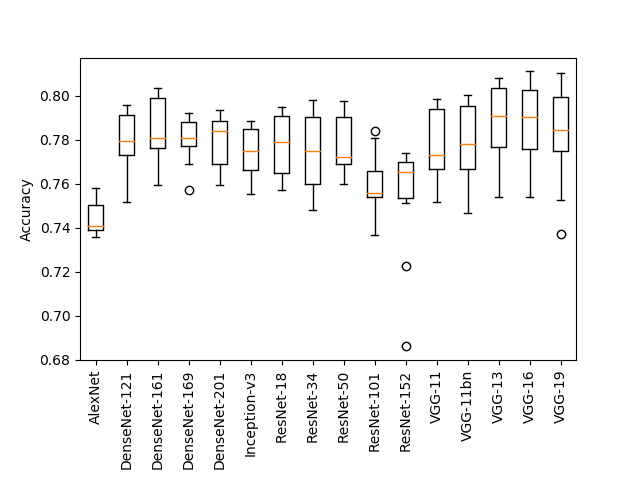}
\caption{Validation Accuracy (Pretrained, Fine-tuning)}
\label{fig:Diabetic_Retinopathy_acc_boxplot_g}
\end{subfigure}
\begin{subfigure}{0.9\textwidth}
\captionsetup{width=0.9\textwidth}
\centering
\includegraphics[width=0.9\linewidth]{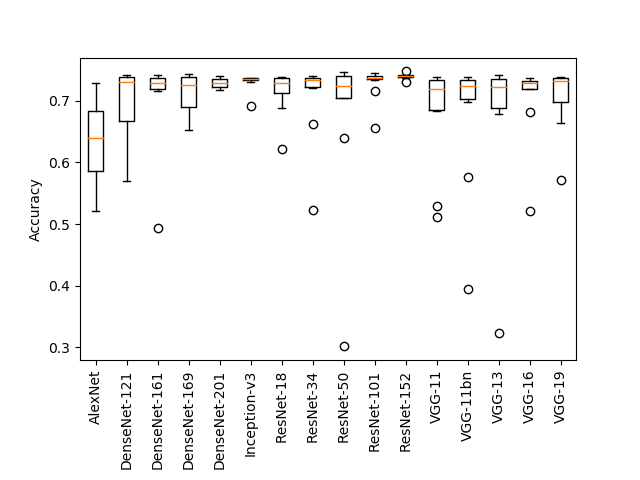}
\caption{Validation Accuracy (Pretrained, Fixed Feature Extractor)}
\label{fig:Diabetic_Retinopathy_acc_boxplot_h}
\end{subfigure}
\caption[Continued - Diabetic Retinopathy: Network accuracy boxplots.]{Continued - Diabetic Retinopathy: Network accuracy boxplots.}
\label{fig:DR_network_acc_boxplots4}
\end{figure}

\begin{figure}[htb]
\begin{subfigure}{0.6\textwidth}
\captionsetup{width=0.6\textwidth}
\centering\includegraphics[width=0.6\linewidth]{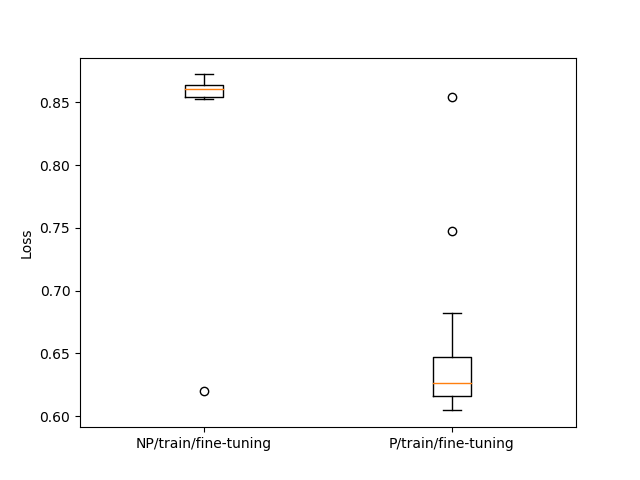}
\caption{p=0.002}
\label{fig:DR_loss_boxplots_a}
\end{subfigure}
\begin{subfigure}{0.6\textwidth}
\captionsetup{width=0.6\textwidth}
\centering\includegraphics[width=0.6\linewidth]{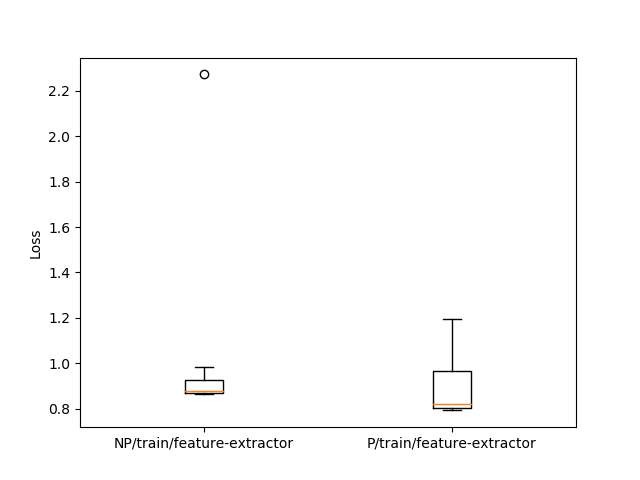}
\caption{p=0.301}
\label{fig:DR_loss_boxplots_b}
\end{subfigure}
\begin{subfigure}{0.6\textwidth}
\captionsetup{width=0.6\textwidth}
\centering\includegraphics[width=0.6\linewidth]{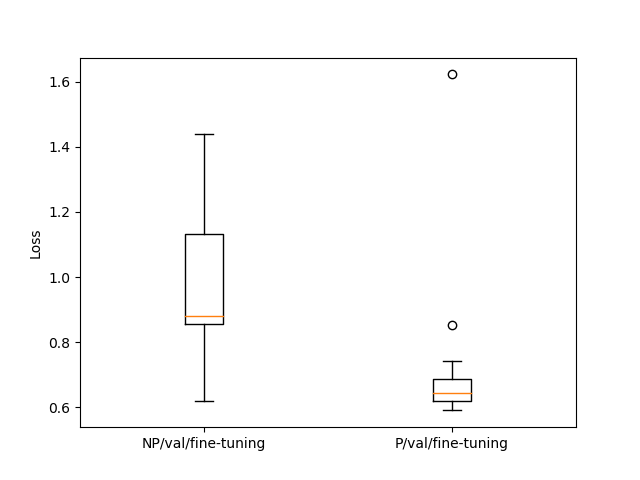}
\caption{p=0.002}
\label{fig:DR_loss_boxplots_c}
\end{subfigure}
\begin{subfigure}{0.6\textwidth}
\captionsetup{width=0.6\textwidth}
\centering\includegraphics[width=0.6\linewidth]{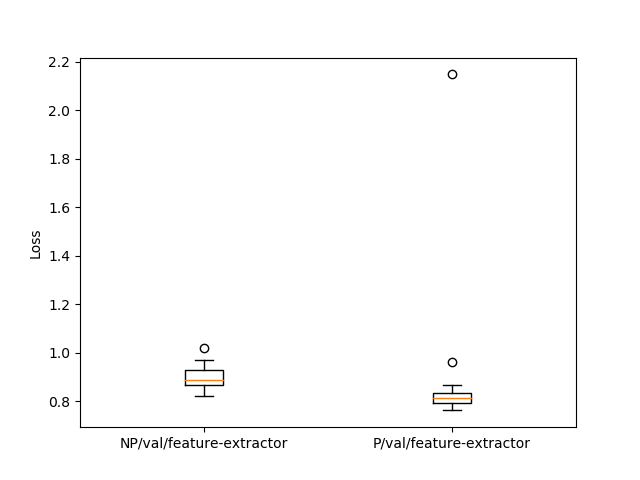}
\caption{p=0.034}
\label{fig:DR_loss_boxplots_d}
\end{subfigure}
\begin{subfigure}{0.6\textwidth}
\captionsetup{width=0.6\textwidth}
\centering\includegraphics[width=0.6\linewidth]{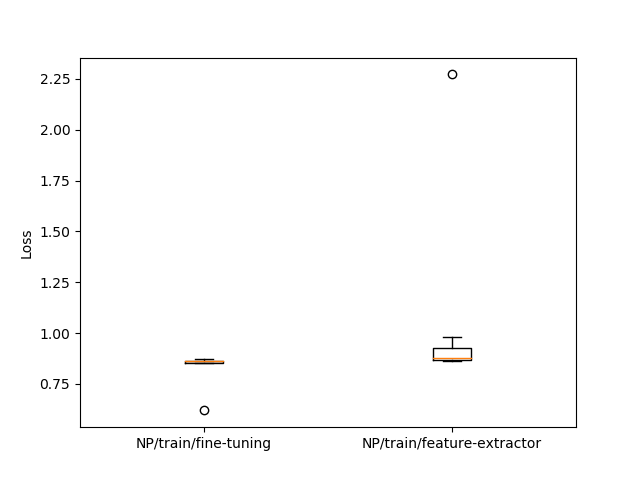}
\caption{p$<$0.001}
\label{fig:DR_loss_boxplots_e}
\end{subfigure}
\begin{subfigure}{0.6\textwidth}
\captionsetup{width=0.6\textwidth}
\centering\includegraphics[width=0.6\linewidth]{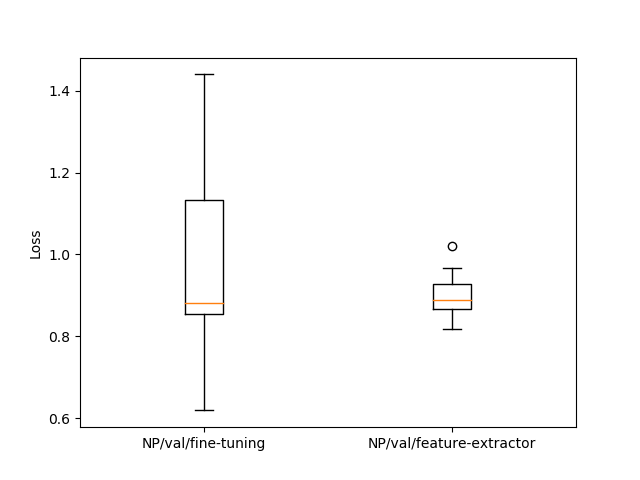}
\caption{p=0.569} 
\label{fig:DR_loss_boxplots_f}
\end{subfigure}
\begin{subfigure}{0.6\textwidth}
\captionsetup{width=0.6\textwidth}
\centering\includegraphics[width=0.6\linewidth]{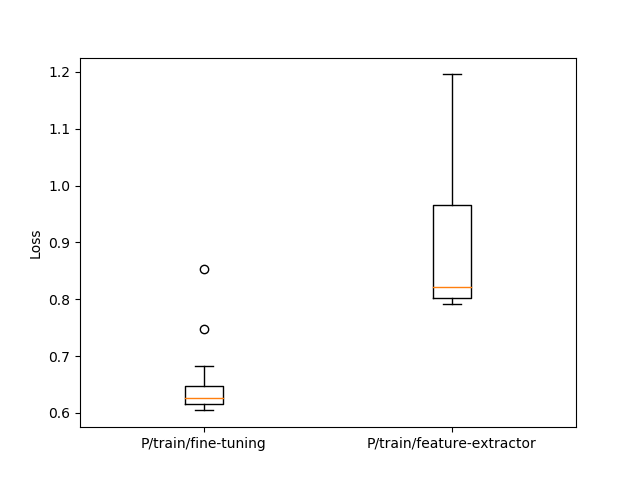}
\label{fig:DR_loss_boxplots_g}
\caption{p$<$0.001} 
\end{subfigure}
\begin{subfigure}{0.6\textwidth}
\captionsetup{width=0.6\textwidth}
\centering\includegraphics[width=0.6\linewidth]{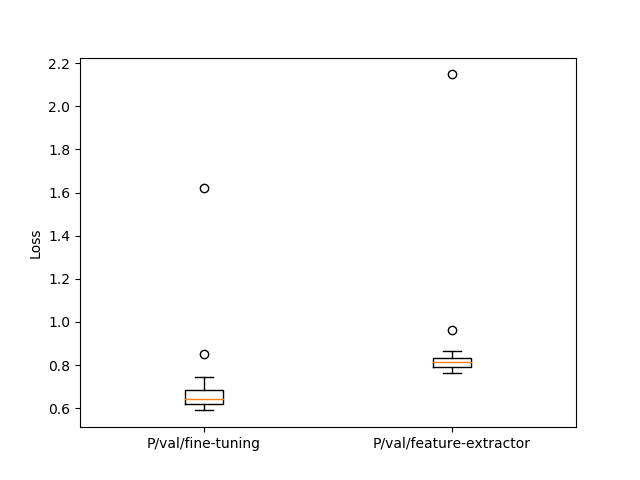}
\caption{p=0.006}
\label{fig:DR_loss_boxplots_h}
\end{subfigure}
\caption[Network loss boxplot comparisons.]{\textbf{Network loss boxplot comparisons.} Each boxplot indicates the two groups compared. Each label indicates 3 variables: if networks were pretrained (P) or not pretrained (P), training (train) or validation phase, and evaluated as a fixed feature extractor or as fine-tuning the networks. p-values indicated are Wilcoxon signed-rank (\ref{fig:DR_loss_boxplots_a}-\ref{fig:DR_loss_boxplots_h}) or Mann-Whitney (\ref{fig:DR_loss_boxplots_i}-\ref{fig:DR_loss_boxplots_l}) tests comparing the two groups, with p $<$ 0.05 indicating a statistically significant difference.}
\label{fig:DR_loss_boxplots}
\end{figure}

\begin{figure}
\ContinuedFloat
\begin{subfigure}{0.6\textwidth}
\captionsetup{width=0.6\textwidth}
\centering\includegraphics[width=0.6\linewidth]{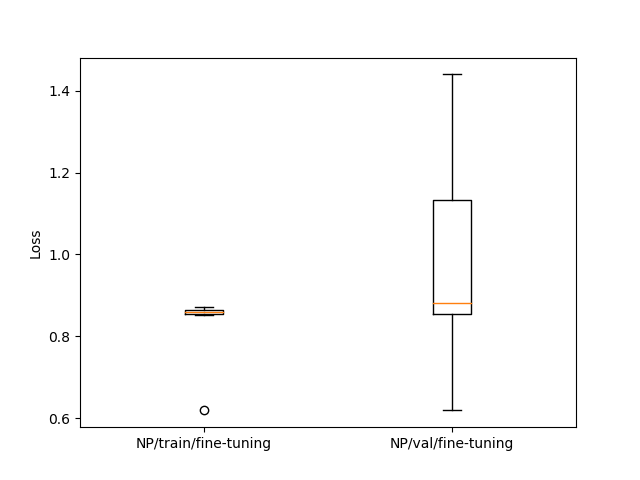}
\caption{p=0.053}
\label{fig:DR_loss_boxplots_i}
\end{subfigure}
\begin{subfigure}{0.6\textwidth}
\captionsetup{width=0.6\textwidth}
\centering\includegraphics[width=0.6\linewidth]{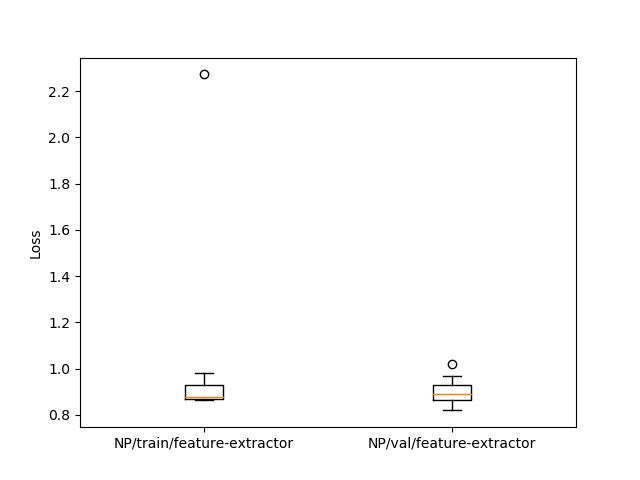}
\caption{p=0.462}
\label{fig:DR_loss_boxplots_j}
\end{subfigure}
\begin{subfigure}{0.6\textwidth}
\captionsetup{width=0.6\textwidth}
\centering\includegraphics[width=0.6\linewidth]{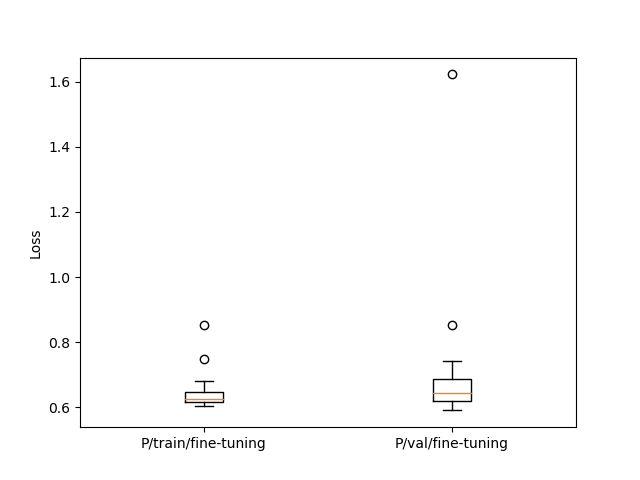}
\caption{p=0.220}
\label{fig:DR_loss_boxplots_k}
\end{subfigure}
\begin{subfigure}{0.6\textwidth}
\captionsetup{width=0.6\textwidth}
\centering\includegraphics[width=0.6\linewidth]{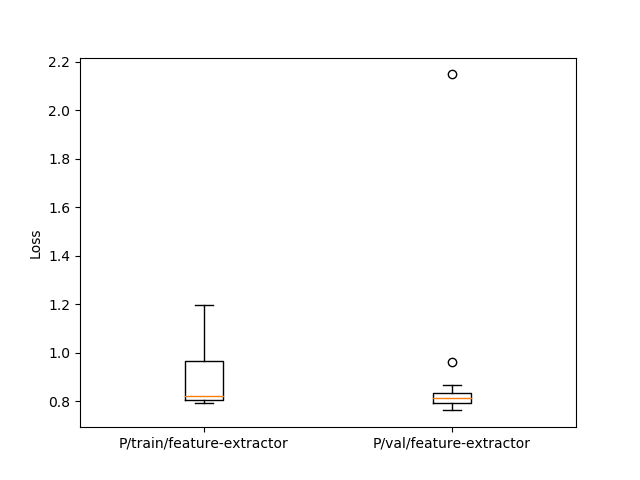}
\caption{p=0.084}
\label{fig:DR_loss_boxplots_l}
\end{subfigure}
\caption[Network loss boxplot comparisons - continued.]{\textbf{Network loss boxplot comparisons - continued.} Each boxplot indicates the two groups compared. Each label indicates 3 variables: if networks were pretrained (P) or not pretrained (P), training (train) or validation phase, and evaluated as a fixed feature extractor or as fine-tuning the networks. p-values indicated are Wilcoxon signed-rank (\ref{fig:DR_loss_boxplots_a}-\ref{fig:DR_loss_boxplots_h}) or Mann-Whitney (\ref{fig:DR_loss_boxplots_i}-\ref{fig:DR_loss_boxplots_l}) tests comparing the two groups, with p $<$ 0.05 indicating a statistically significant difference.}
\label{fig:DR_loss_boxplots2}
\end{figure}

\begin{figure}[htb]
\begin{subfigure}{0.6\textwidth}
\captionsetup{width=0.6\textwidth}
\centering\includegraphics[width=0.6\linewidth]{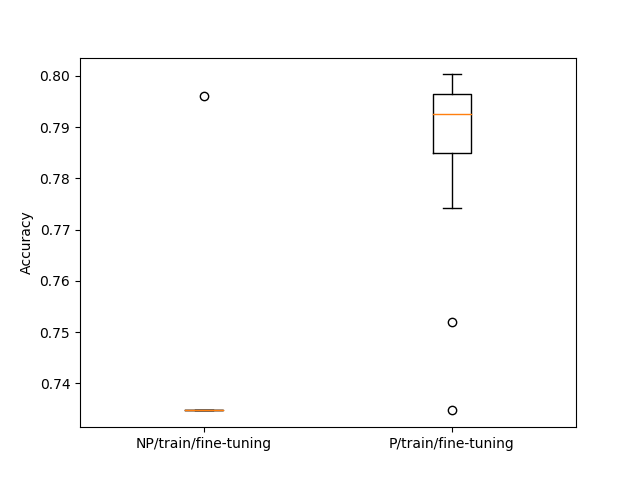}
\caption{p=0.004}
\label{fig:DR_acc_boxplots_a}
\end{subfigure}
\begin{subfigure}{0.6\textwidth}
\captionsetup{width=0.6\textwidth}
\centering\includegraphics[width=0.6\linewidth]{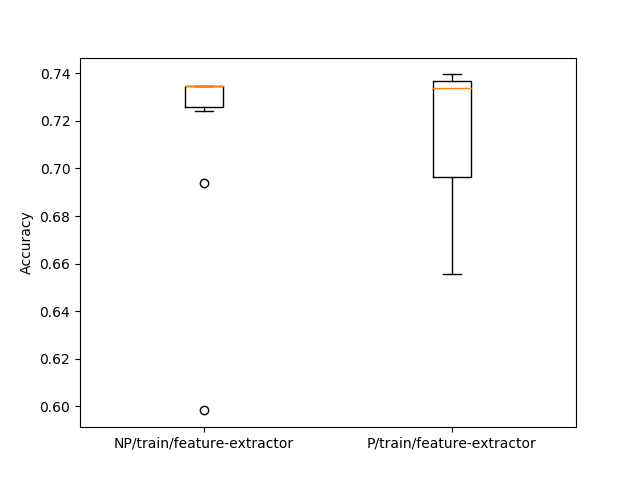}
\caption{p=0.642}
\label{fig:DR_acc_boxplots_b}
\end{subfigure}
\begin{subfigure}{0.6\textwidth}
\captionsetup{width=0.6\textwidth}
\centering\includegraphics[width=0.6\linewidth]{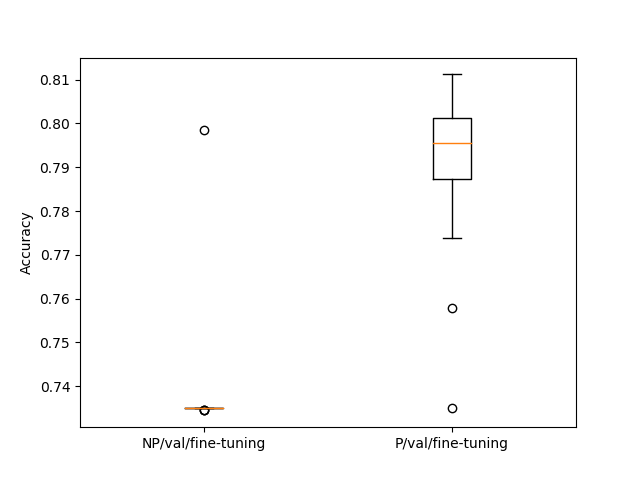}
\caption{p=0.003}
\label{fig:DR_acc_boxplots_c}
\end{subfigure}
\begin{subfigure}{0.6\textwidth}
\captionsetup{width=0.6\textwidth}
\centering\includegraphics[width=0.6\linewidth]{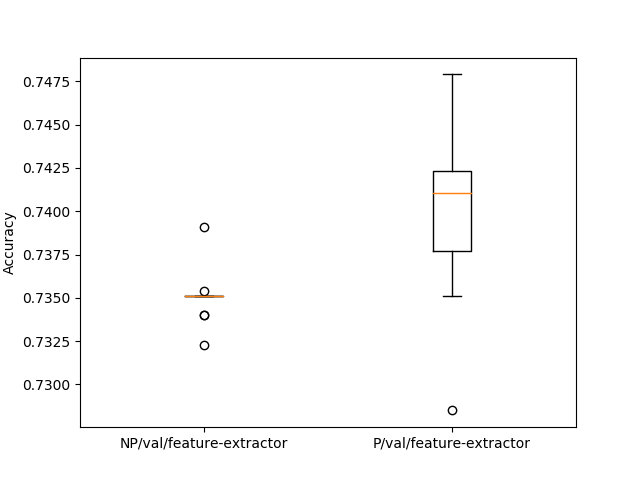}
\caption{p=0.006}
\label{fig:DR_acc_boxplots_d}
\end{subfigure}
\begin{subfigure}{0.6\textwidth}
\captionsetup{width=0.6\textwidth}
\centering\includegraphics[width=0.6\linewidth]{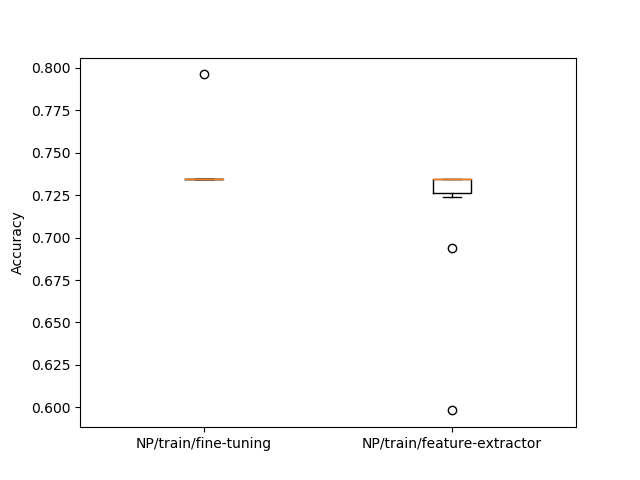}
\caption{p=0.005}
\label{fig:DR_acc_boxplots_e}
\end{subfigure}
\begin{subfigure}{0.6\textwidth}
\captionsetup{width=0.6\textwidth}
\centering\includegraphics[width=0.6\linewidth]{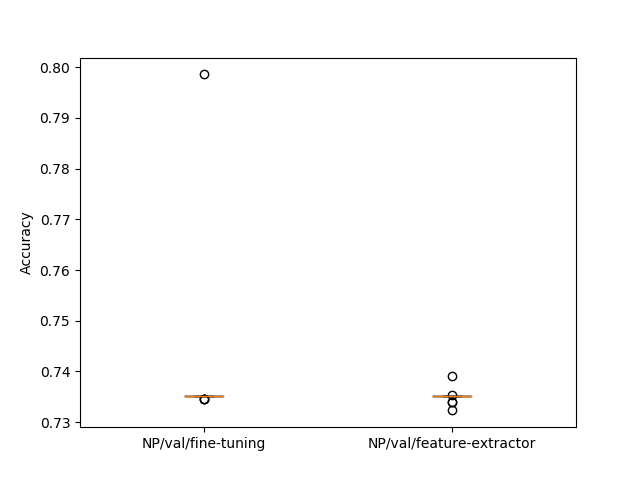}
\caption{p=0.248}
\label{fig:DR_acc_boxplots_f}
\end{subfigure}
\begin{subfigure}{0.6\textwidth}
\captionsetup{width=0.6\textwidth}
\centering\includegraphics[width=0.6\linewidth]{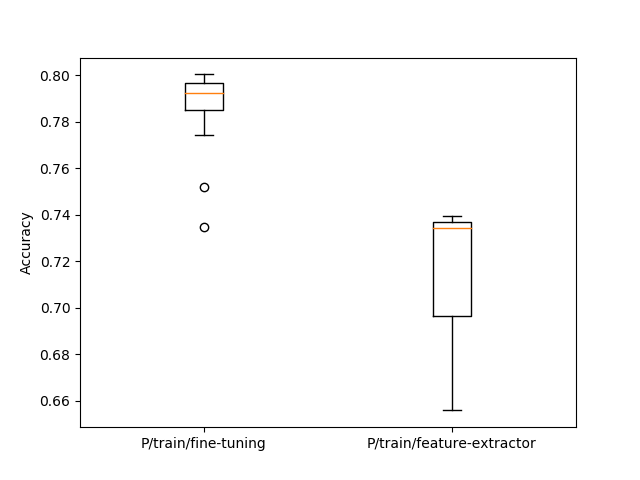}
\caption{p=0.001}
\label{fig:DR_acc_boxplots_g}
\end{subfigure}
\begin{subfigure}{0.6\textwidth}
\captionsetup{width=0.6\textwidth}
\centering\includegraphics[width=0.6\linewidth]{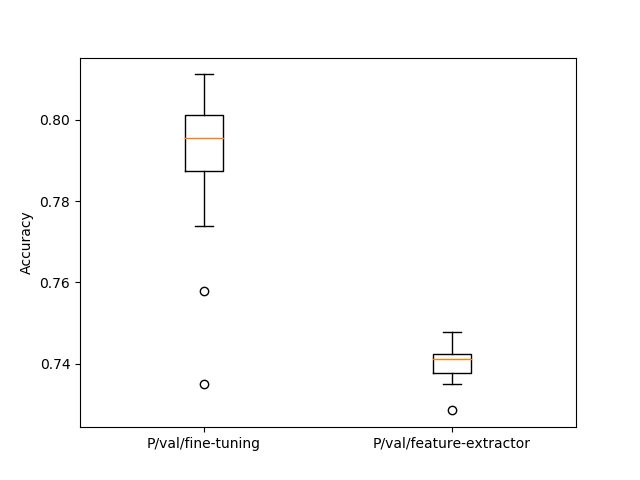}
\caption{p=0.001}
\label{fig:DR_acc_boxplots_h}
\end{subfigure}
\caption[Network accuracy boxplot comparisons.]{\textbf{Network accuracy boxplot comparisons.} Each boxplot indicates the two groups compared. Each label indicates 3 variables: if networks were pretrained (P) or not pretrained (P), training (train) or validation phase, and evaluated as a fixed feature extractor or as fine-tuning the networks. p-values indicated are Wilcoxon signed-rank (\ref{fig:DR_acc_boxplots_a}-\ref{fig:DR_acc_boxplots_h}) or Mann-Whitney (\ref{fig:DR_acc_boxplots_i}-\ref{fig:DR_acc_boxplots_l}) tests comparing the two groups, with p $<$ 0.05 indicating a statistically significant difference.}
\label{fig:DR_acc_boxplots}
\end{figure}

\begin{figure}
\ContinuedFloat
\begin{subfigure}{0.6\textwidth}
\captionsetup{width=0.6\textwidth}
\centering\includegraphics[width=0.6\linewidth]{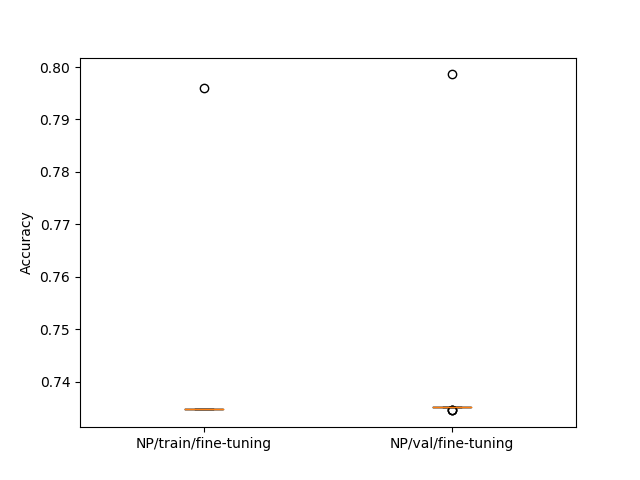}
\caption{p=0.003}
\label{fig:DR_acc_boxplots_i}
\end{subfigure}
\begin{subfigure}{0.6\textwidth}
\captionsetup{width=0.6\textwidth}
\centering\includegraphics[width=0.6\linewidth]{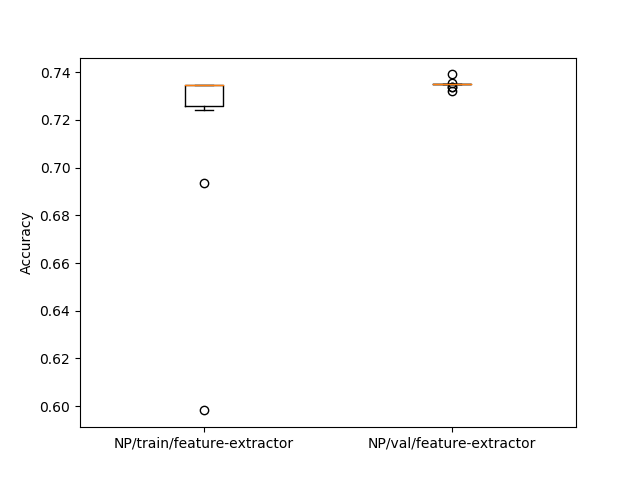}
\caption{p$<$0.001}
\label{fig:DR_acc_boxplots_j}
\end{subfigure}
\begin{subfigure}{0.6\textwidth}
\captionsetup{width=0.6\textwidth}
\centering\includegraphics[width=0.6\linewidth]{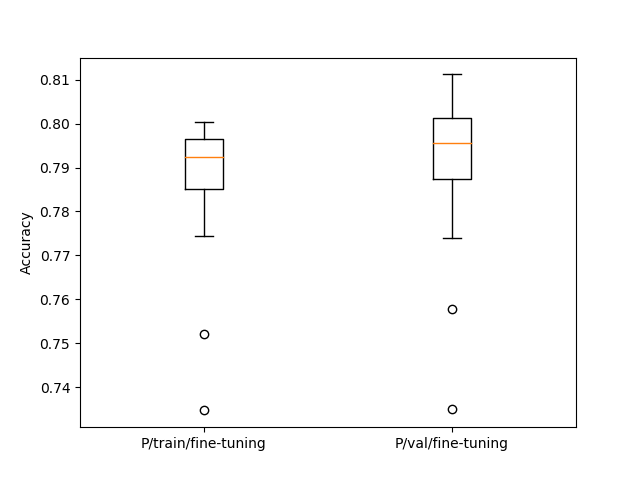}
\caption{p=0.146}
\label{fig:DR_acc_boxplots_k}
\end{subfigure}
\begin{subfigure}{0.6\textwidth}
\captionsetup{width=0.6\textwidth}
\centering\includegraphics[width=0.6\linewidth]{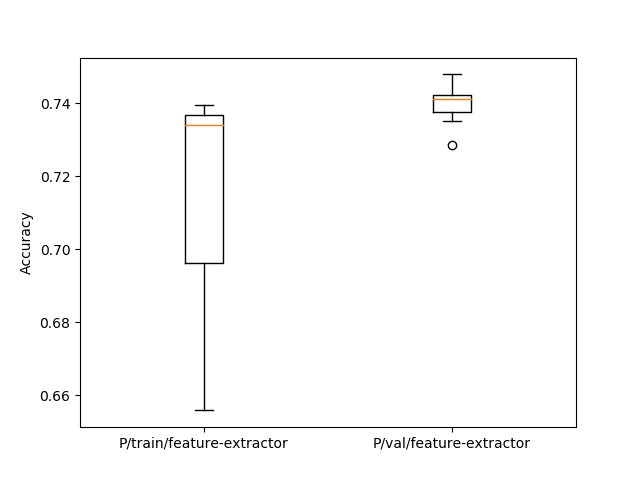}
\caption{p$<$0.001}
\label{fig:DR_acc_boxplots_l}
\end{subfigure}
\caption[Network accuracy boxplot comparisons - continued.]{\textbf{Network accuracy boxplot comparisons - continued.} Each boxplot indicates the two groups compared. Each label indicates 3 variables: if networks were pretrained (P) or not pretrained (P), training (train) or validation phase, and evaluated as a fixed feature extractor or as fine-tuning the networks. p-values indicated are Wilcoxon signed-rank (\ref{fig:DR_acc_boxplots_a}-\ref{fig:DR_acc_boxplots_h}) or Mann-Whitney (\ref{fig:DR_acc_boxplots_i}-\ref{fig:DR_acc_boxplots_l}) tests comparing the two groups, with p $<$ 0.05 indicating a statistically significant difference.}
\label{fig:DR_acc_boxplots2}
\end{figure}

\chapter{OCT-based Classification Confusion Matrices\label{chap:append_CM}}

This appendix shows the OCT-based classification confusion matrices for the networks listed below, where each network was pre-trained and run in fixed-feature extractor and fine-tuning modes. All networks were run for 25 epochs.
\begin{itemize}
\item AlexNet
\item DenseNet-121
\item DenseNet-161
\item Inception\_v3
\item ResNet-18
\item ResNet-34
\item VGG-11
\item VGG-19
\end{itemize}

\begin{figure}[htb]
\begin{subfigure}{0.8\textwidth}
\captionsetup{width=0.8\textwidth}
\centering\includegraphics[width=0.8\linewidth]{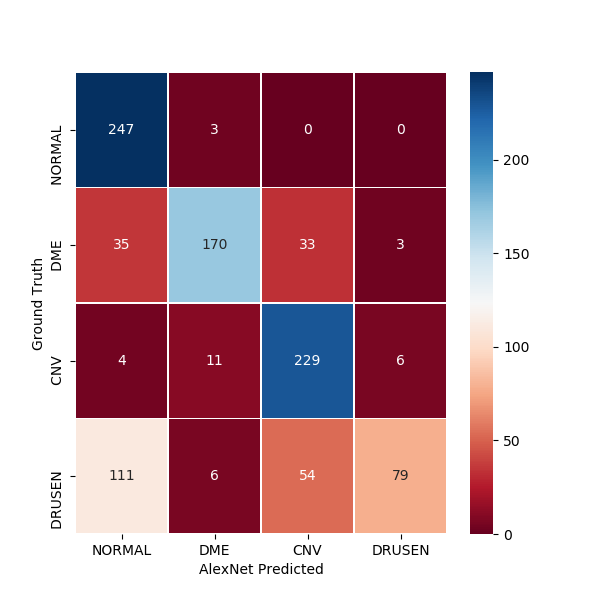}
  \caption{AlexNet: Fixed-feature extractor.}
  \label{fig:alexnet_CM_a}
\end{subfigure}
\begin{subfigure}{0.8\textwidth}
\captionsetup{width=0.8\textwidth}
\centering\includegraphics[width=0.8\linewidth]{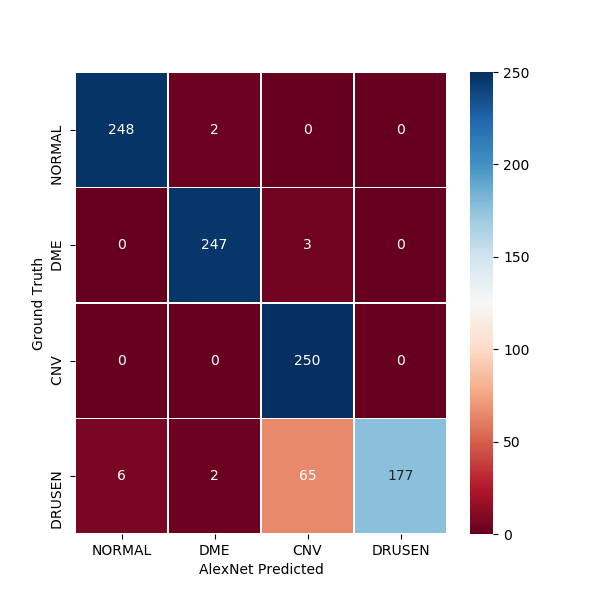}
  \caption{AlexNet: Fine-tuning.}
  \label{fig:alexnet_CM_b}
\end{subfigure}
\caption[OCT-based Classification Confusion Matrices.]{OCT-based Classification  Confusion Matrices.}
\label{fig:oct_CM}
\end{figure}

\begin{figure}[htb!]
\ContinuedFloat
\begin{subfigure}{0.8\textwidth}
\captionsetup{width=0.8\textwidth}
\centering\includegraphics[width=0.8\linewidth]{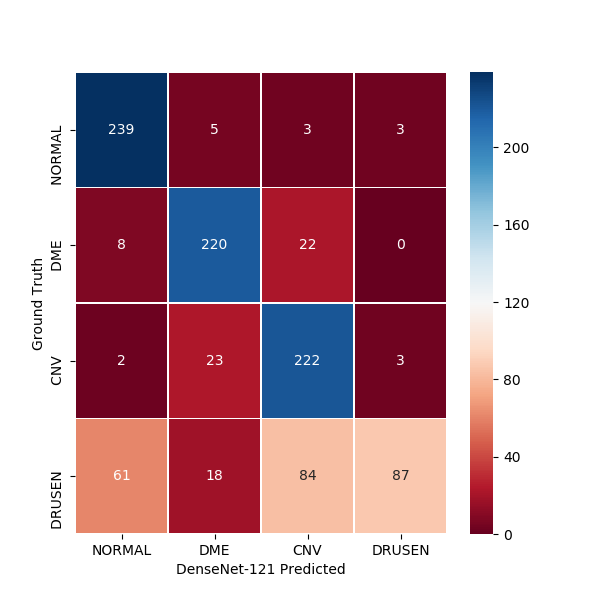}
  \caption{DenseNet-121: Fixed-feature extractor.}
  \label{fig:densenet121_CM_a}
\end{subfigure}
\begin{subfigure}{0.8\textwidth}
\captionsetup{width=0.8\textwidth}
\centering\includegraphics[width=0.8\linewidth]{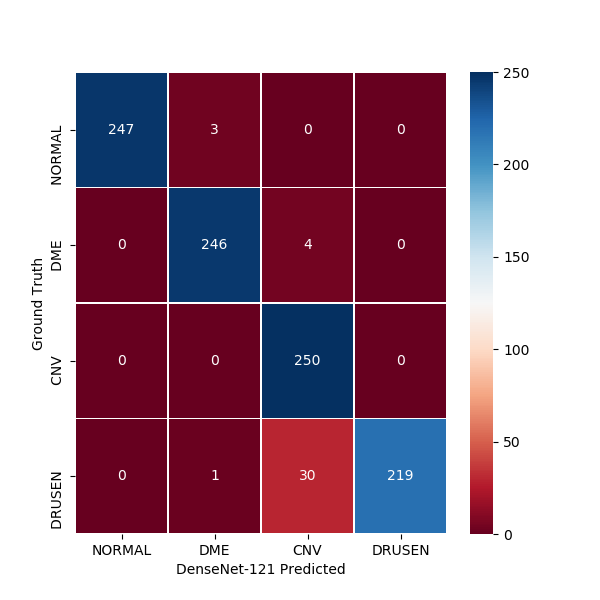}
  \caption{DenseNet-121: Fine-tuning.}
  \label{fig:densenet121_CM_b}
\end{subfigure}
\caption[Continued - OCT-based classification Confusion Matrices.]{OCT-based classification Confusion Matrices.}
\label{fig:oct_CM2}
\end{figure}

\begin{figure}[htb!]
\ContinuedFloat
\begin{subfigure}{0.8\textwidth}
\captionsetup{width=0.8\textwidth}
\centering\includegraphics[width=0.8\linewidth]{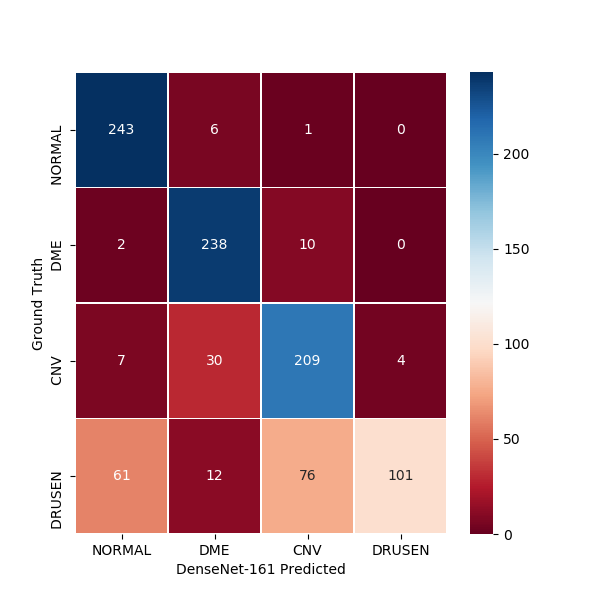}
  \caption{DenseNet-161: Fixed-feature extractor.}
  \label{fig:densenet161_CM_a}
\end{subfigure}
\begin{subfigure}{0.8\textwidth}
\captionsetup{width=0.8\textwidth}
\centering\includegraphics[width=0.8\linewidth]{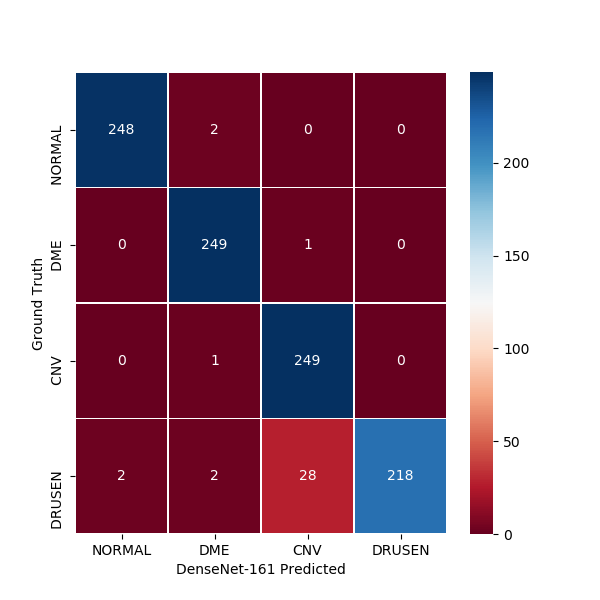}
  \caption{DenseNet-161: Fine-tuning.}
  \label{fig:densenet161_CM_b}
\end{subfigure}
\caption[Continued - OCT-based classification Confusion Matrices.]{OCT-based classification Confusion Matrices.}
\label{fig:oct_CM3}
\end{figure}

\begin{figure}[htb!]
\ContinuedFloat
\begin{subfigure}{0.8\textwidth}
\captionsetup{width=0.8\textwidth}
\centering\includegraphics[width=0.8\linewidth]{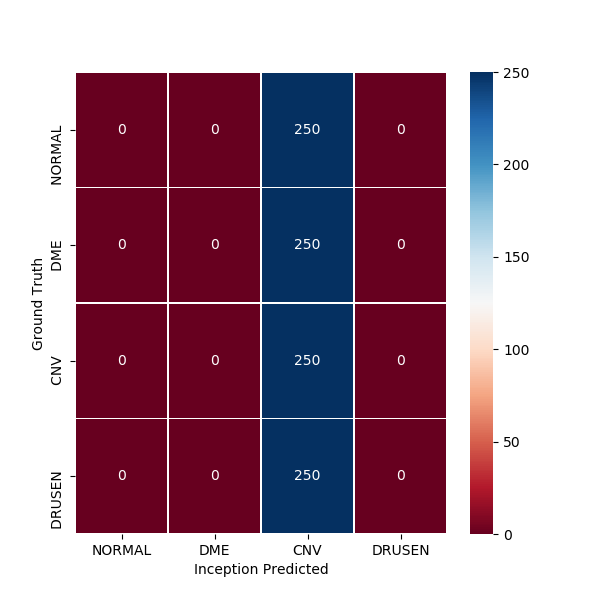}
  \caption{Inception\_v3: Fixed-feature extractor.}
  \label{fig:inception_CM_a}
\end{subfigure}
\begin{subfigure}{0.8\textwidth}
\captionsetup{width=0.8\textwidth}
\centering\includegraphics[width=0.8\linewidth]{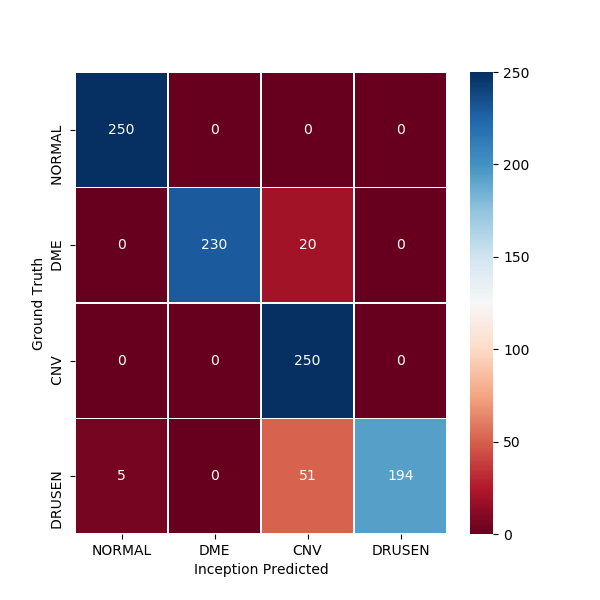}
  \caption{Inception\_v3: Fine-tuning.}
  \label{fig:inception_CM_b}
\end{subfigure}
\caption[Continued - OCT-based classification Confusion Matrices.]{OCT-based classification Confusion Matrices.}
\label{fig:oct_CM4}
\end{figure}

\begin{figure}[htb!]
\ContinuedFloat
\begin{subfigure}{0.8\textwidth}
\captionsetup{width=0.8\textwidth}
\centering\includegraphics[width=0.8\linewidth]{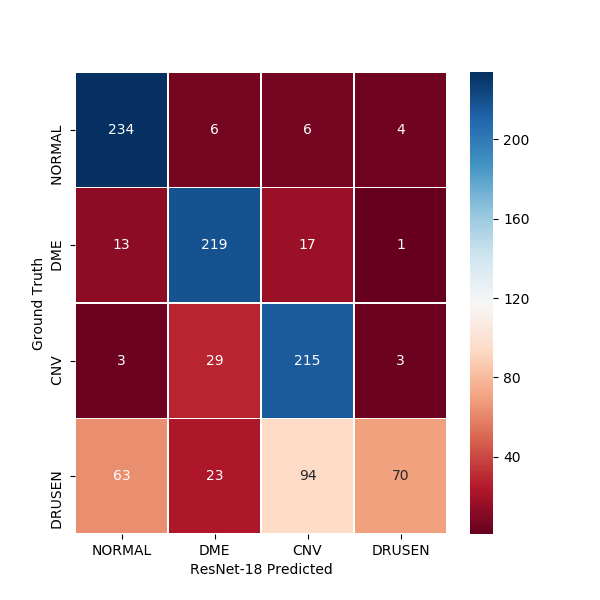}
  \caption{ResNet-18: Fixed-feature extractor.}
  \label{fig:resnet18_CM_a}
\end{subfigure}
\begin{subfigure}{0.8\textwidth}
\captionsetup{width=0.8\textwidth}
\centering\includegraphics[width=0.8\linewidth]{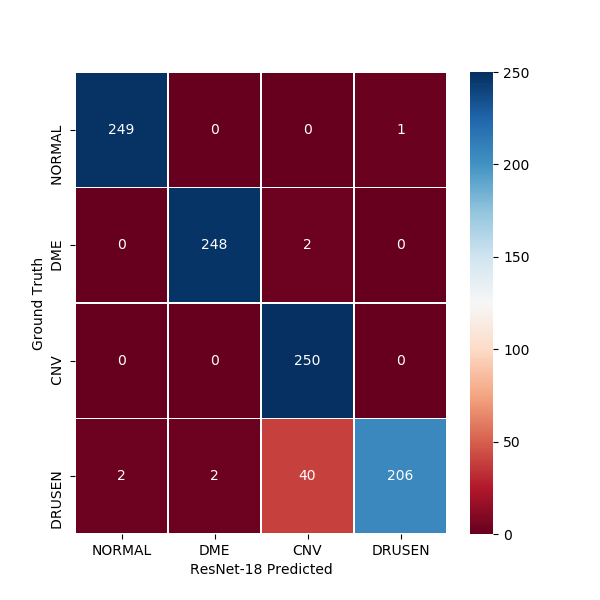}
  \caption{ResNet-18: Fine-tuning.}
  \label{fig:resnet18_CM_b}
\end{subfigure}
\caption[Continued - OCT-based classification Confusion Matrices.]{OCT-based classification Confusion Matrices.}
\label{fig:oct_CM4}
\end{figure}

\begin{figure}[htb!]
\ContinuedFloat
\begin{subfigure}{0.8\textwidth}
\captionsetup{width=0.8\textwidth}
\centering\includegraphics[width=0.8\linewidth]{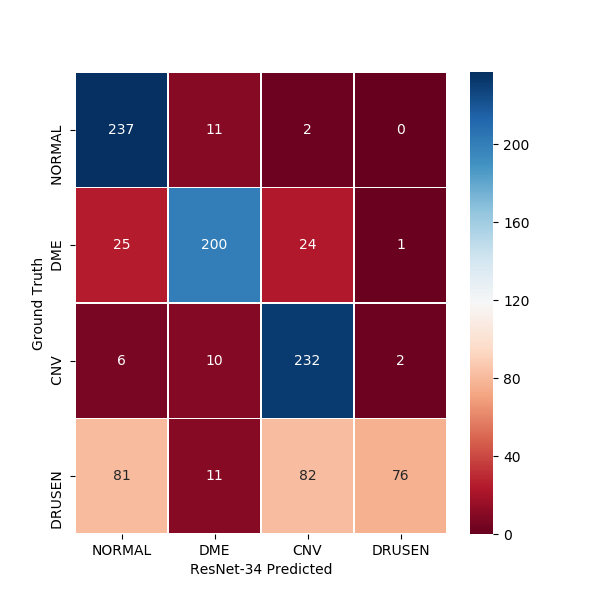}
  \caption{ResNet-34: Fixed-feature extractor.}
  \label{fig:resnet34_CM_a}
\end{subfigure}
\begin{subfigure}{0.8\textwidth}
\captionsetup{width=0.8\textwidth}
\centering\includegraphics[width=0.8\linewidth]{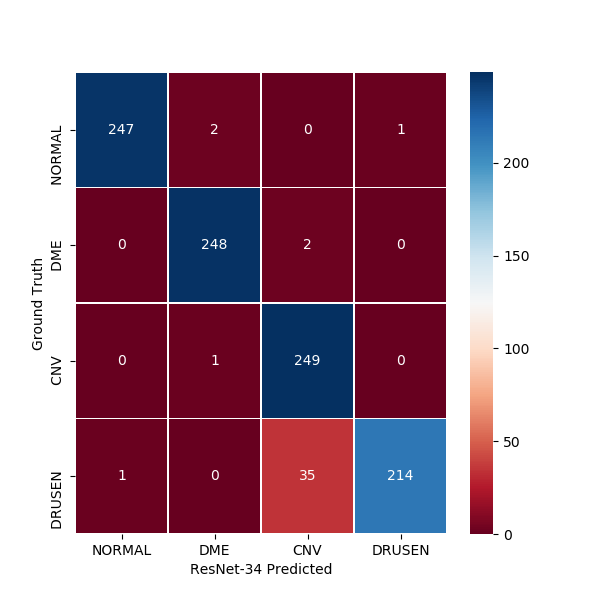}
  \caption{ResNet-34: Fine-tuning.}
  \label{fig:resnet34_CM_b}
\end{subfigure}
\caption[Continued - OCT-based classification Confusion Matrices.]{OCT-based classification Confusion Matrices.}
\label{fig:oct_CM5}
\end{figure}

\begin{figure}[htb!]
\ContinuedFloat
\begin{subfigure}{0.8\textwidth}
\captionsetup{width=0.8\textwidth}
\centering\includegraphics[width=0.8\linewidth]{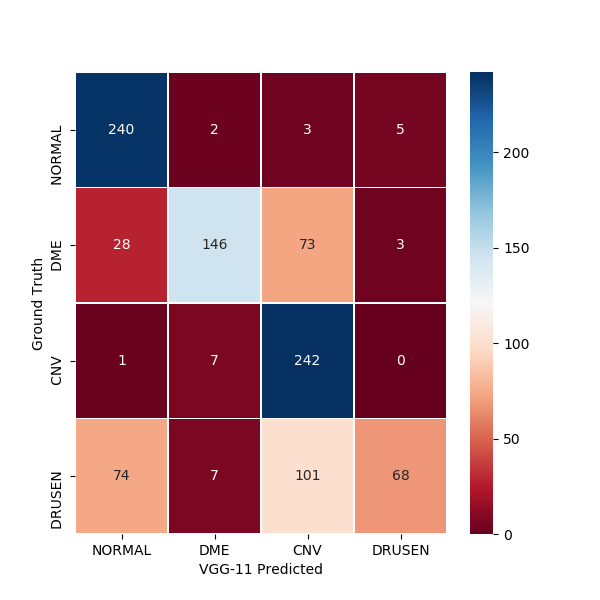}
  \caption{VGG-11: Fixed-feature extractor.}
  \label{fig:vgg11_CM_a}
\end{subfigure}
\begin{subfigure}{0.8\textwidth}
\captionsetup{width=0.8\textwidth}
\centering\includegraphics[width=0.8\linewidth]{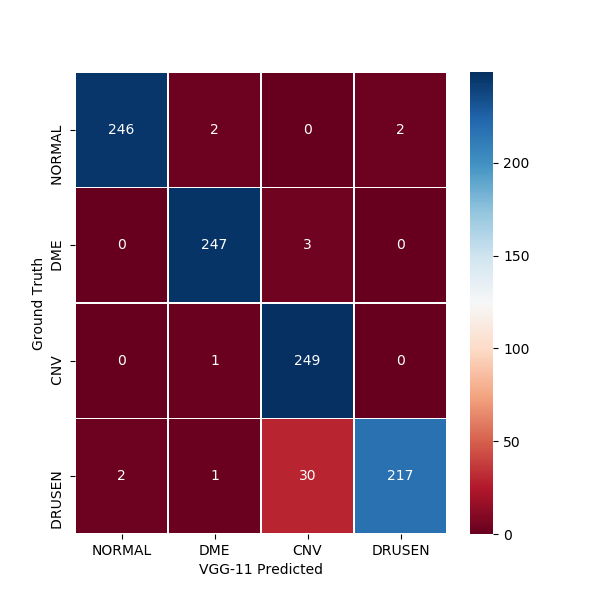}
  \caption{VGG-11: Fine-tuning.}
  \label{fig:vgg11_CM_b}
\end{subfigure}
\caption[Continued - OCT-based classification Confusion Matrices.]{OCT-based classification Confusion Matrices.}
\label{fig:oct_CM6}
\end{figure}

\begin{figure}[htb!]
\ContinuedFloat
\begin{subfigure}{0.8\textwidth}
\captionsetup{width=0.8\textwidth}
\centering\includegraphics[width=0.8\linewidth]{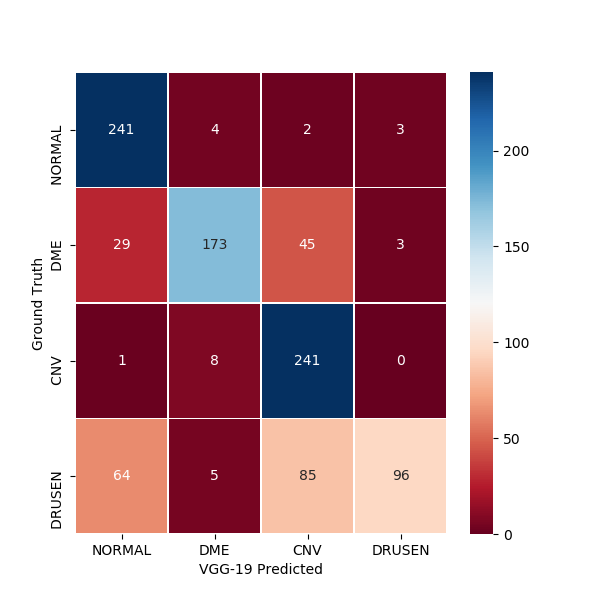}
  \caption{VGG-19: Fixed-feature extractor.}
  \label{fig:vgg19_CM_a}
\end{subfigure}
\begin{subfigure}{0.8\textwidth}
\captionsetup{width=0.8\textwidth}
\centering\includegraphics[width=0.8\linewidth]{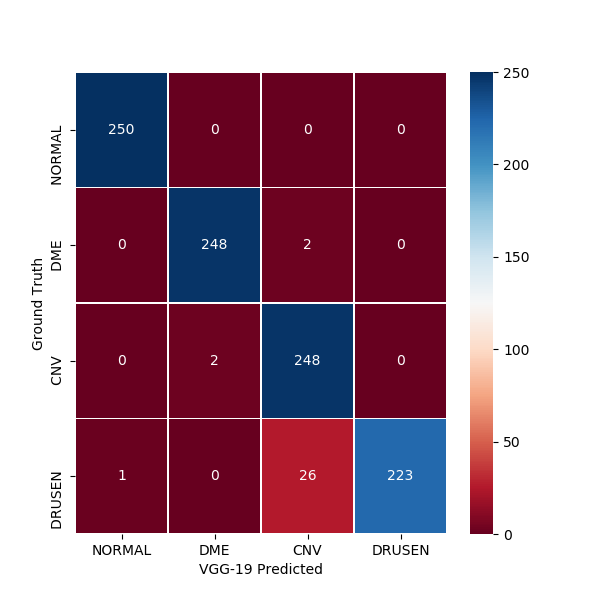}
  \caption{VGG-19: Fine-tuning.}
  \label{fig:vgg19_CM_b}
\end{subfigure}
\caption[Continued - OCT-based classification Confusion Matrices.]{OCT-based classification Confusion Matrices.}
\label{fig:oct_CM7}
\end{figure}

%


\begin{thebibliography}{99}\addcontentsline{toc}{chapter}{Bibliography}

\bibitem{transfer_learning} Online notes for CS231n: Convolutional Neural Networks for Visual Recognition. \url{http://cs231n.github.io/transfer-learning/}. Accessed December 16, 2018.
\bibitem{computer_vision_notes} New York University Computer Vision (CSCI-GA.2271-001) course notes. Slides: Neural nets 1 and 2, Convolutional Networks 1 and 2, Object Classification 1 and 2. \url{https://cs.nyu.edu/~fergus/teaching/vision/index.html} Accessed December 14, 2018.
\bibitem{Goodfellow} Goodfellow I, Bengio Y, Courville A. Deep Learning (online book). \url{http://www.deeplearningbook.org/}. Accessed December 14, 2018.
\bibitem{LeCun} Y. LeCun, L. Bottou, Y. Bengio, \etal  Gradient-Based Learning Applied to Document Recognition, Proceedings of the IEEE, 86(11):2278-2324, November 1998
\bibitem{Ciresan} Ciresan D, Meier U, Schmidhuber J. Multi-column Deep Neural Networks for Image Classification. arXiv:1202.2745 [cs.CV]
\bibitem{Krizhevsky} Krizhevsky A, Sutskever I, Hinton GE. ImageNet Classification with Deep Convolutional Neural Networks. Advances in Neural Information Processing Systems 25.2012.
\bibitem{Resnet} He K, Zhang X, Ren S, \etal Deep Residual Learning for Image Recognition. arXiv:1512.03385 [cs.CV].
\bibitem{Densenet} Huang G, Liu Z, van der Maaten L, \etal Densely Connected Convolutional Networks. arXiv:1608.06993 [cs.CV].
\bibitem{vgg} Simonyan K, Zisserman A. Very Deep Convolutional Networks for Large-Scale Image Recognition. arXiv:1409.1556 [cs.CV].
\bibitem{inception} Szegedy C, Liu W, Jia Y, \etal Going Deeper with Convolutions. arXiv:1409.4842 [cs.CV]. 
\bibitem{DR-PPP} Diabetic Retinopathy Preferred Practice Pattern (2017), published by the American Academy of Ophthalmology. Available at: \url{https://www.aao.org/preferred-practice-pattern/diabetic-retinopathy-ppp-updated-2017}
\bibitem{DR-fact-sheet}Centers for Disease Control and Prevention. National diabetes fact sheet: national estimates and general information on diabetes and prediabetes in the United States, 2011. Atlanta, GA: U.S. Department of Health and Human Services, Centers for Disease Control and Prevention; 2011. Available at: \url{www.cdc.gov/diabetes/pubs/pdf/ndfs_2011.pdf}. Accessed June 11, 2014.
\bibitem{DM-prevalence}Cowie CC, Rust KF, Byrd-Holt DD, \etal Prevalence of diabetes and impaired fasting glucose in adults in the U.S. population: National Health And Nutrition Examination Survey 1999-2002. Diabetes Care 2006;29(6):1263-8.
\bibitem{DR-lifetime-risk}Narayan KM, Boyle JP, Thompson TJ, \etal Lifetime risk for diabetes mellitus in the United States. JAMA 2003;290(14):1884-90.
\bibitem{DR-epi}Klein BE. Overview of epidemiologic studies of diabetic retinopathy. Ophthalmic Epidemiol 2007;14(4):179-83.
\bibitem{DM-prevalence2}Kempen JH, O'Colmain BJ, Leske MC, \etal The prevalence of diabetic retinopathy among adults in the United States. Arch Ophthalmol 2004;122(4):552-63.
\bibitem{DM-prevalence3}Zhang X, Saaddine JB, Chou CF, \etal Prevalence of diabetic retinopathy in the United States, 2005-2008. JAMA 2010;304(6):649-56.
\bibitem{DR-screening1}Williams GA, Scott IU, Haller JA, \etal Single-field fundus photography for diabetic retinopathy screening: a report by the American Academy of Ophthalmology. Ophthalmology 2004;111(5):1055-62.
\bibitem{DR-screening2}Lin DY, Blumenkranz MS, Brothers RJ, \etal The sensitivity and specificity of single-field nonmydriatic monochromatic digital fundus photography with remote image interpretation for diabetic retinopathy screening: a comparison with ophthalmoscopy and standardized mydriatic color photography. Am J Ophthalmol 2002;134(2):204-13.
\bibitem{DR-screening3}Larsen N, Godt J, Grunkin M, \etal Automated detection of diabetic retinopathy in a fundus photographic screening population. Invest Ophthalmol Vis Sci 2003;44(2):767-71.
\bibitem{DR-screening4}Leese GP, Ellis JD, Morris AD, \etal Does direct ophthalmoscopy improve retinal screening for diabetic eye disease by retinal photography? Diabet Med 2002;19(10):867-9.
\bibitem{DR-screening5}Ahmed J, Ward TP, Bursell SE, \etal The sensitivity and specificity of nonmydriatic digital stereoscopic retinal imaging in detecting diabetic retinopathy. Diabetes Care 2006;29(10):2205-9.
\bibitem{DR-screening6}Velez R, Haffner S, Stern MP, \etal Ophthalmologist vs retinal photographs in screening for diabetic retinopathy. Clinical Research 1987;35(3):A363.
\bibitem{DR-screening7}Pugh JA, Jacobson JM, Van Heuven WA, \etal Screening for diabetic retinopathy. The wide-angle retinal camera. Diabetes Care 1993;16(6):889-95.
\bibitem{DR-screening8}Lawrence MG. The accuracy of digital-video retinal imaging to screen for diabetic retinopathy: an analysis of two digital-video retinal imaging systems using standard stereoscopic seven-field photography and dilated clinical examination as reference standards. Trans Am Ophthalmol Soc 2004;102:321-40.
\bibitem{DR-screening9}Abramoff MD, Folk JC, Han DP, \etal Automated analysis of retinal images for detection of referable diabetic retinopathy. JAMA Ophthalmol 2013;131(3):351-7.
\bibitem{Brady_2014} Brady CJ, Villanti AC, Pearson JL, \etal Rapid grading of fundus photographs for diabetic retinopathy using crowdsourcing. J Med Internet Res. 2014;16(10):e233.
\bibitem{Mitry_2013} Mitry D, Peto T, Hayat S, \etal Crowdsourcing as a novel technique for retinal fundus photography classification: analysis of images in the EPIC Norfolk Cohort on behalf of the UKBiobank eye and vision consortium. PLoS One. 2013;8(8):e71154.
\bibitem{Brady_2017} Mudie LI, Wang X, Friedman DS, \etal Crowdsourcing and Automated Retinal Image Analysis for Diabetic Retinopathy. Curr Diab Rep. 2017 Sep 23;17(11):106. doi: 10.1007/s11892-017-0940-x. Review.  
\bibitem{Trucco} Trucco E, Ruggeri A, Karnowski T, \etal Validating retinal fundus image analysis algorithms: issues and a proposal. Invest Ophthalmol Vis Sci.  2013;54(5):3546–59.
\bibitem{Kaggle_DR} Kaggle Diabetic Retinopathy Detection competition. \url{https://www.kaggle.com/c/diabetic-retinopathy-detection}. Accessed November 21, 2018.
\bibitem{Graham} Competition report (min-pooling) for Kaggle Diabetic Retinopathy Detection competition. \url{https://storage.googleapis.com/kaggle-forum-message-attachments/88655/2795/competitionreport.pdf}. Accessed November 30, 2018.
\bibitem{team_o_O} Team o\_O Competition report and code. \url{https://storage.googleapis.com/kaggle-forum-message-attachments/88693/2797/report.pdf}. Accessed November 30, 2018.
\bibitem{pytorch_tutorial} Pytorch transfer learning tutorial. \url{https://pytorch.org/tutorials/beginner/transfer_learning_tutorial.html}. Accessed December 3, 2018.
\bibitem{ratio_source} Github blog entry: Question about the training loss and validation loss. \url{https://github.com/karpathy/char-rnn/issues/160}. Accessed December 3, 2018.
\bibitem{Gulshan} Gulshan, V, Peng, L, Coram, M, \etal Development and Validation of a Deep Learning Algorithm for Detection of Diabetic Retinopathy in Retinal Fundus Photographs. JAMA. 2016;316(22):2402-2410.
\bibitem{Pratt} Pratt, H, Coenenb, F, Broadbent, DM, \etal Convolutional Neural Networks for Diabetic Retinopathy. Procedia Computer Science 90(2016)200–205.
\bibitem{Schwartz} Schwartz B, Martin SW. Principles of Epidemiology and Public Health. Principles and Practice of Pediatric Infectious Diseases. 4th Edition. 2012.
\bibitem{zago} Zago GT, Andreao RV, Dorizzi B, \etal Retinal image quality assessment using deep learning. Comput Biol Med. 2018 Oct 11;103:64-70.
\bibitem{AMD-PPP} Age-Related Macular Degneration Preferred Practice Pattern (2015), published by the American Academy of Ophthalmology. Available at: \url{https://www.aao.org/preferred-practice-pattern/age-related-macular-degeneration-ppp-2015}.
\bibitem{OCT_review} Adhi M, Duker JS. Optical Coherence Tomography - Current and Future Applications. Curr Opin Ophthalmol. 2013 May ; 24(3): 213–221.
\bibitem{OCT_paper1} Huang D, Swanson EA, Lin CP, \etal Optical coherence tomography. Science. 1991; 254:1178– 1181.
\bibitem{OCT_paper2} Sull AC, Vuong LN, Price LL, \etal Comparison of spectral/Fourier domain optical coherence tomography instruments for assessment of normal macular thickness. Retina. 2010; 30:235–245.
\bibitem{OCT_book} Schuman JS, Puliafito CA, Fujimoto JG, \etal Optical Coherence Tomography of Ocular Diseases. SLACK Incorporated. Third Edition. 2013.
\bibitem{OCT_numbers} Swanson EA, Huang D. Ophthalmic OCT reaches \$1 billion per year. Retinal Physician. 2011; 45: 58–59.
\bibitem{Machine_learning_DME_paper} Alsaih K, Lemaitre G, Rastgoo M, \etal Machine learning techniques for diabetic macular edema (DME) classification on SD‑OCT images. BioMed Eng OnLine (2017) 16:68.
\bibitem{Automated_DME_analysis} Kamble RM, Chan GCY, Perdomo O, \etal. Automated Diabetic Macular Edema (DME) Analysis Using Fine Tuning with Inception-Resnet-v2 on OCT Images. Conf Proc IEEE Eng Med Biol Soc. 2018 Jul;2018:2715-2718.
\bibitem{inception-resnet} Szegedy C, Ioffe S, Vanhoucke V, \etal. Inception-v4, Inception-ResNet and the Impact of Residual Connections on Learning. arXiv:1602.07261.
\bibitem{automated_staging} Venhuizen FG, van Ginneken B, van Asten F, \etal. Automated Staging of Age-Related Macular Degeneration Using Optical Coherence Tomography. Invest Ophthalmol Vis Sci. 2017 Apr 1;58(4):2318-2328.
\bibitem{OCT_paper} Kermany DS, Goldbaum M, Cai W, \etal Identifying Medical Diagnoses and Treatable Diseases by Image-Based Deep Learning. Cell 172, 1122–1131.
\bibitem{stats-book} Pagano, RR. Understanding Statistics in the Behavioral Sciences. Cengage Learning; 10th edition (January 1, 2012).
\bibitem{wilcoxon} Wilcoxon signed-rank test Wikipedia entry. \url{https://en.wikipedia.org/wiki/Wilcoxon_signed-rank_test}. Accessed December 5, 2018.
\bibitem{mann-whitney} Mann–Whitney U test Wikipedia entry. \url{https://en.wikipedia.org/wiki/Mann%E2%80%93Whitney_U_test}. Accessed December 5, 2018. 
\bibitem{kappa} Cohen, J. Weighted kappa: Nominal scale agreement with provision for scaled disagreement or partial credit. Psychological Bulletin. 70(4):213–220.
\bibitem{wiki_kappa} Cohen's kappa Wikipedia entry. \url{https://en.wikipedia.org/wiki/Cohen%27s_kappa}. Accessed December 16, 2018.
\end{thebibliography}
\end{document}